\documentclass{article} % For LaTeX2e
\usepackage{iclr2027_conference,times}

\usepackage{amsmath,amsfonts,bm}

\def\eqref#1{equation~\ref{#1}}
\def\1{\bm{1}}

\DeclareMathAlphabet{\mathsfit}{\encodingdefault}{\sfdefault}{m}{sl}
\SetMathAlphabet{\mathsfit}{bold}{\encodingdefault}{\sfdefault}{bx}{n}

\usepackage{hyperref}
\usepackage{url}

\usepackage[utf8]{inputenc} % allow utf-8 input
\usepackage[T1]{fontenc}    % use 8-bit T1 fonts
\usepackage{hyperref}       % hyperlinks
\usepackage{url}            % simple URL typesetting
\usepackage{booktabs}       % professional-quality tables
\usepackage{amsfonts}       % blackboard math symbols
\usepackage{nicefrac}       % compact symbols for 1/2, etc.
\usepackage{microtype}      % microtypography
\usepackage{xcolor}         % colors

\usepackage{graphicx}
\usepackage[accsupp]{axessibility}  
\usepackage[table]{xcolor} % \rowcolor, \cellcolor 사용
\usepackage{multirow}
\usepackage{float}
\usepackage{placeins}
\usepackage{wrapfig}
\usepackage{makecell}
\usepackage[figuresright]{rotating}
\usepackage{orcidlink}

\usepackage{array}
\usepackage{tikz}
\usepackage{amsthm}

\newtheorem{theorem}{Theorem}
\newtheorem{proposition}{Proposition}
\newtheorem*{proposition*}{Proposition}
\usepackage{subcaption}
\usepackage{adjustbox}

\title{Anchor to Expand: Semantic Anchoring for Personalized Text-to-Image Diffusion Models}
\author{
Seoyun Yang\textsuperscript{*} \quad
Gihoon Kim\textsuperscript{*} \quad
Taesup Kim\textsuperscript{\textdagger} \\
Graduate School of Data Science, Seoul National University \\
\textsuperscript{*}Equal contribution. \quad
\textsuperscript{\textdagger}Corresponding author.
}

\newcolumntype{C}[1]{>{\centering\arraybackslash}m{#1}}

\newcommand{\resultimg}[1]{%
    \includegraphics[
        width=\linewidth,
        height=18mm,
        keepaspectratio
    ]{#1}
}

\newcommand{\methodheader}[1]{%
    {\fontsize{5.7}{6.0}\selectfont\bfseries
    \makecell[c]{#1}}%
}

\newcolumntype{C}[1]{%
    >{\centering\arraybackslash}m{#1}%
}

\newcommand{\figfiveheader}[1]{%
    {\fontsize{7.0}{7.5}\selectfont\bfseries #1}%
}

\newcommand{\figfivesubheader}[1]{%
    {\fontsize{6.5}{7.0}\selectfont\bfseries #1}%
}

\newcommand{\figfiveprompt}[1]{%
    {\scriptsize\itshape #1}%
}

\newcommand{\anchorimg}[1]{%
    \includegraphics[width=\linewidth]{#1}%
}

\newcommand{\anchorheader}[1]{%
    {\fontsize{5.7}{6.2}\selectfont\bfseries #1}%
}

\newcommand{\anchorlabel}[1]{%
    {\fontsize{5.3}{5.8}\selectfont\bfseries
    \makecell[c]{#1}}%
}

\newcommand{\suppheader}[1]{%
    {\scriptsize\bfseries
    \begin{tabular}{c}
        #1
    \end{tabular}}%
}

\newcommand{\suppbackbone}[1]{%
    {\scriptsize\bfseries #1}%
}

\newcommand{\suppprompt}[1]{%
    {\fontsize{6}{6.5}\selectfont\itshape #1}%
}

\newcommand{\suppimg}[1]{%
    \includegraphics[
        width=\linewidth,
        height=\linewidth,
        keepaspectratio
    ]{#1}%
}

\newcommand{\wheader}[1]{%
    {\scriptsize\bfseries #1}%
}

\newcommand{\wprompt}[1]{%
    {\fontsize{6}{6.5}\selectfont\itshape
    \begin{tabular}{c}
        #1
    \end{tabular}}%
}

\newcommand{\paheader}[1]{%
    {\scriptsize\bfseries #1}%
}

\newcommand{\palabel}[1]{%
    {\scriptsize\bfseries
    \begin{tabular}{c}
        #1
    \end{tabular}}%
}

\newcommand{\paprompt}[1]{%
    {\fontsize{6}{6.5}\selectfont\itshape #1}%
}

\newcommand{\paimg}[1]{%
    \includegraphics[
        width=\linewidth,
        height=\linewidth,
        keepaspectratio
    ]{#1}%
}

\newcommand{\infheader}[1]{%
    {\scriptsize\bfseries
    \begin{tabular}{c}
        #1
    \end{tabular}}%
}

\newcommand{\infprompt}[1]{%
    {\fontsize{6}{6.5}\selectfont\itshape
    \begin{tabular}{c}
        #1
    \end{tabular}}%
}

\newcommand{\infimg}[1]{%
    \includegraphics[
        width=\linewidth,
        height=\linewidth,
        keepaspectratio
    ]{#1}%
}

\iclrfinalcopy % Uncomment for camera-ready version, but NOT for submission.
\begin{document}

\maketitle
% \lhead{}
\fancyhead{}
\renewcommand{\headrulewidth}{0pt}

% One of the ultimate goals in text-to-image 

\begin{abstract}

Personalizing text-to-image diffusion models extends pretrained models to represent novel user-specific concepts from only a few reference images. However, learning a new concept while building on the prior knowledge of the pretrained model remains a key challenge. When personalization focuses on learning the target concept, the model tends to overfit the reference examples and degrade its general capability. In contrast, emphasizing prior preservation can hinder capturing distinctive personalized attributes. In this paper, we address this challenge by viewing a personalized concept as an underrepresented concept whose semantic counterpart is well represented in the pretrained model. Rather than treating the learning of a new concept and prior preservation as separate objectives, we reformulate them as a single anchored learning problem. We therefore introduce \textit{Semantic Anchoring Personalization} (SAP), which keeps concept learning grounded in the pretrained semantic structure while capturing subject-specific attributes. The proposed objective offers a simple yet effective formulation that can be applied across different model backbones without architectural modifications or auxiliary networks. Extensive experiments across various settings demonstrate that SAP achieves a better balance between subject fidelity and text-image alignment than baseline methods. Further ablation studies validate the contribution of semantic anchoring to personalization.

% A central challenge in text-to-image diffusion is to generate user-specified details that extend beyond the generic generative capability of pretrained models. Personalization emerges as a task for introducing novel concepts into pretrained models. However, adapting a model to personalized semantics while preserving its prior knowledge remains unstable. The key challenge lies in learning a new concept from a limited number of reference images. When the model focuses on subject fidelity, it tends to overfit the few examples and deviates from the original distribution. In contrast, emphasizing prior preservation prevents the model from learning new personalized attributes. In this paper, we show that this issue arises when the model learns a personalized concept in isolation from the pretrained semantics. To address this, we propose a new objective that enables interactive expansion from the prior distribution during personalization. The proposed approach reformulates personalization as learning a novel concept guided by its general counterpart through semantic anchoring. Our anchoring strategy offers a simple yet effective approach that integrates with existing personalization settings without additional computational resources, making it applicable across different methods and setups. As a result, the proposed method achieves stable adaptation and consistent improvements in both subject fidelity and text-image alignment compared to baseline methods. Extensive experiments and ablations further validate the robustness and effectiveness of our method.

\end{abstract}

\section{Introduction}
\label{sec:intro}

Recent advancements in text-to-image diffusion models~\citep{ramesh2021zero, ramesh2022hierarchical, rombach2022high, schuhmann2022laion, podell2023sdxl, esser2024scaling} have established powerful foundation models for generic content generation. Despite their strong performance, these models still struggle to depict user-preferred details that are not captured by textual conditioning. To address this gap, personalized generation has emerged as a major application of text-to-image generative models~\citep{gal2022image,ruiz2023dreambooth}. Rather than relying on samples drawn from the learned distribution, personalization aims to introduce user-specific attributes into the generation process.

However, a key challenge in personalization is capturing user-desired attributes from only a handful of examples while leveraging the prior knowledge of the pretrained model. This few-shot personalization often leads the model to overfit to limited data and lose the knowledge encoded in the pretrained prior. To mitigate this issue, existing methods constrain the update components of the pretrained model to prevent such drift. These approaches include selective layer updates~\citep{kumari2023multi, tewel2023key}, additional normalization~\citep{han2023svdiff, qiu2023controlling}, and parameter-efficient modules~\citep{ hu2022lora, yeh2023navigating, chenpersonalizing}. While restricting parameter updates enhances training stability, it also limits sufficient learning of target attributes. 

Alongside efforts to preserve the prior knowledge of the pretrained model, recent studies have explored ways to better capture fine-grained personalized attributes. Building on Classifier-Free Guidance~\citep{dhariwal2021diffusion, ho2022classifier}, some methods leverage guidance from the pretrained model~\citep{park2025steering, phunyaphibarn2025unconditional}. Other approaches disentangle the target concept from irrelevant attributes using image-specific descriptions obtained from additional vision-language models~\citep{Kim_2026_CVPR, liu2026accord}. These approaches improve the representation of personalized attributes using additional semantic information. Prior studies have explored these complementary directions independently, treating prior preservation and personalized attribute learning as separate problems.

% 방법 칭찬&한계 % The two complementary directions of preserving pretrained semantics and learning personalized attributes have been studied in parallel.
% 학습과 인퍼런스에서 해당 시맨틱을 잘 표현할 수 있도록
% these approaches consider guidance mechanisms that improve the representation of intended personalized semantics. 

% Building on Classifier-Free Guidance~\citep{dhariwal2021diffusion, ho2022classifier}, some methods leverage guidance from the pretrained model~\citep{park2025steering, phunyaphibarn2025unconditional}, while others use additional captions to separate the target concept from irrelevant attributes in the reference images~\citep{Kim_2026_CVPR, liu2026accord}. The two complementary directions of preserving pretrained semantics and learning personalized attributes have been studied in parallel.

% The core question has thus shifted from how well the model reconstructs the concept to how effectively it can guide generation toward the personalized semantics. Recent studies on personalization go beyond learning the denoising process for novel concepts and instead focus on guidance mechanisms that steer generation toward the intended semantics during inference~\citep{park2025steering, phunyaphibarn2025unconditional}. Based on Classifier-Free Guidance~\citep{dhariwal2021diffusion, ho2022classifier, park2025steering, liu2026accord, Kim_2026_CVPR}, they investigate how to guide personalized generation using the unconditional guidance of the original pretrained model.

In this paper, we view learning personalized attributes and preserving pretrained semantics through the lens of rare concept generation~\citep{samuel2023norm, samuel2024generating, um2025minority}. Rare concept generation aims to improve the generation of specific concepts that are underrepresented in pretrained models. This setting resembles personalization, where the model learns fine-grained user-specific attributes from only a few examples. In particular, Rare-to-Frequent~\citep{park2024rare} provides a theoretical basis by showing that a concept underrepresented in the pretrained model can benefit from guidance provided by its more common semantic counterpart. We extend this insight to personalization by treating a personalized concept represented by only a few reference images as a rare concept and anchoring its learning to its more common semantic counterpart. 
% However, the relationship between personalized concepts and pretrained knowledge has received limited attention in personalization~\citep{soboleva2025beyond}.

%이러한 지점은 personalization에서 limited scope에서만 적용 fully leverage 되지 않았음 ~\citep{soboleva2025beyond}

\begin{figure}[t]
    \centering
    \includegraphics[width=0.5\linewidth]{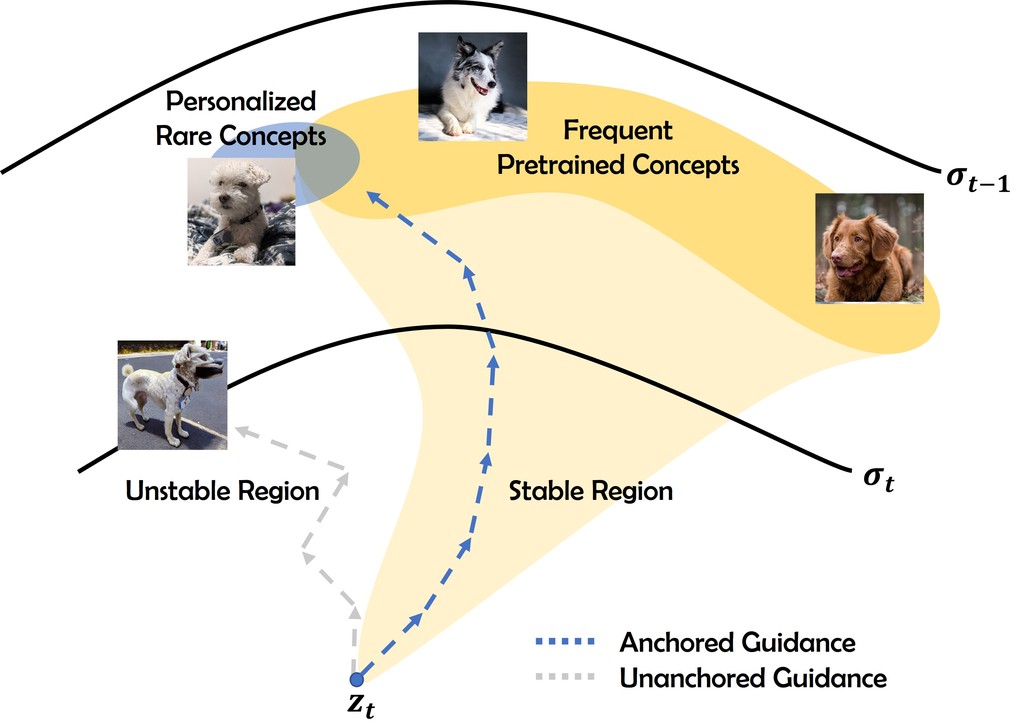}
    \caption{Conceptual illustration of our method, Semantic Anchoring Personalization (SAP). Pretrained semantics provide stable guidance, whereas guidance in newly introduced regions remains unstable. Our approach anchors personalization to pretrained semantics, enabling stable guidance as the model expands toward novel concepts.}
    \label{fig:concept}
\end{figure}

Therefore, we propose \textit{Semantic Anchoring Personalization}~(SAP), a reformulated personalization objective that models the relation between the target concept and its pretrained semantic counterpart. As illustrated in Fig.~\ref{fig:concept}, SAP learns the personalized concept as an extension of the pretrained semantic structure with the general semantic counterpart serving as an anchor. Our analysis further shows that the relation between a personalized concept and its general semantic counterpart weakens as personalization proceeds. This divergence indicates that the personalized concept loses alignment with the pretrained semantic structure. This observation suggests that SAP can mitigate such divergence by grounding subject-specific learning in pretrained semantics. Quantitative and qualitative results demonstrate consistent improvements across different model backbones, while ablation studies validate the role of semantic anchoring in balancing subject fidelity and text-image alignment.
In summary, our main contributions are as follows:
\begin{itemize}
    % \item We take a joint view of learning a new concept and preserving pretrained semantics through the lens of rare concept generation.
    % Inspired by the relation between underrepresented and well-represented concepts, we reformulate the learning target to incorporate the corresponding semantic counterpart of the target concept.
    % \item We introduce a perspective on concept learning and pretrained semantic preservation in personalization through the lens of the relation between underrepresented and well-represented concepts.
    \item We introduce a perspective on concept learning and pretrained semantic preservation in personalization based on the relationship between underrepresented and well-represented concepts.
    \item We propose \textit{Semantic Anchoring Personalization} (SAP), a reformulated personalization objective that incorporates the relation between the target concept and its pretrained semantic counterpart.
    \item Our experiments and analyses demonstrate that our method achieves consistent improvements in fidelity and generalization across diverse backbones and outperforms existing baselines.
\end{itemize}

\section{Related Work}
\label{sec:RelatedWork}

% \subsection{Personalized Text-to-Image Generation}

Personalization methods have explored which model components should be adapted for a new concept. DreamBooth~\citep{ruiz2023dreambooth} trains all model parameters, which can make personalization sensitive to overfitting on limited reference images. In contrast, Textual Inversion~\citep{gal2022image} optimizes a token-level representation, which often suffers from limited expressiveness in capturing the target concept. These challenges have motivated subsequent studies to identify which parameters should be adapted for personalization. Custom Diffusion~\citep{kumari2023multi} trains the attention layers, while other studies modify specific components such as singular values~\citep{han2023svdiff} or angular features~\citep{qiu2023controlling} of weight matrices. Low-rank adaptation strategies~\citep{hu2022lora} are also widely studied for personalization. They introduce a small set of trainable parameters while keeping the original weights fixed. Recent work investigates how to improve the initialization~\citep{yeh2023navigating} or training efficiency~\citep{chenpersonalizing}. While restricting parameter updates can stabilize personalization, selecting which components to adapt may need to be reconsidered for different model architectures. Moreover, these approaches preserve pretrained semantics through frozen parameters rather than incorporating them into the learning of the target concept.

Recent studies have shifted their focus from parameter adaptation toward better capturing fine-grained personalized attributes. \citet{park2025steering} and \citet{phunyaphibarn2025unconditional} leverage pretrained unconditional guidance as an inference-time control signal. However, these methods do not address the learning of the personalized representation itself~\citep{soboleva2025beyond}. \citet{liu2026accord} and \citet{Kim_2026_CVPR} use image-specific captions to disentangle target attributes from irrelevant factors in the reference images. However, these approaches rely on additional vision-language models to obtain such descriptions during personalization. Some methods instead use superclass semantics available within the personalization setup~\citep{ruiz2023dreambooth, tewel2023key}. \citet{huang2024classdiffusion} aligns personalized token representations with the superclass, while \citet{qiao2024facechain} regularizes personalized predictions to remain within the pretrained semantic space using unconditional and class-level outputs. These approaches use general semantics as preservation constraints during adaptation. In contrast, we view the personalized concept and its general counterpart as a rare-to-frequent pair and incorporate their relationship into the learning objective.

\section{Method}
%\vspace{-5pt}
\label{sec:method}
In this section, we first introduce the notation and the baseline formulation used in personalization. Next, we show that the learned personalized representation diverges from the pretrained semantics throughout personalization. Motivated by this observation, we reformulate the personalization objective through semantic anchoring to retain its connection to the pretrained semantic structure.

\subsection{Background}
\label{sec:Preliminary}
% \paragraph{\textbf{Text-to-Image Diffusion.}}
Text-to-image (T2I) diffusion models learn to generate an image latent $z_0$ from a textual condition $c$ by reversing a gradual noising process~\citep{song2020denoising, song2020score, nichol2021improved, luo2022understanding}. At each diffusion timestep $t$, a noisy latent $z_t$ is obtained as $z_t=\alpha_t z_0+\sigma_t\epsilon$, where $\epsilon\sim\mathcal{N}(0,I)$, and $\alpha_t$ and $\sigma_t$ control the signal and noise scales.
A denoising network $\epsilon_{\theta}(z_t,c,t)$ is trained to predict the added noise $\epsilon$ through the noise-prediction objective~\citep{ho2020denoising}:
\begin{equation}
\mathcal{L}_{\text{diffusion}}
=
\mathbb{E}_{z_t,c,\epsilon,t}
\left[
\|\epsilon-\epsilon_{\theta}(z_t,c,t)\|_2^2
\right].
\label{eq:diffusion}
\end{equation}

% \paragraph{\textbf{General Setup.}}
Personalization extends a pretrained T2I diffusion model parameterized by $\theta'$ into a personalized model $\theta$ that captures a specific subject.
Let $c^{\text{sbj}}$ denote the subject prompt (e.g., ``\texttt{a photo of a [V*] dog}''), where \texttt{[V*]} denotes a special token representing the subject attribute~\citep{ruiz2023dreambooth}.
For a subject image, let $z_t$ denote its noisy latent at timestep $t$. The personalized model learns the new subject from $(z_t,c^{\mathrm{sbj}})$ following Eq.~\ref{eq:diffusion}:
\begin{equation}
\mathcal{L}_{\text{sbj}} =
\mathbb{E}_{z_{t}, c^{\text{sbj}}, \epsilon, t}
\left[\|\epsilon - \epsilon_{\theta}(z_{t}, c^{\text{sbj}}, t)\|_2^2\right].
\label{eq:recon}
\end{equation}

\textit{Prior Preservation Loss} (PPL) introduces an additional objective using class-level samples generated by the pretrained model~\citep{ruiz2023dreambooth}. The pretrained model generates $z^{\text{cls}}$ from a class prompt $c^{\text{cls}}$ (e.g., ``\texttt{a photo of a dog}''), encouraging the personalized model to preserve the pretrained behavior on these samples during personalization. Combining Eq.~\ref{eq:recon} with PPL has become a standard training scheme for diffusion-based personalization. However, the two objectives treat subject learning and prior preservation separately without considering their semantic relation.

\newpage

\subsection{Semantic Divergence in Personalization}
\label{sec:motivation}

% In this paper, we treat pretrained semantics not as an independent preservation target, but as guidance for learning the personalized concept. Our motivation comes from rare concept generation, which aims to improve the generation of underrepresented concepts. In particular, Rare-to-Frequent~\citep{park2024rare} shows that such a concept can benefit from guidance provided by a related frequent concept. This approach formalizes this relation between a rare concept $c^{\mathrm{rar}}$ and its related frequent concept $c^{\mathrm{frq}}$:
In this paper, we treat pretrained semantics not as an independent preservation target, but as guidance for learning the personalized concept. Our motivation comes from rare concept generation, which aims to improve the generation of underrepresented concepts. Such concepts are less reliably represented because they receive limited support in the training distribution. In particular, Rare-to-Frequent~\citep{park2024rare} studies whether a related frequent concept can provide guidance for such underrepresented concepts. It formalizes the relation between a rare concept $c^{\mathrm{rar}}$ and its related frequent concept $c^{\mathrm{frq}}$ through score interpolation:
\begin{theorem}
\label{thm:r2f}
Under a Gaussian model where the frequent concept is well represented and the rare concept has estimation uncertainty characterized by $\sigma$, consider the interpolated score
\begin{equation}
\begin{aligned}
s_{\alpha}(z_t)
&=
\alpha \nabla_{z_t}\log p_{\theta}(z_t|c^{\mathrm{rar}})
+
(1-\alpha)\nabla_{z_t}\log p_{\theta}(z_t|c^{\mathrm{frq}})
\end{aligned}
\label{eq:r2f_noise}
\end{equation}
Let $p_{\mathrm{lerp}}(x|c^{\mathrm{rar}};\alpha)$ denote the distribution corresponding to this interpolated score, and let $\mu_{\mathrm{rar}}$ and $\mu_{\mathrm{frq}}$ denote the means of the rare and frequent concept distributions. If
$
\sigma
\ge
1+\sqrt{\|\mu_{\mathrm{frq}}-\mu_{\mathrm{rar}}\|_2^2+0.2},
$
then
\begin{equation}
\min_{\alpha}
W_2\!\left(
p_{\mathrm{lerp}}(x|c^{\mathrm{rar}};\alpha),
p_{\mathrm{data}}(x|c^{\mathrm{rar}})
\right)
<
W_2\!\left(
p_{\theta}(x|c^{\mathrm{rar}}),
p_{\mathrm{data}}(x|c^{\mathrm{rar}})
\right),
\label{eq:r2f_theorem}
\end{equation}
where $W_2$ denotes the 2-Wasserstein distance.
\end{theorem}
Eq.~\ref{eq:r2f_theorem} shows that an underrepresented concept can be better approximated by leveraging a related well-represented concept rather than using the rare concept alone. This highlights the importance of considering an underrepresented concept in relation to its well-represented semantic counterpart.

% -------------------------------------
\begin{wrapfigure}{r}{0.45\textwidth}
    \vspace{-10pt}
    \centering
    \includegraphics[width=1.0\linewidth]{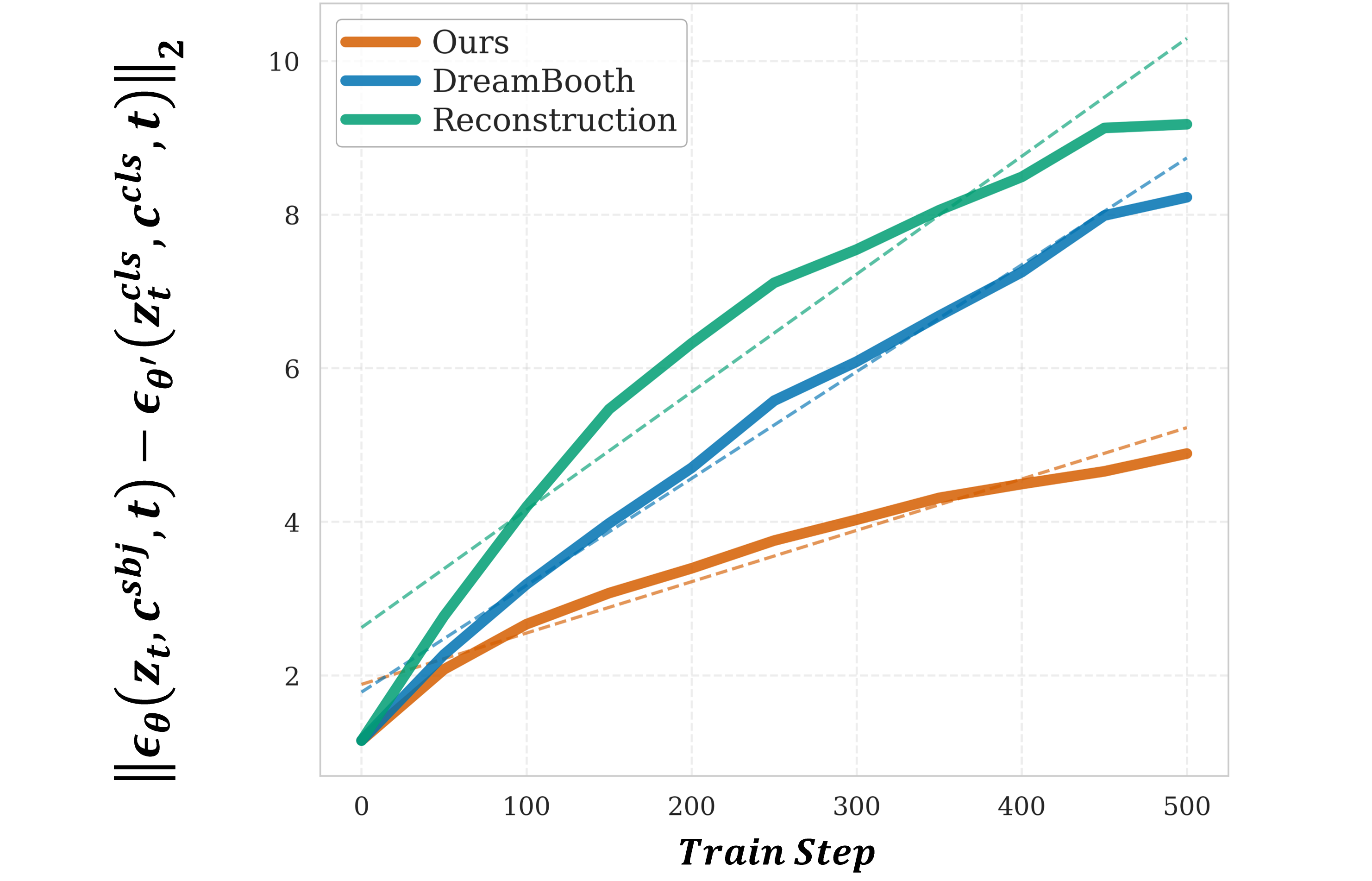}
    %\vspace{-17pt}
    \caption{$L_2$ distance between the personalized subject prediction and its pretrained class-level counterpart throughout the personalization process.}
    \label{fig:subject_anchor}
    \vspace{-10pt}
\end{wrapfigure}

% -------------------------------------

A similar relation arises in personalization, where a personalized concept learned from only a few reference images can be regarded as an underrepresented concept, while its general semantic counterpart is already well represented in the pretrained model. Under this correspondence, we assume that the personalized concept should maintain its relation to its pretrained semantic counterpart. To examine how this relation evolves, we measure
$\|\epsilon_{\theta}(z_t,c^{\text{sbj}},t)-\epsilon_{\theta'}(z_t^{\text{cls}},c^{\text{cls}},t)\|_2$
between the personalized subject prediction and its pretrained class-level counterpart throughout personalization. As shown by the green curve in Fig.~\ref{fig:subject_anchor}, the distance increases when optimizing the objective in Eq.~\ref{eq:recon}. Even with PPL (blue curve), the increasing trend persists as personalization proceeds. This observation indicates that prior preservation does not ensure that the personalized concept remains aligned with its pretrained semantic counterpart. This divergence can isolate the personalized concept from the pretrained semantic structure as illustrated in Fig.~\ref{fig:concept}. From the perspective of Theorem~\ref{thm:r2f}, preserving this relation is essential for leveraging guidance from the well-represented semantic counterpart.

% This motivates a personalization objective that not only learns the target concept but also accounts for its relation to the pretrained semantic counterpart.

% This observation suggests that preserving class-level semantics through an independent objective does not maintain their relation to the personalized concept. 

% Motivated by the theoretical view of R2F and this observation, we incorporate the general semantic counterpart directly into personalized concept learning. 

% The theorem shows that, when the rare concept is insufficiently represented, incorporating a related frequent concept can better approximate the target rare distribution than relying on the rare concept alone. This suggests that the frequent semantic counterpart can serve not only as a semantic reference to preserve, but also as guidance for forming the rare concept itself.

%그래서 우리는 이 관계가 personalization 중에 벌어지는지? 유지되는지 확인 )To examine the preservation of semantic alignment during personalization, we analyze the behavior of $c^{\text{sbj}}$ and its prior $c^{\text{cls}}$ as shown in \autoref{fig:subject_anchor}.) 그 결과는 \autoref{fig:concept}. 와 같음. adaptation이 진행됨에 따라 그 class-level과 subject와의 거리가 점점 멀어지는 쪽으로 PPL도 그 slope(변화)의 정도가 완화되지는 않는 모습. 이를 통해 이 관계를 고려해야할 필요성 업

% Accordingly, rather than preserving pretrained semantics through an independent objective, we incorporate the general semantic counterpart into the learning of the personalized concept. We refer to this formulation as \textit{Semantic Anchoring Personalization} (SAP).

\FloatBarrier
\subsection{Semantic Anchoring Personalization}
\label{sec:method}

% Motivated by the analysis above, personalization objective는 필요로 한다 that not only learns the target concept but also accounts for its relation to the pretrained semantic counterpart.  However, The reconstruction objective in \autoref{eq:recon} 는 subject만 고려한다. To reflect the Theorem, 새로 배우는 personalized guidance는 이것만 배워야하는게 아니라 기존 가지고있던 그 관계성을 만들어내야 한다는 것임. 그래서 우리는  We reformulate the 식 \autoref{eq:r2f_score}를 녹인다 a training objective에 for personalization. Since this relation is formulated in the score space in \autoref{eq:r2f_score}, we first express it under the noise-prediction parameterization.

Motivated by the analysis above, we consider a personalization objective that learns the target concept while accounting for its relation to the pretrained semantic counterpart. To this end, we reformulate the rare-to-frequent guidance in Eq.~\ref{eq:r2f_noise} as a training objective for personalization. We first express the score-based guidance under the noise-prediction parameterization. By Tweedie's formula~\citep{Efron2011Tweedie, song2020score}, the conditional score at timestep $t$ can be written as $\nabla_{z_t}\log p(z_t|c)=-\epsilon(z_t,c,t)/\sigma_t$. Substituting this relation into Eq.~\ref{eq:r2f_noise}, we obtain
\begin{equation}
s_{\alpha}(z_t)
\propto
-\left[
\lambda\epsilon^{\mathrm{rar}}
+
(1-\lambda)\epsilon^{\mathrm{frq}}
\right].
\label{eq:sap_noise}
\end{equation}

We then set the interpolated noise prediction in Eq.~\ref{eq:sap_noise} as the training target itself. This formulation incorporates the pretrained counterpart into the training target rather than adding it as a separate preservation term. The following proposition shows that optimizing this target can be reformulated as rare-concept reconstruction anchored to its frequent semantic counterpart:
\begin{proposition}
\label{eq:Proposition}
Let $\lambda\in(0,1]$ and define
$\epsilon^{*}
=
\lambda\epsilon^{\mathrm{rar}}
+
(1-\lambda)\epsilon^{\mathrm{frq}}$.
Using $\epsilon^{*}$ as the training target yields the same optimal parameters as
\begin{equation}
\theta^{*}
=
\arg\min_{\theta}
\mathbb{E}
\left[
\left\|
\epsilon^{\mathrm{rar}}-\epsilon_{\theta}
\right\|_2^2
+
\frac{1-\lambda}{\lambda}
\left\|
\epsilon_{\theta}-\epsilon^{\mathrm{frq}}
\right\|_2^2
\right].
\label{eq:anchor_objective}
\end{equation}
The proof is provided in the appendix~\ref{app:proof}.
\end{proposition}
Eq.~\ref{eq:anchor_objective} shows that learning a rare concept requires not only reconstructing its target representation but also remaining anchored to its frequent semantic counterpart.

Following the correspondence established above, we instantiate the derived formulation for personalization by treating the personalized subject as the rare concept and its class-level semantics as the frequent counterpart. Accordingly, $\epsilon^{\mathrm{rar}}$ corresponds to the ground-truth noise $\epsilon$ used as the subject target in Eq.~\ref{eq:recon}, while $\epsilon^{\mathrm{frq}}$ is given by the prediction of the pretrained model $\epsilon_{\theta'}(z_t,c^{\mathrm{cls}},t)$. Substituting these terms into Eq.~\ref{eq:anchor_objective} yields:
\begin{equation}
\label{eq:final}
\begin{aligned}
\mathcal{L}_{\mathrm{SAP}}
=
\mathbb{E}_{z_t,c^{\mathrm{sbj}},c^{\mathrm{cls}},\epsilon,t}
\Big[
&
\|\epsilon-\epsilon_{\theta}(z_t,c^{\mathrm{sbj}},t)\|_2^2
\\
&+
w
\|\epsilon_{\theta}(z_t,c^{\mathrm{sbj}},t)
-
\epsilon_{\theta'}(z_t,c^{\mathrm{cls}},t)\|_2^2
\Big],
\end{aligned}
\end{equation}
where $w=(1-\lambda)/\lambda$ and $\theta'$ remains frozen during personalization.

We introduce the objective in Eq.~\ref{eq:final} as \textit{Semantic Anchoring Personalization} (SAP). The first term learns the subject-specific attributes, and the second term anchors the personalized prediction to its pretrained class-level counterpart. Anchoring the personalized prediction to its pretrained class-level counterpart preserves its connection to the pretrained semantic structure as illustrated in Fig.~\ref{fig:concept}. Unlike prior approaches that separate concept learning from the preservation of pretrained semantics, our objective reframes personalization as learning the target concept in relation to its pretrained semantic counterpart. As a result, it reduces the semantic divergence observed in Fig.~\ref{fig:subject_anchor} (orange line). Moreover, this derivation yields an objective that can be applied to the standard personalization setting without additional components. In practice, SAP leads to consistent performance improvements across the settings in Sec.~\ref{sec:experiments}, suggesting that preserving this semantic relation contributes to effective personalization.

\begin{figure*}[t]
    \centering

    \setlength{\tabcolsep}{0pt}

    % =========================================================
    % Reference 4장: 간격 없는 2×2 배열
    % 전체 크기 = 결과 이미지 한 장 크기
    % =========================================================
    \newcommand{\fourrefimgs}[4]{%
        \vbox{%
            \offinterlineskip
            \hbox{%
                \includegraphics[
                    width=0.0445\textwidth,
                    height=0.0445\textwidth
                ]{#1}%
                \includegraphics[
                    width=0.0445\textwidth,
                    height=0.0445\textwidth
                ]{#2}%
            }%
            \hbox{%
                \includegraphics[
                    width=0.0445\textwidth,
                    height=0.0445\textwidth
                ]{#3}%
                \includegraphics[
                    width=0.0445\textwidth,
                    height=0.0445\textwidth
                ]{#4}%
            }%
        }%
    }

    % 결과 이미지 크기
    \renewcommand{\resultimg}[1]{%
        \includegraphics[
            width=0.089\textwidth,
            height=0.089\textwidth
        ]{#1}%
    }

    \begin{tabular}{
        @{}
        C{0.089\textwidth}
        @{\hspace{3pt}}
        *{10}{C{0.089\textwidth}}
        @{}
    }

    % =========================================================
    % Header
    % =========================================================
    \methodheader{References}
    &
    \methodheader{Ours}
    &
    \methodheader{DreamBooth}
    &
    \methodheader{DreamBooth\\-LoRA}
    &
    \methodheader{Textual\\Inversion}
    &
    \methodheader{SuDe}
    &
    \methodheader{SVDiff}
    &
    \methodheader{Custom\\Diffusion}
    &
    \methodheader{Steering\\Guidance}
    &
    \methodheader{IP-\\Adapter}
    &
    \methodheader{BLIP-\\Diffusion}
    \\[0pt]

    % =========================================================
    % Row 1: Bear
    % =========================================================
    \fourrefimgs
        {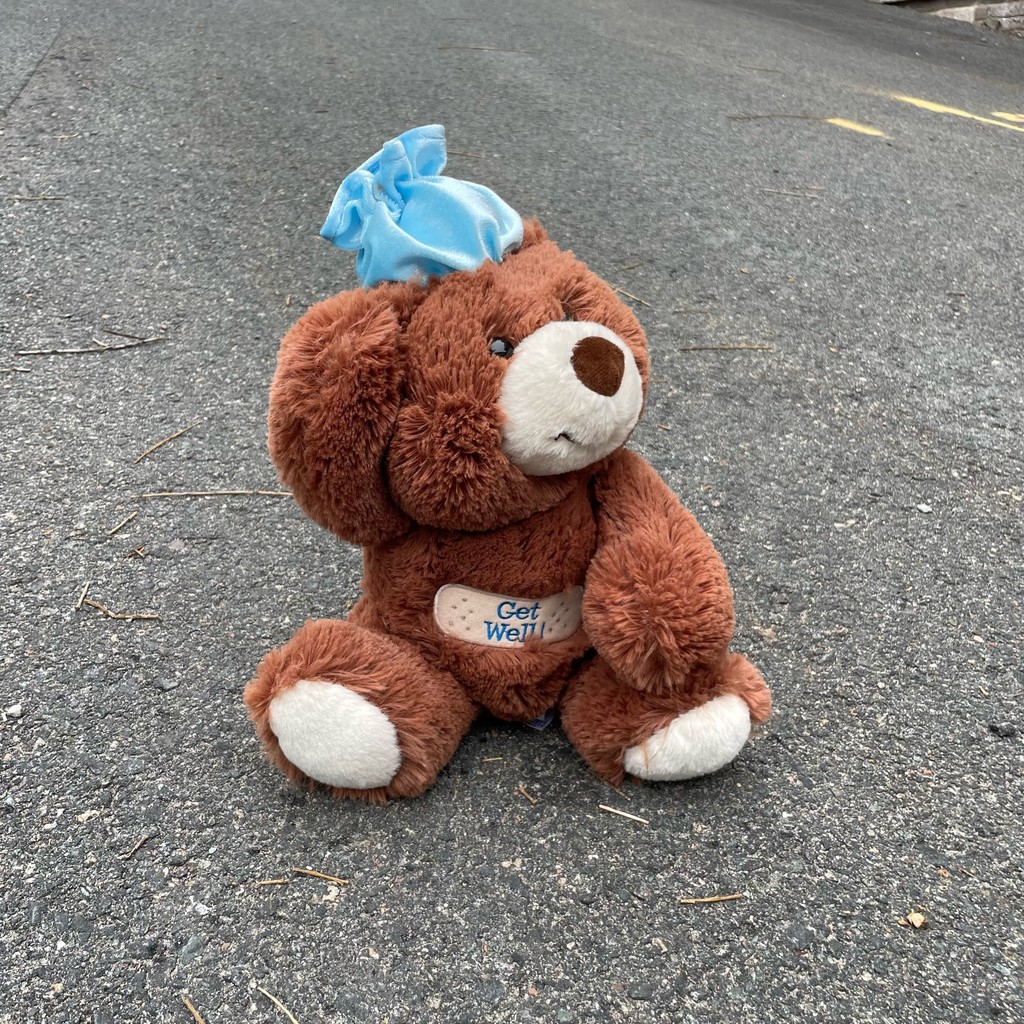}
        {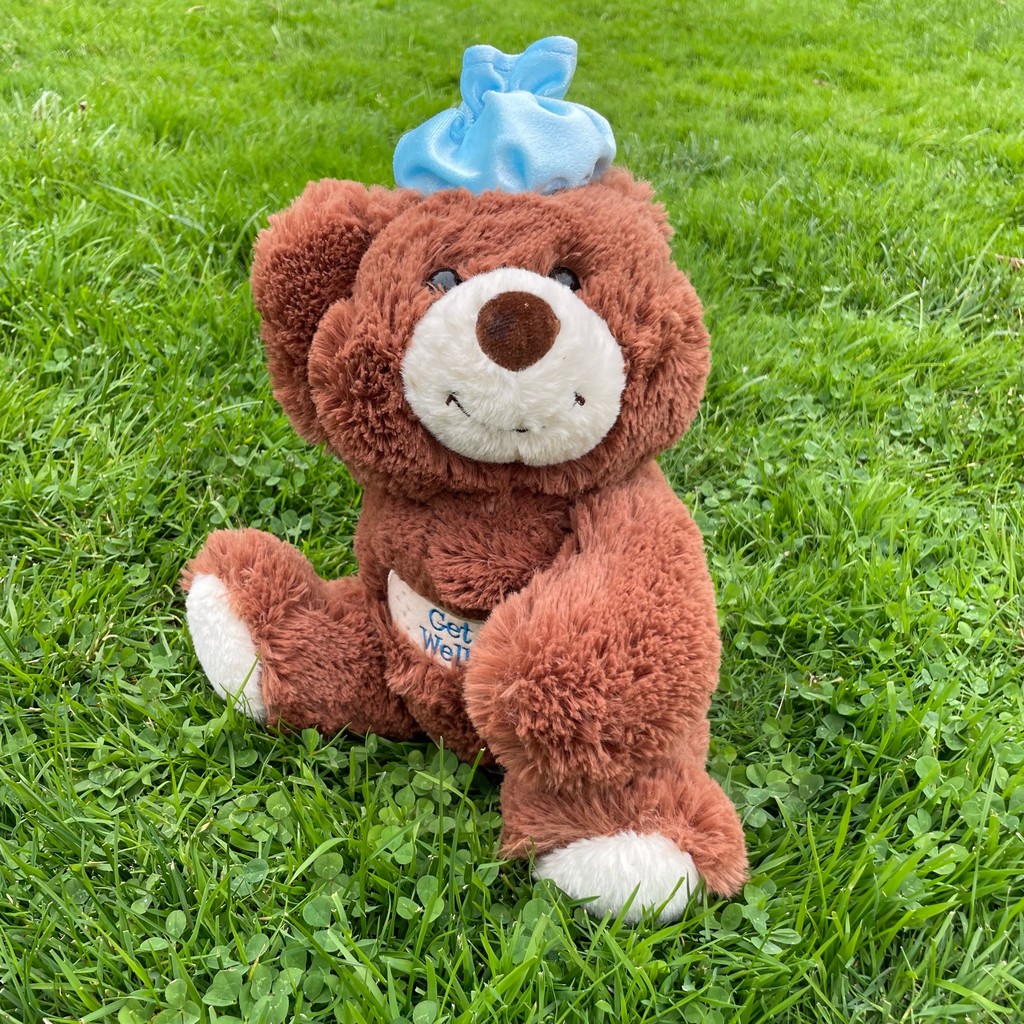}
        {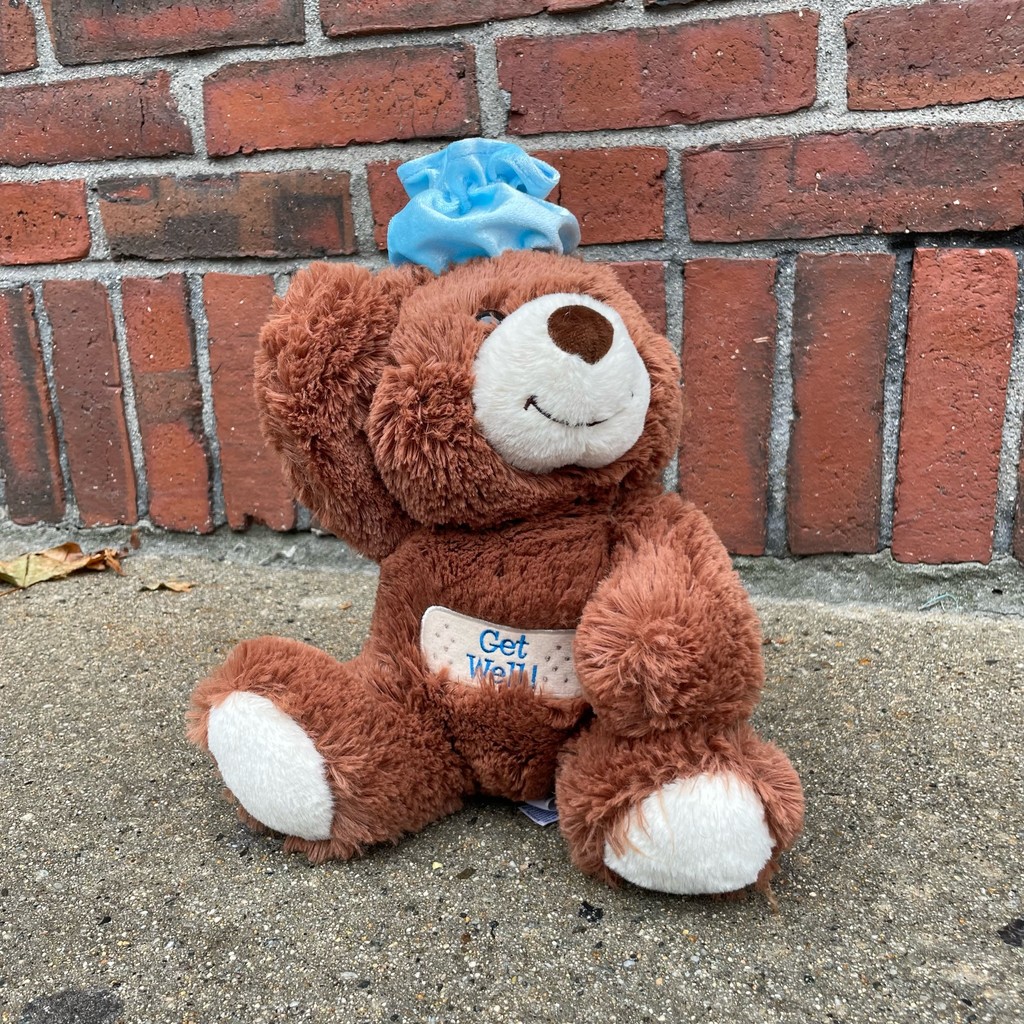}
        {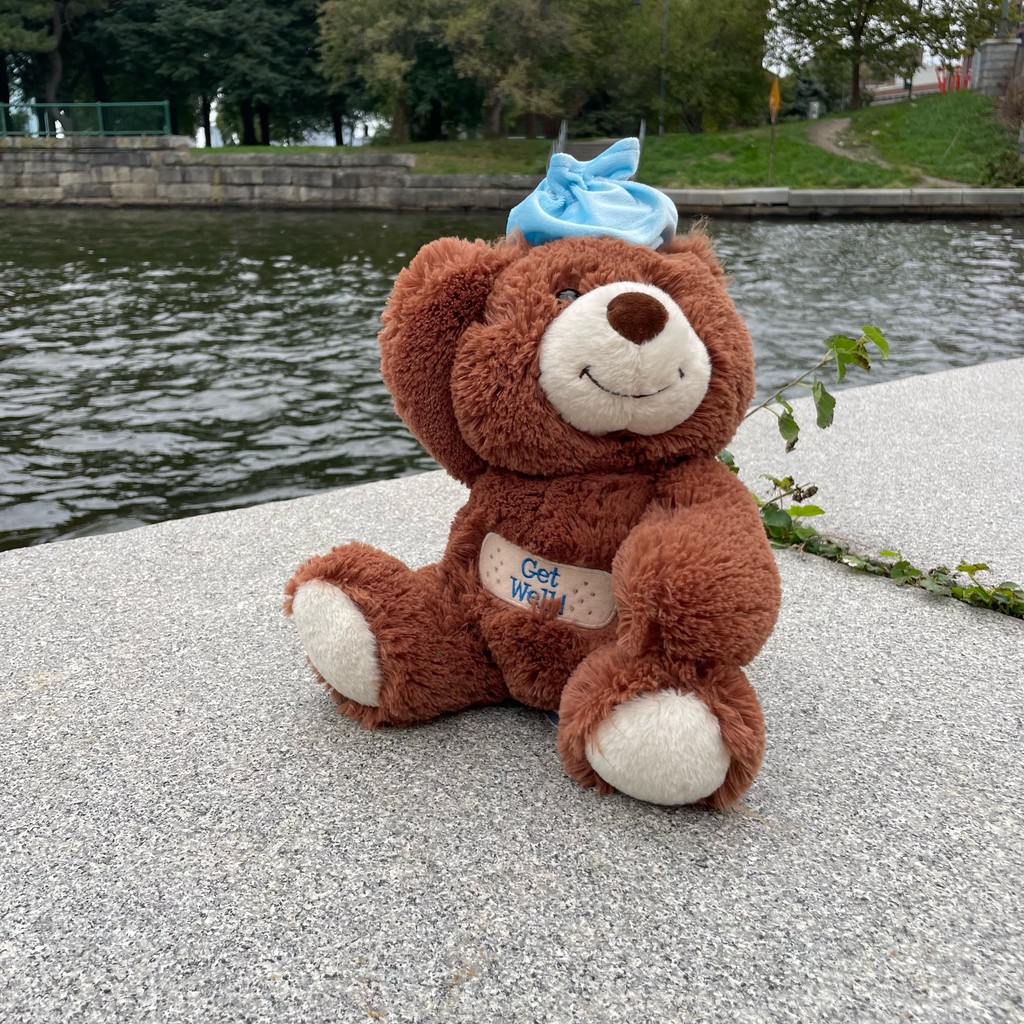}
    &
    \resultimg{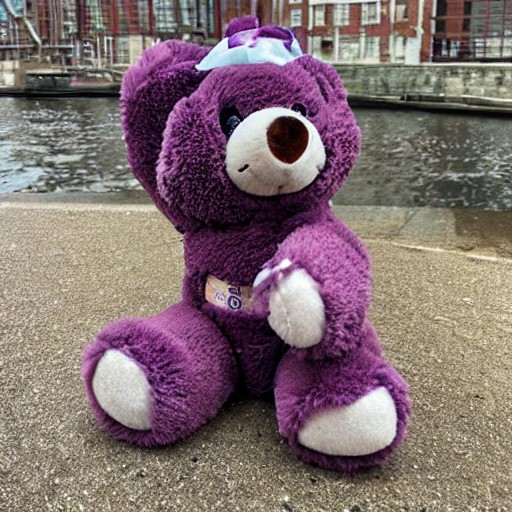}
    &
    \resultimg{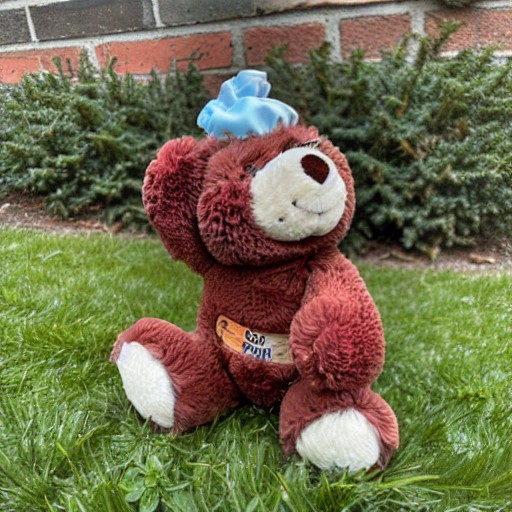}
    &
    \resultimg{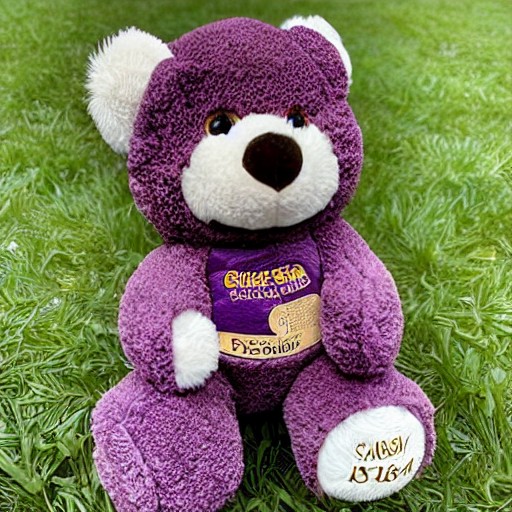}
    &
    \resultimg{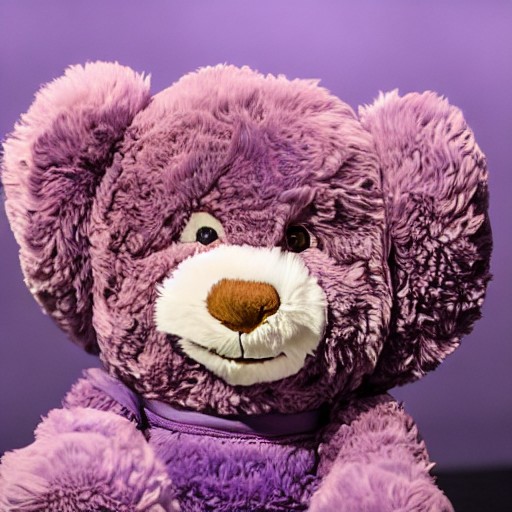}
    &
    \resultimg{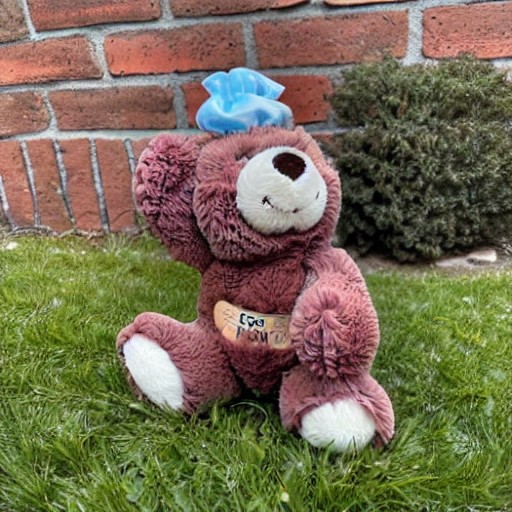}
    &
    \resultimg{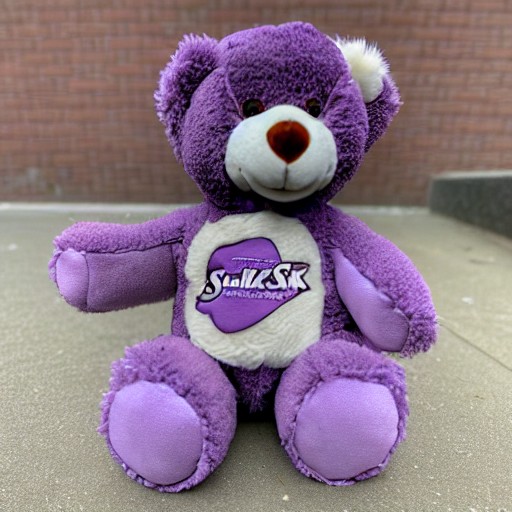}
    &
    \resultimg{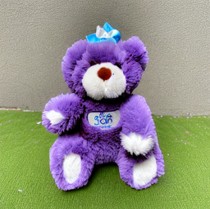}
    &
    \resultimg{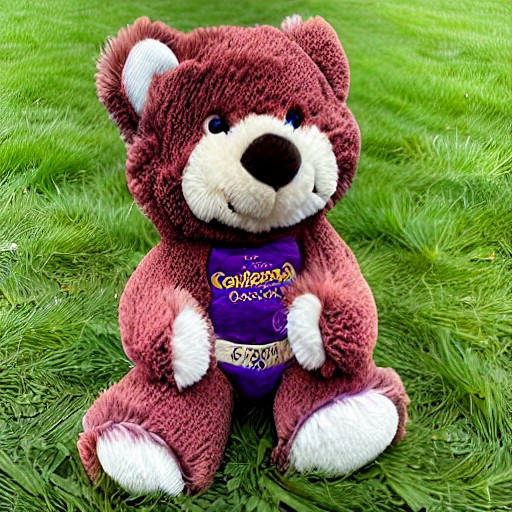}
    &
    \resultimg{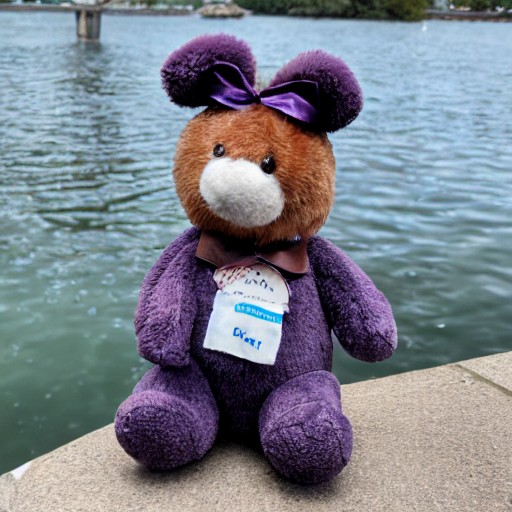}
    &
    \resultimg{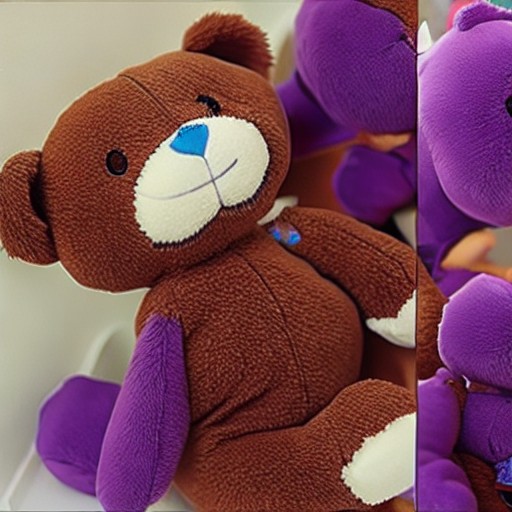}
    \\[-3pt]

    \multicolumn{11}{c}{
        {\scriptsize\itshape ``a purple [V*]''}
    }
    \\[1pt]

    % =========================================================
    % Row 2: Cat
    % =========================================================
    \fourrefimgs
        {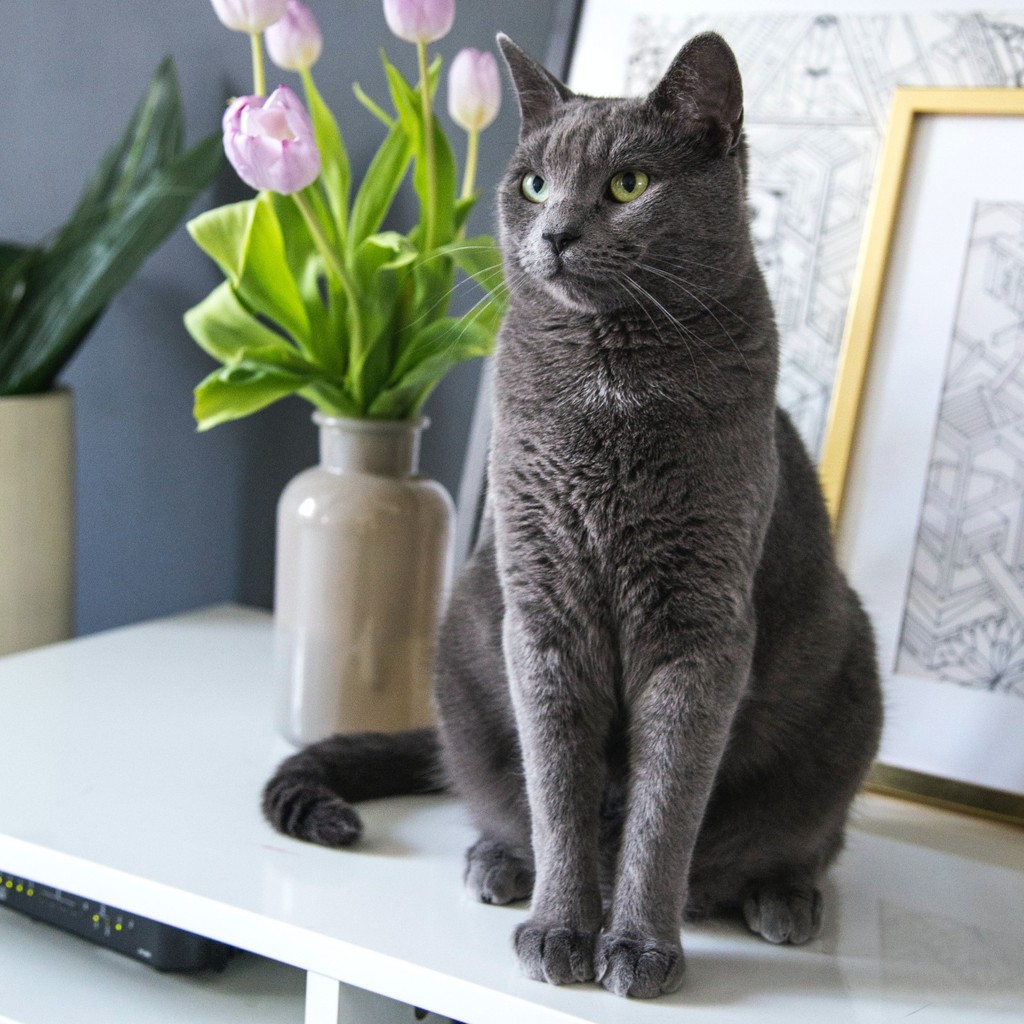}
        {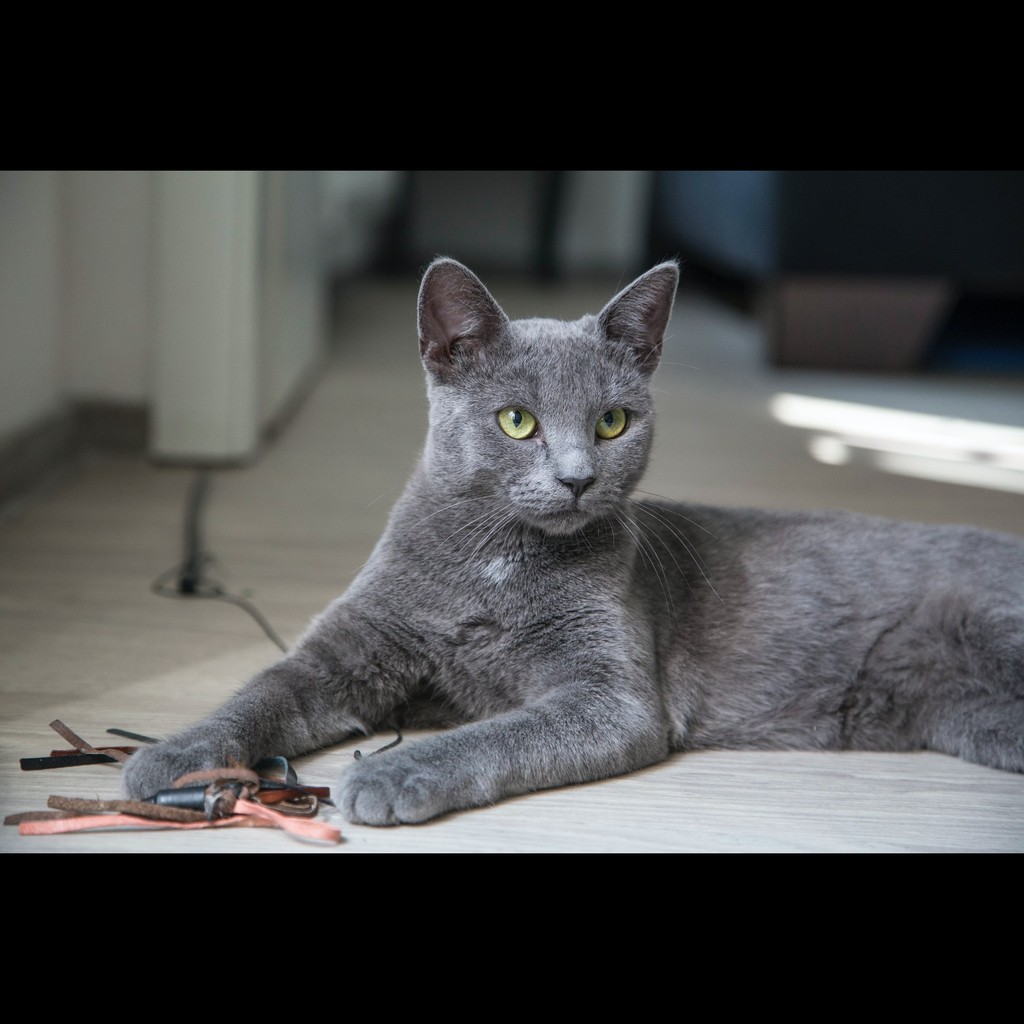}
        {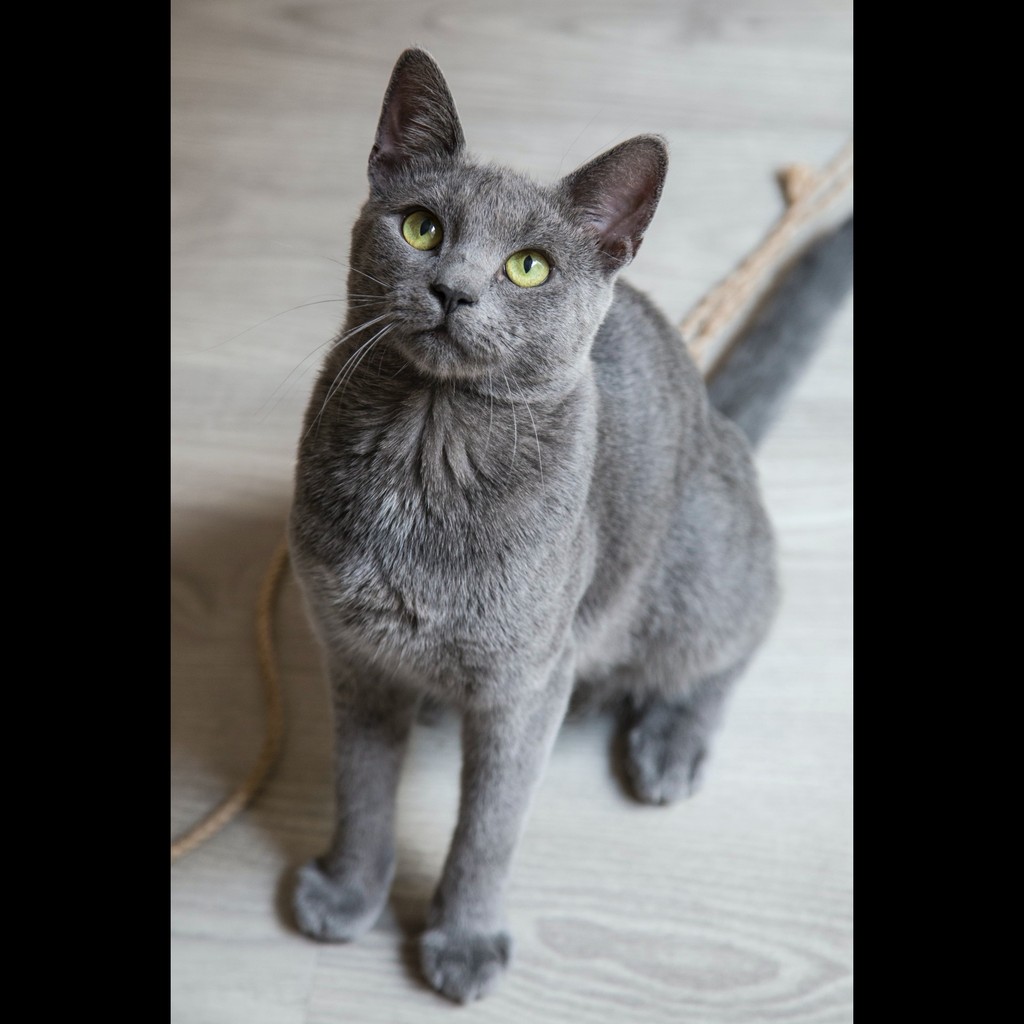}
        {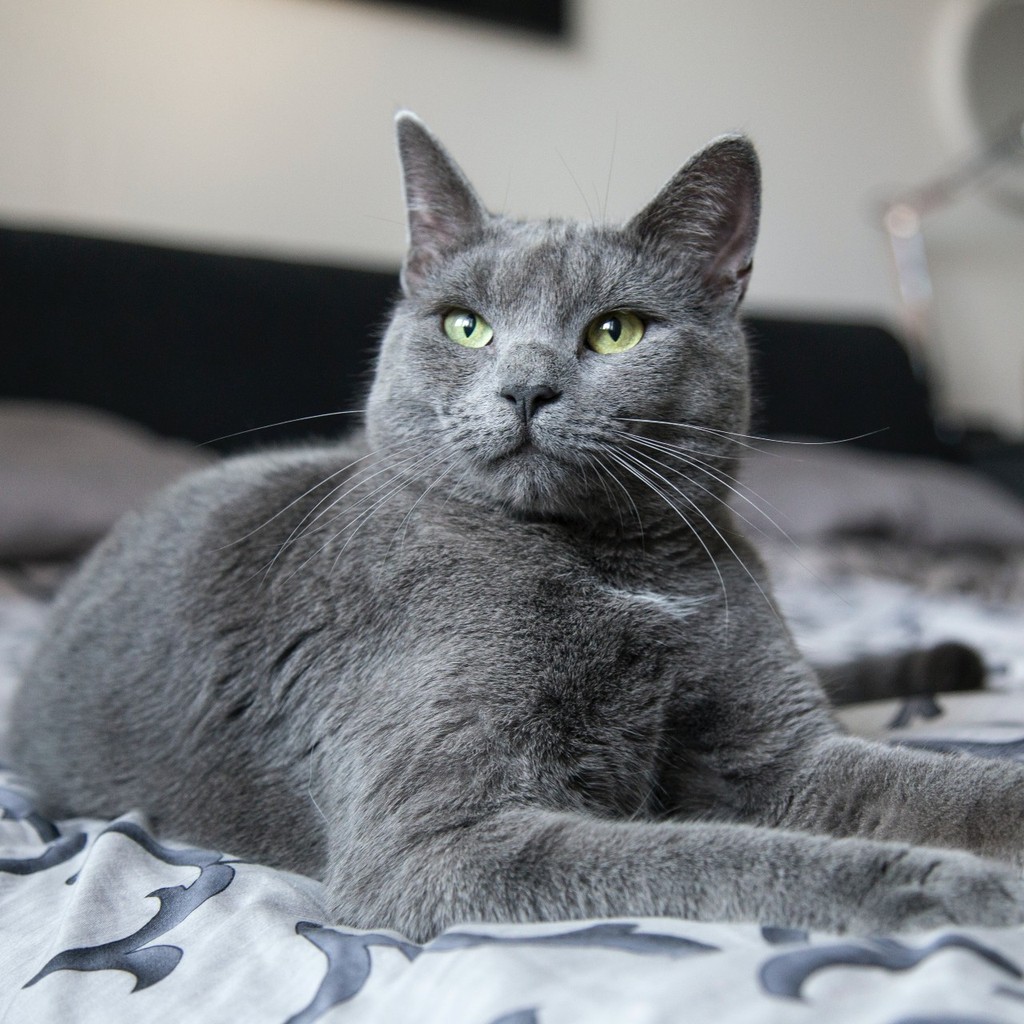}
    &
    \resultimg{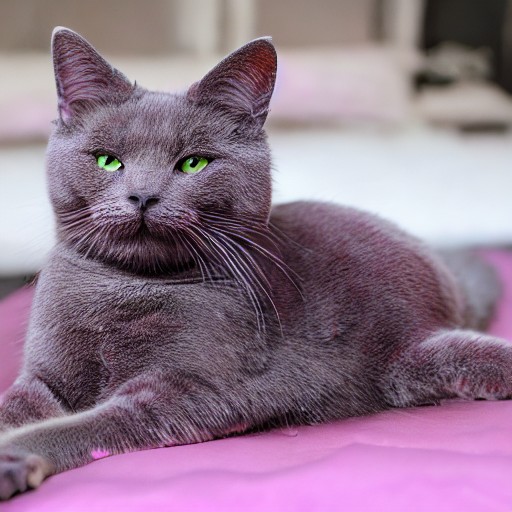}
    &
    \resultimg{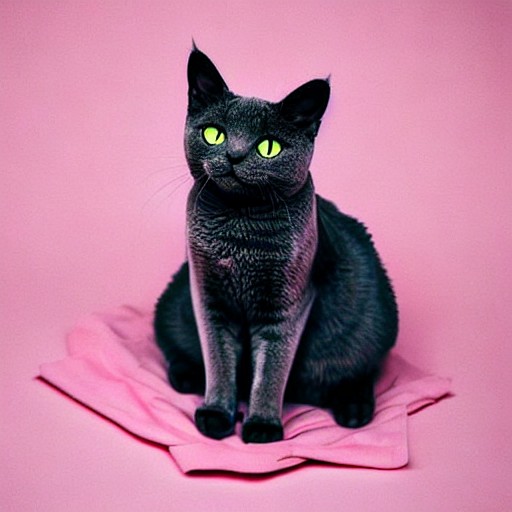}
    &
    \resultimg{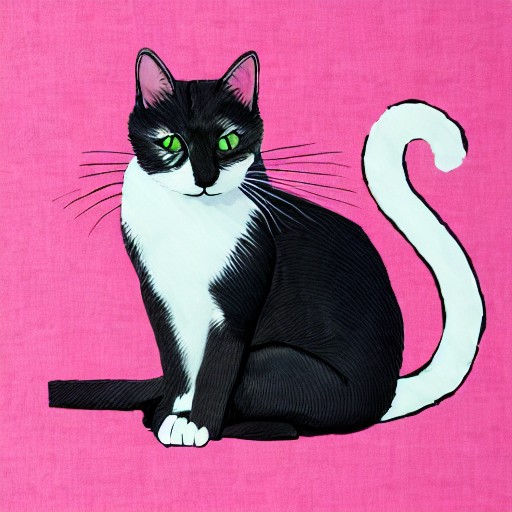}
    &
    \resultimg{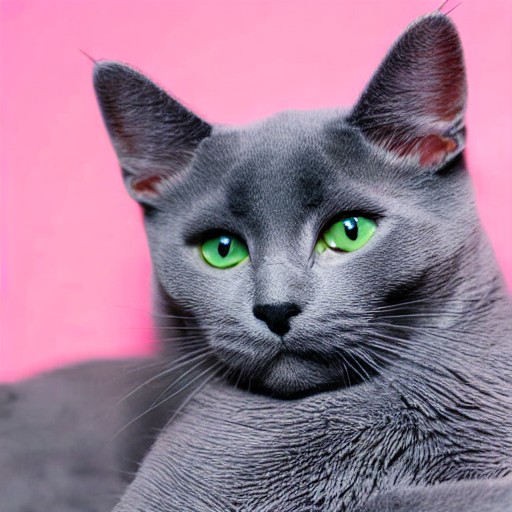}
    &
    \resultimg{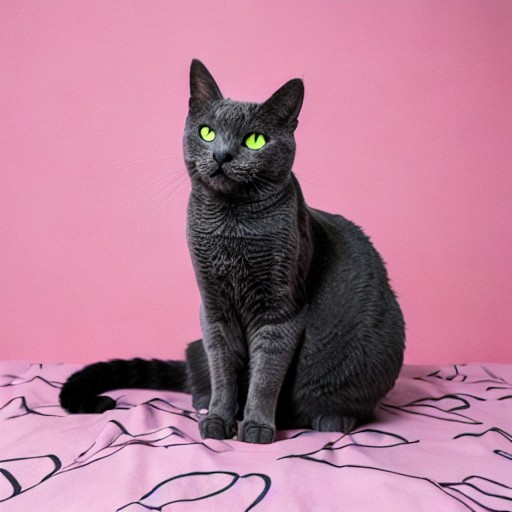}
    &
    \resultimg{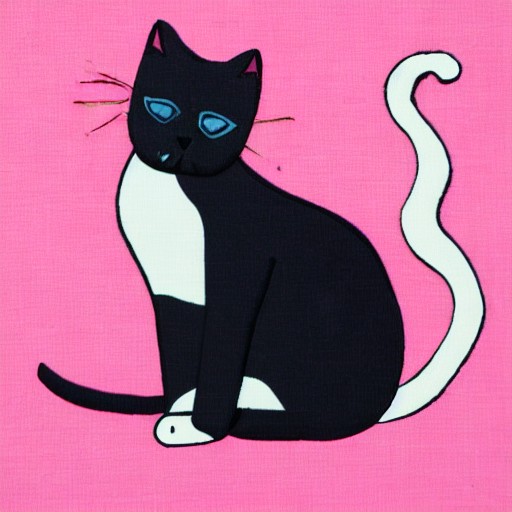}
    &
    \resultimg{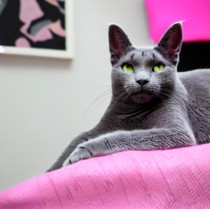}
    &
    \resultimg{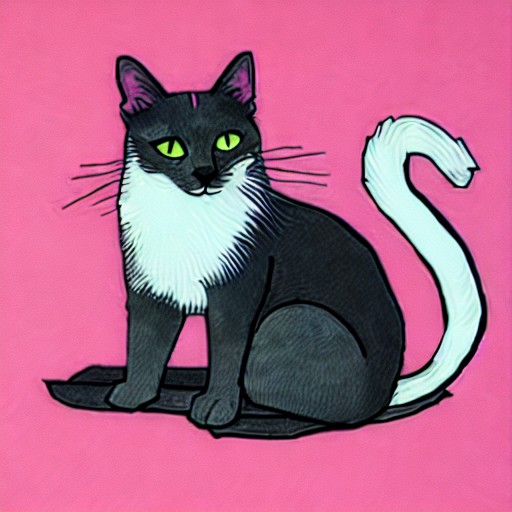}
    &
    \resultimg{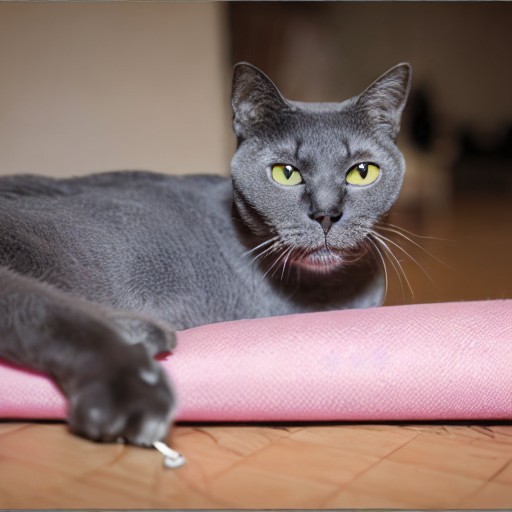}
    &
    \resultimg{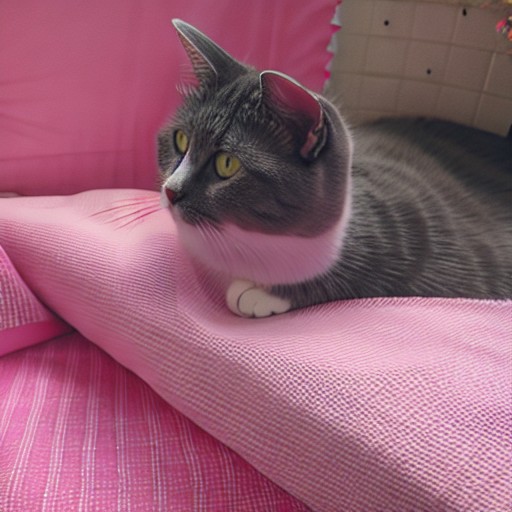}
    \\[-3pt]

    \multicolumn{11}{c}{
        {\scriptsize\itshape ``a [V*] on top of pink fabric''}
    }
    \\[1pt]

    % =========================================================
    % Row 3: Shoes
    % =========================================================
    \fourrefimgs
        {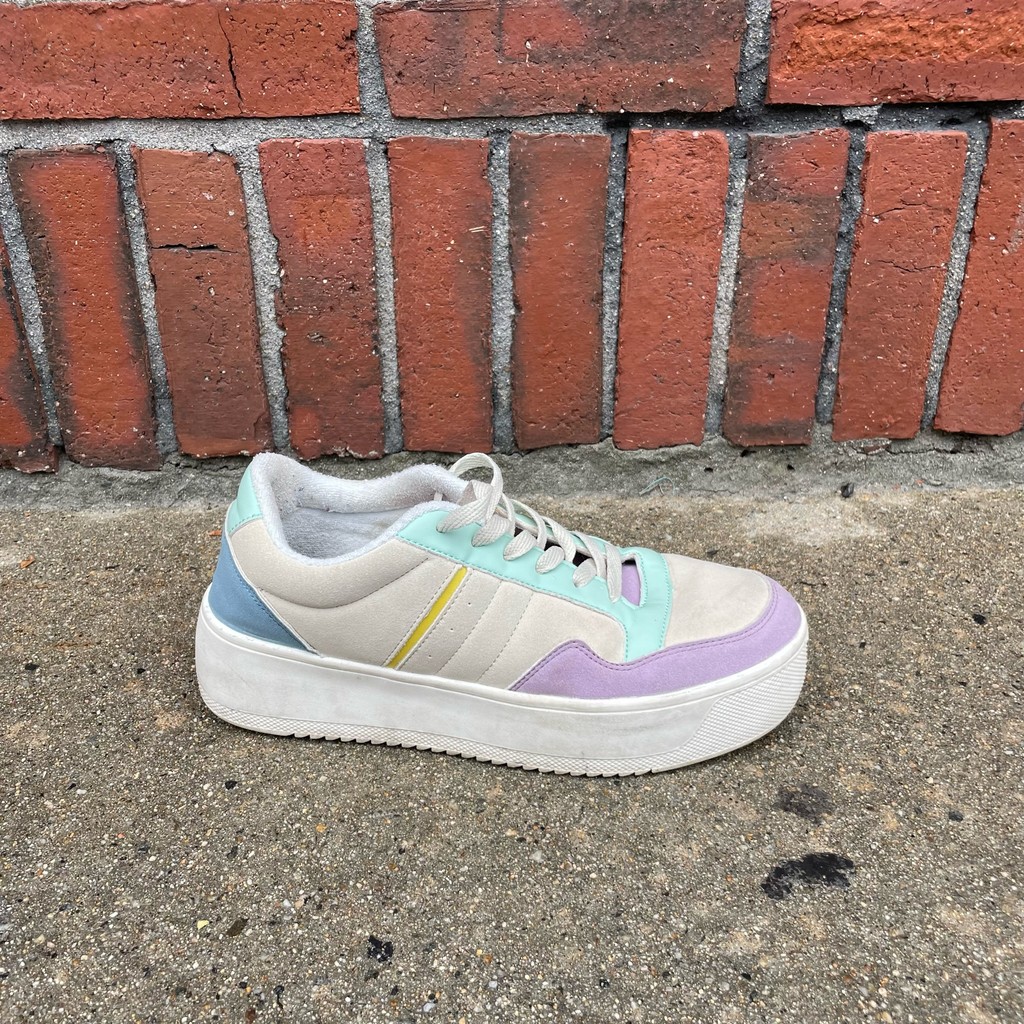}
        {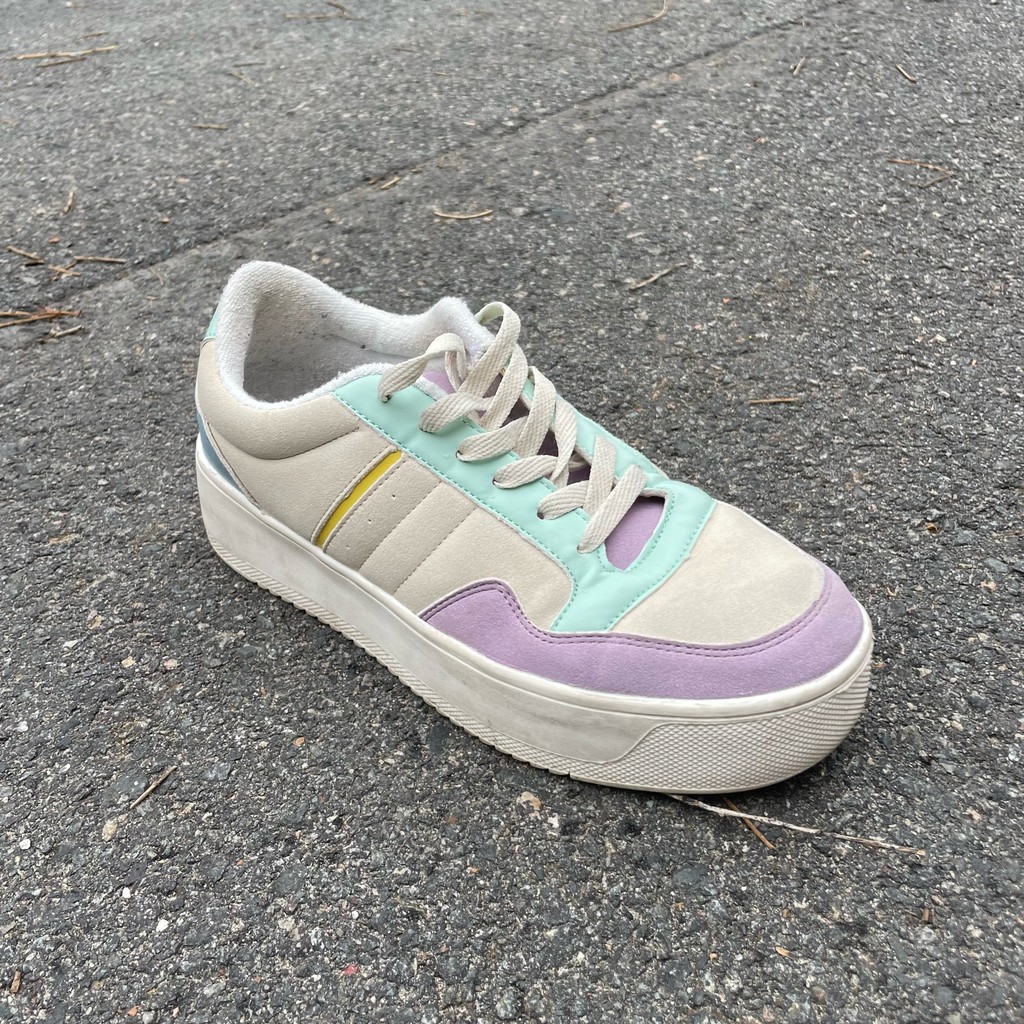}
        {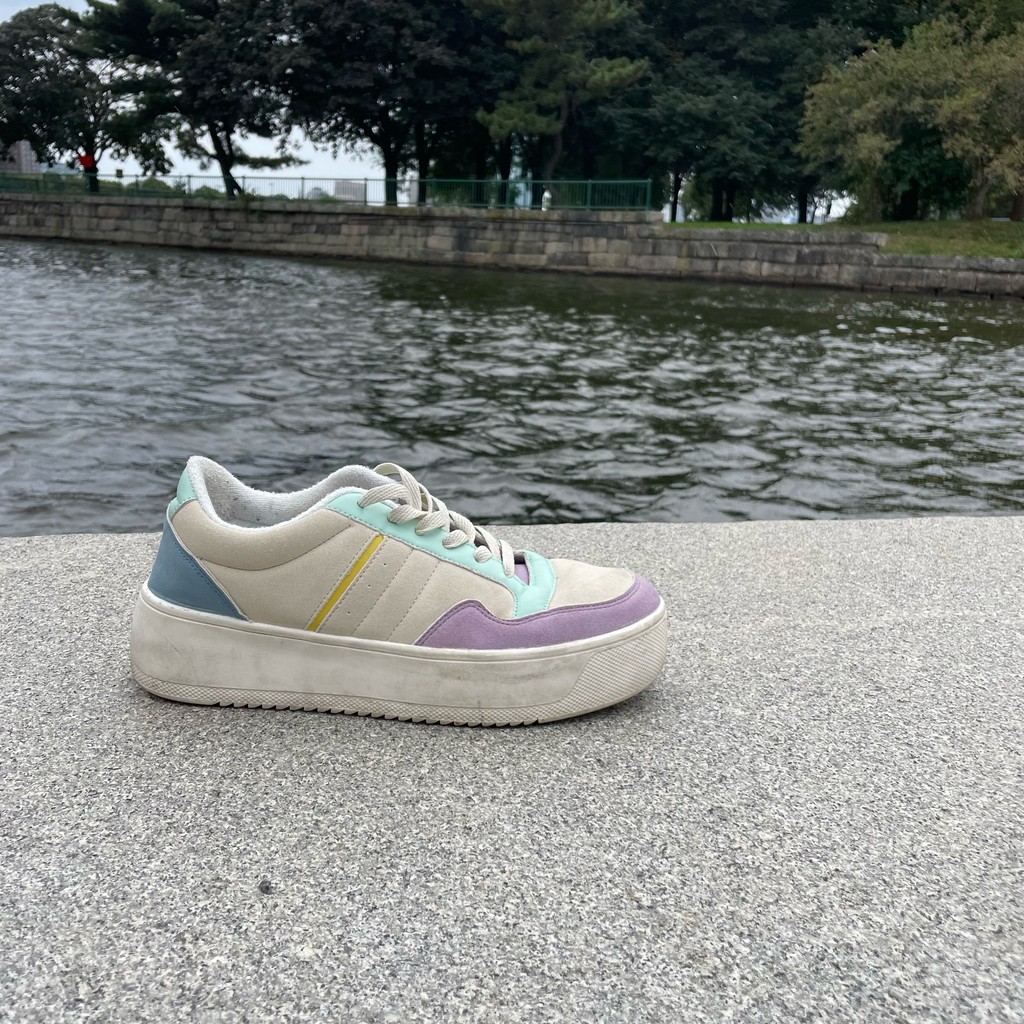}
        {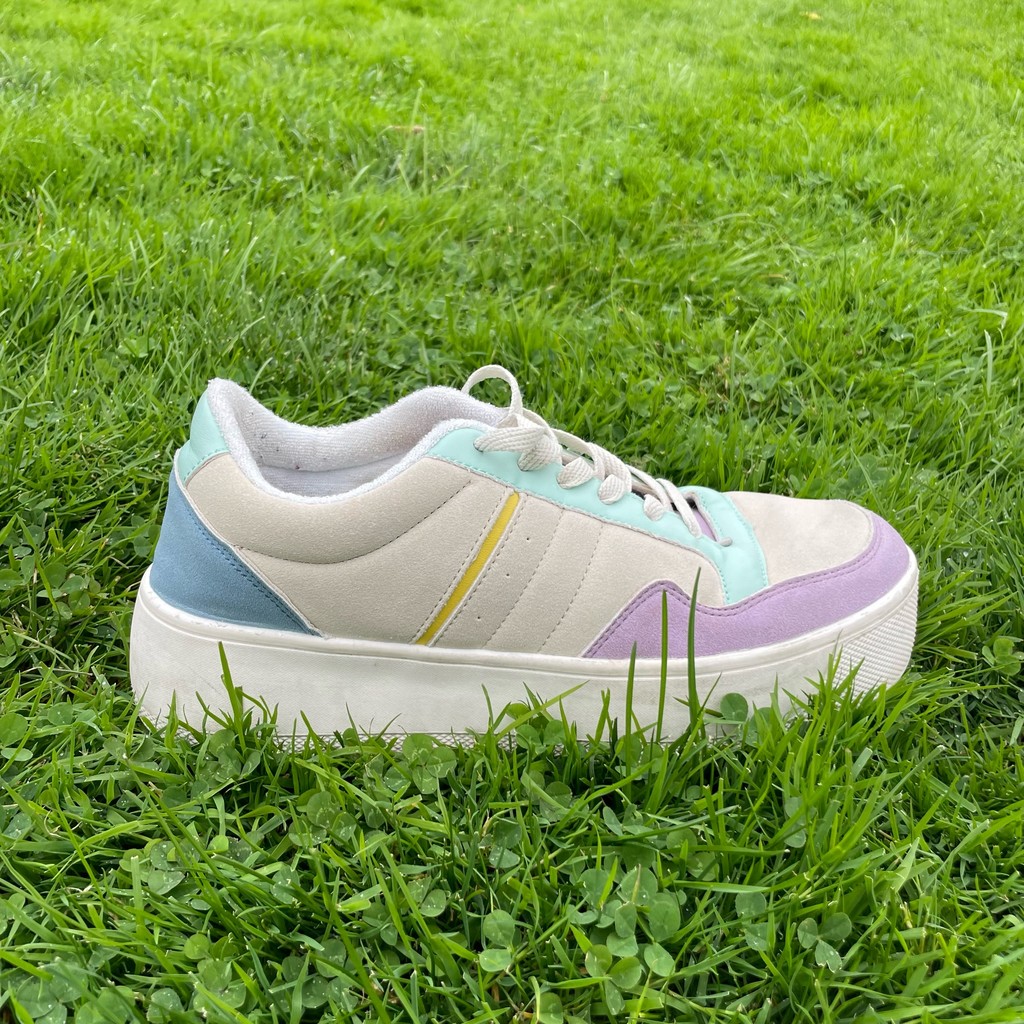}
    &
    \resultimg{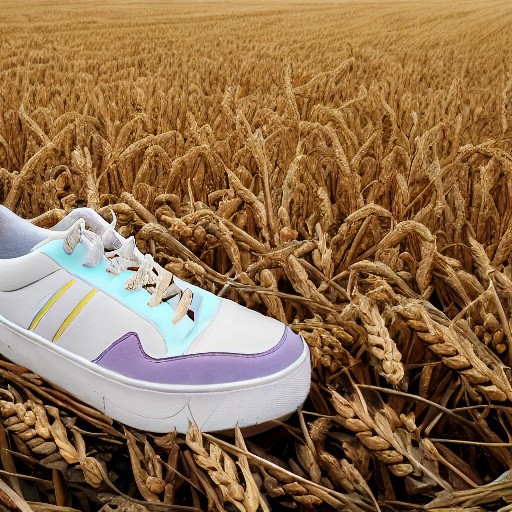}
    &
    \resultimg{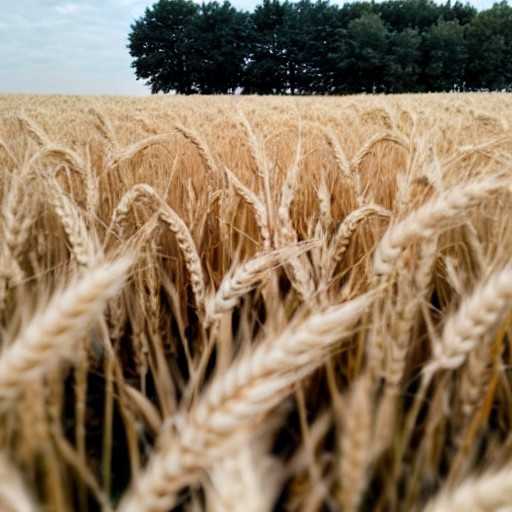}
    &
    \resultimg{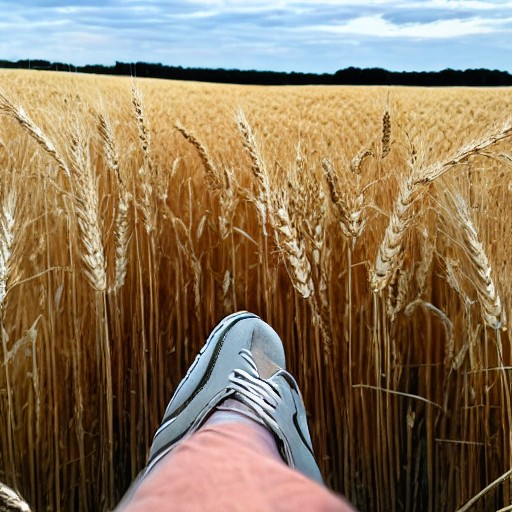}
    &
    \resultimg{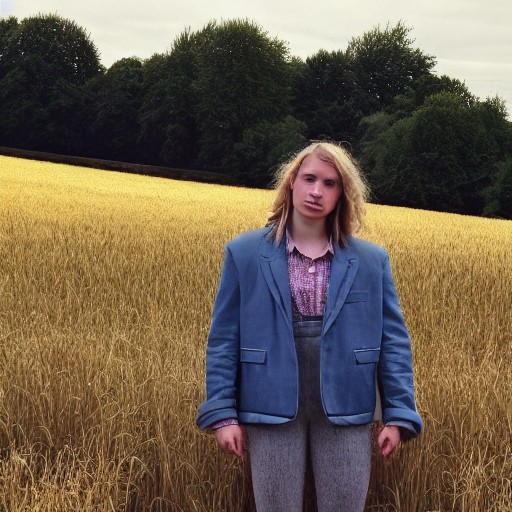}
    &
    \resultimg{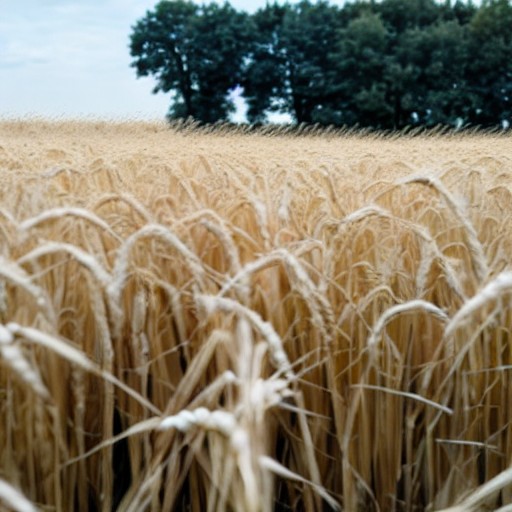}
    &
    \resultimg{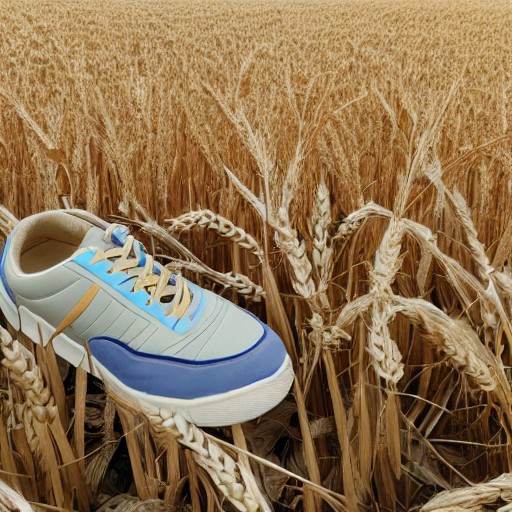}
    &
    \resultimg{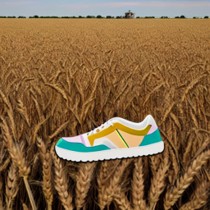}
    &
    \resultimg{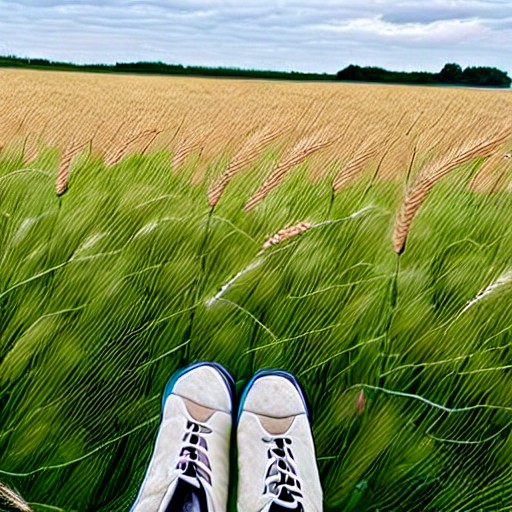}
    &
    \resultimg{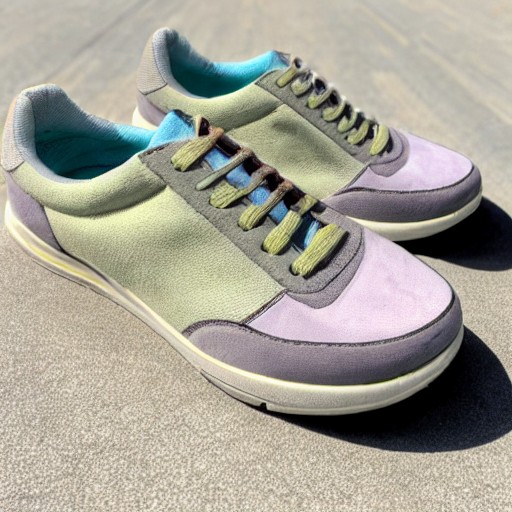}
    &
    \resultimg{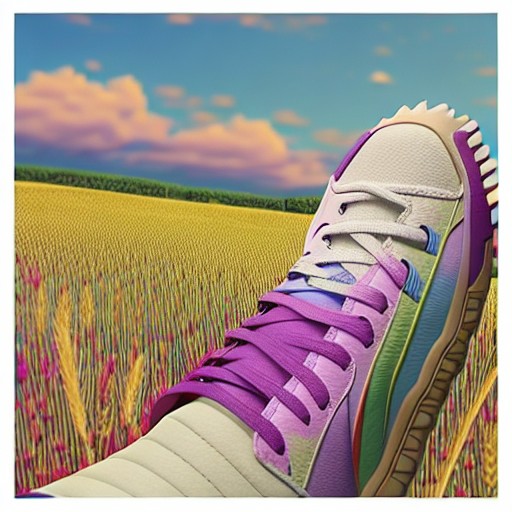}
    \\[-3pt]

    \multicolumn{11}{c}{
        {\scriptsize\itshape
        ``a [V*] with a wheat field in the background''}
    }
    \\[1pt]

    % =========================================================
    % Row 4: Backpack
    % =========================================================
    \fourrefimgs
        {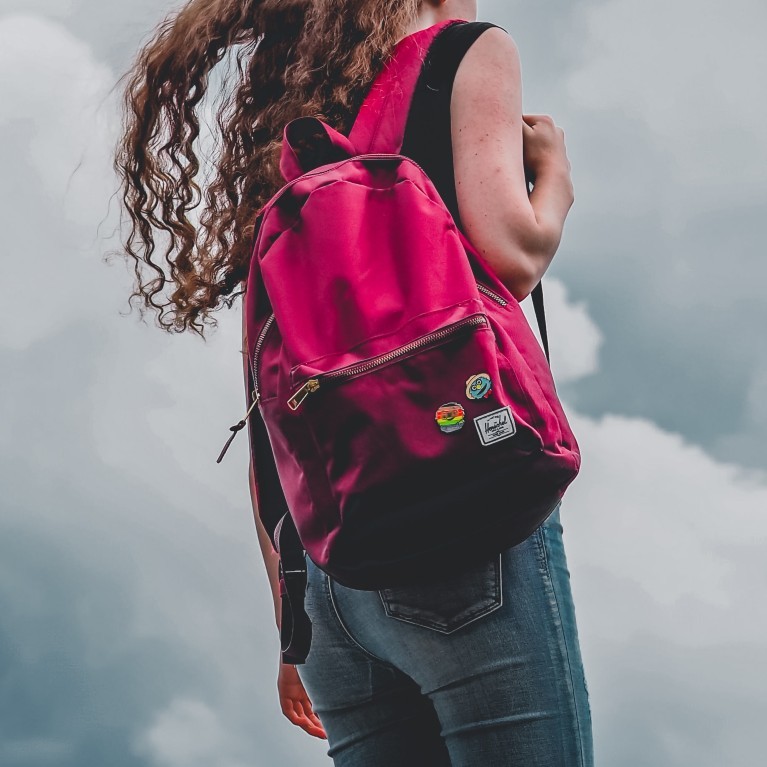}
        {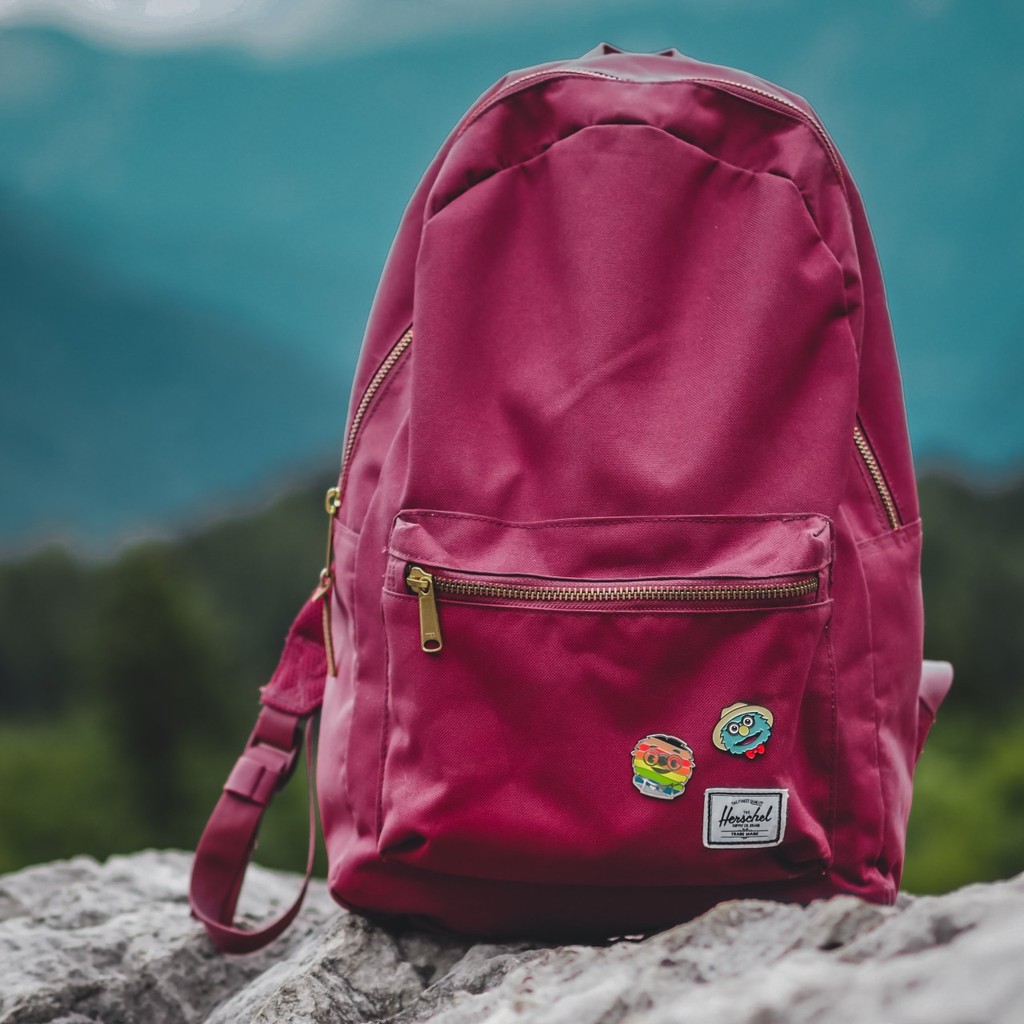}
        {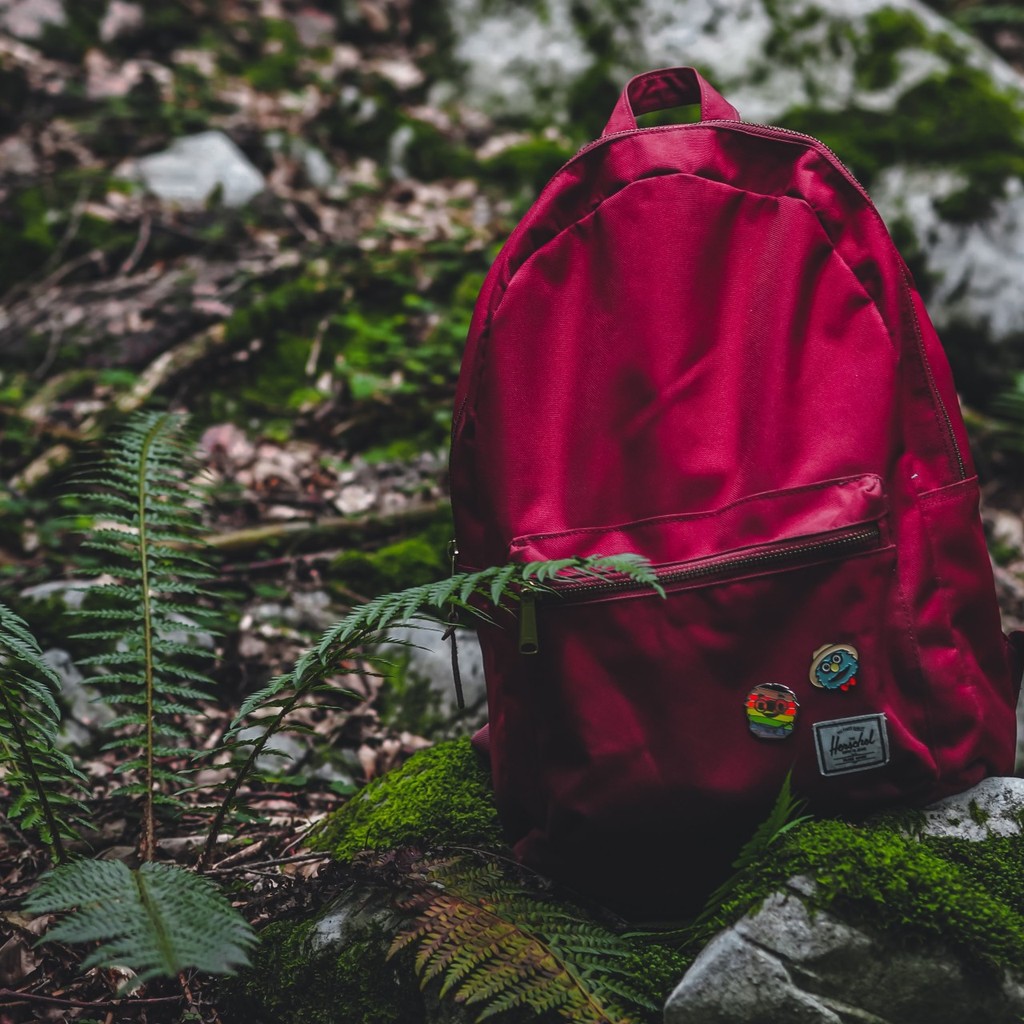}
        {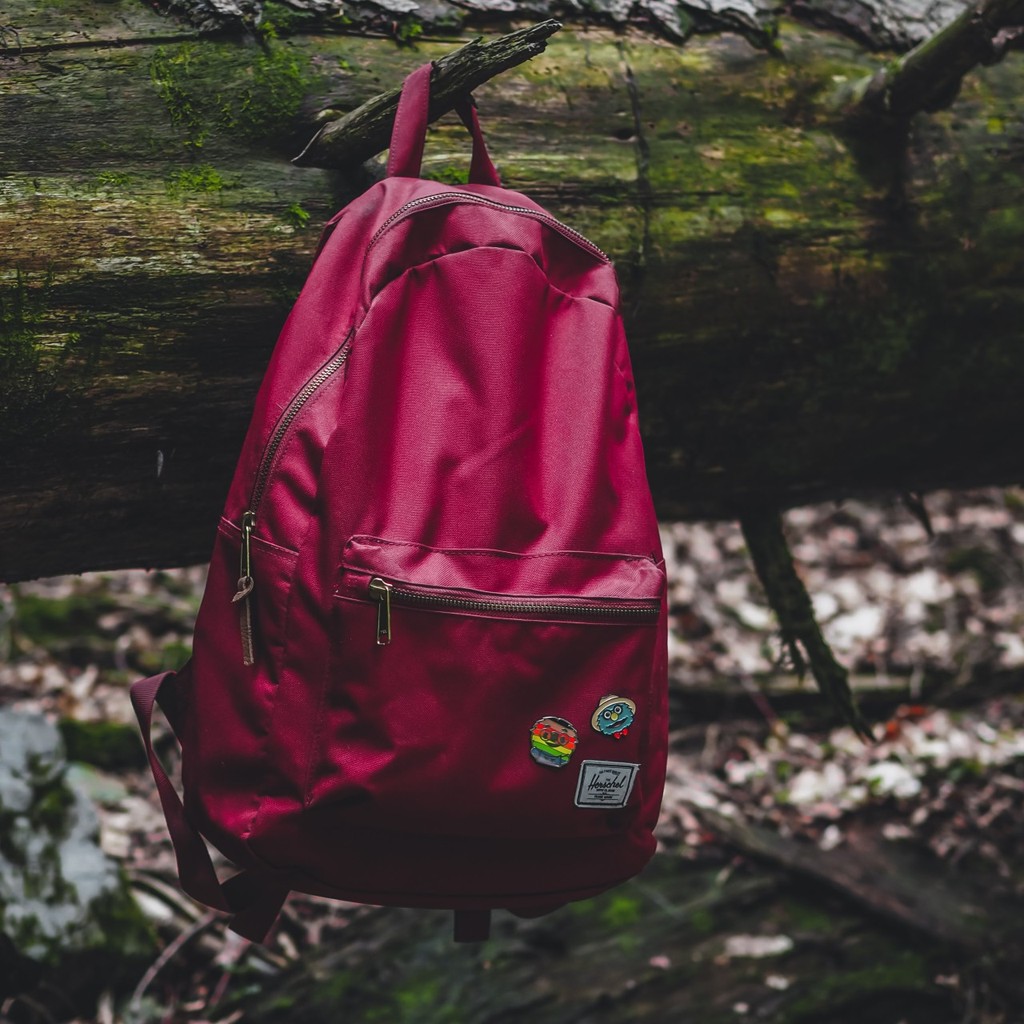}
    &
    \resultimg{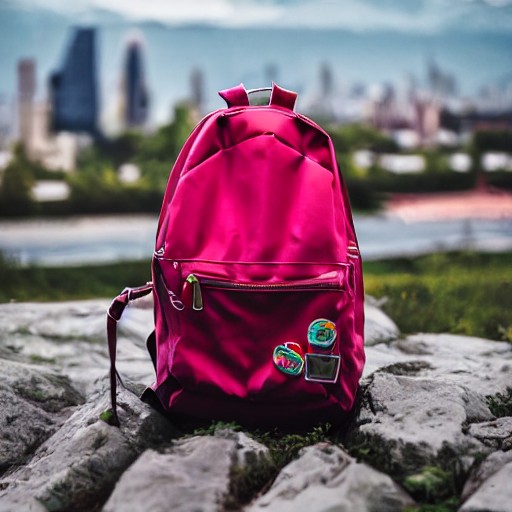}
    &
    \resultimg{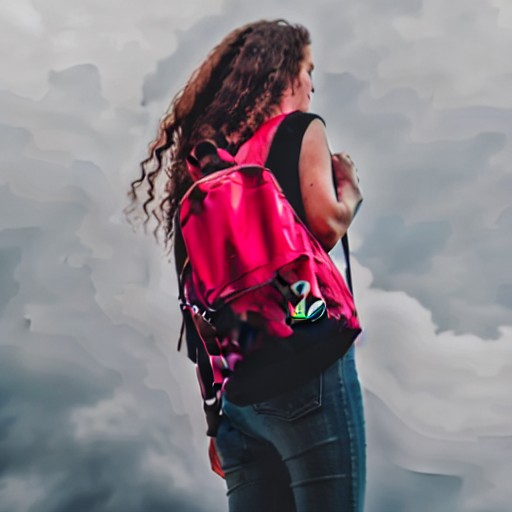}
    &
    \resultimg{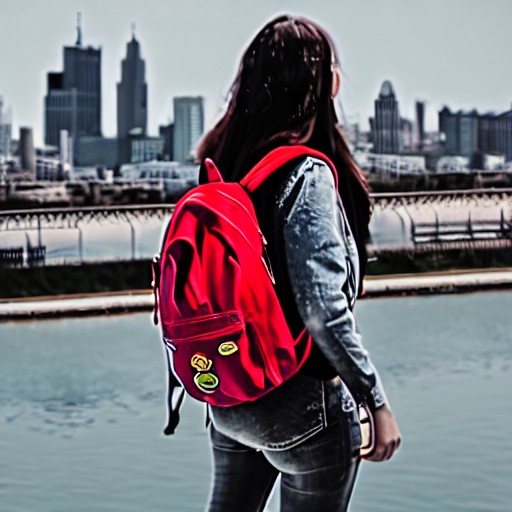}
    &
    \resultimg{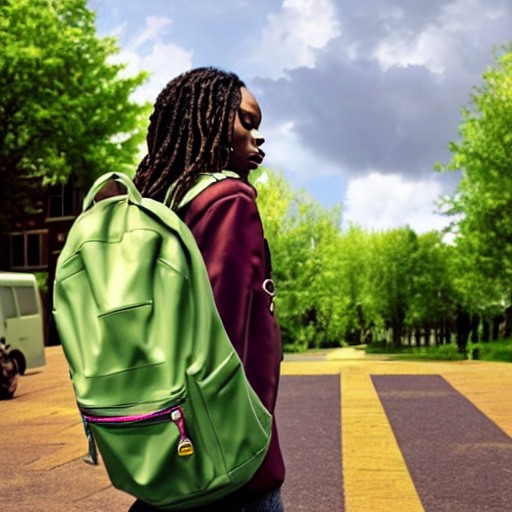}
    &
    \resultimg{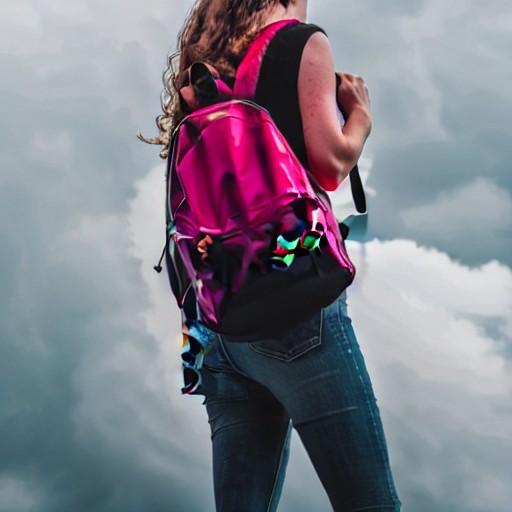}
    &
    \resultimg{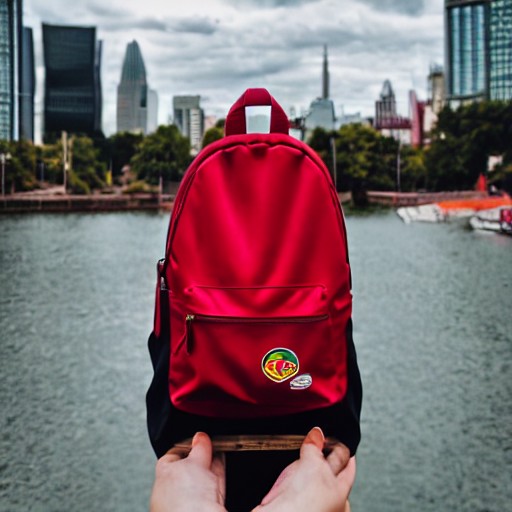}
    &
    \resultimg{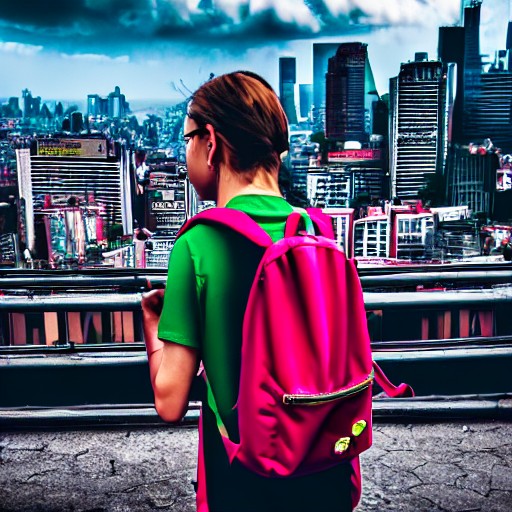}
    &
    \resultimg{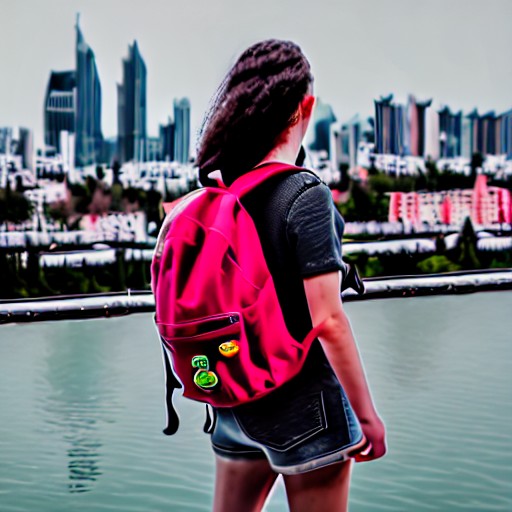}
    &
    \resultimg{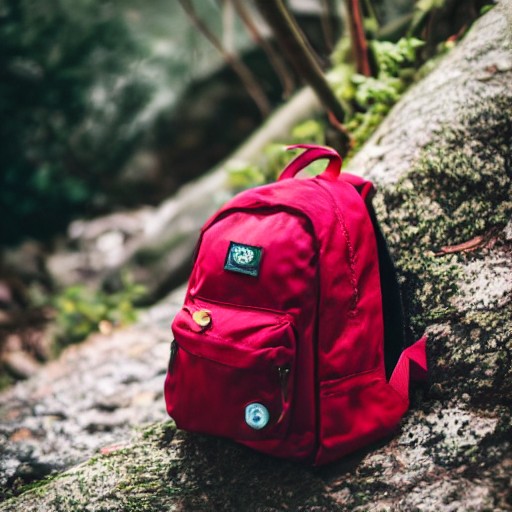}
    &
    \resultimg{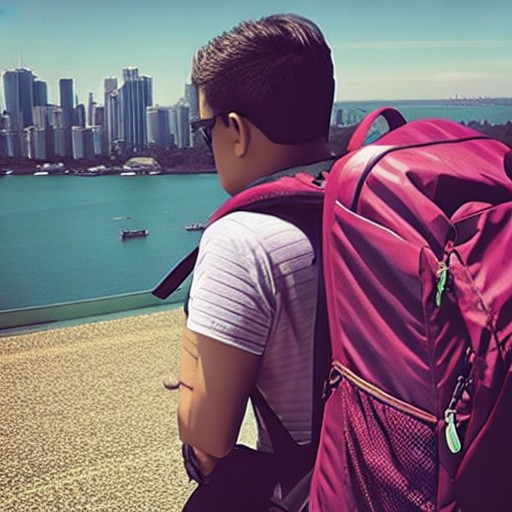}
    \\[-3pt]

    \multicolumn{11}{c}{
        {\scriptsize\itshape
        ``a [V*] with a city in the background''}
    }
    \\[1pt]

    % =========================================================
    % Row 5: Boot
    % =========================================================
    \fourrefimgs
        {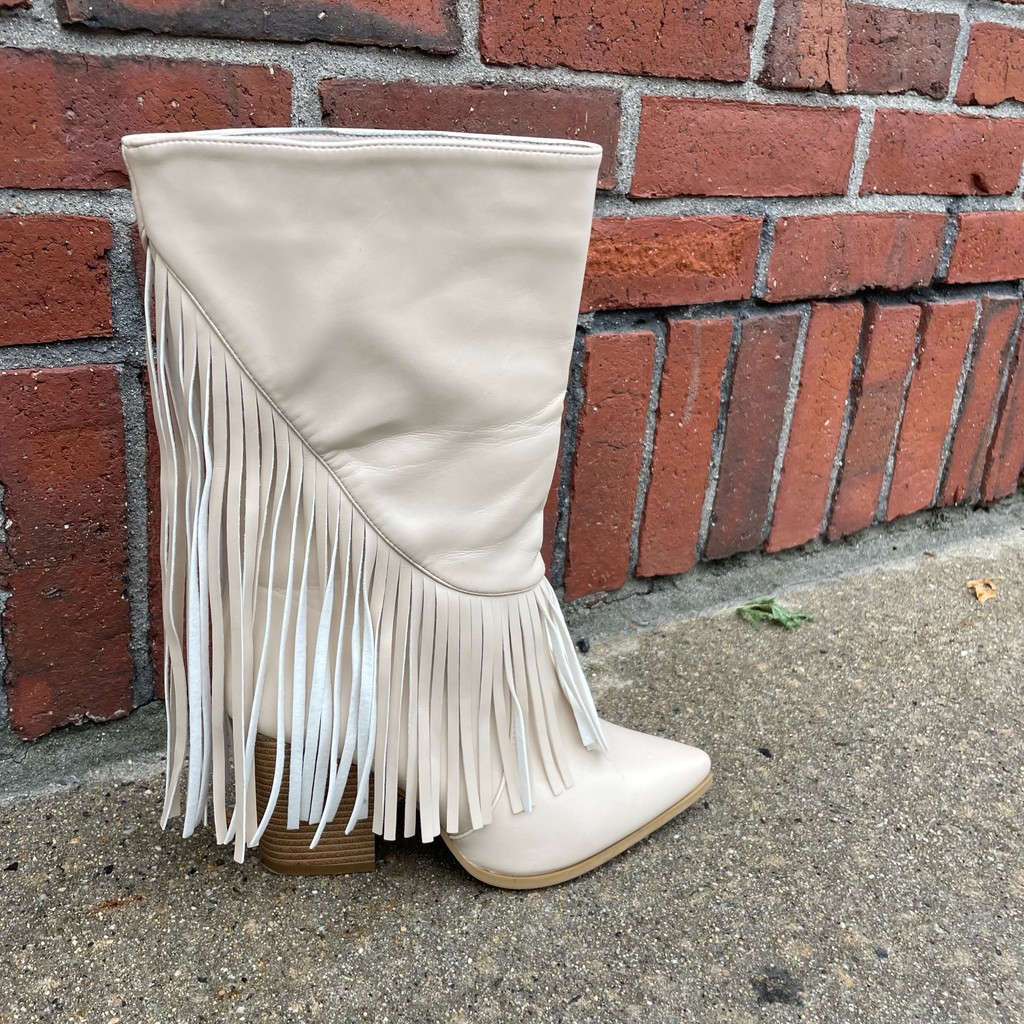}
        {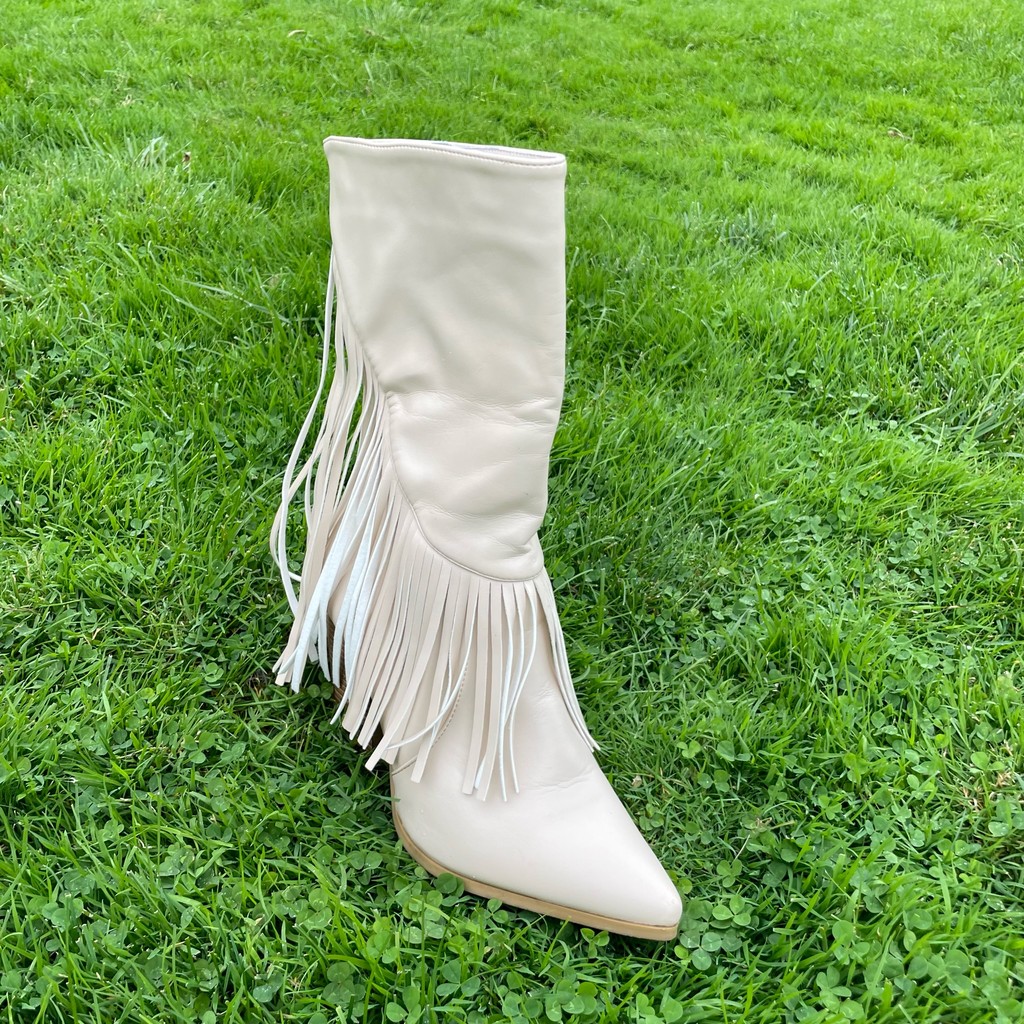}
        {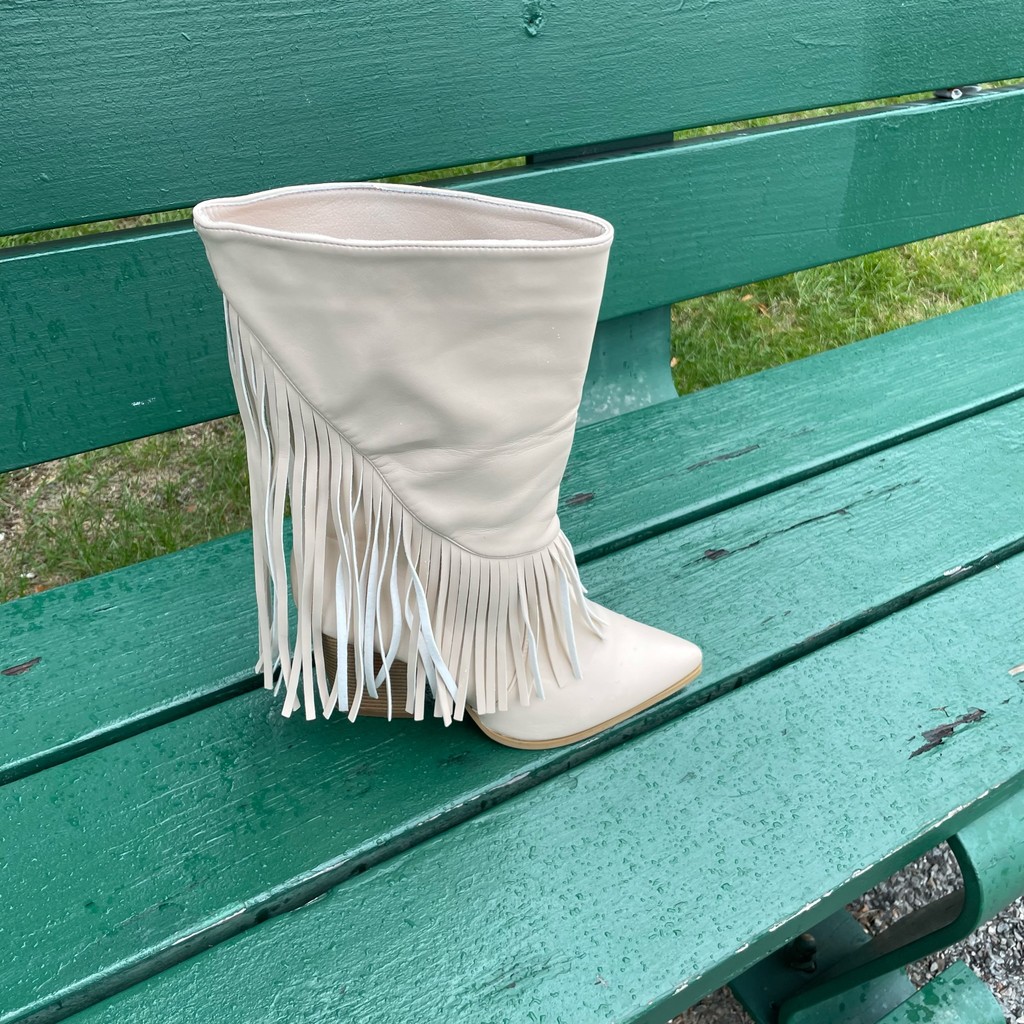}
        {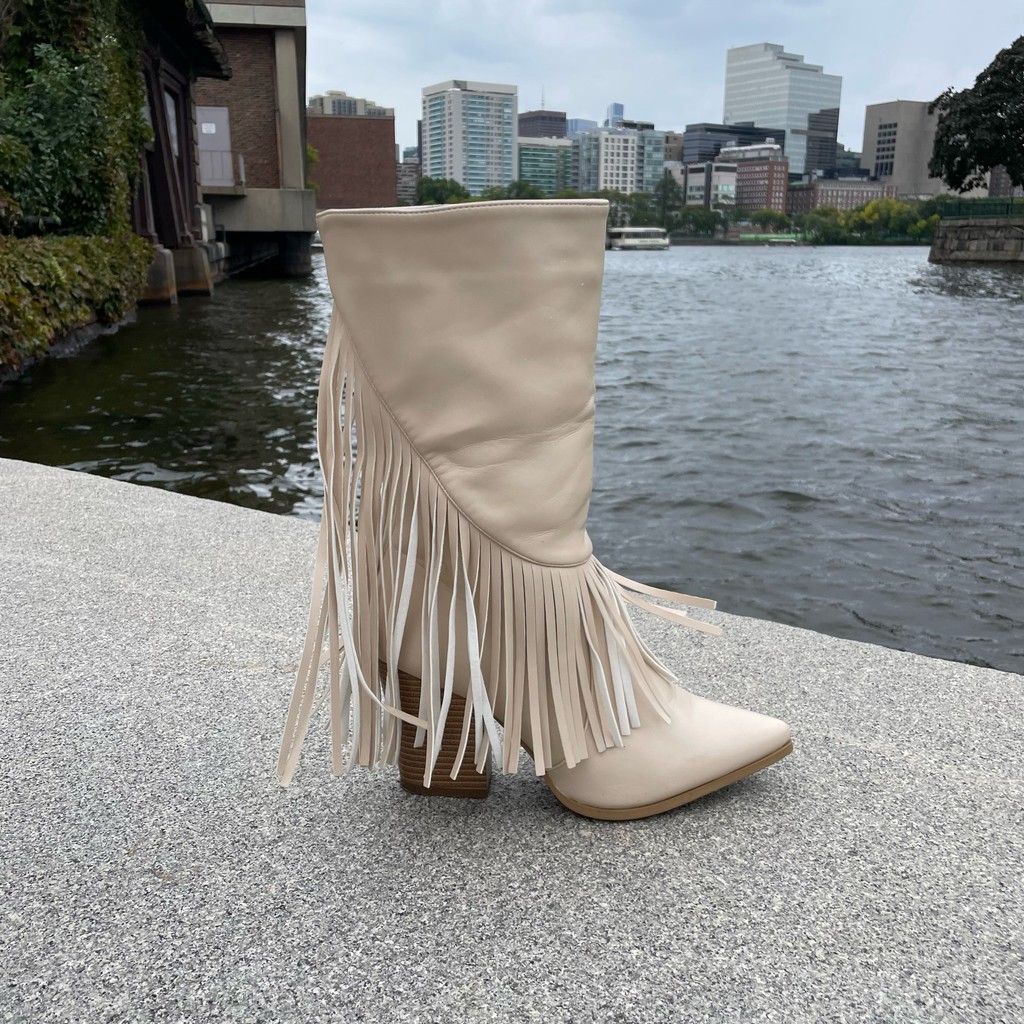}
    &
    \resultimg{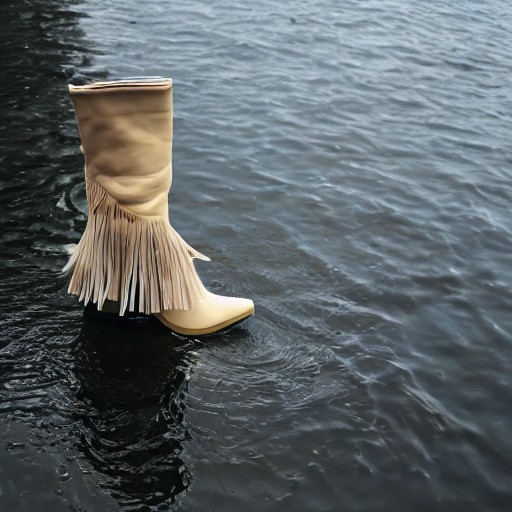}
    &
    \resultimg{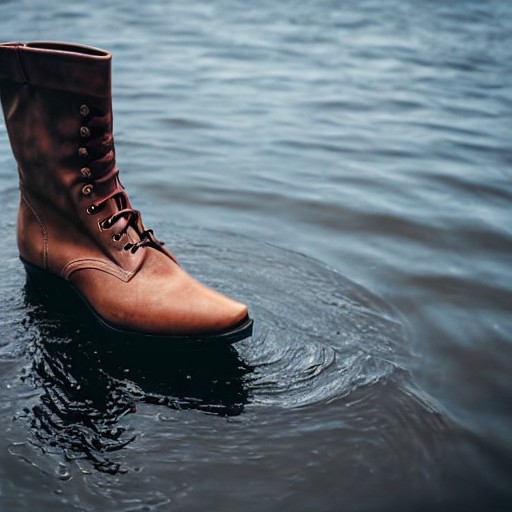}
    &
    \resultimg{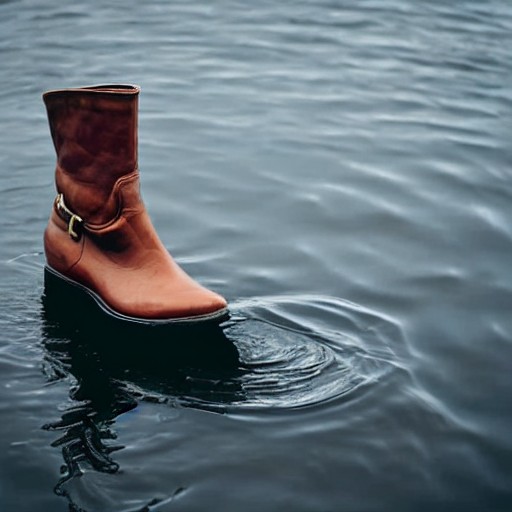}
    &
    \resultimg{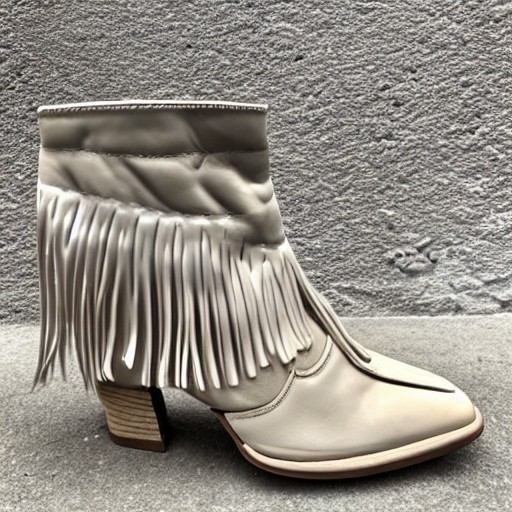}
    &
    \resultimg{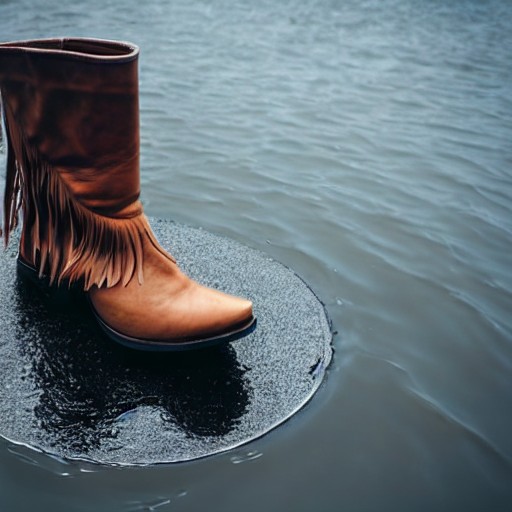}
    &
    \resultimg{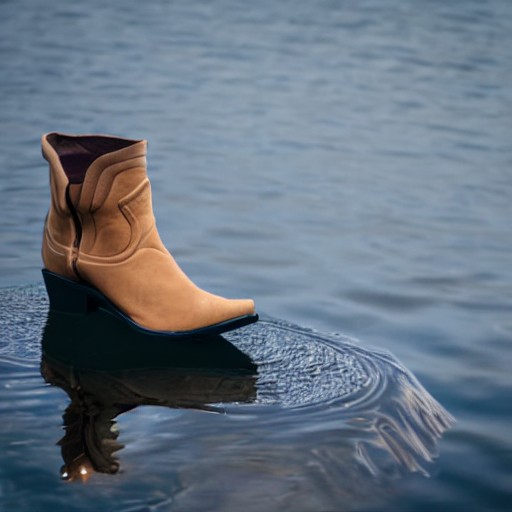}
    &
    \resultimg{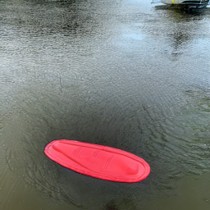}
    &
    \resultimg{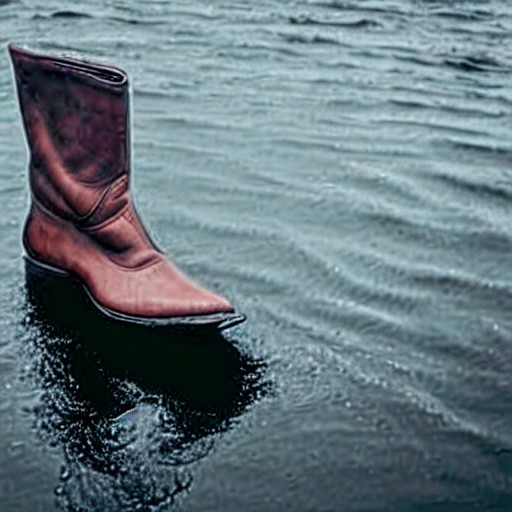}
    &
    \resultimg{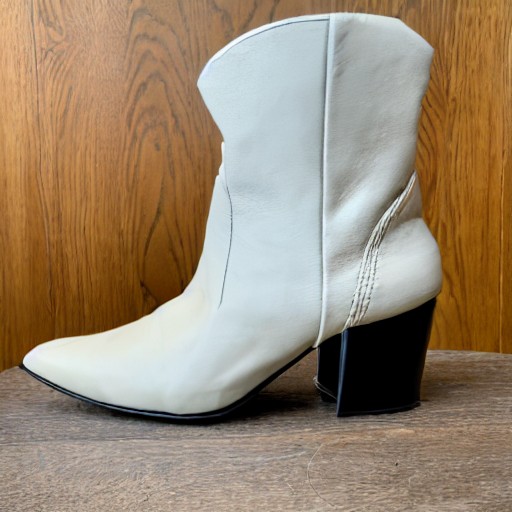}
    &
    \resultimg{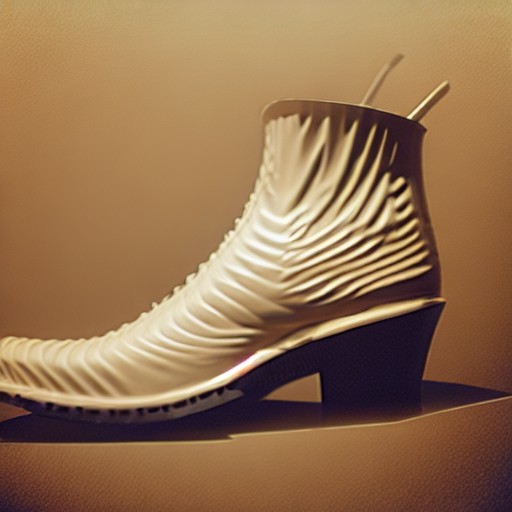}
    \\[-3pt]

    \multicolumn{11}{c}{
        {\scriptsize\itshape
        ``a [V*] floating on top of water''}
    }
    \\[0pt]

    \end{tabular}

    \caption{
        Visual comparison of few-shot personalization and
        encoder-based methods on SD1.5.
    }
    \label{fig:SD1.5_final}
    \vspace{-10pt}
\end{figure*}

%------------------------------
\begin{table}[t!]
  \centering
  % 첫 번째 표 (Table 1)
  \begin{minipage}[t]{0.47\linewidth}
    \centering
    \caption{Quantitative comparison with baseline methods on SD1.5. The Rank column is computed by averaging the per-metric ranks.}
    %\vspace{-8pt} 
    \label{tab:main_1}
    \setlength{\tabcolsep}{6pt} % 간격 미세 조정
    \renewcommand{\arraystretch}{1.05}
    \resizebox{\linewidth}{!}{%
      \begin{tabular}{lccc c}
        \toprule
        \textbf{Method} \rule{0pt}{12pt} & \textbf{CLIP-I $\uparrow$} & \textbf{CLIP-T $\uparrow$} & \textbf{DINO $\uparrow$} & \textbf{Rank} \\
        \midrule\midrule
        \multicolumn{5}{c}{\textit{Few-shot personalization methods}} \\
        \midrule
        \rowcolor{gray!15}\textbf{Ours } & 0.7833 & 0.3064 & 0.6051 & \textbf{1} \\
        DreamBooth    & 0.7761 & 0.3039 & 0.5895 & 4 \\
        DreamBooth-LoRA        & 0.7314 & 0.3195 & 0.4866 & 6 \\
        SuDe              & 0.7800 & 0.3047 & 0.5953 & 3 \\
        SVDiff            & 0.7470 & 0.3167 & 0.5171 & 4 \\
        Textual Inversion & 0.7599 & 0.2578 & 0.5330 & 9 \\
        Custom Diffusion  & 0.6479 & 0.3154 & 0.5568 & 6 \\
        Steering Guidance  & 0.7381 & 0.3133 & 0.5064 & 8 \\
        \midrule\midrule
        \multicolumn{5}{c}{\textit{Encoder-based methods}} \\
        \midrule
        IP-Adapter        & 0.8311 & 0.2646 & 0.6290 & 2 \\
        BLIP-Diffusion    & 0.7016 & 0.2949 & 0.4544 & 10 \\
        \bottomrule
      \end{tabular}%
    }
  \end{minipage}
  \hfill % 두 표 사이의 간격 생성
  % 두 번째 표 (Table 2)
  \begin{minipage}[t]{0.49\linewidth}
    \centering
    \renewcommand{\arraystretch}{1.42}
    \caption{Quantitative comparison on SD1.5, SDXL, and SD3. The Rank column is computed by averaging the per-metric ranks.}
    %\vspace{-8pt} 
    \label{tab:main_2}
    \setlength{\tabcolsep}{4pt} % 간격 미세 조정
    \resizebox{\linewidth}{!}{%
      \begin{tabular}{l l c c c c}
        \toprule
        \textbf{Model} \rule{0pt}{12pt} & \textbf{Method} 
        & \textbf{CLIP-I ${\uparrow}$} 
        & \textbf{CLIP-T ${\uparrow}$} 
        & \textbf{DINO ${\uparrow}$} 
        & \textbf{Rank} \\
        \midrule\midrule

        \multirow{3}{*}{\textbf{SD1.5}} 
         & DreamBooth-LoRA & 0.7314 & 0.3195 & 0.4866 & 2 \\
         & Beyond-Finetuning & 0.7294 & \textbf{0.3212} & 0.4786 & 3 \\
         & \cellcolor{gray!15}\textbf{Ours } 
           & \cellcolor{gray!15}\textbf{0.7833} 
           & \cellcolor{gray!15}0.3064 
           & \cellcolor{gray!15}\textbf{0.6051} 
           & \cellcolor{gray!15}\textbf{1} \\
        \midrule

        \multirow{3}{*}{\textbf{SDXL}}
         & DreamBooth-LoRA & 0.7939 & 0.3107 & 0.6586 & 2 \\
         & Beyond-Finetuning & 0.7872 & 0.3141 & 0.6429 & 3 \\
         & \cellcolor{gray!15}\textbf{Ours } 
           & \cellcolor{gray!15}\textbf{0.8005} 
           & \cellcolor{gray!15}\textbf{0.3154} 
           & \cellcolor{gray!15}\textbf{0.6661} 
           & \cellcolor{gray!15}\textbf{1} \\
        \midrule

        \multirow{3}{*}{\textbf{SD3}} 
         & DreamBooth-LoRA & 0.7449 & 0.3185 & 0.5109 & 2 \\
         & Beyond-Finetuning & 0.7443 & \textbf{0.3226} & 0.5037 & 2 \\
         & \cellcolor{gray!15}\textbf{Ours } 
           & \cellcolor{gray!15}\textbf{0.7579} 
           & \cellcolor{gray!15}0.3187 
           & \cellcolor{gray!15}\textbf{0.5679} 
           & \cellcolor{gray!15}\textbf{1} \\
        \bottomrule
      \end{tabular}%
    }
  \end{minipage}
    % \vspace{-10pt}
  
\end{table}
\vspace{-10pt}
\section{Experiments}
\label{sec:experiments}
\vspace{-5pt}

In this section, we present experiments and analyses to validate the effectiveness of our method across various subjects and diffusion backbones. For further analysis beyond metrics, we also conduct a user study to assess perceptual quality and preference. Additional qualitative results and detailed analyses are provided in Appendix~\ref{sec:supp_add_visual_results}.

\vspace{-5pt}
\subsection{Experimental Setup}
\label{sec:experimental_setup}
\vspace{-5pt}
\paragraph{\textbf{Datasets.}} We conduct experiments on the widely used personalization benchmark~\citep{ruiz2023dreambooth}. The benchmark contains 30 subjects, including both objects and live subjects. Each subject provides 4--6 reference images, and each model is evaluated using 25 distinct textual templates with four generated samples per prompt to measure subject and text consistency. We use
$c^\text{sbj}$: ``\texttt{a photo of [V*] [Subject]}'' as the personalized prompt and
$c^\text{cls}$: ``\texttt{a photo of a [Subject]}'' as the class-level anchor prompt.
This setup follows the standard personalization setting, and we introduce no additional prompt modifications.

\vspace{-5pt}
\paragraph{\textbf{Evaluation Metrics.}}
We evaluate the generated results along two dimensions. Subject fidelity, quantified by CLIP-I~\citep{radford2021learning} and DINO~\citep{caron2021emerging}, measures the cosine similarity between image embeddings of generated images and reference samples. Text fidelity, measured by CLIP-T, evaluates text-image alignment through the cosine similarity between CLIP text and image embeddings. Since subject fidelity and text fidelity often exhibit a trade-off, evaluating either dimension alone may not reflect the overall personalization performance. We therefore report an aggregate rank by averaging the per-metric ranks across CLIP-I, CLIP-T, and DINO, where a lower rank indicates a better balance between subject fidelity and text alignment.

% We conduct experiments on the widely used personalization benchmark \cite{ruiz2023dreambooth}. The benchmark contains 30 subjects encompassing both objects and live subjects. Each subject provides 4-6 reference images and every model is prompted with 25 distinct textual templates, generating four samples per prompt to evaluate both subject and text consistency.
% We follow the standard personalization protocol~\citep{ruiz2023dreambooth}, using
% $c^\text{sbj}$: ``\texttt{a photo of [V*] [Subject]}'' as the personalized prompt.
% For SAP, we adopt the class prompt used for class image generation as the anchor prompt,
% $c^\text{anc}$: ``\texttt{a photo of a [Subject]}''.
% The \texttt{Subject} token follows the benchmark setting~\citep{ruiz2023dreambooth}, with no additional prompt modification.
% For the personalized and anchoring prompts,  we define the target prompt as $c^\text{sbj}$: ``\texttt{a photo of [V*] [Subject]}'' and the anchor prompt as $c^\text{anc}$: ``\texttt{a photo of a [Subject]}''. Here, the \texttt{Subject} for each subject is configured based on the general settings ~\citep{ruiz2023dreambooth}.

\vspace{-5pt}
\paragraph{\textbf{Baselines.}}
We compare our method against several personalization approaches. The baselines include DreamBooth~\citep{ruiz2023dreambooth} with both full fine-tuning and LoRA-based variants, SuDe~\citep{qiao2024facechain}, SVDiff~\citep{han2023svdiff}, Textual Inversion~\citep{gal2022image}, Custom Diffusion~\citep{kumari2023multi}, and Steering Guidance~\citep{park2025steering}. We also include encoder-based methods, including IP-Adapter~\citep{ye2023ip} and BLIP-Diffusion~\citep{li2023blip}. For comparisons across personalization methods, we adopt SD1.5, where official implementations and training recipes are available, facilitating consistent comparison across methods. Furthermore, we evaluate our method across SD1.5~\citep{rombach2022high}, SDXL~\citep{podell2023sdxl}, and SD3~\citep{esser2024scaling} to examine its applicability across different model backbones. For this cross-backbone evaluation, we compare with Beyond-Finetuning~\citep{soboleva2025beyond}, which applies class-level guidance at inference time, allowing comparison with our training-time formulation. For all experiments, our method builds upon LoRA-based adaptation.

% All fine-tuning experiments adopt LoRA-based adaptation \cite{hu2022lora} for efficient personalization.

%\vspace{-3pt}
\subsection{Comparison with Baseline Methods}
\label{sec:experiments_baseline}
%\vspace{-3pt}

Table~\ref{tab:main_1} summarizes quantitative comparisons with baseline methods across subject fidelity and text alignment metrics. Our method obtains the top rank under this evaluation, indicating the best balance between subject fidelity and text alignment among the compared methods. In particular, SAP improves upon DreamBooth and SuDe across all three metrics. Compared with DreamBooth-LoRA, SAP yields substantial gains in CLIP-I and DINO under the same LoRA adaptation setting, highlighting the effect of the proposed training objective on learning subject-specific attributes. Although DreamBooth-LoRA achieves a higher CLIP-T score, further qualitative analysis suggests that this gain is associated with underfitting to the personalized subject, leading to weaker subject fidelity despite stronger text alignment.

The importance of balancing subject fidelity and text alignment becomes more evident in the qualitative results as illustrated in Fig.~\ref{fig:SD1.5_final}. In the third and fifth rows, SVDiff, Custom Diffusion, and Steering Guidance follow the target prompts but fail to preserve distinctive subject details. In contrast, DreamBooth and SuDe preserve the subject appearance, but the first and fourth rows show cases where the generated images remain close to the reference images and do not reflect the target prompt. These examples illustrate the trade-off between subject fidelity and prompt alignment and motivate evaluating both aspects jointly. SAP maintains a better balance across these cases. Although subject--prompt mixing appears in the second row, the generated results preserve subject identity while following the target prompt. We further support these findings with a user study and additional analyses provided in the Appendices~\ref{sec:supp_add_visual_results} and ~\ref{sec:supp_user_study}.

\begin{figure*}[t]
    \centering
    \setlength{\tabcolsep}{0pt}

    % =========================================================
    % 크기 설정: minipage 밖에서 확정
    % =========================================================
    \newlength{\ffResultSize}
    \newlength{\ffResultsWidth}
    \newlength{\ffDogRefSize}
    \newlength{\ffTeapotRefSize}
    \newlength{\ffRefColumnWidth}

    \setlength{\ffResultSize}{0.115\textwidth}
    \setlength{\ffResultsWidth}{6\ffResultSize}

    % Reference 전체 높이 비율
    % 0.98: 살짝 작게 / 1.00: 동일하게
    \newcommand{\ffRefScale}{0.7}

    % 결과 이미지
    \newcommand{\ffResult}[1]{%
        \includegraphics[
            width=\ffResultSize,
            height=\ffResultSize
        ]{#1}%
    }

    % Reference 4장: 간격 없는 2×2 배열
    % #1: 배열 전체 한 변 길이
    % #2~#5: 이미지 경로
    \newcommand{\ffFourRefs}[5]{%
        \vbox{%
            \offinterlineskip
            \hbox{%
                \includegraphics[
                    width=\dimexpr#1/2\relax,
                    height=\dimexpr#1/2\relax
                ]{#2}%
                \includegraphics[
                    width=\dimexpr#1/2\relax,
                    height=\dimexpr#1/2\relax
                ]{#3}%
            }%
            \hbox{%
                \includegraphics[
                    width=\dimexpr#1/2\relax,
                    height=\dimexpr#1/2\relax
                ]{#4}%
                \includegraphics[
                    width=\dimexpr#1/2\relax,
                    height=\dimexpr#1/2\relax
                ]{#5}%
            }%
        }%
    }

    % =========================================================
    % 결과 블록 저장
    % 이미지 1행 + 중간 prompt + 이미지 2행
    % 마지막 prompt는 포함하지 않음
    % =========================================================
    \newsavebox{\ffDogResults}
    \newsavebox{\ffTeapotResults}

    % ---------------------------------------------------------
    % DOG
    % ---------------------------------------------------------
    \sbox{\ffDogResults}{%
        \begin{minipage}[c]{\ffResultsWidth}
            \centering
            \begin{tabular}{
                @{}
                *{6}{C{\ffResultSize}}
                @{}
            }

                % 이미지 행 1
                \ffResult{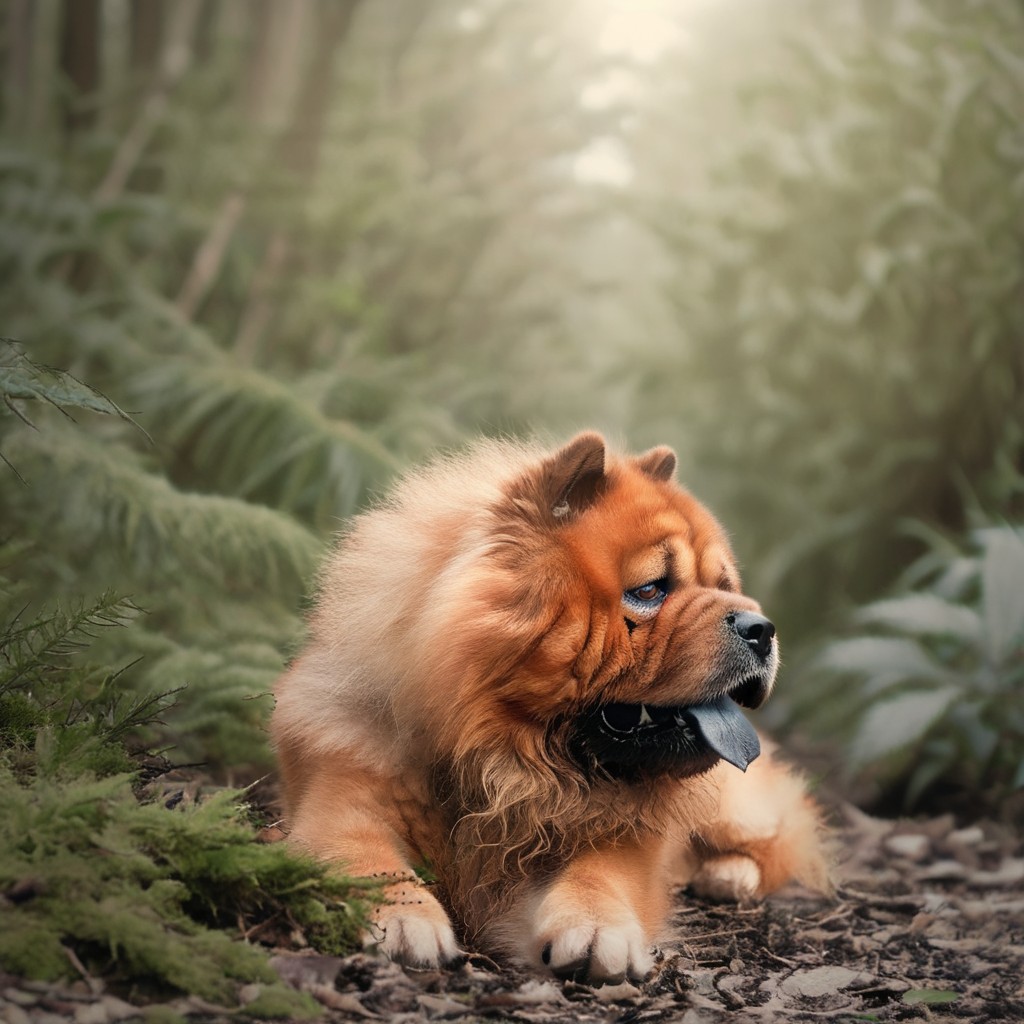}
                &
                \ffResult{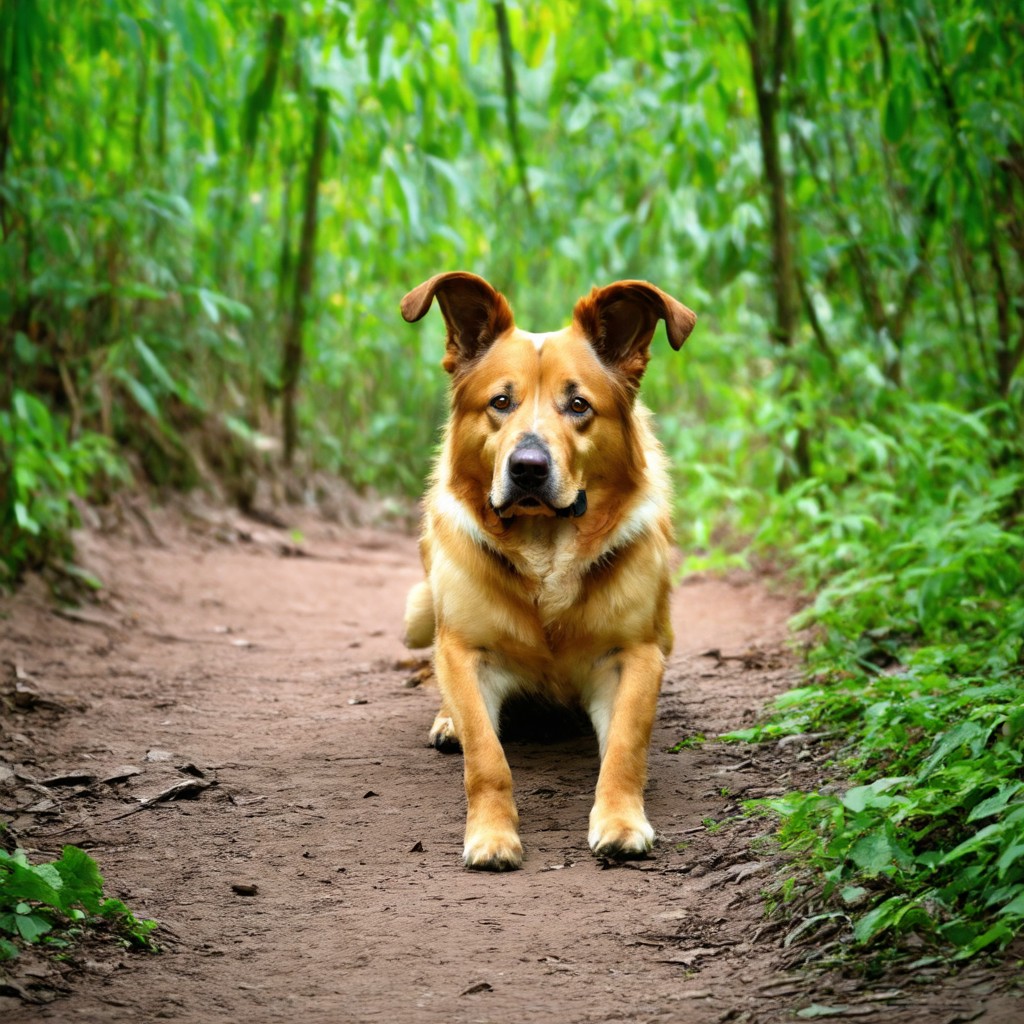}
                &
                \ffResult{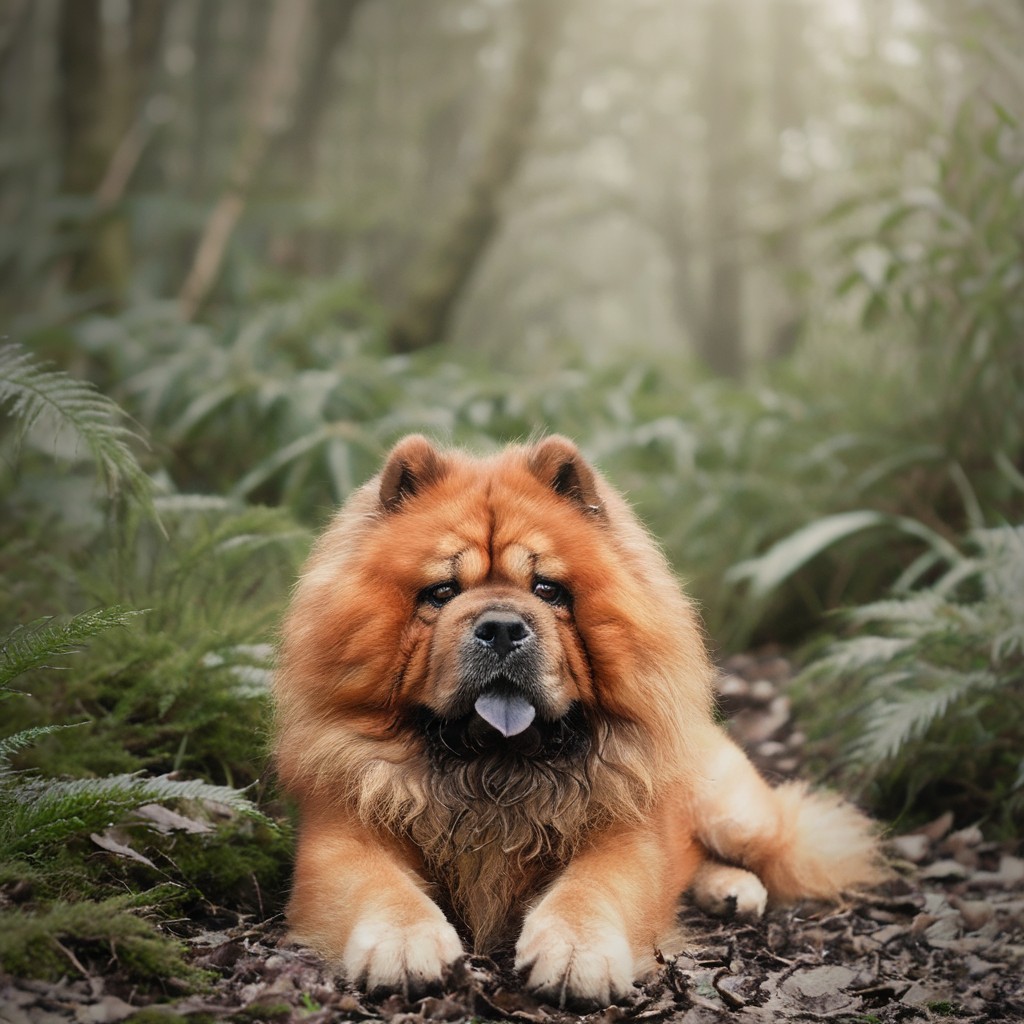}
                &
                \ffResult{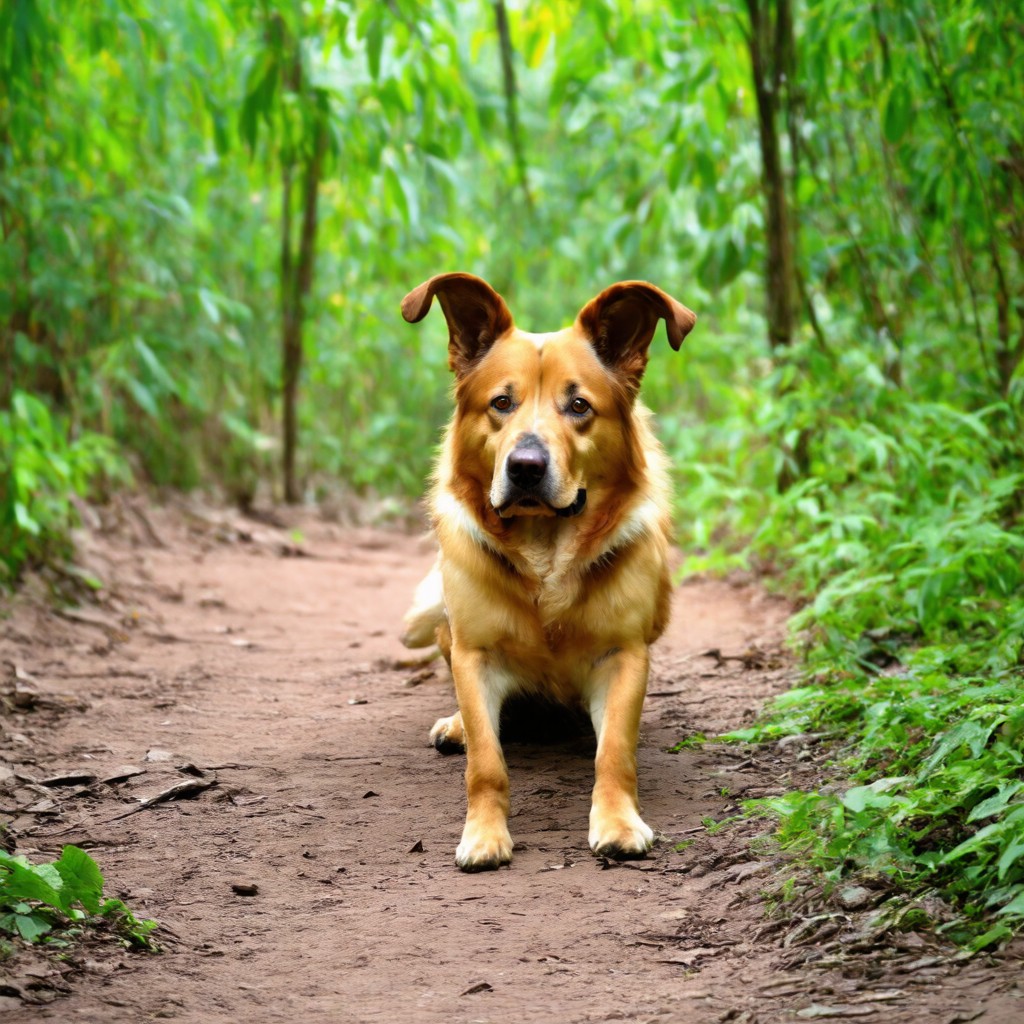}
                &
                \ffResult{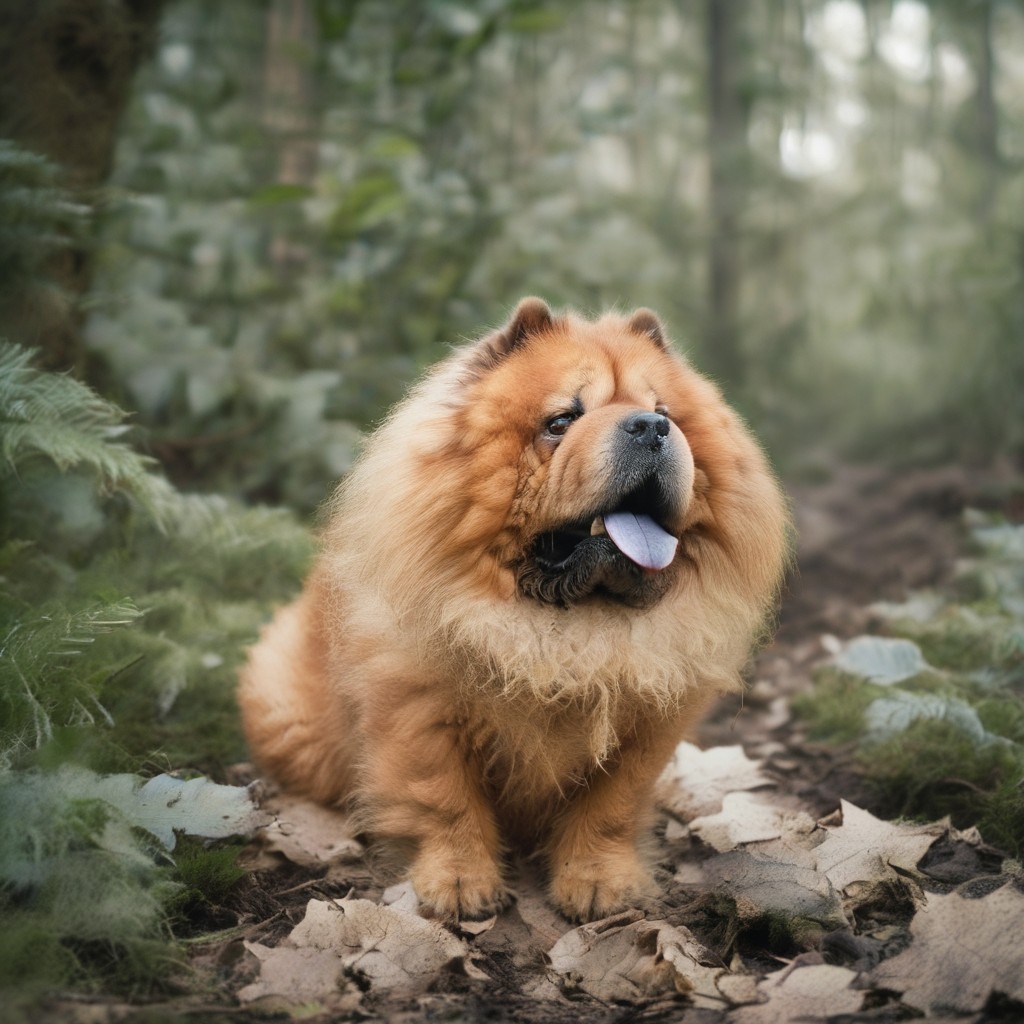}
                &
                \ffResult{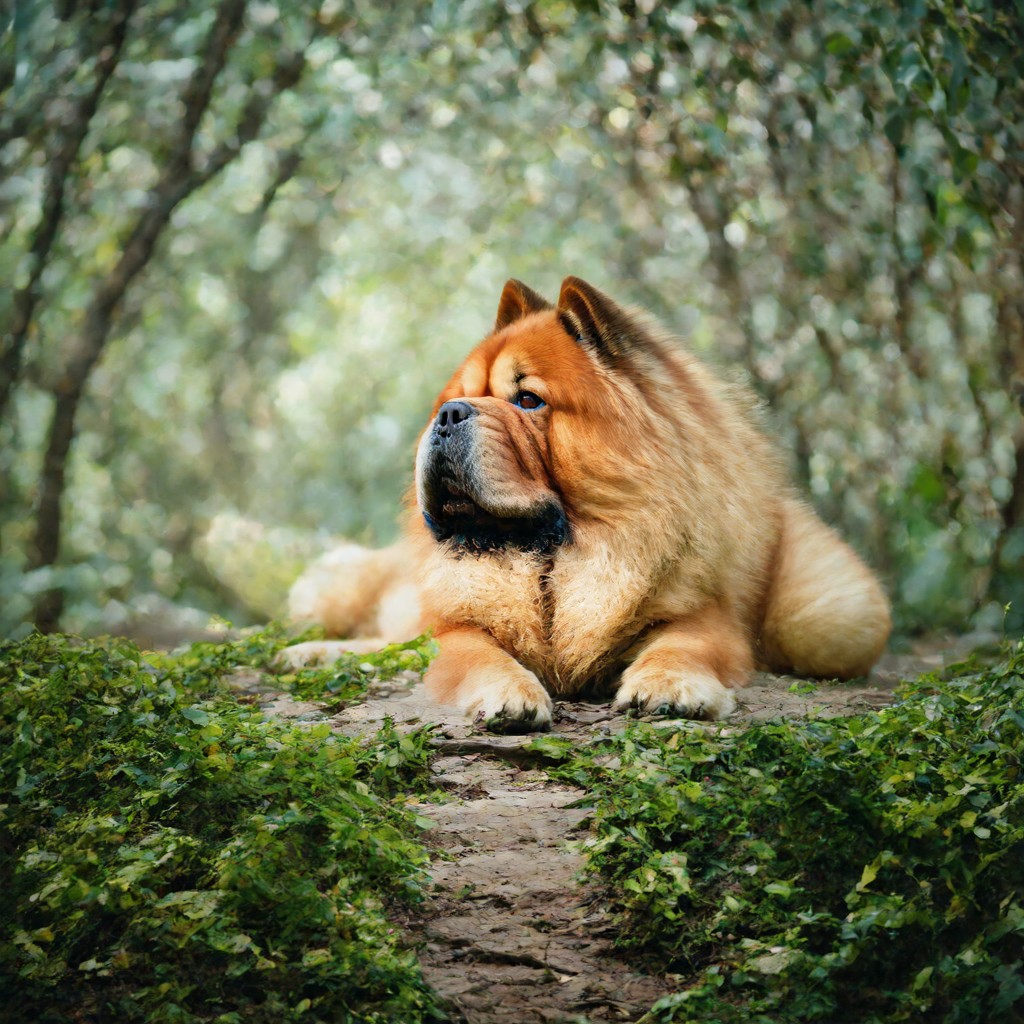}
                \\[-3pt]

                % 중간 prompt
                \multicolumn{6}{c}{%
                    \figfiveprompt{``a [V*] in the jungle''}
                }
                \\[1pt]

                % 이미지 행 2
                \ffResult{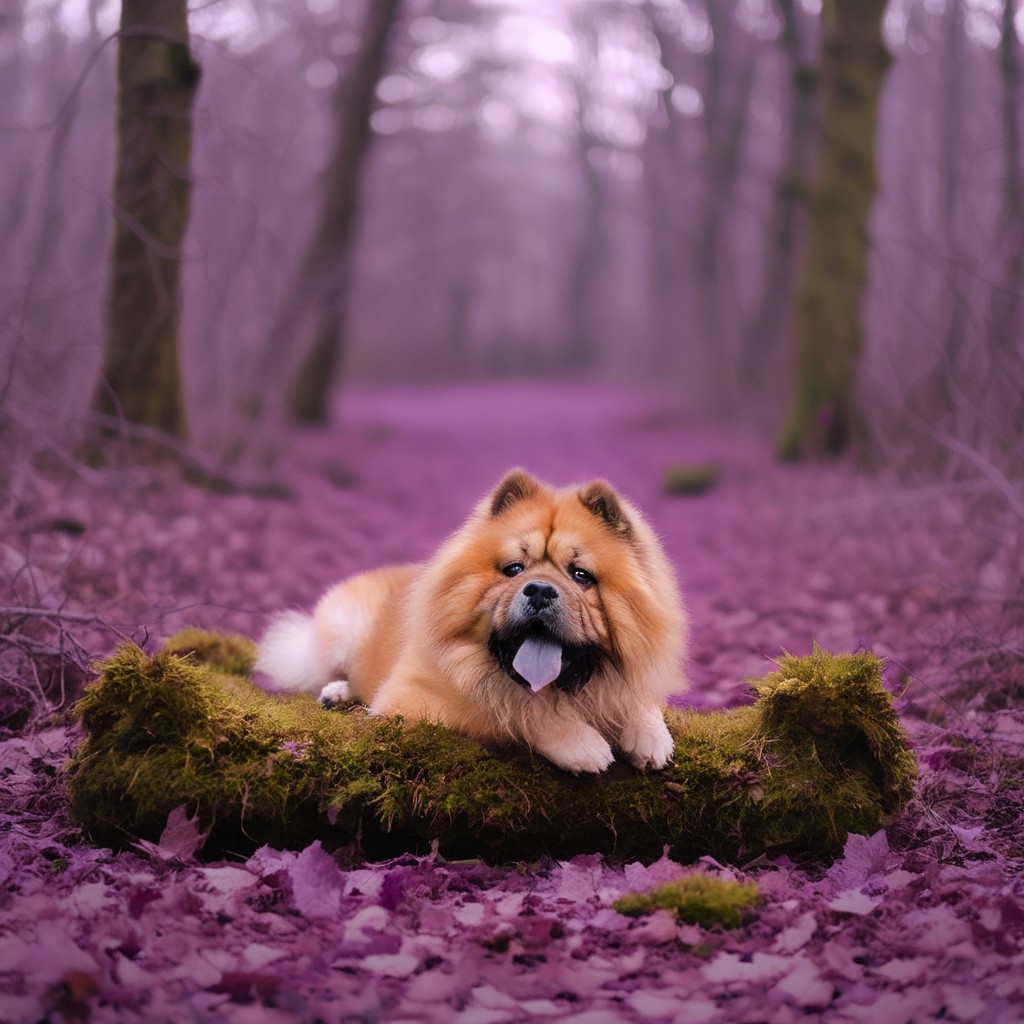}
                &
                \ffResult{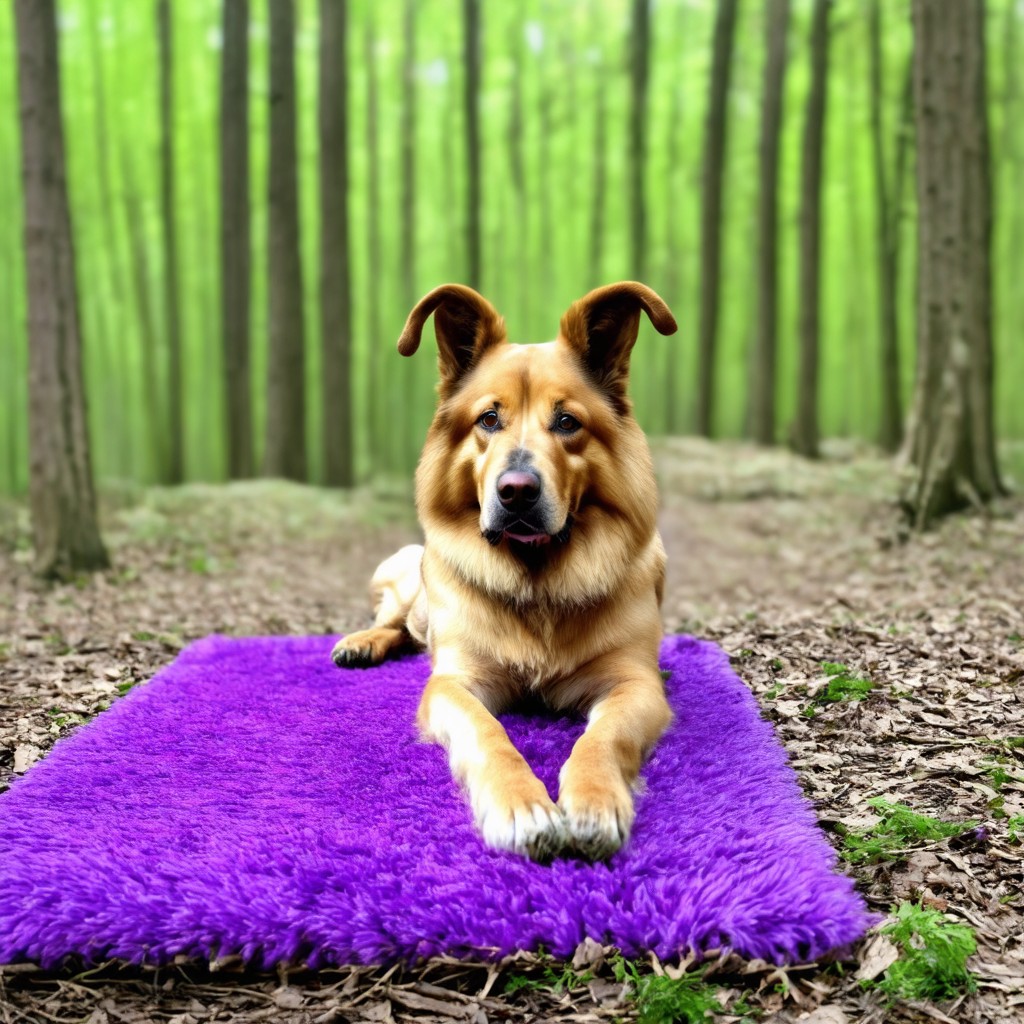}
                &
                \ffResult{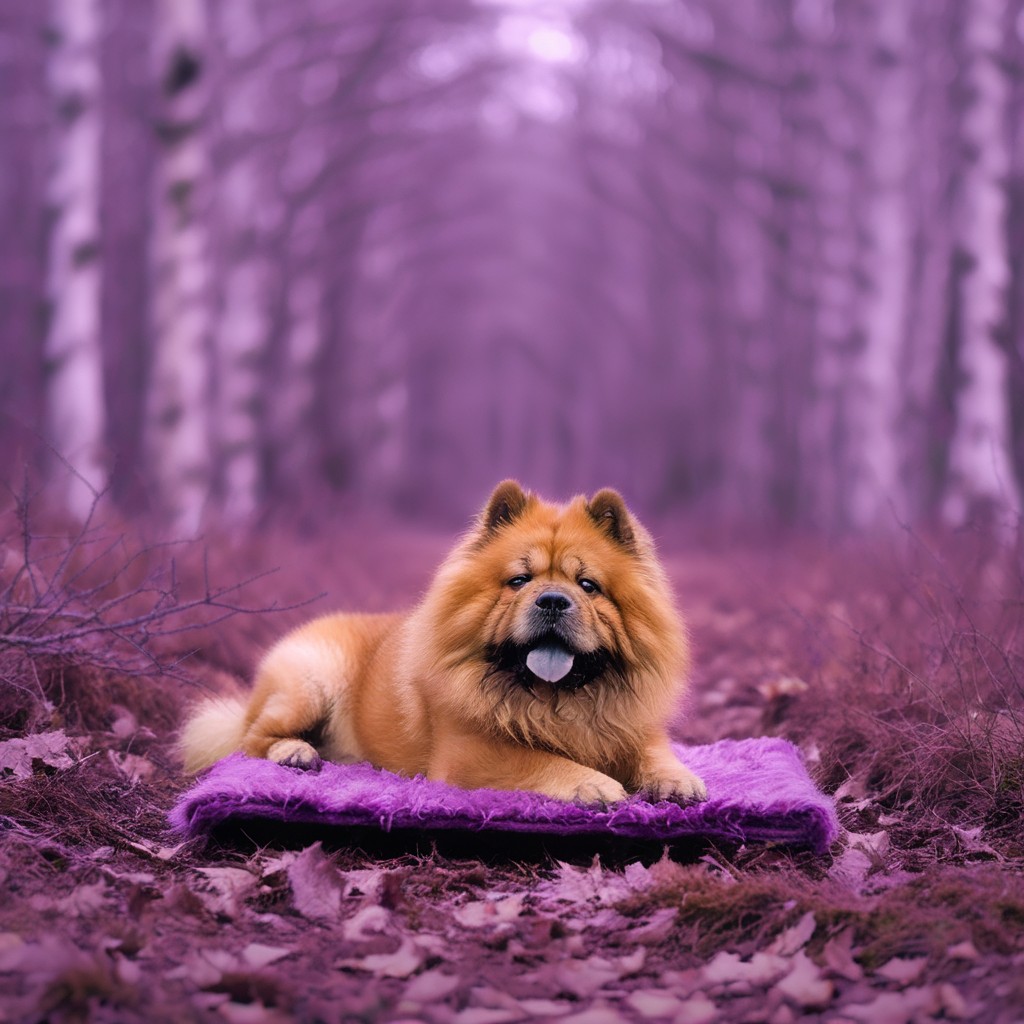}
                &
                \ffResult{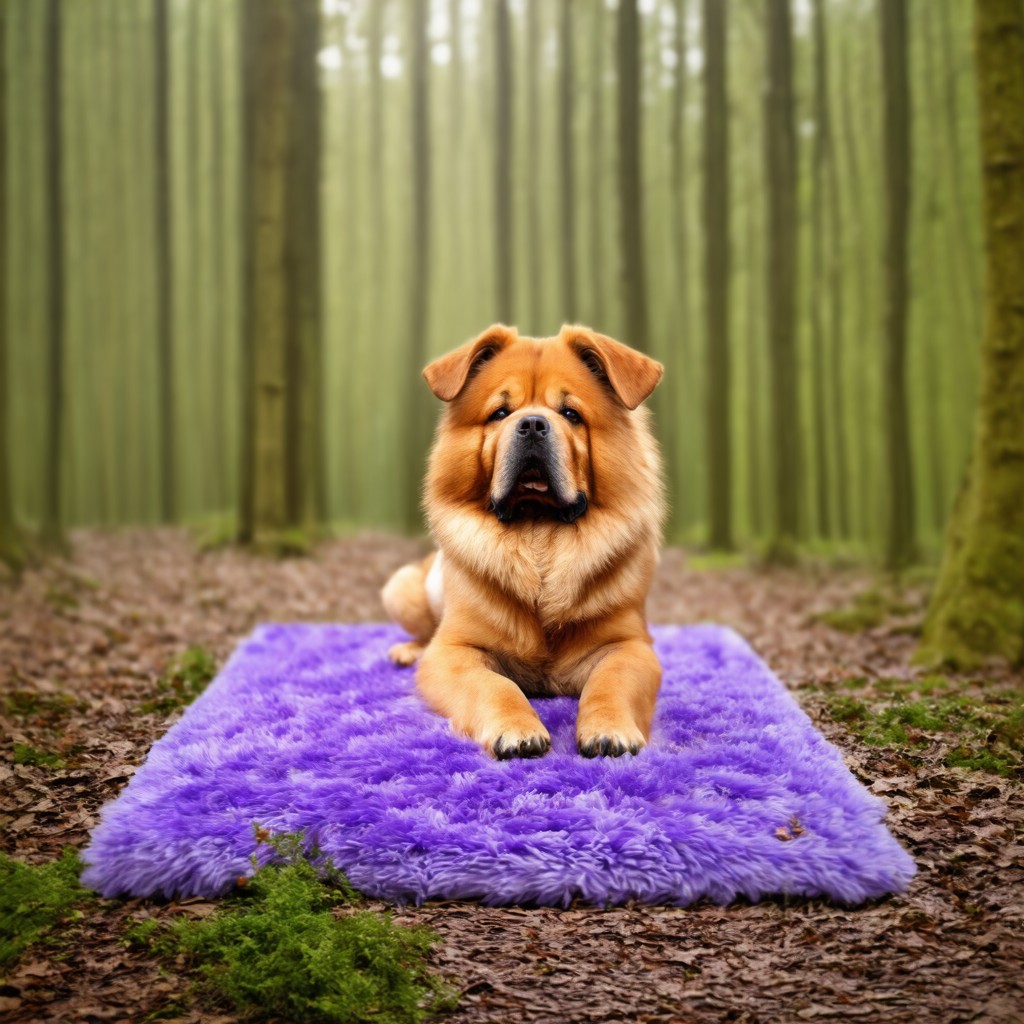}
                &
                \ffResult{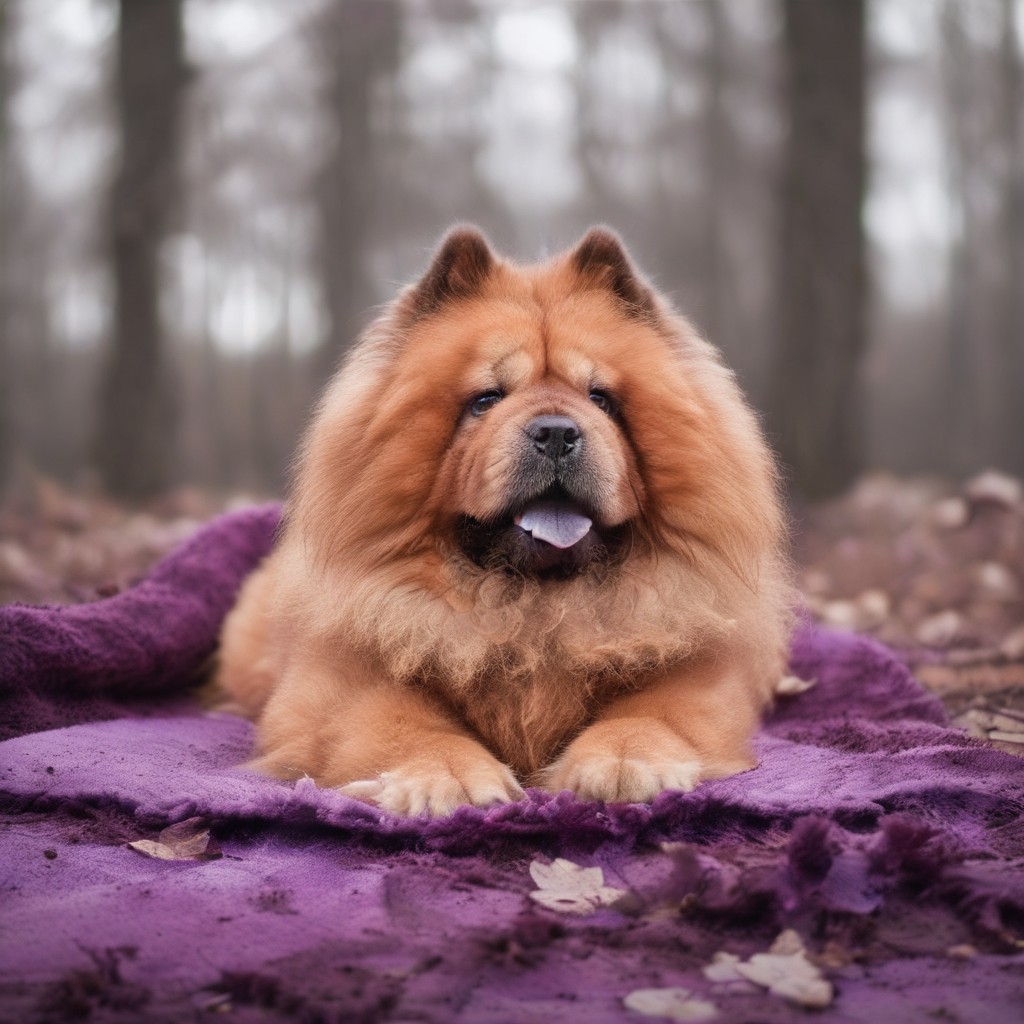}
                &
                \ffResult{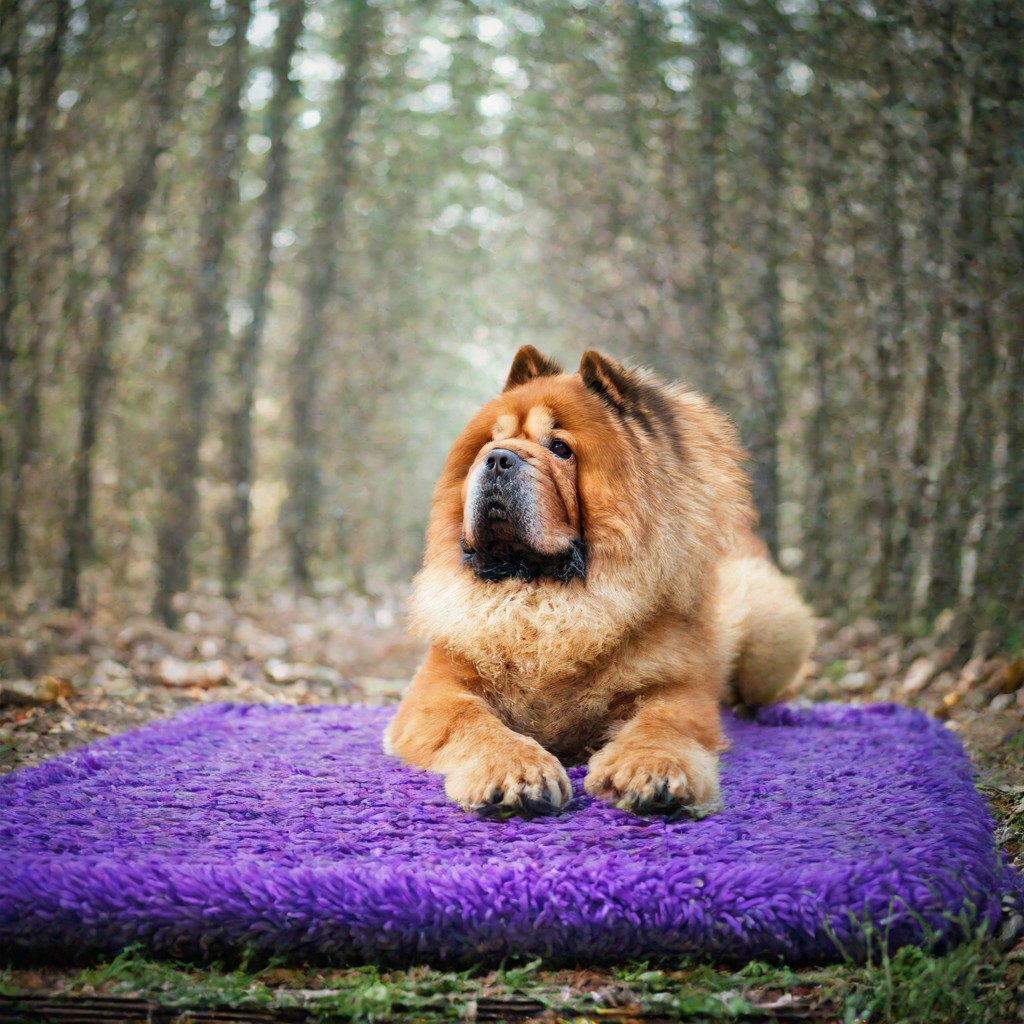}

            \end{tabular}
        \end{minipage}%
    }

    % ---------------------------------------------------------
    % TEAPOT
    % ---------------------------------------------------------
    \sbox{\ffTeapotResults}{%
        \begin{minipage}[c]{\ffResultsWidth}
            \centering
            \begin{tabular}{
                @{}
                *{6}{C{\ffResultSize}}
                @{}
            }

                % 이미지 행 1
                \ffResult{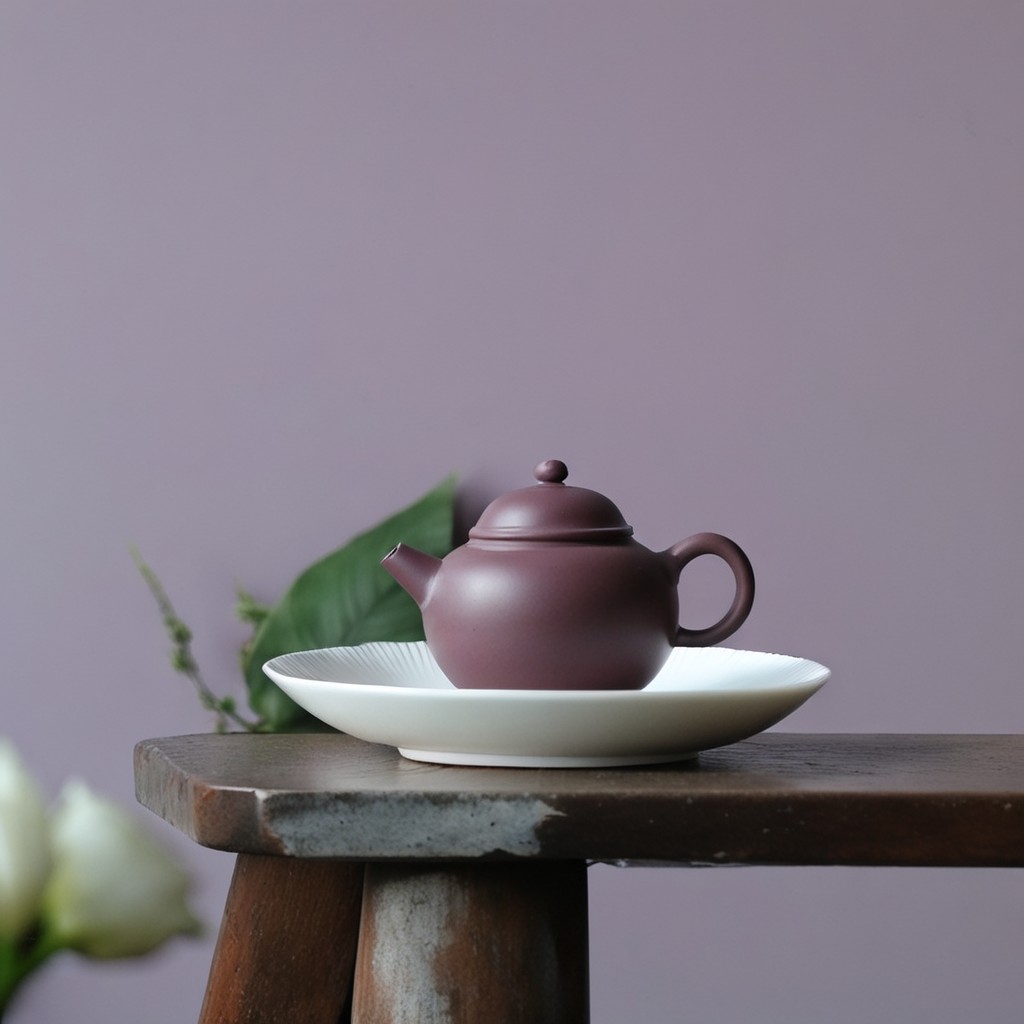}
                &
                \ffResult{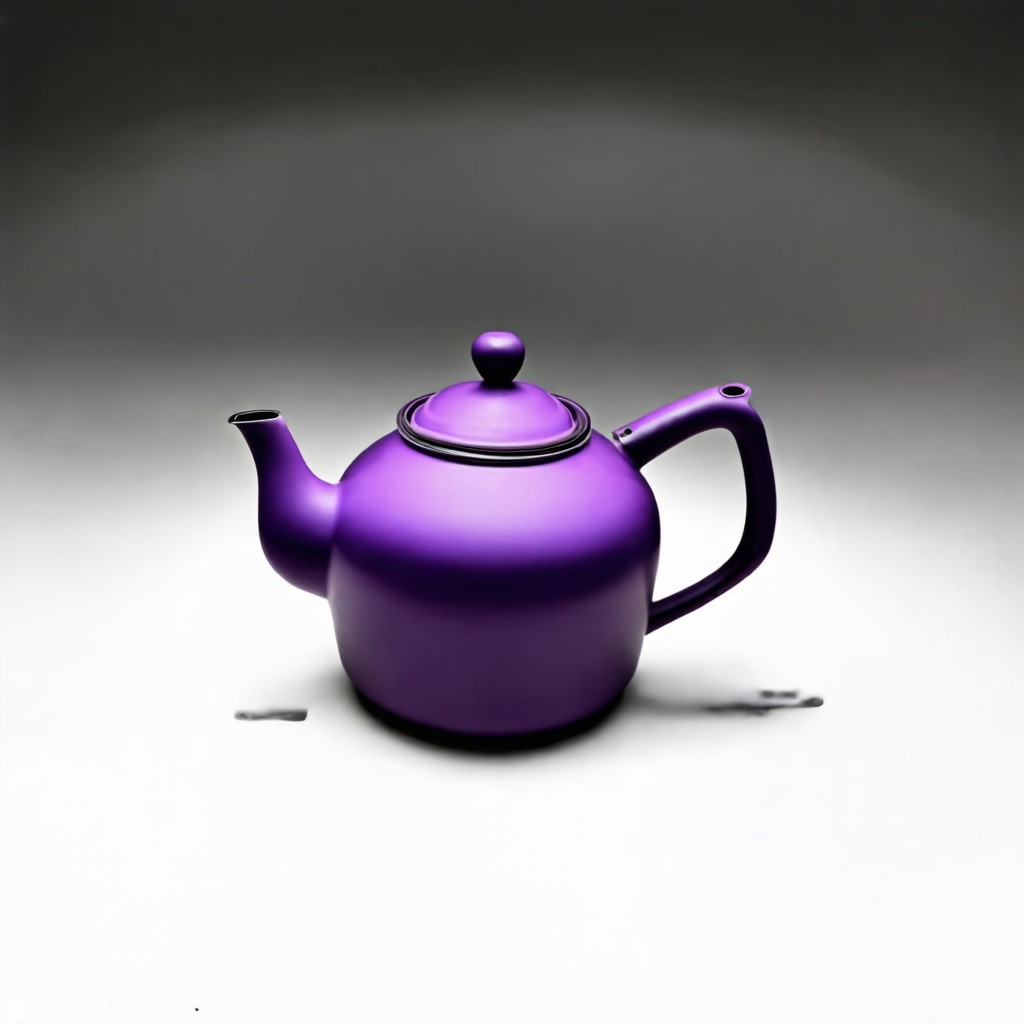}
                &
                \ffResult{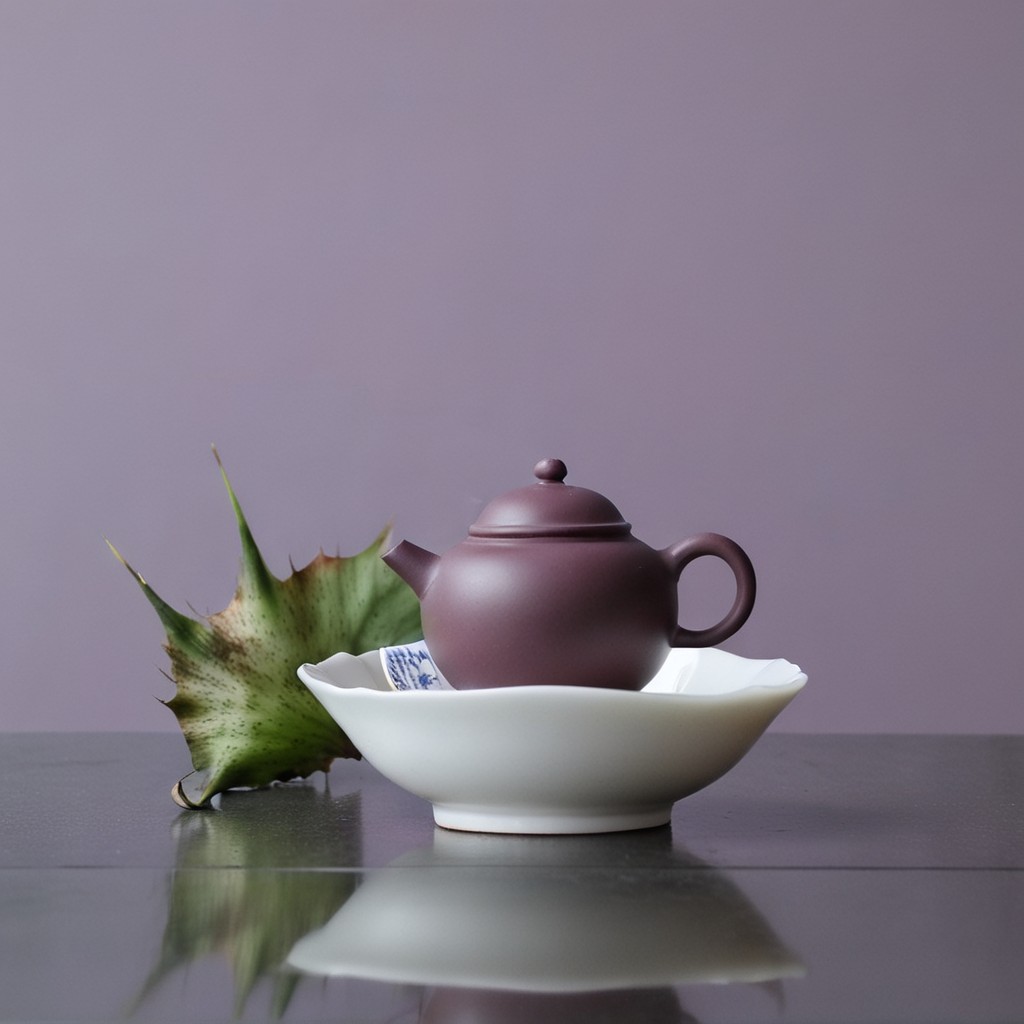}
                &
                \ffResult{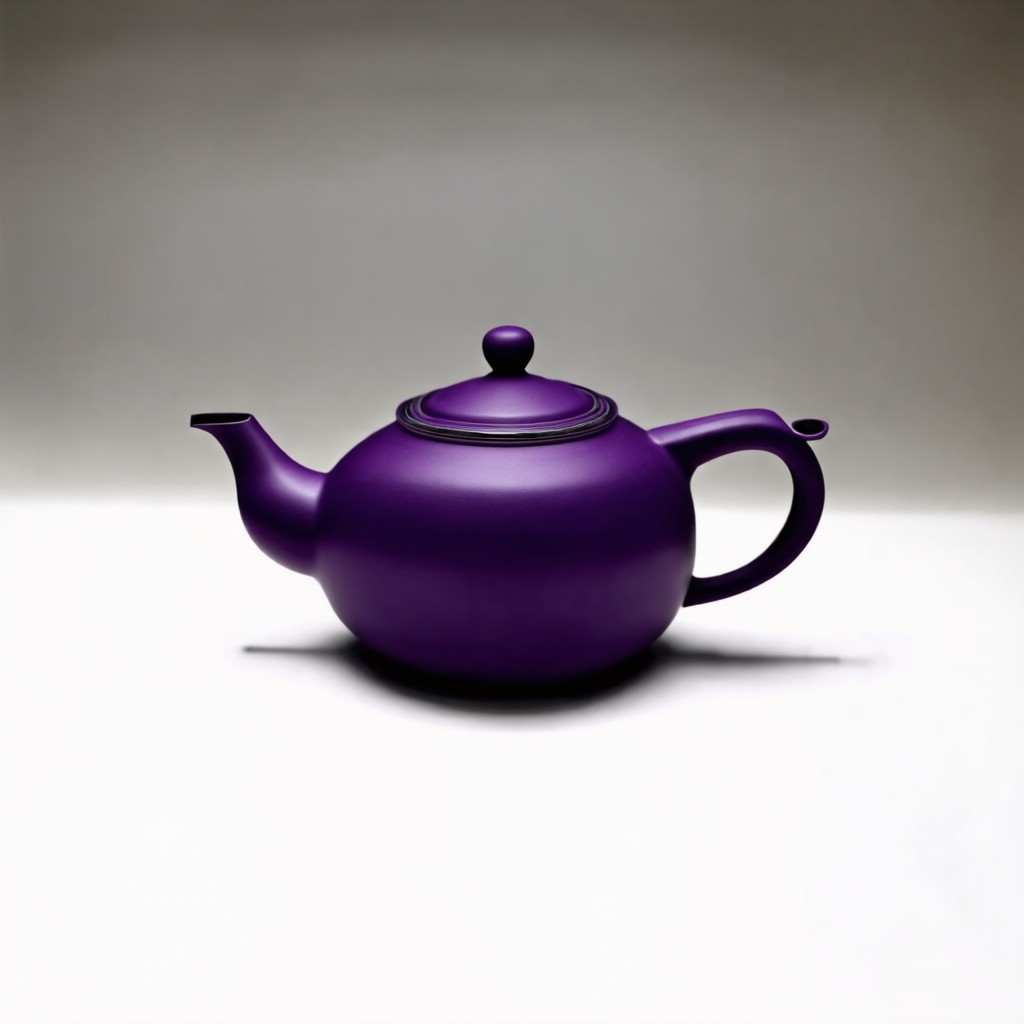}
                &
                \ffResult{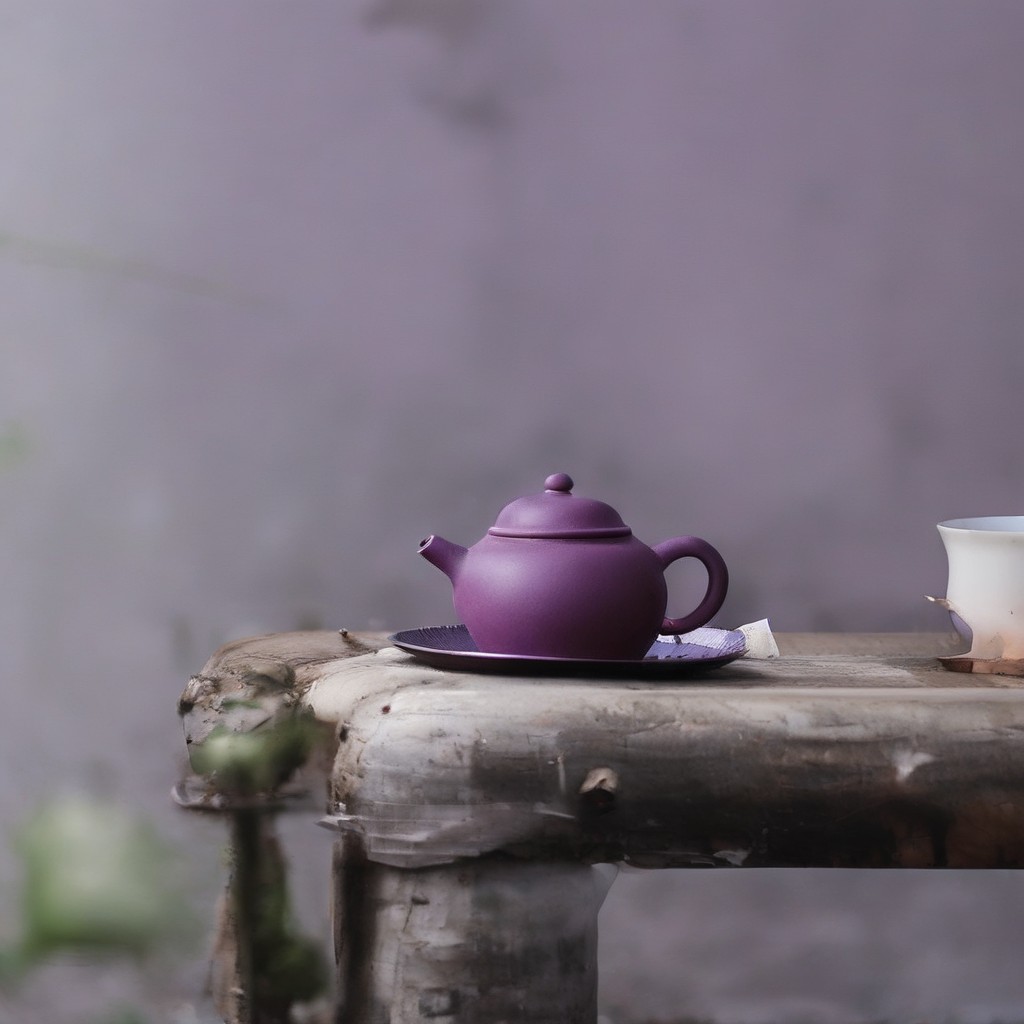}
                &
                \ffResult{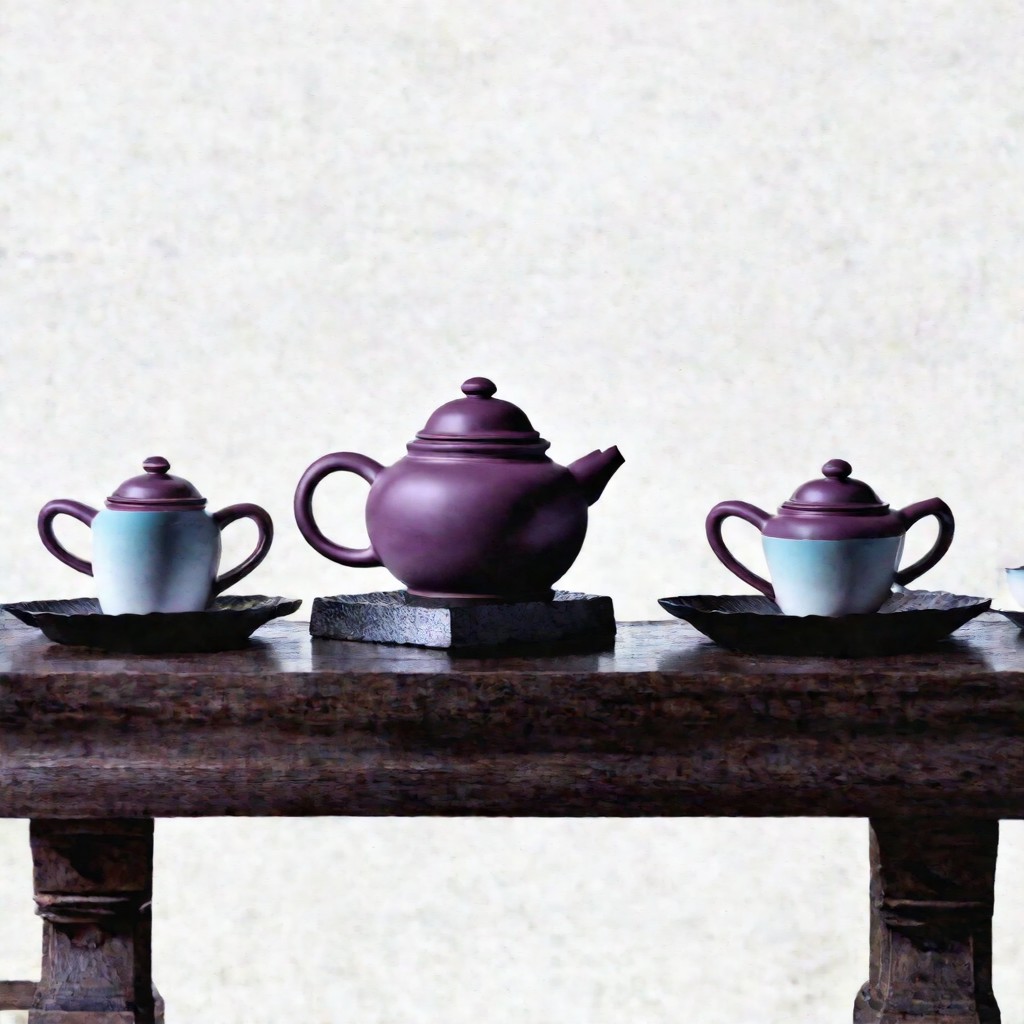}
                \\[-3pt]

                % 중간 prompt
                \multicolumn{6}{c}{%
                    \figfiveprompt{``a purple [V*]''}
                }
                \\[1pt]

                % 이미지 행 2
                \ffResult{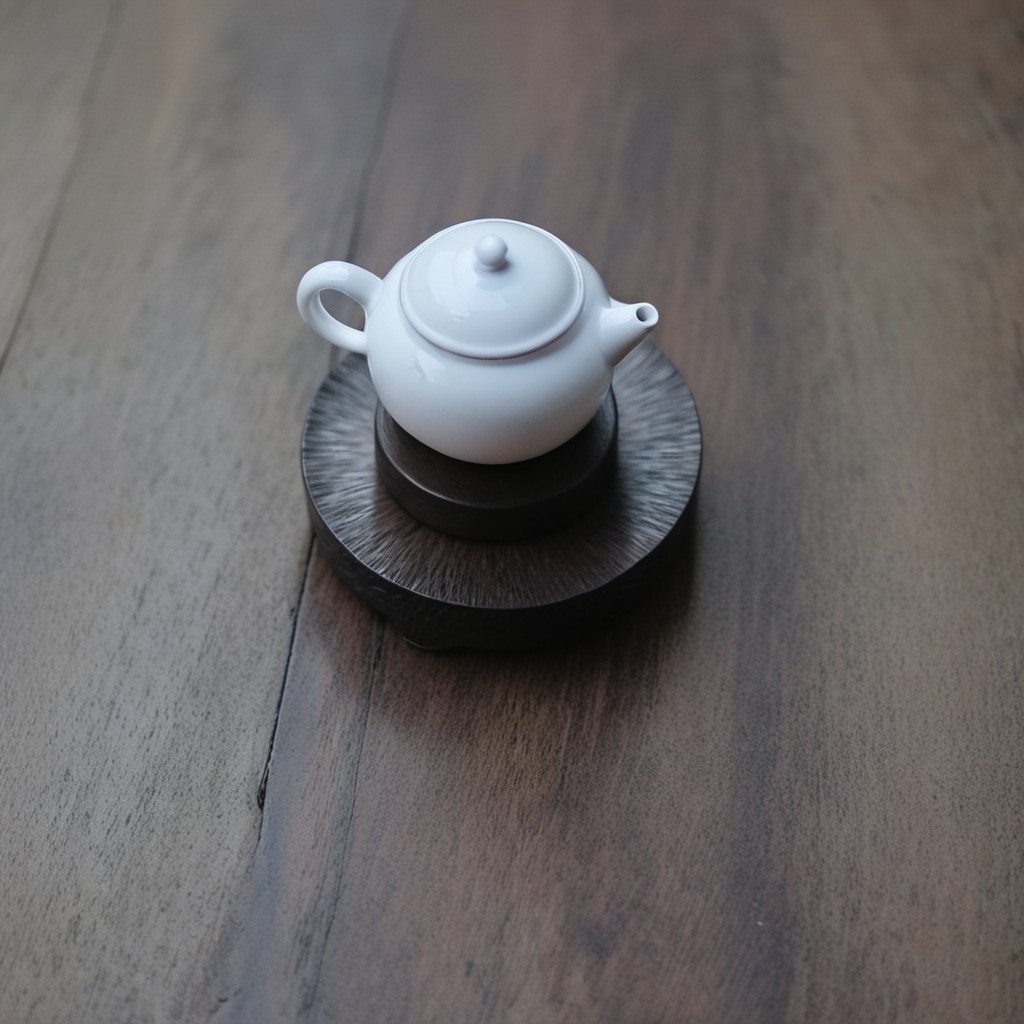}
                &
                \ffResult{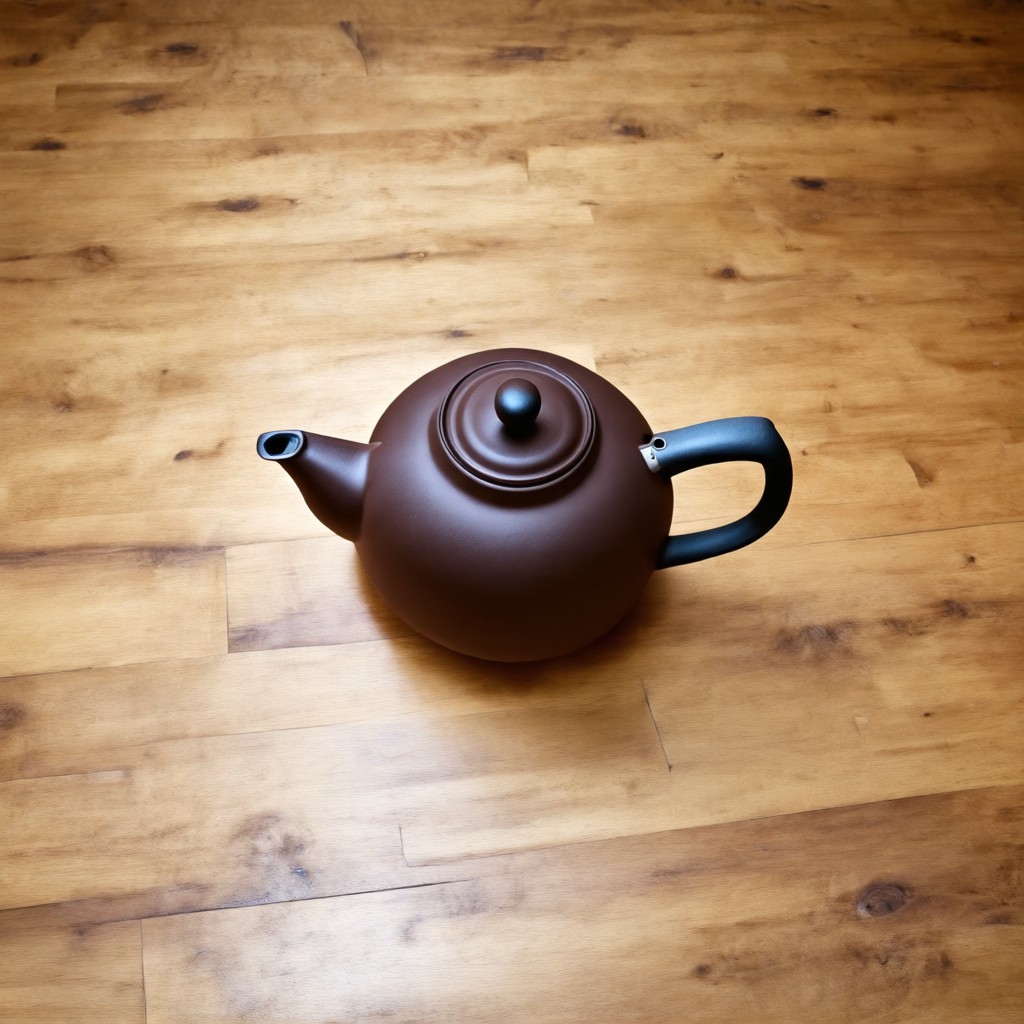}
                &
                \ffResult{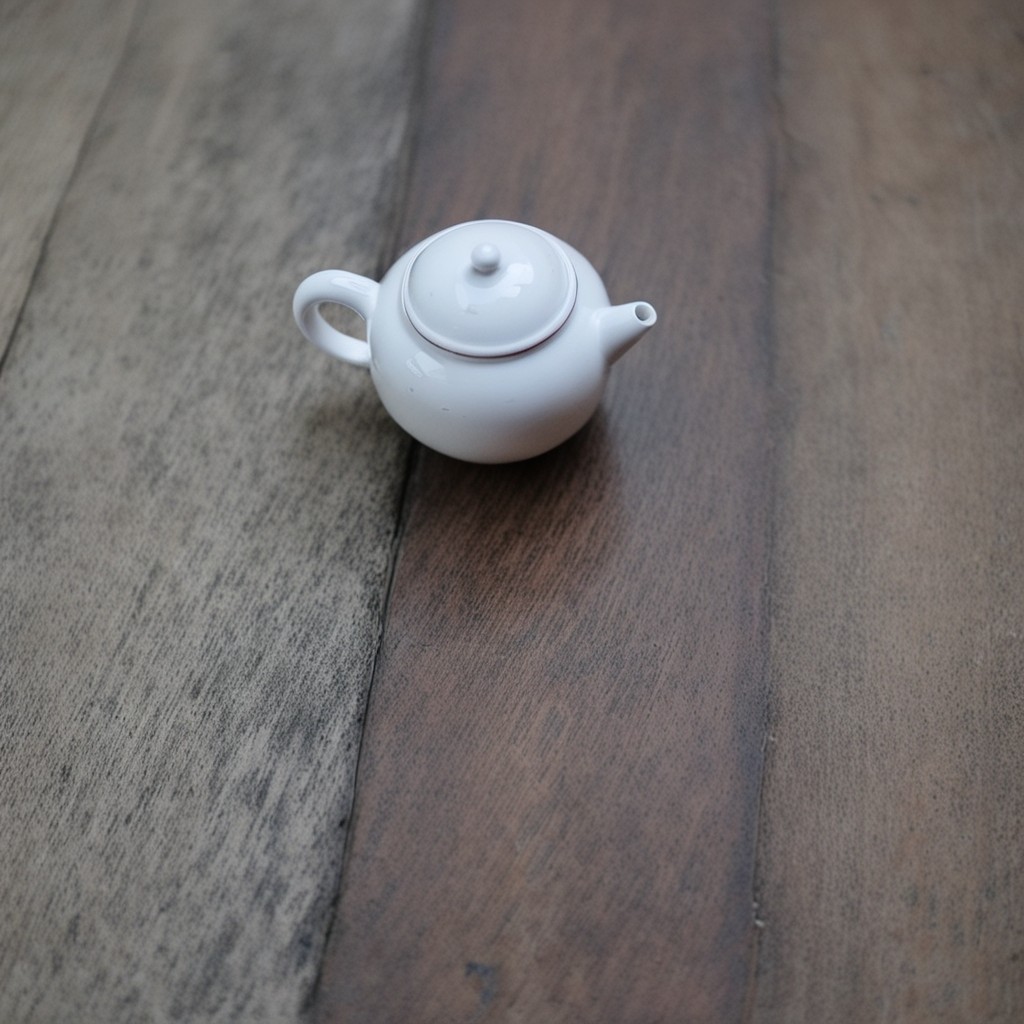}
                &
                \ffResult{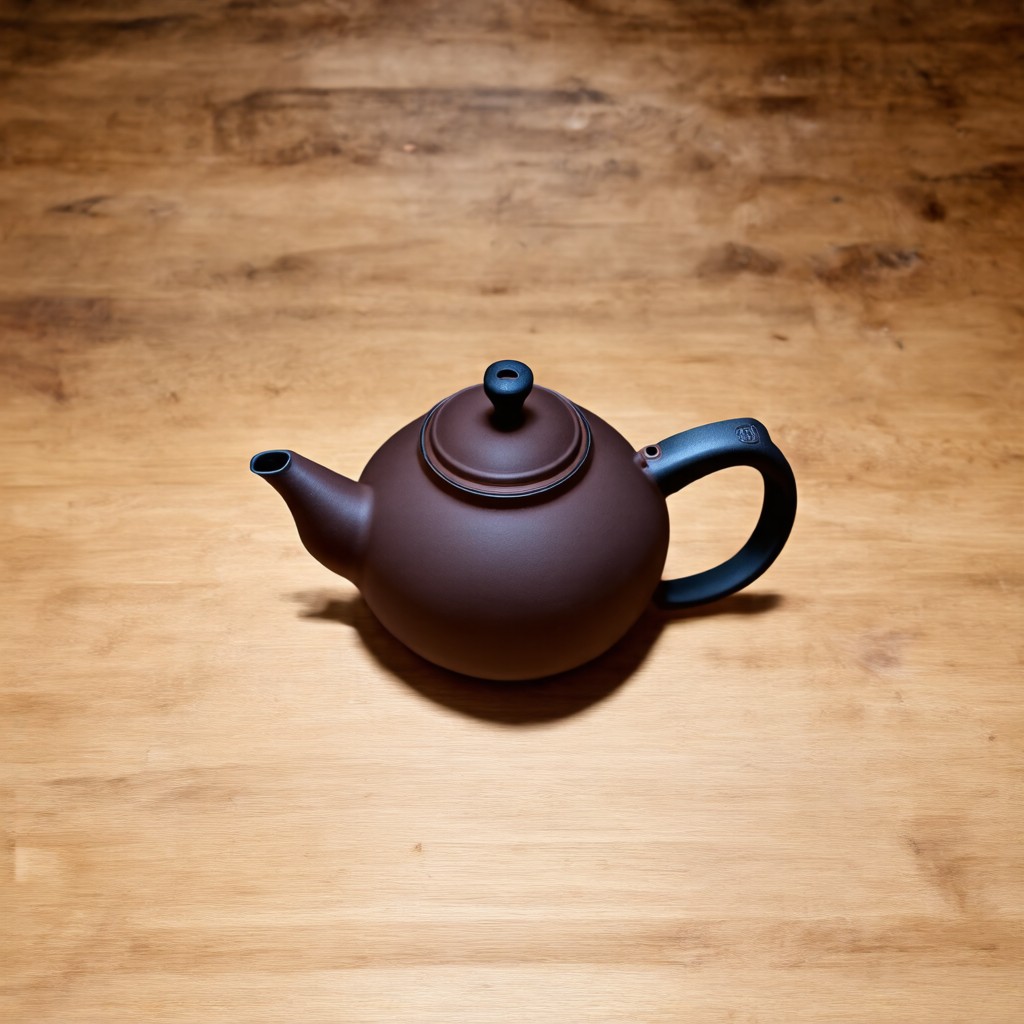}
                &
                \ffResult{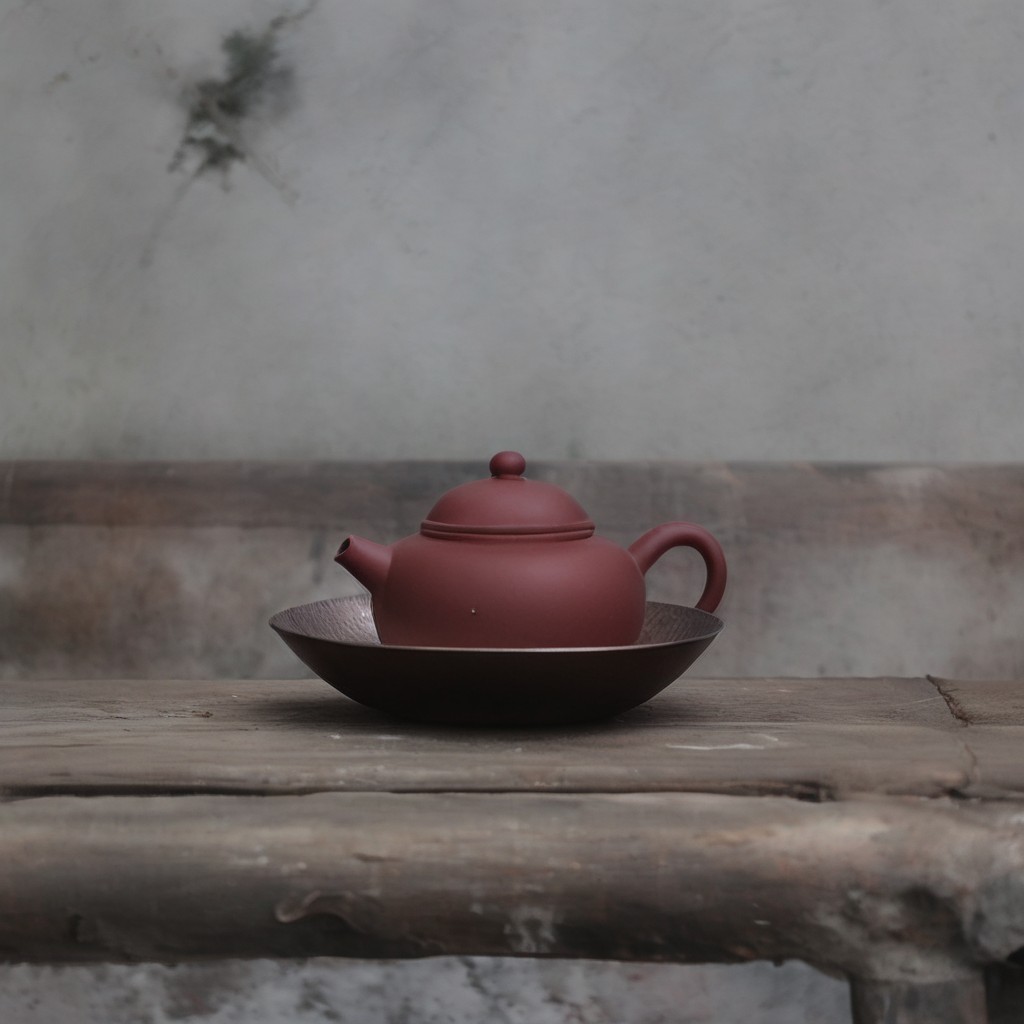}
                &
                \ffResult{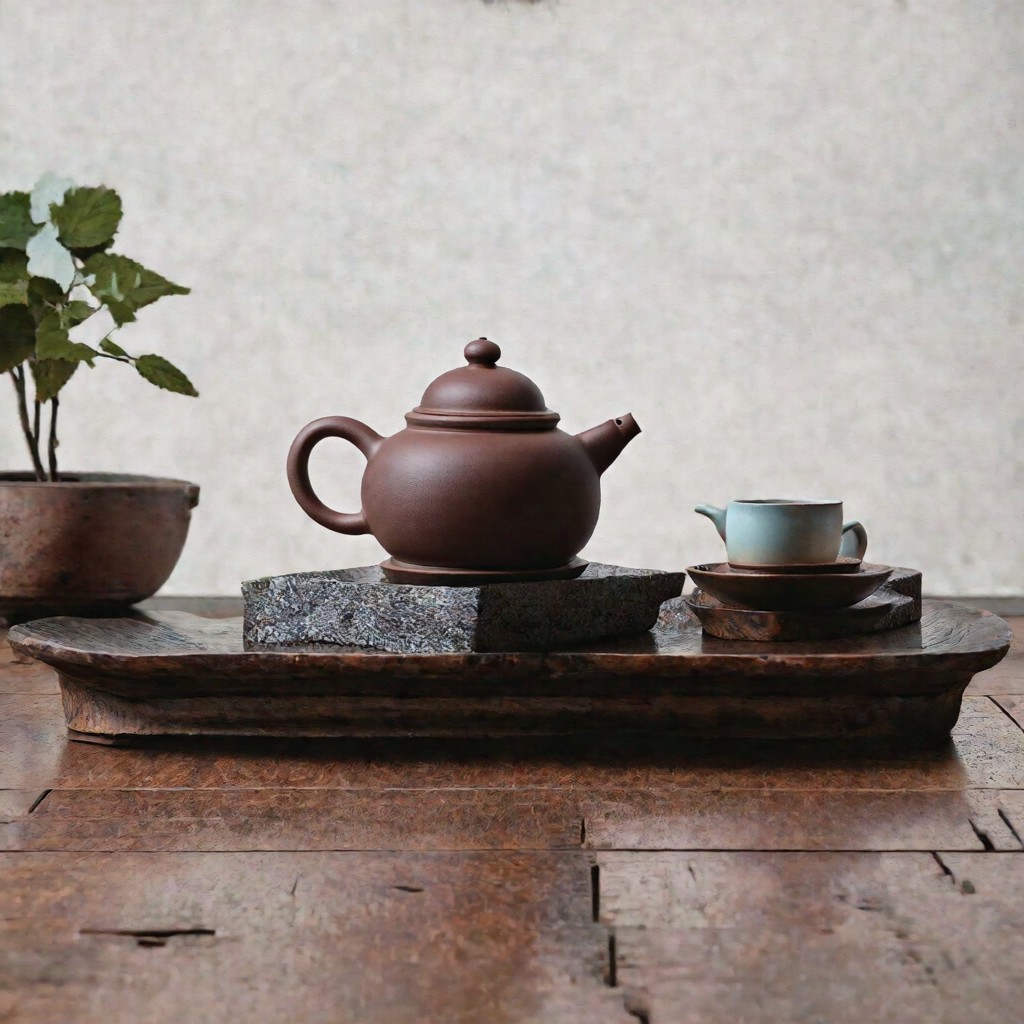}

            \end{tabular}
        \end{minipage}%
    }

    % =========================================================
    % 실제 블록 높이 측정
    % Reference 높이 = (이미지 2행 + 중간 prompt) × 0.98
    % =========================================================
    \setlength{\ffDogRefSize}{%
        \dimexpr\ht\ffDogResults+\dp\ffDogResults\relax
    }
    \setlength{\ffDogRefSize}{\ffRefScale\ffDogRefSize}

    \setlength{\ffTeapotRefSize}{%
        \dimexpr\ht\ffTeapotResults+\dp\ffTeapotResults\relax
    }
    \setlength{\ffTeapotRefSize}{\ffRefScale\ffTeapotRefSize}

    % Reference 열 너비
    \setlength{\ffRefColumnWidth}{\ffDogRefSize}
    \ifdim\ffTeapotRefSize>\ffRefColumnWidth
        \setlength{\ffRefColumnWidth}{\ffTeapotRefSize}
    \fi

    % =========================================================
    % 최종 배치
    % =========================================================
    \begin{tabular}{
        @{}
        c
        % !{\hspace{6pt}\vrule width 0.6pt\hspace{6pt}}
        !{\hspace{12.6pt}}
        c
        @{}
    }

        % -----------------------------------------------------
        % Header
        % -----------------------------------------------------
        \begin{minipage}[c]{\ffRefColumnWidth}
            \centering
            \figfiveheader{References}
        \end{minipage}
        &
        \begin{minipage}[c]{\ffResultsWidth}
            \centering
            \begin{tabular}{
                @{}
                *{6}{C{\ffResultSize}}
                @{}
            }
                \multicolumn{2}{c}{
                    \figfiveheader{DreamBooth-LoRA}
                }
                &
                \multicolumn{2}{c}{
                    \figfiveheader{Beyond Finetuning}
                }
                &
                \multicolumn{2}{c}{
                    \figfiveheader{Ours}
                }
                \\[0pt]

                \figfivesubheader{SDXL}
                &
                \figfivesubheader{SD3}
                &
                \figfivesubheader{SDXL}
                &
                \figfivesubheader{SD3}
                &
                \figfivesubheader{SDXL}
                &
                \figfivesubheader{SD3}
            \end{tabular}
        \end{minipage}
        \\[1pt]

        % -----------------------------------------------------
        % DOG: Reference와 결과 블록의 세로 중앙 정렬
        % -----------------------------------------------------
        \begin{minipage}[c]{\ffRefColumnWidth}
            \centering
            \ffFourRefs{\ffDogRefSize}
                {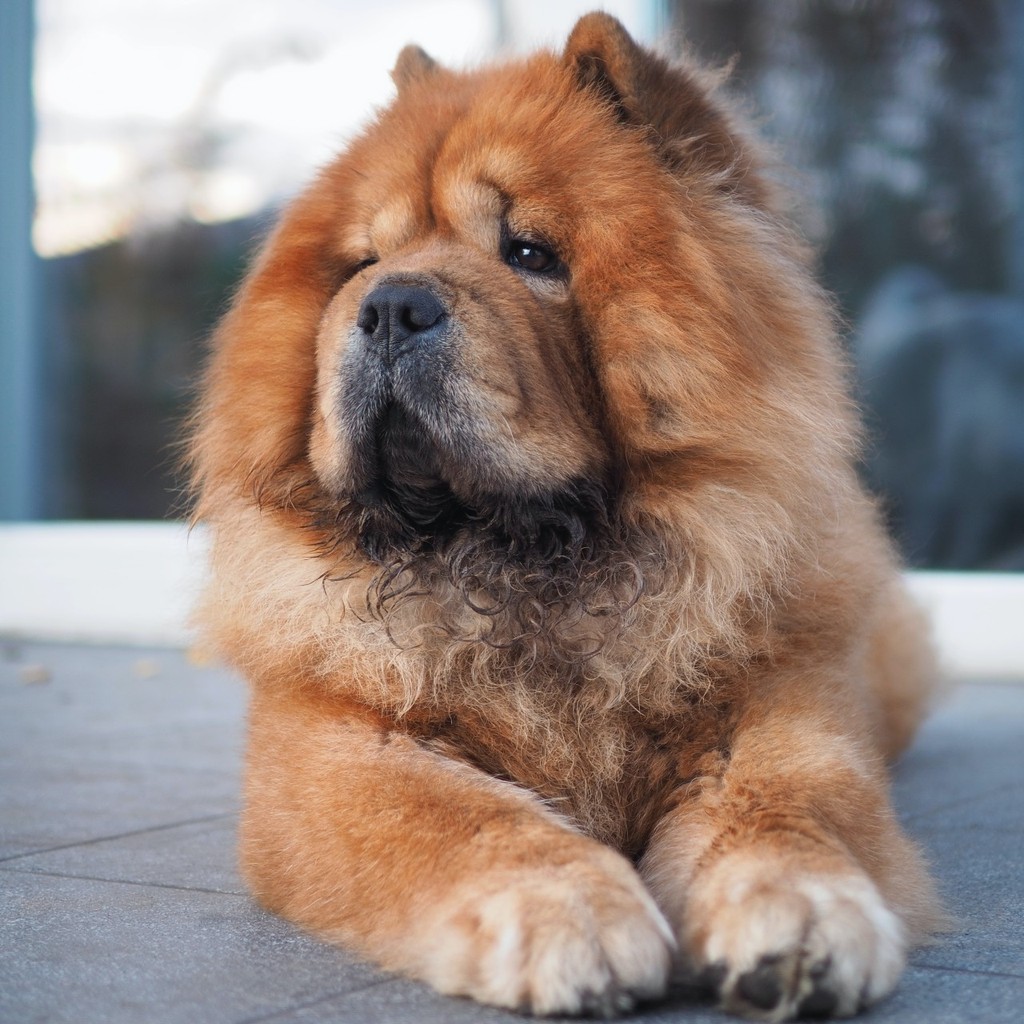}
                {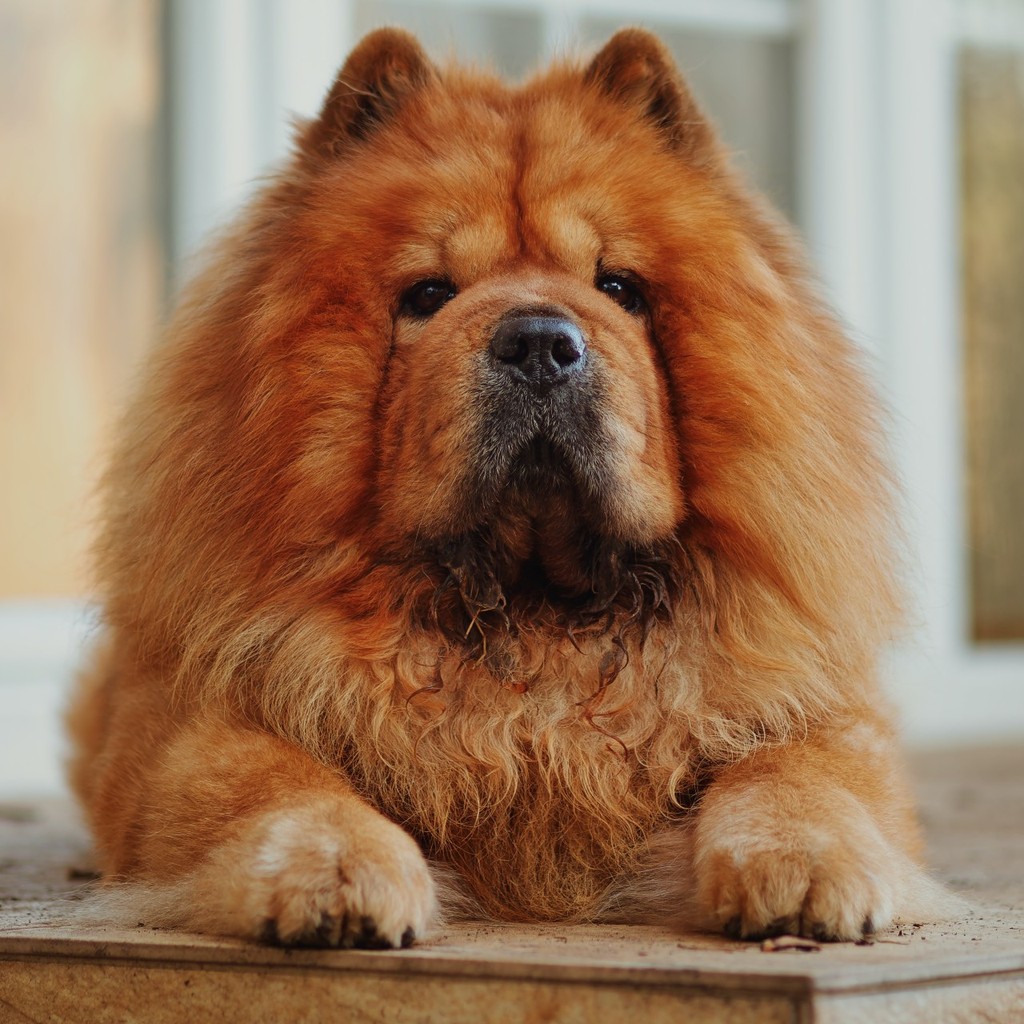}
                {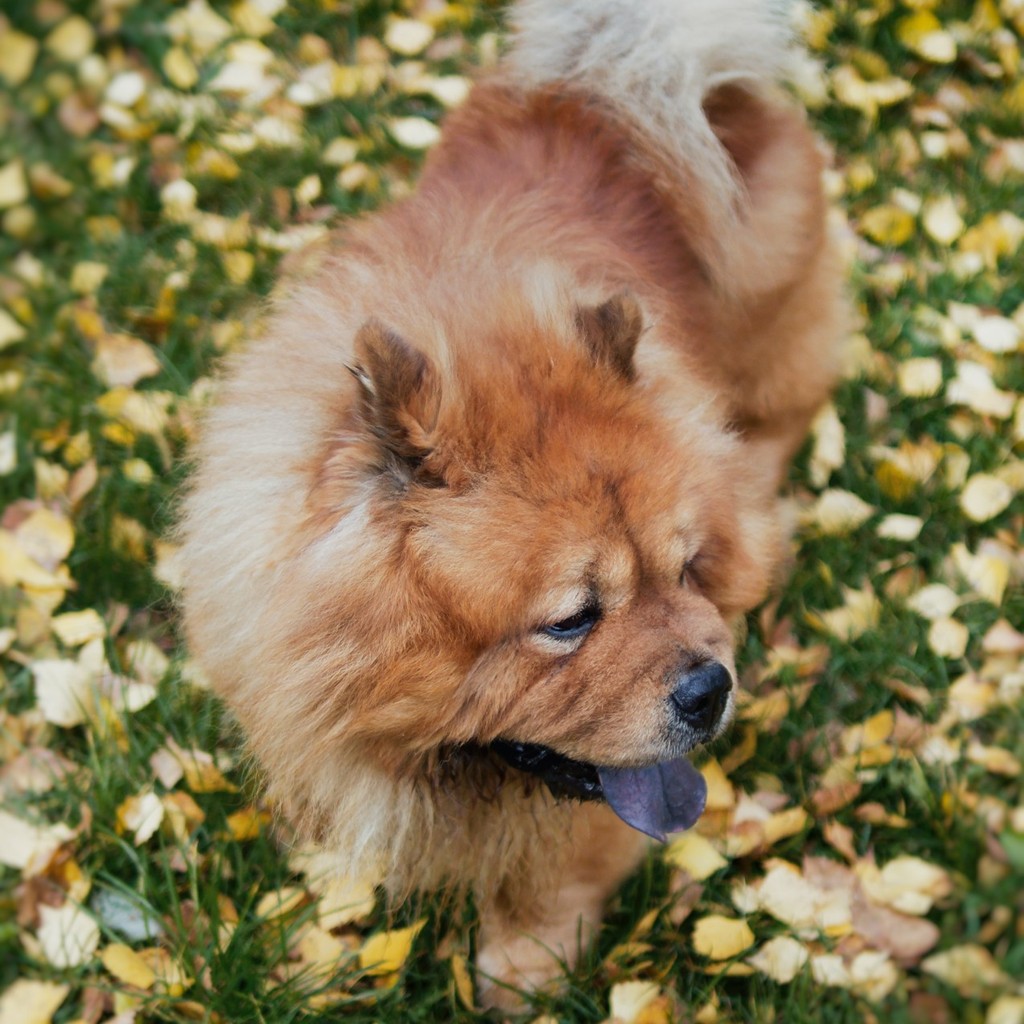}
                {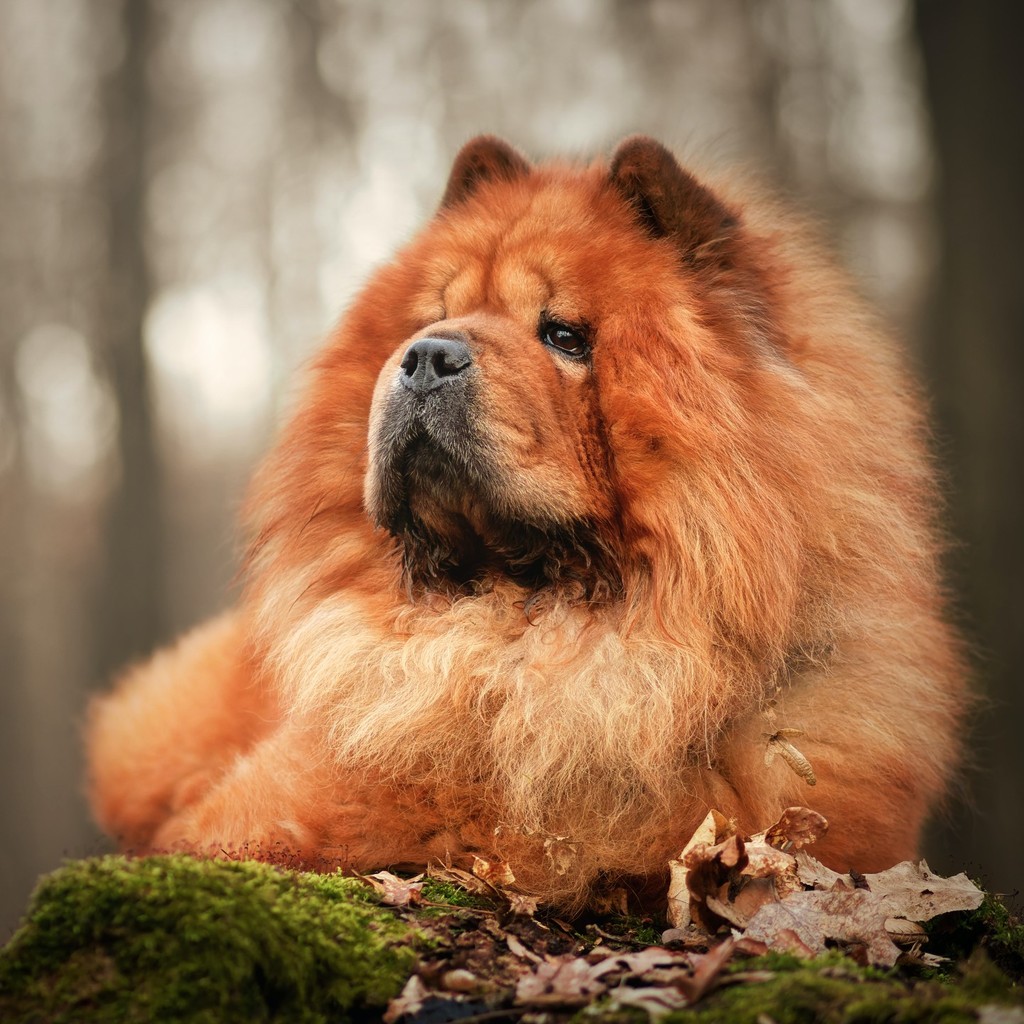}
        \end{minipage}
        &
        \usebox{\ffDogResults}
        \\[-3pt]

        % DOG 마지막 prompt: 정렬 기준에서 제외
        % DOG 마지막 prompt: 구분선 없음
        \multicolumn{2}{@{}r@{}}{%
            \makebox[\ffResultsWidth][c]{%
                \figfiveprompt{
                    ``a [V*] on top of a purple rug in a forest''
                }
            }%
        }
        \\[2pt]

        % -----------------------------------------------------
        % TEAPOT: Reference와 결과 블록의 세로 중앙 정렬
        % -----------------------------------------------------
        \begin{minipage}[c]{\ffRefColumnWidth}
            \centering
            \ffFourRefs{\ffTeapotRefSize}
                {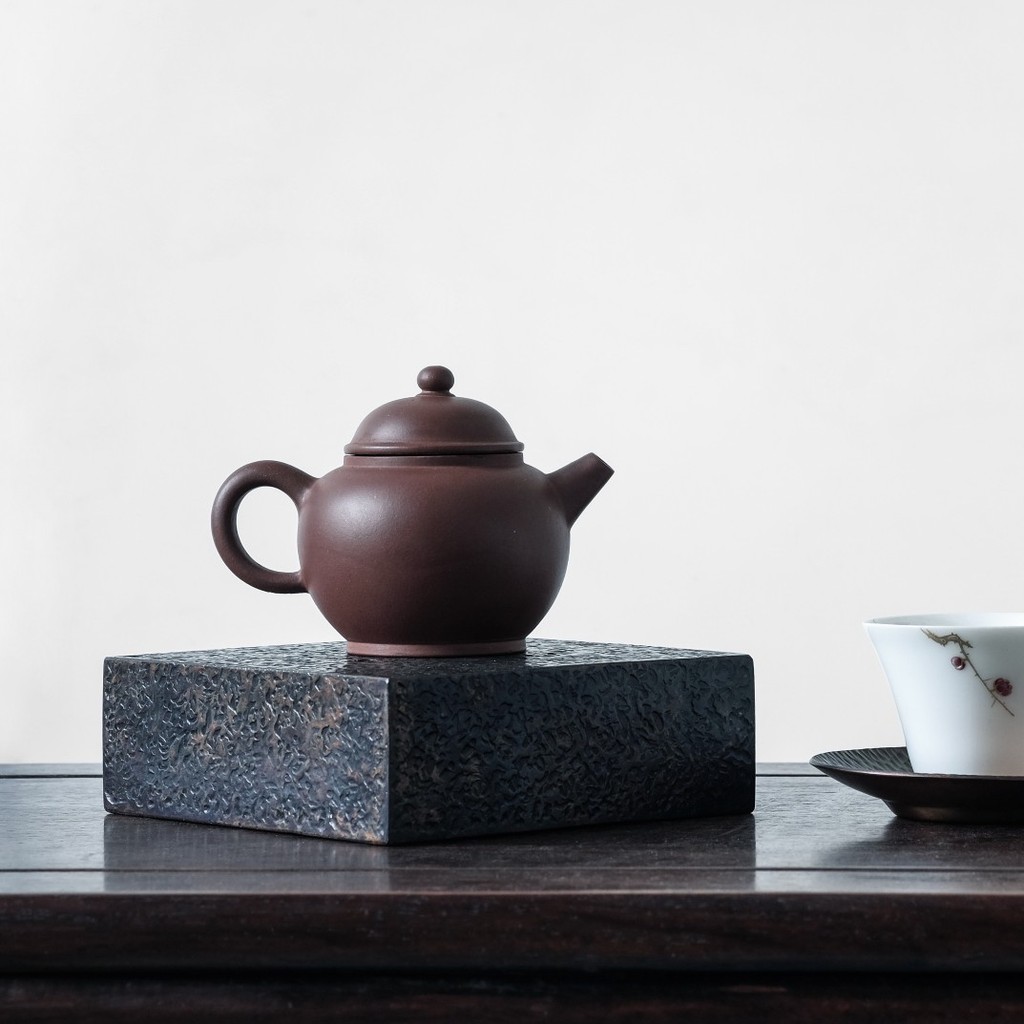}
                {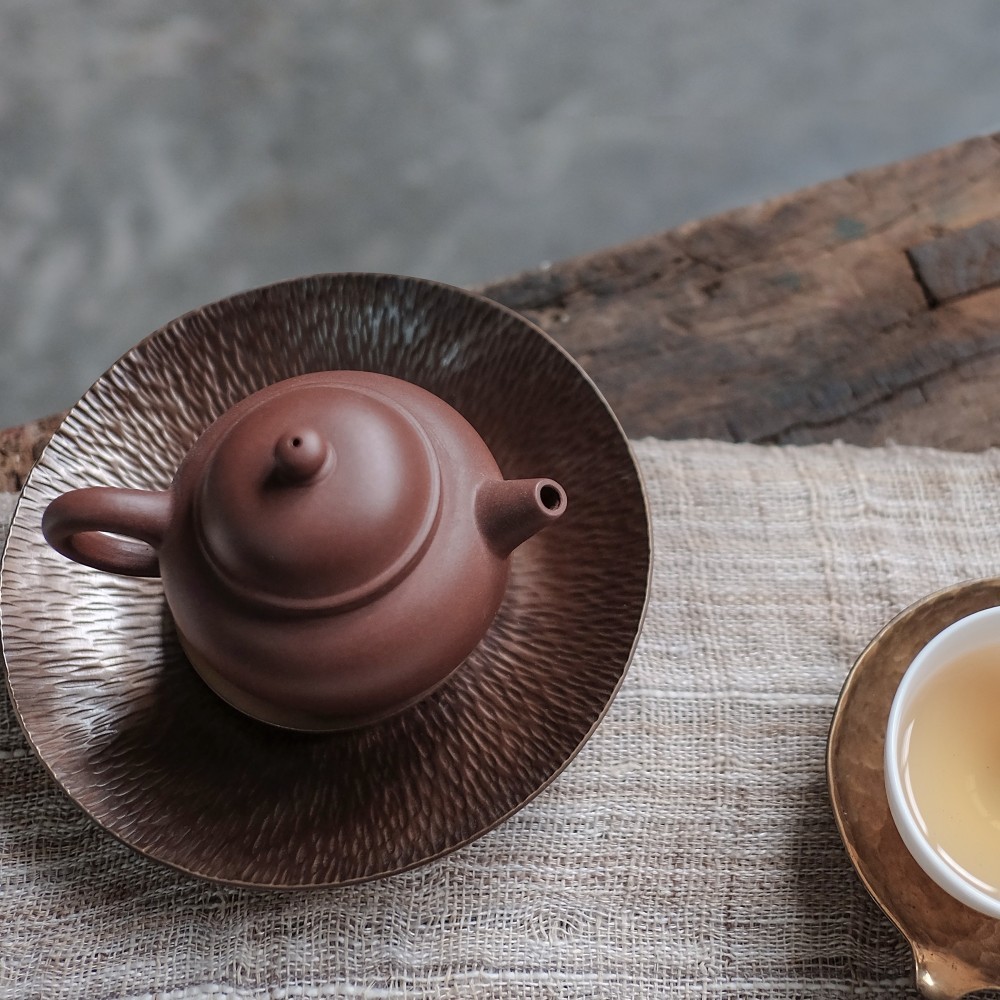}
                {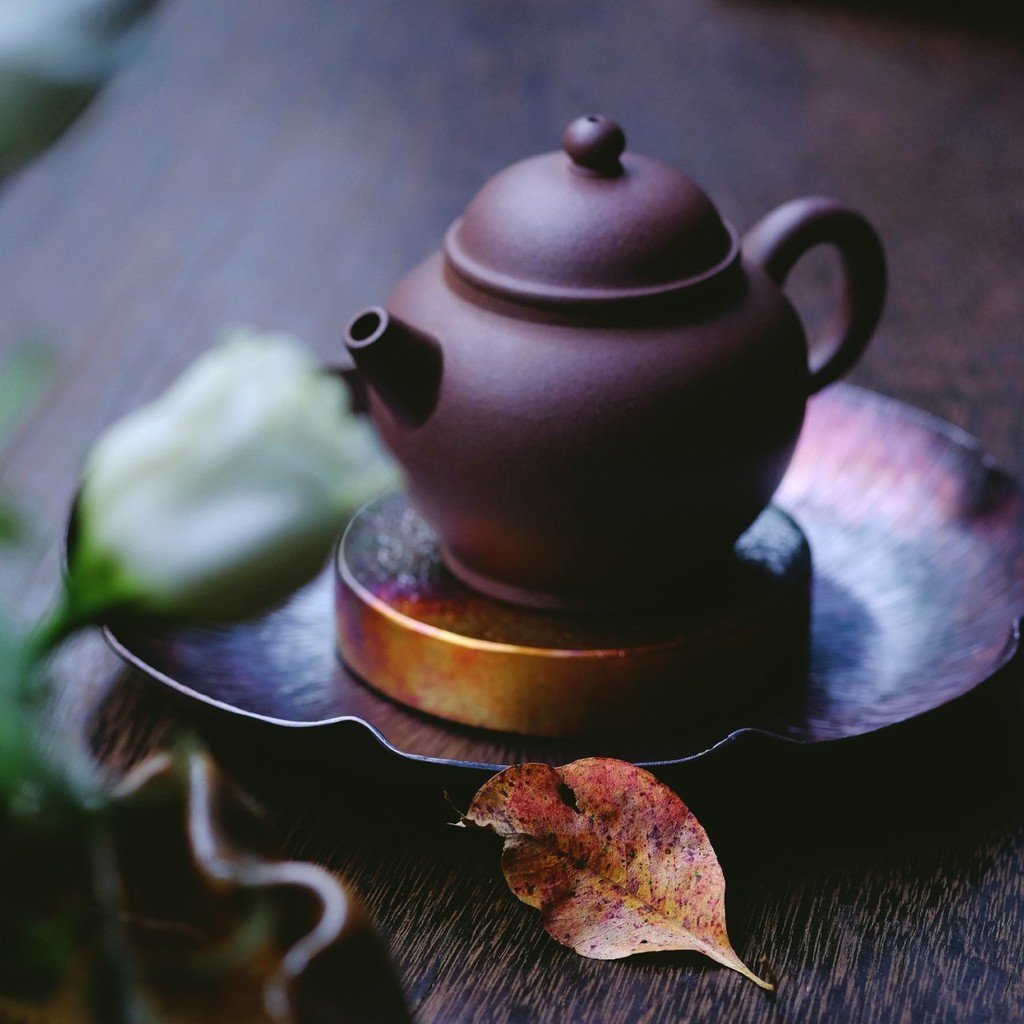}
                {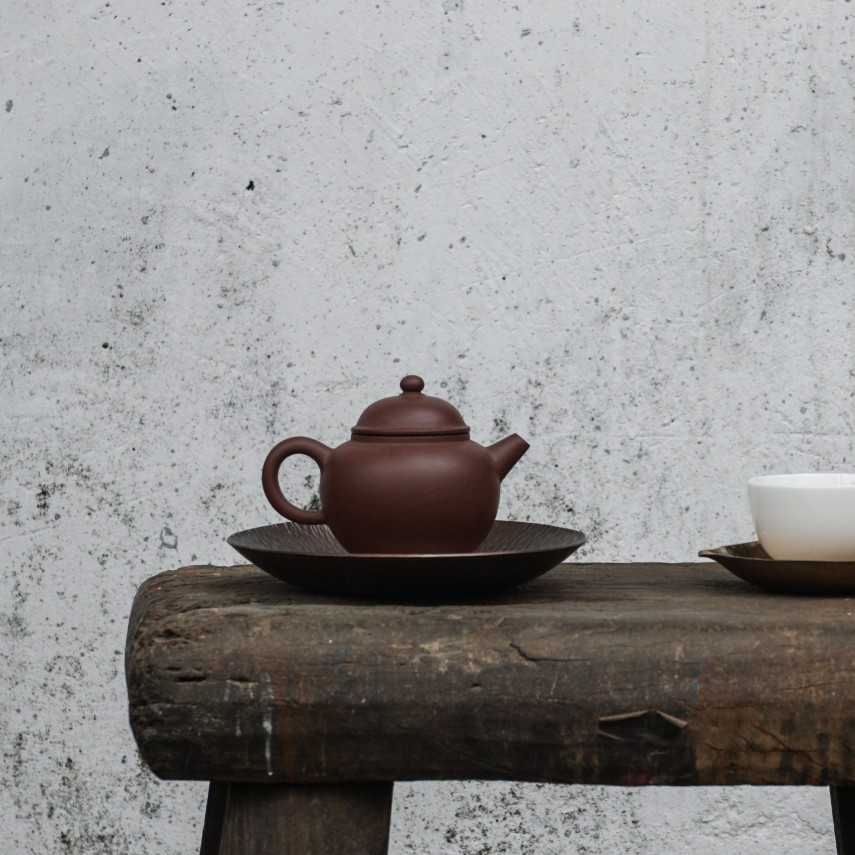}
        \end{minipage}
        &
        \usebox{\ffTeapotResults}
        \\[-3pt]

        % TEAPOT 마지막 prompt: 정렬 기준에서 제외
        % TEAPOT 마지막 prompt: 구분선 없음
        \multicolumn{2}{@{}r@{}}{%
            \makebox[\ffResultsWidth][c]{%
                \figfiveprompt{
                    ``a [V*] on top of a wooden floor''
                }
            }%
        }

    \end{tabular}

    \caption{
        Qualitative comparison on SDXL and SD3 backbones among
        DreamBooth-LoRA, Beyond-Finetuning, and our method.
    }
    \label{fig:SD3_XL_final_2}
    \vspace{-10pt}
\end{figure*}

%-------------------------------

% \input{table/main_2}

\subsection{Comparison across Backbone Models}
\label{sec:experiments_backbone}

We further evaluate SAP across SD1.5, SDXL, and SD3 backbones to examine whether its effectiveness extends beyond a single model architecture, as shown in Table~\ref{tab:main_2}. SAP attains the top rank across all three backbones, based on the same ranking procedure used in Table~\ref{tab:main_1}. On SDXL, SAP achieves the highest scores across all metrics. On SD1.5 and SD3, SAP yields higher subject fidelity while maintaining comparable text alignment with DreamBooth-LoRA and Beyond-Finetuning. The gains over DreamBooth-LoRA demonstrate the benefit of the proposed training objective across different backbones. The comparison with Beyond-Finetuning further suggests that incorporating class-level guidance into the training objective provides benefits beyond applying it only at inference time.

Qualitative results further support these findings as illustrated in Fig.~\ref{fig:SD3_XL_final_2}. For the dog subject, DreamBooth-LoRA and Beyond-Finetuning produce faithful personalized results on SDXL, but fail to preserve subject identity on SD3. For the teapot subject, both methods show the opposite trend, producing appropriate results on SD3 but either failing to reflect the target prompt or losing subject identity on SDXL. In contrast, SAP preserves distinctive subject details such as fur color, snout shape, and posture, resulting in the highest CLIP-I and DINO scores. It also captures contextual cues such as ``\texttt{on top of a purple rug}'' and ``\texttt{in a forest}''. A similar pattern is observed in the teapot examples (third and fourth rows). These results indicate that semantic anchoring yields more consistent personalization across diffusion backbones.

\begin{figure}[t]
    \centering

    % Figure 5
    \begin{minipage}[t]{0.45\linewidth}
        \vspace{0pt}
        \centering
        \includegraphics[
            width=\linewidth,
            height=0.55\linewidth
        ]{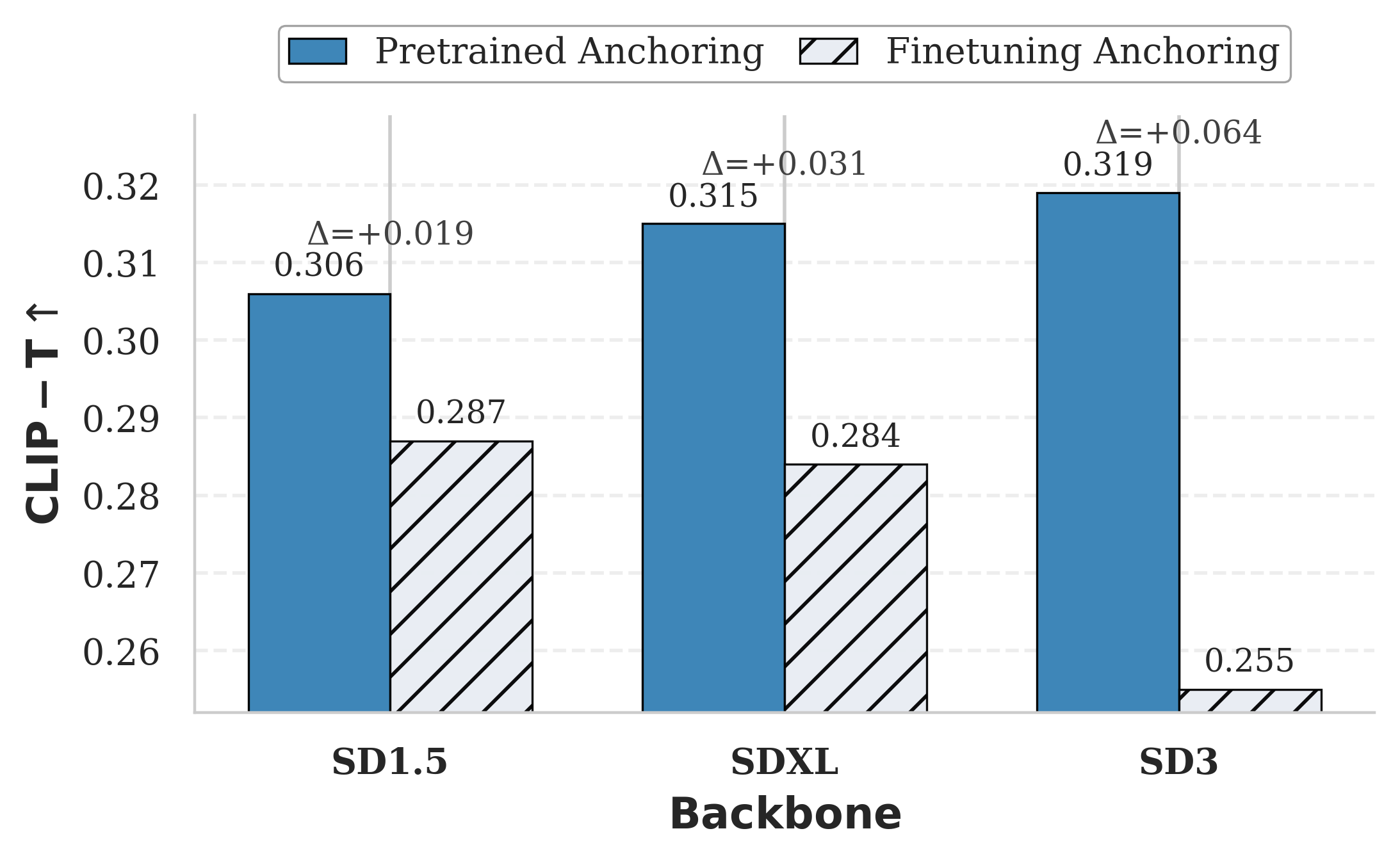}
        %\vspace{-2.00em}
        \caption{
            Comparison of semantic alignment (CLIP-T) using two different anchoring strategies. Pretrained anchoring outperforms finetuning anchoring.
        }
        \label{fig:clip_t_pre_fine}
    \end{minipage}
    \hfill
    % Figure 6
    \begin{minipage}[t]{0.53\linewidth}
        \vspace{-1mm}
        \centering
        \centering
\setlength{\tabcolsep}{0pt}

% Reference 한 장 크기
% 2×2 배열 전체 너비 = 0.22\linewidth
\newlength{\anchorRefTileSize}
\setlength{\anchorRefTileSize}{0.11\linewidth}

% Reference 4장: 위 ref1·ref2 / 아래 ref3·ref4
\newcommand{\anchorrefsFour}[4]{%
    \vbox{%
        \offinterlineskip
        \hbox{%
            \includegraphics[
                width=\anchorRefTileSize,
                height=\anchorRefTileSize
            ]{#1}%
            \includegraphics[
                width=\anchorRefTileSize,
                height=\anchorRefTileSize
            ]{#2}%
        }%
        \hbox{%
            \includegraphics[
                width=\anchorRefTileSize,
                height=\anchorRefTileSize
            ]{#3}%
            \includegraphics[
                width=\anchorRefTileSize,
                height=\anchorRefTileSize
            ]{#4}%
        }%
    }%
}

\begin{tabular}{
    @{}
    C{2\anchorRefTileSize}
    !{\hspace{5.5pt}}
    C{0.14\linewidth}
    @{\hspace{2pt}}
    *{3}{C{0.19\linewidth}}
    @{}
}

% =========================================================
% Header
% =========================================================
\anchorheader{References}
&
{}
&
\anchorheader{SD1.5}
&
\anchorheader{SDXL}
&
\anchorheader{SD3}
\\[1pt]

% =========================================================
% Finetuning Anchoring
% =========================================================
\multirow[c]{2}{*}{%
    \anchorrefsFour
        {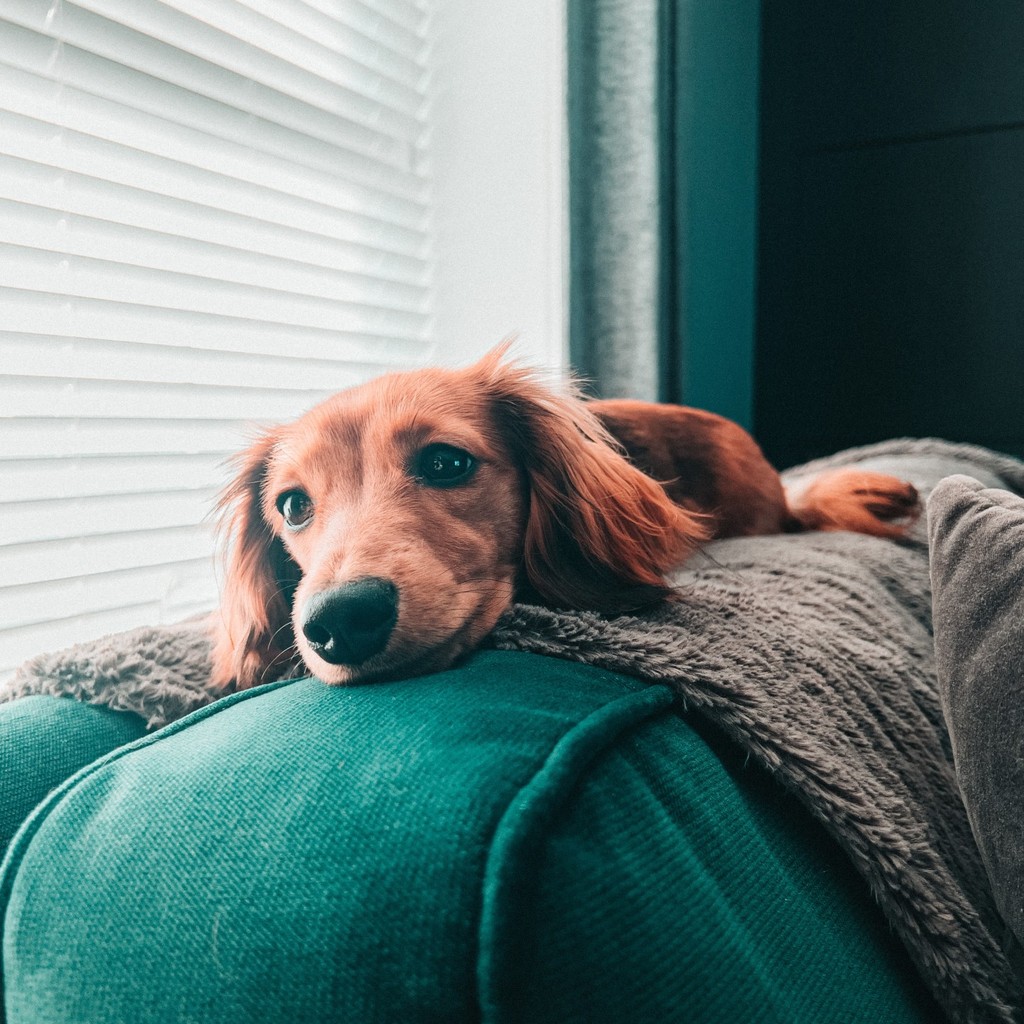}
        {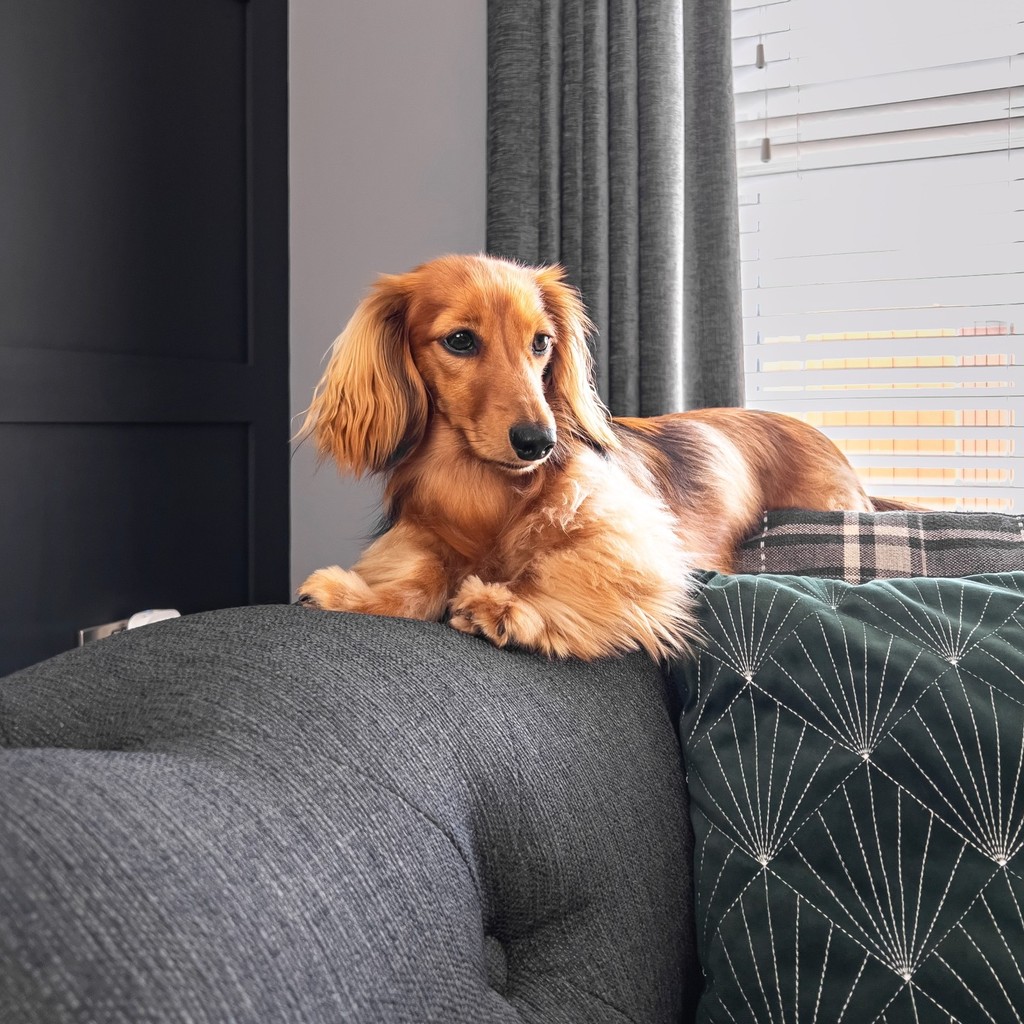}
        {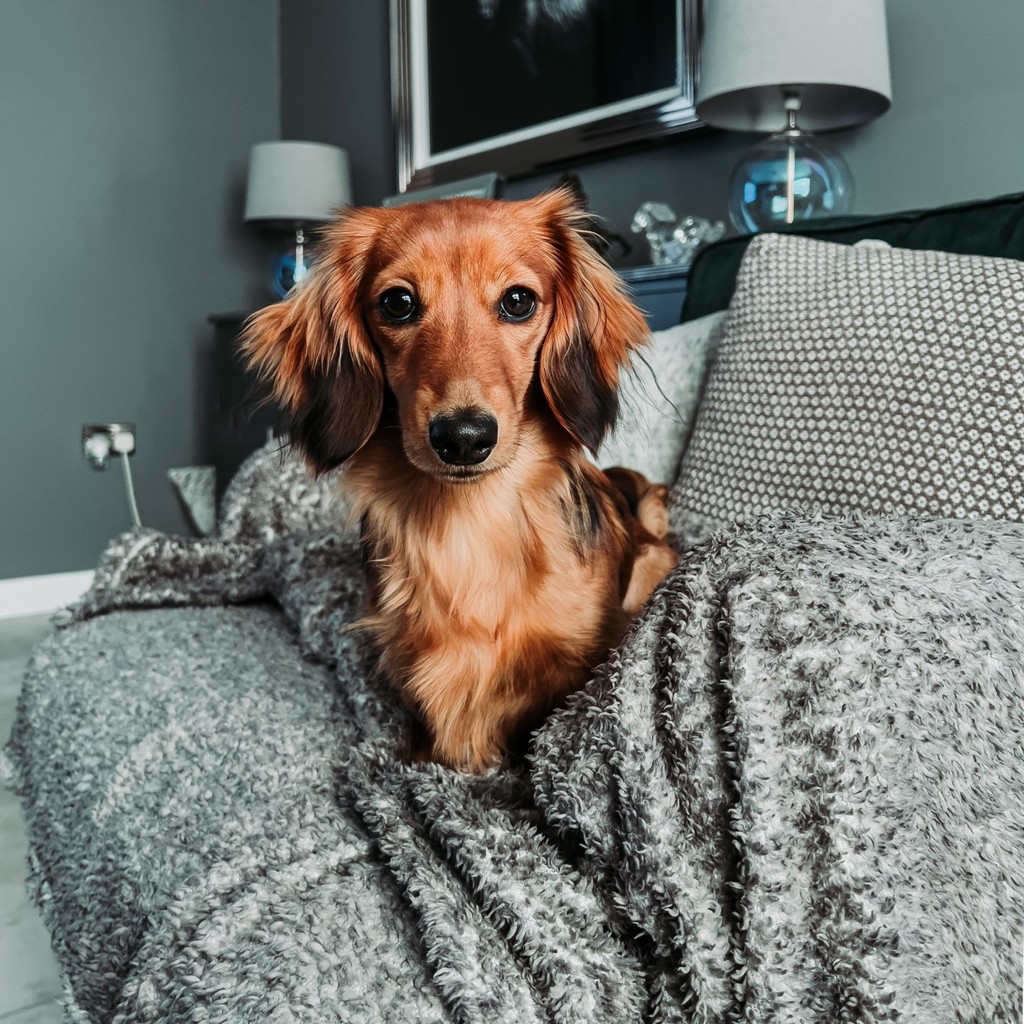}
        {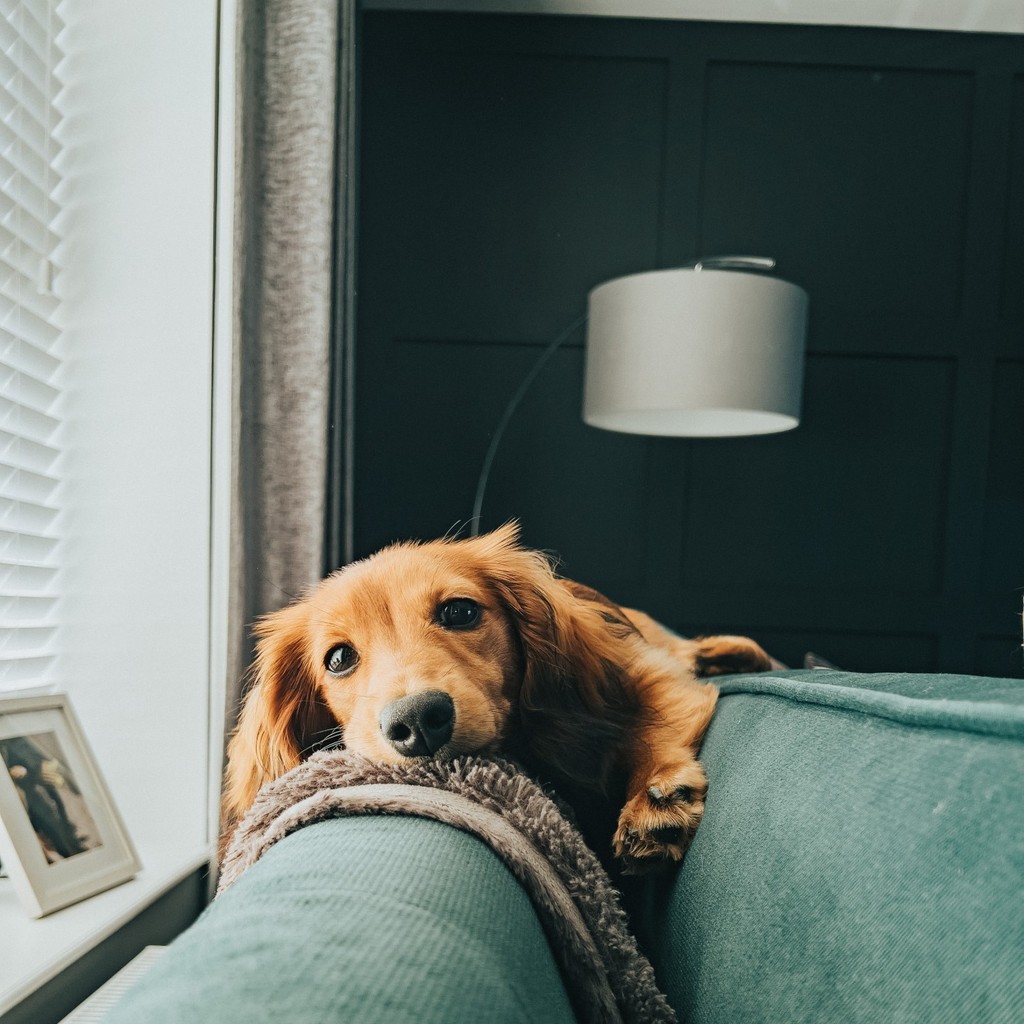}
}
&
\anchorlabel{Finetuning\\Anchoring}
&
\anchorimg{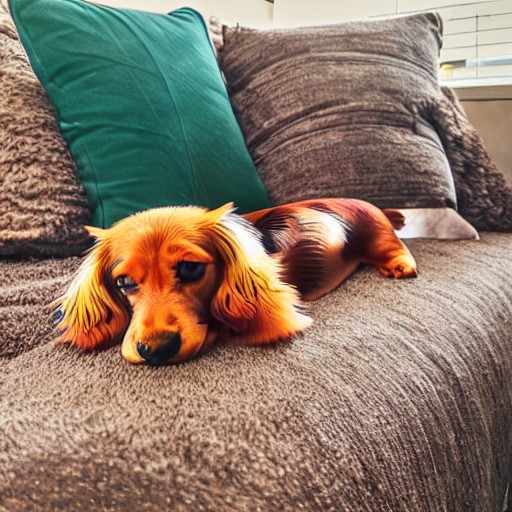}
&
\anchorimg{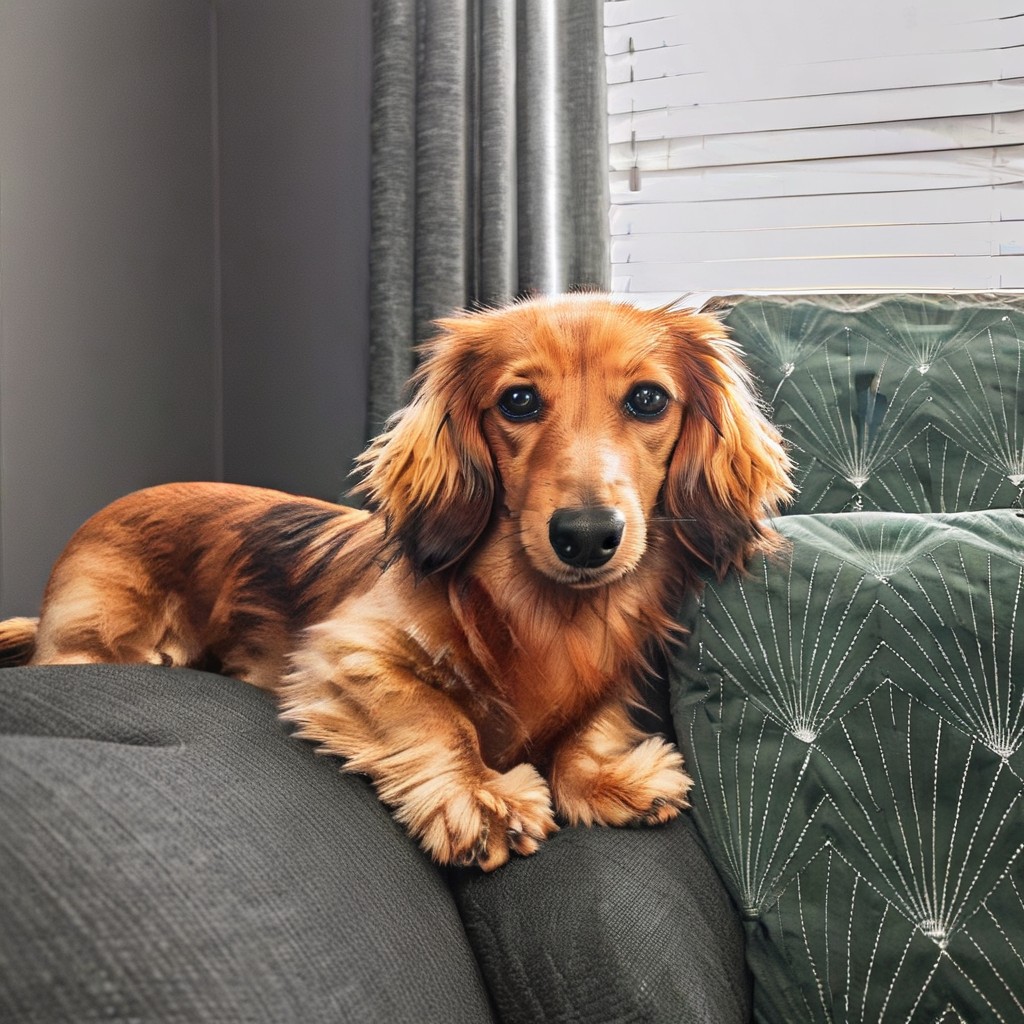}
&
\anchorimg{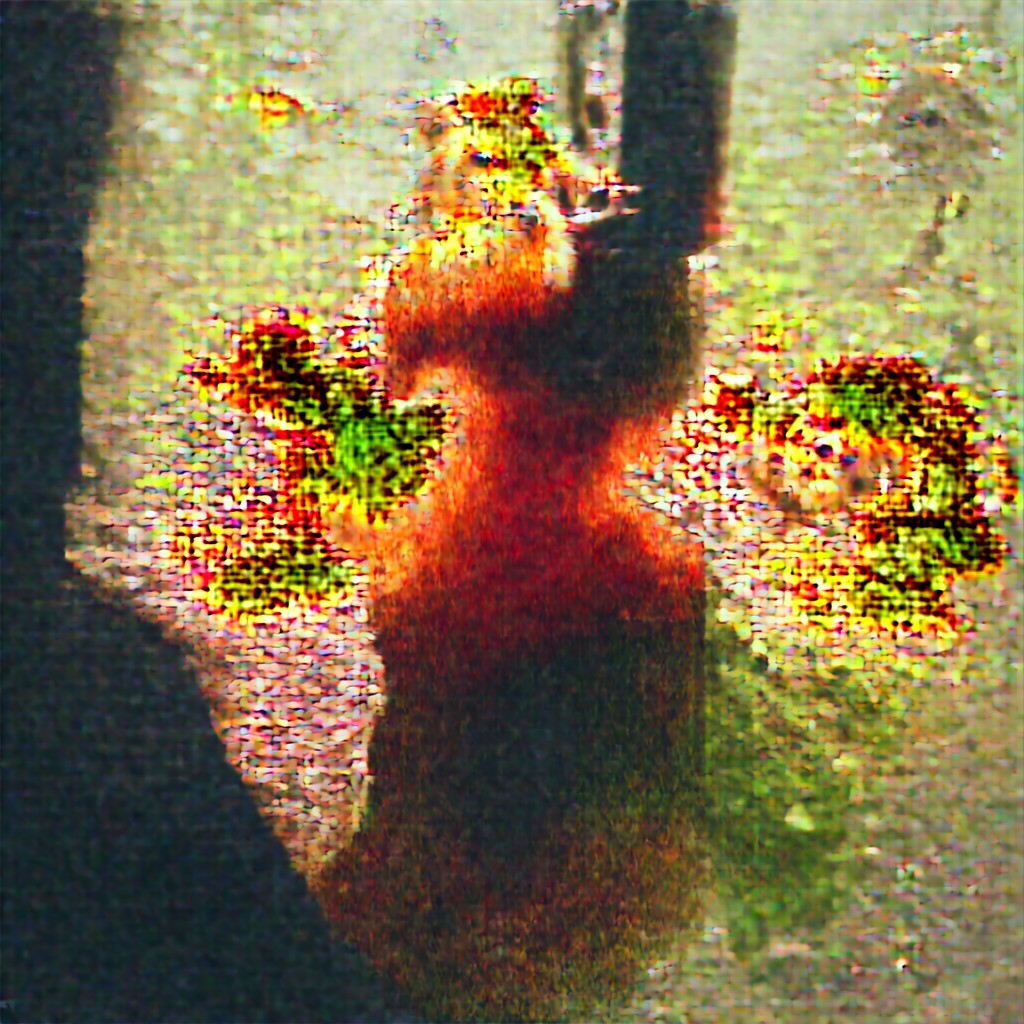}
\\[-2pt]

% =========================================================
% Pretrained Anchoring
% =========================================================
{}
&
\anchorlabel{Pretrained\\Anchoring}
&
\anchorimg{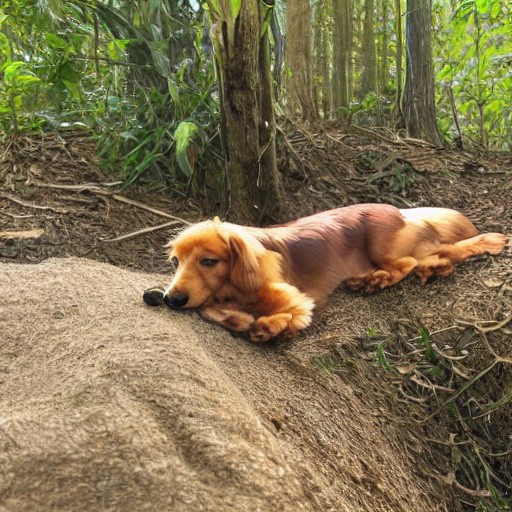}
&
\anchorimg{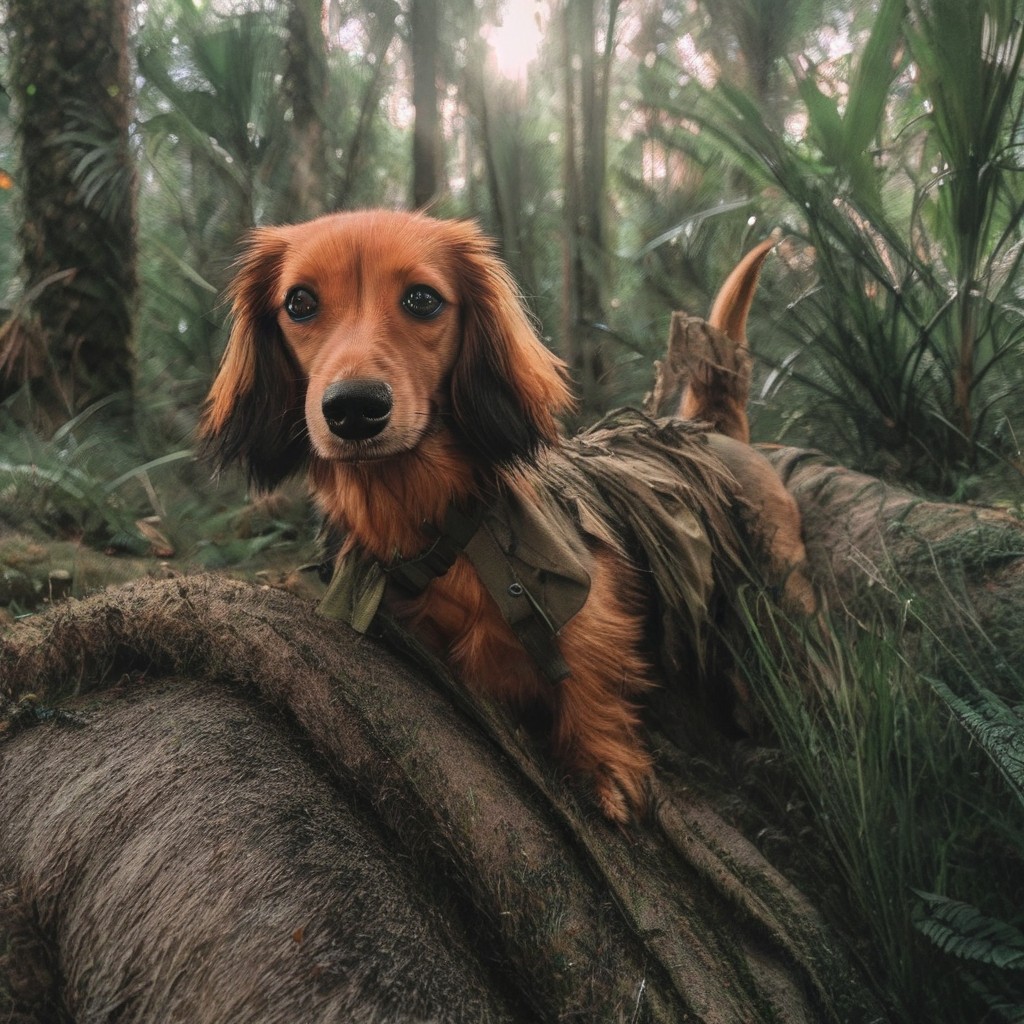}
&
\anchorimg{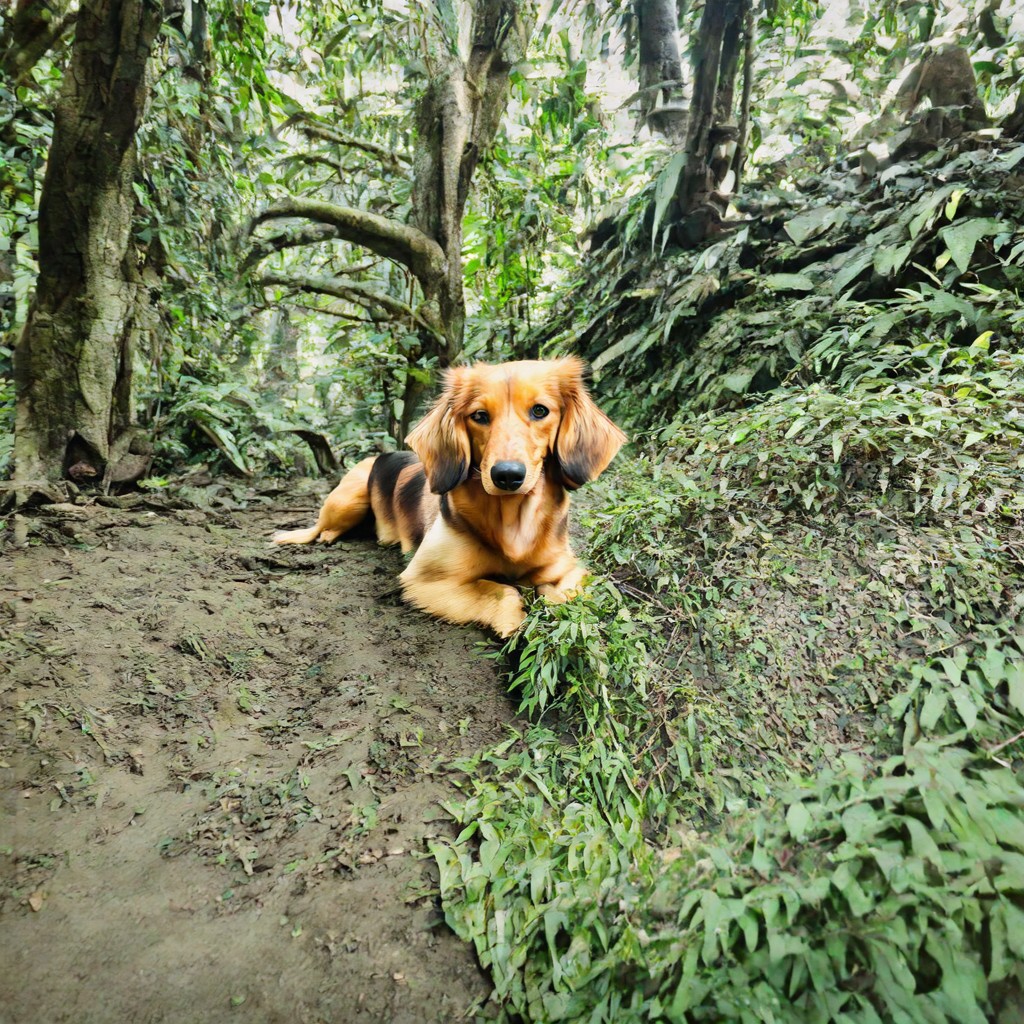}
\\[-3pt]

% =========================================================
% Prompt
% =========================================================
{}
&
\multicolumn{4}{c}{%
    {\fontsize{5.5}{6.0}\selectfont\itshape
    ``a [V*] in the jungle''}
}

\end{tabular}

\vspace{-0.5em}

\caption{
    Comparison of qualitative results using two different
    anchoring strategies. Both preserve subject fidelity,
    and the pretrained anchoring shows better consistency
    to the text description.
}
\label{fig:anchoring_compare}
    \end{minipage}

\end{figure}
%-----------------------------------------
%\vspace{-5pt}
\section{Ablation Studies}
\label{sec:5_ablation}
%\vspace{-5pt}
In this section, we conduct ablation studies to analyze the contribution of each component in semantic anchoring and its impact on the overall personalization process. We also provide additional ablation studies in Appendix~\ref{sec:supp_add_ablation_study}.

%We first examine how varying the anchoring strength influences performance, providing insight into the role of the anchoring term. Next, we investigate whether the anchor should be fixed or jointly finetuned. Furthermore, we explore the semantic relatedness of anchors, examining how the choice of the anchor prompt affects guidance quality. Finally, we analyze the personalization trajectory to understand how the proposed objective shapes the semantic space dynamics and its resulting implications. Additional ablation studies are provided in Section~\ref{sec:supp_add_ablation_study}.

%-------------------------------------

\begin{figure*}[t]
  \centering
  \begin{minipage}[t]{0.66\textwidth}
    \centering
    \begin{minipage}[t]{0.49\linewidth}
      \centering
      \includegraphics[width=\linewidth]{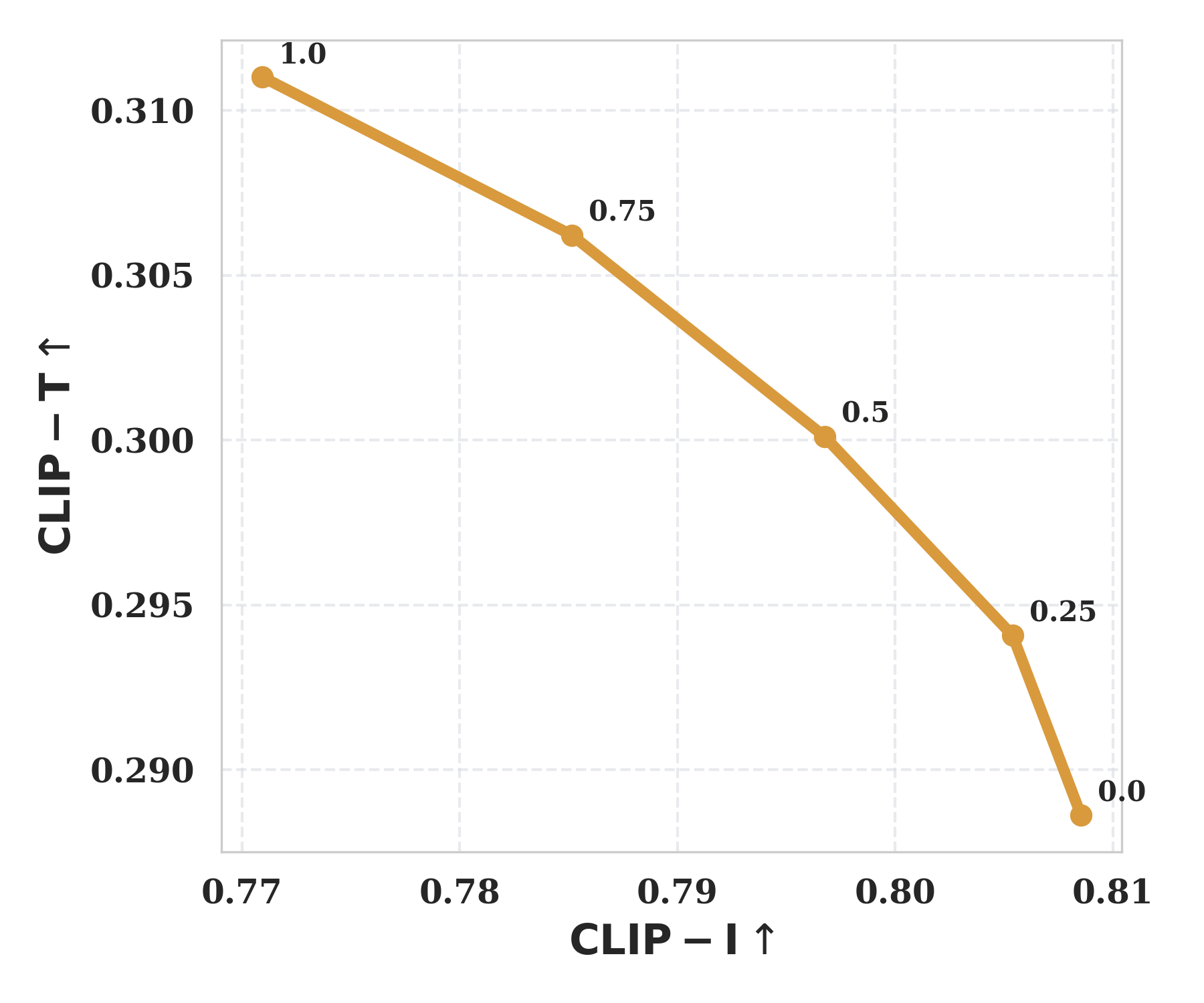}
      \label{fig:clipi_vs_clipt_lines}
    \end{minipage}\hfill
    \begin{minipage}[t]{0.49\linewidth}
      \centering
      \includegraphics[width=\linewidth]{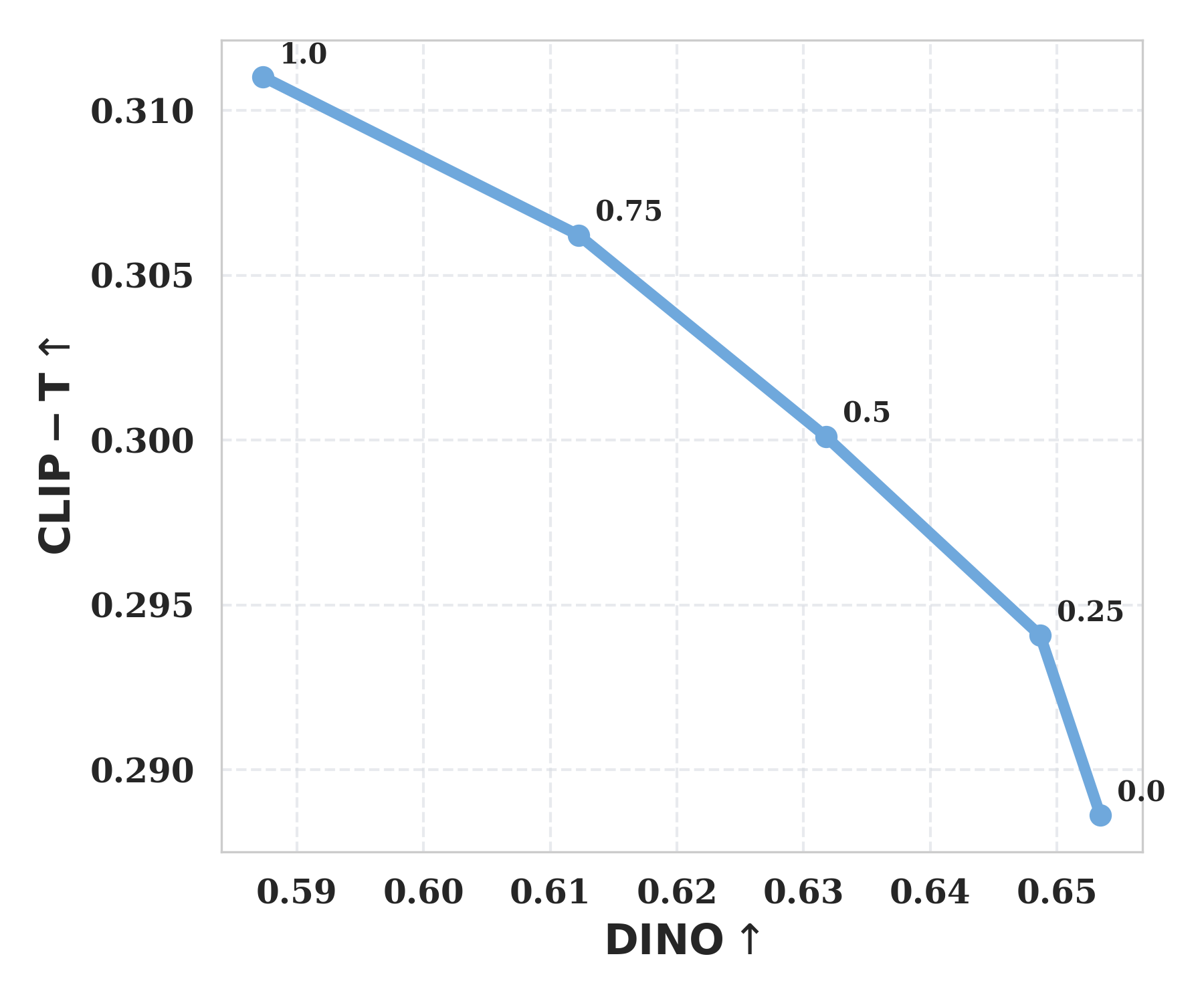}
      \label{fig:dino_vs_clipt_lines}
    \end{minipage}
    \vspace{-10pt}
    %\vspace{-0.70em}
    \captionof{figure}{Analysis of subject fidelity and text alignment under different weighting parameters $w$. Left: relationship between CLIP-I and CLIP-T. Right: relationship between DINO and CLIP-T.}
    \label{fig:side_by_side}
  \end{minipage}\hfill
  \begin{minipage}[t]{0.32\textwidth}
    \centering
      \includegraphics[width=\linewidth]{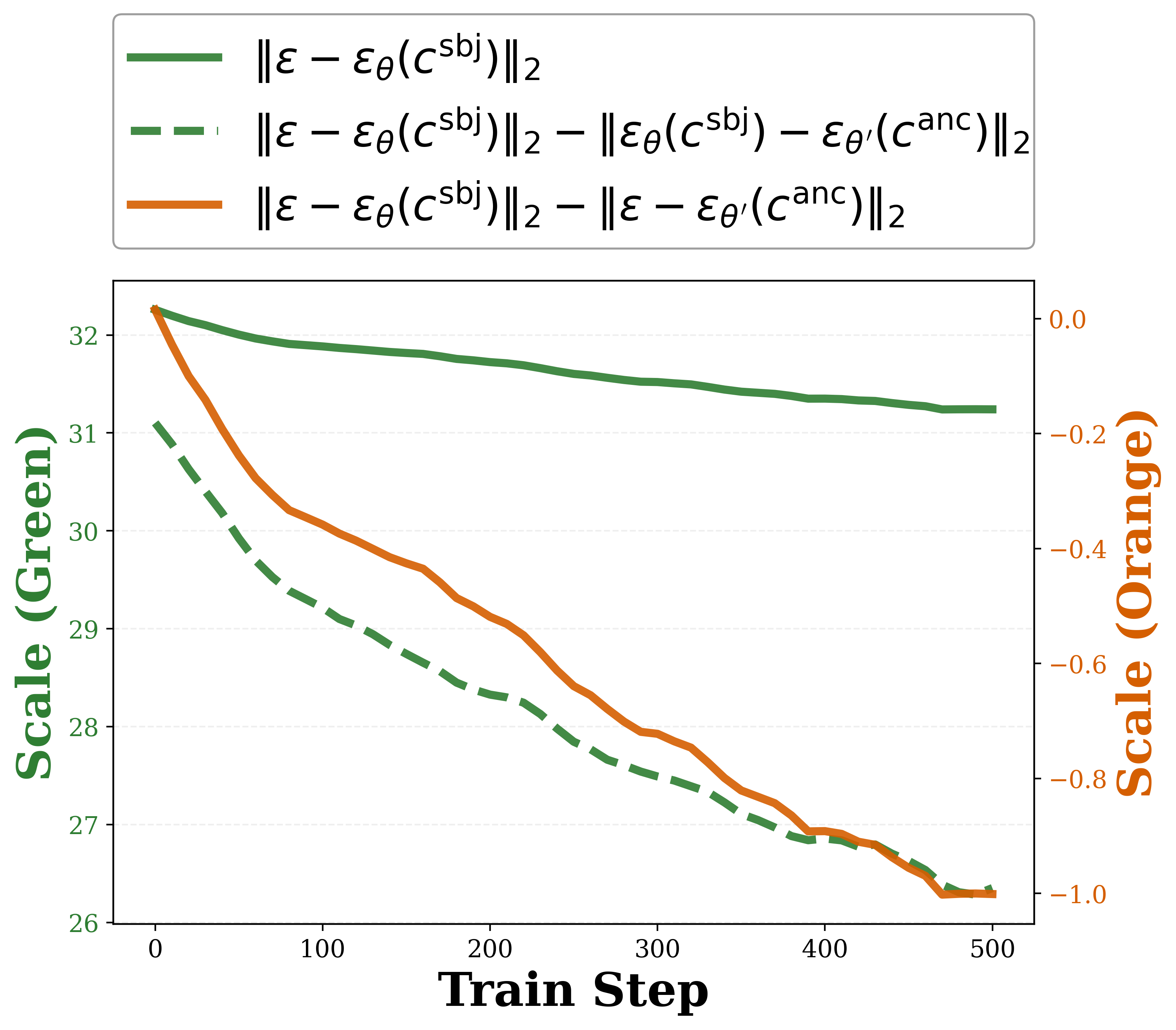}
    %\vspace{-10pt}
    %\vspace{-0.70em}
    \captionof{figure}{Analysis of distance relationships among reference, subject, and anchor predictions.}
    % \vspace{-20pt}
    
    \label{fig:four_distances_row}
  \end{minipage}
\end{figure*}

% \vspace{-20pt}
%-------------------------------------

%\vspace{-5pt}
\subsection{Anchoring Effect on Personalization}
\label{sec:w_ablation}
%\vspace{-5pt}

We investigate how different balances between learning the personalized subject and maintaining its relation to the pretrained semantic counterpart affect the final personalization results. In Eq.~\ref{eq:final}, we vary the anchoring weight $w \in \{0.0, 0.25, 0.5, 0.75, 1.0\}$ to control the relative strength of subject learning and semantic anchoring. As shown in Fig.~\ref{fig:side_by_side}, increasing $w$ improves CLIP-T but decreases CLIP-I and DINO, whereas reducing $w$ produces the opposite trend. At $w=0$, CLIP-I exceeds 0.8 whereas CLIP-T falls below 0.3. This setting favors subject fidelity over text alignment. At $w=1.0$, CLIP-T exceeds 0.3 whereas DINO falls below 0.6, showing the opposite trade-off. An intermediate value of $w=0.5$ maintains both visual fidelity and semantic consistency. This behavior indicates that the proposed objective balances personalized learning with pretrained semantics across different weighting regimes, rather than relying on a single operating point.

% We investigate how varying the anchoring weight affects the balance between subject fidelity and semantic alignment. In \autoref{eq:final}, $w$ controls the relative contribution between the reconstruction loss and the semantic anchoring loss. We vary the weight ($w \in \{0.0, 0.25, 0.5, 0.75, 1.0\}$) to investigate its influence. As illustrated in \autoref{fig:side_by_side}, increasing $w$ enhances CLIP-T while reducing CLIP-I and DINO, revealing a smooth balance trajectory between semantic alignment and subject fidelity. This indicates that $w$ determines the interpolation strength between the subject-specific distribution and the pretrained anchor representation. The optimal region appears around $w=0.5$, where the interpolation between subject and pretrained anchor achieves the best compromise, maintaining both visual fidelity and semantic consistency. This tendency supports our formulation, showing that balancing subject-specific semantics with the pretrained prior stabilizes the personalization process.

%-------------------------------------------
\vspace{-5pt}
\subsection{Reference Model Selection for Semantic Anchoring}
\vspace{-5pt}
\label{sec:anchoring_strategy}
To examine reference model selection for semantic anchoring, we vary the source of the anchor prediction between the pretrained weights $\theta'$ and the current personalized weights $\theta$. This comparison evaluates which reference better serves as a semantic anchor for personalization. Fig.~\ref{fig:clip_t_pre_fine} shows that the pretrained anchor setting achieves higher CLIP-T scores across all backbones. CLIP-T remains above 0.3 with the pretrained anchor, whereas it falls below 0.3 with the jointly finetuned anchor. Since text coherence reflects the pretrained semantic knowledge, these results indicate that preserving the pretrained reference is essential for maintaining the original contextual semantics. When the anchor reference is also finetuned, both the subject and anchor predictions evolve simultaneously, causing the semantic reference point to drift. Consequently, the model loses text alignment capability. This observation is consistent with our formulation in Eq.~\ref{eq:final}, encouraging the personalized prediction to remain connected to the pretrained semantic structure during adaptation.

%This observation aligns with the principle formalized in \autoref{eq:blend-target}, where the frequent concept should remain stable while the rare concept is newly learned. 

% This observation is consistent with our formulation in \autoref{eq:final}, encouraging the personalized prediction to remain connected to the pretrained semantic structure during adaptation.

% In this section, we investigate the appropriate anchoring state for our method. In \autoref{eq:final}, the anchor term $\epsilon_{\theta'}(z_t, c^{\text{anc}}, t)$ is obtained from the pretrained model $\theta'$ throughout personalization. To examine the role of anchoring, we conduct an additional experiment in which the anchor is instead taken from the current personalized model $\theta$. As a result, the pretrained anchor configuration consistently achieves higher CLIP-T scores across all backbones in \autoref{fig:clip_t_pre_fine}. Since text coherence reflects the pretrained semantic knowledge, these results indicate that preserving the pretrained reference is essential for maintaining the original contextual semantics. When the anchor reference is also finetuned, both the subject and anchor predictions evolve simultaneously, causing the semantic reference point to drift. Consequently, the model loses text alignment capability. This observation aligns with the principle formalized in \autoref{eq:blend-target}, where the frequent concept should remain stable while the rare concept is newly learned. 

%----------------------------------------
\vspace{-5pt}
\subsection{Ablation on the Semantic Anchor Choice}

\begin{wraptable}{r}{0.48\textwidth} 
  \centering
  \vspace{-12pt}
  \caption{Ablation study on anchor semantics.}
  \label{tab:summary_dog_vase}
  \resizebox{0.48\textwidth}{!}{ 
    \begin{tabular}{llccc}
    \toprule
    \textbf{Group} & \textbf{Setting} & \textbf{CLIP-I $\uparrow$} & \textbf{CLIP-T $\uparrow$} & \textbf{DINO $\uparrow$} \\
    \midrule
    \midrule
    \multirow{4}{*}{\textbf{dog}} 
    & \cellcolor{gray!15}\textbf{dog$\rightarrow$dog (Ours)}
    & \cellcolor{gray!15}\textbf{0.8108}
    & \cellcolor{gray!15}\textbf{0.3059}
    & \cellcolor{gray!15}\textbf{0.6230} \\
    & dog$\rightarrow$cat                    & 0.8076 & 0.3031 & 0.6170 \\
    & dog$\rightarrow$animal                & 0.8027 & 0.3039 & 0.6130 \\
    & dog$\rightarrow$shark                  & 0.7965 & 0.3026 & 0.5918 \\
    \midrule
    \multirow{4}{*}{\textbf{teapot}} 
    & \cellcolor{gray!15}\textbf{teapot$\rightarrow$teapot (Ours)}	
    & \cellcolor{gray!15}\textbf{0.8329}
    & \cellcolor{gray!15}\textbf{0.3400}
    & \cellcolor{gray!15}\textbf{0.5002} \\
    & teapot$\rightarrow$kettle                & 0.8024 &	0.3318	& 0.4488 \\
    & teapot$\rightarrow$container             & 0.8184	& 0.3287 & 0.4488 \\
    & teapot$\rightarrow$clock                 & 0.8026	& 0.3270 & 0.4331 \\
    \bottomrule
    \end{tabular}
  }
  \vspace{-10pt}
\end{wraptable}
\vspace{-5pt}

To verify that SAP benefits from the semantic relation to the anchor rather than from arbitrary anchoring, we vary the anchor semantics. Compared to using the corresponding class concept as the anchor (e.g., dog$\rightarrow$dog), we vary the anchor to a related category (e.g., dog$\rightarrow$cat), a broader superclass (e.g., dog$\rightarrow$animal), or an unrelated concept (e.g., dog$\rightarrow$shark). In our experiments, for a live subject (dog), we test a set of anchors, including \emph{dog}, \emph{cat}, \emph{animal}, and \emph{shark}. We further conduct the same analysis for an object concept (teapot). The results are summarized in Table~\ref{tab:summary_dog_vase}. Using an unrelated semantic anchor (teapot$\rightarrow$clock) leads to a clear performance drop. Similarly, overly broad anchors (teapot$\rightarrow$container) also degrade performance. These results indicate that our method leverages the pretrained semantic prior through the anchor. They also show that the corresponding class concept provides an effective and reliable reference for semantic anchoring.

% In this section, we study how varying the anchor semantics affects performance through an ablation analysis. We analyze the resulting trends and confirm the effectiveness of our setting. Compared to using the exact concept as the anchor (e.g., dog$\rightarrow$dog), we vary the anchor to a related category (e.g., dog$\rightarrow$cat), a broader superclass (e.g., dog$\rightarrow$animal), or even an unrelated concept (e.g., dog$\rightarrow$shark). In our experiments, for a live subject (dog), we test a set of anchors, including \emph{dog}, \emph{cat}, \emph{animal}, and \emph{shark}. We further conduct the same analysis for an object concept (teapot). The results are summarized in \autoref{tab:summary_dog_vase}. Using an unrelated semantic anchor (teapot$\rightarrow$clock) leads to a clear performance drop. Similarly, overly broad anchors (teapot$\rightarrow$container) also degrade performance. These results indicate that our method interacts with the pretrained semantic prior through the anchor, and that anchoring on the corresponding class concept provides an effective and reliable setting.

\vspace{-5pt}
\subsection{Semantic Space Dynamics}
\vspace{-5pt}
To better understand the semantic dynamics throughout personalization, we analyze the distances among the components of our objective. As shown in Fig.~\ref{fig:four_distances_row}, the distance
$\|\epsilon-\epsilon_{\theta}(z_t,c^{\text{sbj}},t)\|_2$
decreases as training progresses. This indicates that the personalized representation learns subject-specific attributes. The dashed green line shows that the personalized prediction moves away from the pretrained class-level anchor as the subject is learned. At the same time, the orange line shows that the personalized prediction becomes closer to the subject target than the pretrained class-level prediction. However, as shown in Fig.~\ref{fig:subject_anchor}, SAP exhibits the smallest divergence from the pretrained counterpart among the compared objectives. Together, these observations are consistent with the intended behavior of SAP. Rather than constraining the personalized concept within pretrained semantics~\citep{qiao2024facechain}, SAP allows it to expand from the pretrained semantic reference toward subject-specific semantics while maintaining its connection to the anchor as illustrated in Fig.~\ref{fig:concept}.

\section{Conclusion}
\label{sec:conclusion}
% \vspace{-5pt}

In this paper, we take a joint view of concept learning and prior preservation, which are often treated as separate objectives in personalization. Inspired by the relation between underrepresented and well-represented concepts, we reformulate the learning target to incorporate the corresponding semantic counterpart of the target concept. The proposed method anchors the personalized concept to its pretrained semantic counterpart throughout personalization, allowing subject-specific semantics to develop from the pretrained semantic structure rather than in isolation. Experiments show that the proposed method achieves a better balance between subject fidelity and text alignment than existing baselines. Results across multiple diffusion backbones further demonstrate the effectiveness of the proposed objective across different model architectures. Our ablation studies validate the role of semantic anchoring and provide further analysis of its key design choices. In conclusion, we present Semantic Anchoring Personalization, a simple yet effective personalization framework that learns novel user-specific concepts as extensions of pretrained semantics.

% In this paper, we reformulate personalization as the task of learning novel concepts grounded in general semantics. Rather than optimizing subject-specific reconstruction and pretrained knowledge preservation independently, we propose a relational objective that explicitly anchors the novel concept to its semantic prior. Across a wide range of baselines, our experiments show that the proposed method achieves strong improvements in both text alignment and image fidelity. Experiments on various backbone models further demonstrate that the proposed objective can be applied in a general and model-agnostic manner. Through ablation studies, we analyze the impact of the proposed method and validate the soundness of the proposed setting. In conclusion, we present Semantic Anchoring Personalization, a novel method that leverages the relationship between general and subject-specific concepts, offering a more reliable and semantically consistent personalization scheme.

% The proposed anchoring strategy enables personalization to expand from the pretrained semantic structure while retaining its contextual coherence. 

 % In addition, user studies confirm that the proposed approach also improves perceptual quality.

 % We further investigate the internal dynamics of our objective and show that it behaves as predicted by the formulation. 

\nocite{*}
\bibliography{iclr2027_conference}
\bibliographystyle{iclr2027_conference}

\clearpage
\appendix
\setcounter{page}{1}
% \maketitlesupplementary

\section{Discussion and Future Work}
\label{sec:supp_discussion}
Our findings demonstrate that introducing a semantic anchor offers a principled way to structure the interaction between pretrained and subject-specific representations. While our formulation anchors each corresponding semantics, the appropriate granularity and adaptiveness of this anchoring remain open questions. Exploring richer semantic structures and understanding how different semantic factors contribute to each subject's adaptation dynamics presents a promising direction for future work.

\section{Societal Impact}
\label{sec:supp_societalimpact}
This work improves personalization in text-to-image diffusion models, enabling more controllable and expressive image generation. However, such capabilities may increase the risk of misuse, particularly in the creation of deepfakes or misleading visual content. To address these concerns, responsible use is essential, including transparency, detection tools, and safeguards such as watermarking. We emphasize the importance of considering these societal risks when deploying such systems.

% \section{Proof}
% % proof==================================================================================

\section{Proof of Proposition 1}
\label{app:proof}

\begin{proposition*}[Restatement of Proposition~\ref{thm:r2f}]
Let $\lambda\in(0,1]$ and define
$\epsilon^{*}
=
\lambda\epsilon^{\mathrm{rar}}
+
(1-\lambda)\epsilon^{\mathrm{frq}}$.
Using $\epsilon^{*}$ as the training target yields the same optimal parameters as
\[
\theta^{*}
=
\arg\min_{\theta}
\mathbb{E}
\left[
\left\|
\epsilon^{\mathrm{rar}}-\epsilon_{\theta}
\right\|_2^2
+
\frac{1-\lambda}{\lambda}
\left\|
\epsilon_{\theta}-\epsilon^{\mathrm{frq}}
\right\|_2^2
\right].
\]
\end{proposition*}

\begin{proof}
For clarity, let
\begin{equation*}
A
=
\epsilon_{\theta}-\epsilon^{\mathrm{rar}},
\qquad
B
=
\epsilon_{\theta}-\epsilon^{\mathrm{frq}},
\end{equation*}
where we omit the input arguments of $\epsilon_{\theta}$ for brevity.
Since
$\epsilon^{*}
=
\lambda\epsilon^{\mathrm{rar}}
+
(1-\lambda)\epsilon^{\mathrm{frq}}$,
the prediction error can be written as
\begin{equation*}
\begin{aligned}
\epsilon_{\theta}-\epsilon^{*}
&=
\epsilon_{\theta}
-
\left(
\lambda\epsilon^{\mathrm{rar}}
+
(1-\lambda)\epsilon^{\mathrm{frq}}
\right) \\
&=
\lambda
\left(
\epsilon_{\theta}-\epsilon^{\mathrm{rar}}
\right)
+
(1-\lambda)
\left(
\epsilon_{\theta}-\epsilon^{\mathrm{frq}}
\right) \\
&=
\lambda A+(1-\lambda)B.
\end{aligned}
\end{equation*}

Therefore,
\begin{align}
\left\|
\epsilon_{\theta}-\epsilon^{*}
\right\|_2^2
&=
\left\|
\lambda A+(1-\lambda)B
\right\|_2^2 \notag\\
&=
\lambda^2\|A\|_2^2
+
(1-\lambda)^2\|B\|_2^2
+
2\lambda(1-\lambda)\langle A,B\rangle.
\label{eq:expand_step}
\end{align}

Using
\begin{equation*}
\langle A,B\rangle
=
\frac{1}{2}
\left(
\|A\|_2^2
+
\|B\|_2^2
-
\|A-B\|_2^2
\right),
\end{equation*}
Eq.~\ref{eq:expand_step} becomes
\begin{align}
\left\|
\epsilon_{\theta}-\epsilon^{*}
\right\|_2^2
&=
\lambda\|A\|_2^2
+
(1-\lambda)\|B\|_2^2 \notag\\
&\quad
-
\lambda(1-\lambda)\|A-B\|_2^2.
\label{eq:expand_step2}
\end{align}

Substituting $A$ and $B$ into Eq.~\ref{eq:expand_step2} gives
\begin{align}
\left\|
\epsilon_{\theta}-\epsilon^{*}
\right\|_2^2
&=
\lambda
\left\|
\epsilon_{\theta}-\epsilon^{\mathrm{rar}}
\right\|_2^2
+
(1-\lambda)
\left\|
\epsilon_{\theta}-\epsilon^{\mathrm{frq}}
\right\|_2^2
\notag\\
&\quad
-
\lambda(1-\lambda)
\left\|
\epsilon^{\mathrm{rar}}-\epsilon^{\mathrm{frq}}
\right\|_2^2.
\label{eq:expanded_objective}
\end{align}

Since $\epsilon^{\mathrm{rar}}$ and $\epsilon^{\mathrm{frq}}$ do not depend on the trainable parameters $\theta$, the last term in Eq.~\ref{eq:expanded_objective} is constant with respect to $\theta$. Taking the expectation gives
\begin{align*}
\mathbb{E}
\left[
\left\|
\epsilon_{\theta}-\epsilon^{*}
\right\|_2^2
\right]
&=
\lambda
\mathbb{E}
\left[
\left\|
\epsilon_{\theta}-\epsilon^{\mathrm{rar}}
\right\|_2^2
+
\frac{1-\lambda}{\lambda}
\left\|
\epsilon_{\theta}-\epsilon^{\mathrm{frq}}
\right\|_2^2
\right]
+
C,
\end{align*}
where $C$ is independent of $\theta$.

Since $\lambda>0$, the additive constant and positive scaling factor do not change the minimizer. Therefore,
\begin{equation*}
\arg\min_{\theta}
\mathbb{E}
\left[
\left\|
\epsilon^{*}-\epsilon_{\theta}
\right\|_2^2
\right]
=
\arg\min_{\theta}
\mathbb{E}
\left[
\left\|
\epsilon^{\mathrm{rar}}-\epsilon_{\theta}
\right\|_2^2
+
\frac{1-\lambda}{\lambda}
\left\|
\epsilon_{\theta}-\epsilon^{\mathrm{frq}}
\right\|_2^2
\right]
=
\theta^{*}.
\end{equation*}
\end{proof}

\section{Additional Implementation Details}
\label{sec:supp_implementation_detail}
This section provides additional implementation details for both the proposed method and the baseline setups in Sec.~\ref{sec:experiments}. All experiments were conducted on a single RTX 3090 or a single A6000 GPU.

\paragraph{\textbf{Ours.}} We apply the proposed method to multiple backbones, and the corresponding hyperparameters are as follows. For SD~1.5, we use a learning rate of $5\times10^{-4}$, a semantic anchoring weight of $w=0.8$, and a guidance scale of $7.5$ during inference. For SDXL, the learning rate is also set to $5\times10^{-4}$, with $w=1.0$ and a guidance scale of $7.0$. For SD3, we use a learning rate of $2.5\times10^{-4}$, a semantic anchoring weight of $w=0.05$, and an inference guidance scale of $10.0$. The special token for all training prompts is set to ``\texttt{sks}'', and the number of inference steps is fixed to $50$. Our method is implemented on top of the Diffusers library\footnote{\url{https://github.com/huggingface/diffusers}}, and all settings follow the default configurations unless otherwise specified.

\paragraph{\textbf{Baselines.}} We follow the configurations specified in the original papers or the official libraries when conducting the baseline experiments. For Textual Inversion~\citep{gal2022image}\footnote{\url{https://github.com/huggingface/diffusers/tree/main/examples/textual_inversion}}, Custom Diffusion~\citep{kumari2023multi}\footnote{\url{https://github.com/huggingface/diffusers/tree/main/examples/custom_diffusion}}, SVDiff~\citep{han2023svdiff}\footnote{\url{https://github.com/mkshing/svdiff-pytorch}}, DreamBooth full fine-tuning, and DreamBooth~\citep{ruiz2023dreambooth} with LoRA~\citep{hu2022lora}\footnote{\url{https://github.com/huggingface/diffusers/tree/main/examples/dreambooth}}, we use the publicly available implementations provided in the corresponding repositories and follow their default training settings. For BLIP-Diffusion~\citep{li2023blip}\footnote{\url{https://github.com/salesforce/LAVIS/tree/main/projects/blip-diffusion}} and IP-Adapter~\citep{ye2023ip}\footnote{\url{https://github.com/tencent-ailab/IP-Adapter}}, we use the publicly released pretrained models and evaluate them through inference. For SuDe~\citep{qiao2024facechain} and Steering Guidance~\citep{park2025steering}, we re-implemented the methods following the descriptions in the papers based on the Diffusers library. For SuDe, we observed training instability, and therefore removed the denominator term $\sigma_t$ in the objective. The modified version was used for training and subsequent evaluation.

% chen2023subject

\section{Additional Visual Results in Main Experiments}
\label{sec:supp_add_visual_results}

In this section, we present additional visual results and offer extended interpretation. We further include a detailed examination of failure cases.

%-----------------------------
\subsection{Additional Comparison with Baseline Methods}
\label{app:add_base}

Fig.~\ref{fig:supp_sd15_1} presents additional visual comparisons for Sec.~\ref{sec:experiments_baseline}. Our method demonstrates a balanced trade-off between subject fidelity and prompt coherence across a wide range of subjects. In the first, second, and third rows, the model trained with our approach preserves the subject identity while faithfully reflecting the given prompts. In contrast, DreamBooth introduces artifacts in the first row or fails to retain identity in cases such as the second row, and similar issues are also observed in SuDe. These observations are consistent with the improved image-level metrics reported in Table~\ref{tab:main_1}. SVDiff and Custom Diffusion generally follow the text prompts well; however, SVDiff tends to lose subject identity, as seen in the second row, while Custom Diffusion misses fine-grained details such as color in the first and third rows. Steering Guidance similarly reflects the requested attributes and contexts, but does not consistently preserve the distinctive appearance of the reference subjects. This aligns with their relatively high text-alignment scores but lower image-fidelity scores in Table~\ref{tab:main_1}. These results indicate that our model achieves a more balanced optimization across both dimensions, which is reflected in its highest position in the comprehensive ranking.

Although our method achieves meaningful improvements, several cases remain challenging. We examine these examples to analyze the limitations of the approach. While our method provides improved image fidelity and preserves subject identity more effectively than other baselines, the first row in Fig.~\ref{fig:supp_sd15_2} shows that the prompt is not fully reflected. In the second row, the generated results primarily capture the subject’s texture and the prompt, but fail to represent identity-specific characteristics. The last row further illustrates that the model struggles to capture detailed attributes for rare concepts such as cartoon-style subjects. These observations indicate that, despite the overall improvements and balanced performance, the effectiveness of our method varies across subjects. As discussed in the Future Work section, this highlights the need for subject-dependent anchoring strategies and more adaptive balancing during the personalization process.

%-----------------------------

\subsection{Additional Comparison across Backbone Models}
Fig.~\ref{fig:supp_sd15sdxlsd3} demonstrates additional visual comparisons for Sec.~\ref{sec:experiments_backbone}. The proposed method produces the most balanced results across different backbones. For the first subject, DreamBooth-LoRA and Beyond-Finetuning fail to preserve subject-specific characteristics in SD1.5 and SD3, while our approach generates coherent results across all backbones. This observation aligns with the substantial gains in the image fidelity scores (CLIP-I and DINO) reported in Table~\ref{tab:main_2}. A similar trend can be seen for the second subject in Fig.~\ref{fig:supp_sd15sdxlsd3}. 

For the third subject, our method achieves the strongest subject fidelity in SDXL and SD3. Notably, SDXL yields strong results across all methods, which is consistent with its superior performance in Table~\ref{tab:main_2}. However, the SD1.5–based result of our method exhibits insufficient prompt adherence, corresponding to its lower text-alignment score. In contrast, the results of other methods fail to preserve subject characteristics, again reflecting the improvements in image fidelity enabled by our method. Taken together with the observations discussed in Sec.~\ref{app:add_base}, this analysis illustrates that personalization remains challenging and highly subject-dependent, highlighting the need for more advanced techniques that adapt to the characteristics of each subject.

%-----------------------------

\section{Extended Analyses in Ablation Studies}
\label{sec:supp_add_ablation_study}
%-----------------------------
In this section, we provide extended analyses of the ablation studies. We expand the experiments in Sec.~\ref{sec:5_ablation} with additional visual comparisons and further evaluations of alternative inference-time strategies.

\subsection{Extended Analysis of Anchoring Effects on Personalization}
Fig.~\ref{fig:supp_w_ablation} presents additional visual results that correspond to the analysis in Sec.~\ref{sec:w_ablation}. As shown in Fig.~\ref{fig:supp_w_ablation}, varying the weighting parameter $w$ produces a transition between subject fidelity and semantic alignment. For many subjects, the intermediate region around $w=0.5$ yields the most coherent balance between retaining the reference identity and incorporating the desired attributes. Lower values of $w$ retain more subject-specific appearance, while increasing $w$ strengthens adherence to the textual attributes introduced by the pretrained anchor. The smooth transition observed across different $w$ values empirically supports the interpretation in Fig.~\ref{fig:side_by_side} that the proposed objective behaves as an interpolation between subject-specific semantics and the pretrained anchor prior. This also justifies our design choice that the training objective should guide the model toward an ideal blended target direction, reflecting the intended balance of identity and prompt attributes.

However, the bottom row in Fig.~\ref{fig:supp_w_ablation} shows that the optimal choice of $w$ is not universal. In this example, higher values of $w$ capture the prompt more faithfully without fully compromising identity, suggesting that different subjects may require different interpolation strengths. These subject-dependent variations indicate that no single fixed $w$ is optimal across all cases. This observation motivates future work on adaptive anchoring mechanisms, where the effective anchor strength is dynamically adjusted based on the characteristics of each reference subject.

%-----------------------------

\begin{table}[h]
\centering
\centering
\caption{Comparison on SD1.5, SDXL, SD3 using two different anchoring strategies.}
\label{tab:ablation_anchoring}
\small
\setlength{\tabcolsep}{5pt}
\resizebox{0.85\textwidth}{!}{
\begin{tabular}{c l | c c c}
\toprule
\textbf{Model} & \textbf{Method} & \textbf{CLIP-I ${\uparrow}$} & \textbf{CLIP-T ${\uparrow}$} & \textbf{DINO ${\uparrow}$} \\
\midrule\midrule
% ---------- SD 1.5 ----------
\multirow{2}{*}{{\textbf{SD1.5}}}
 & Finetuning Anchoring & 0.8112 & 0.2870 & 0.6563 \\		
 & \cellcolor{gray!15}Pretrained Anchoring (Ours) & 0.7833	& 0.3064 & 0.6051\\
\midrule
% ---------- SD XL ----------
\multirow{2}{*}{{\textbf{SDXL}}}
 & Finetuning Anchoring & 0.8407 & 0.2838 & 0.7109 \\		
 & \cellcolor{gray!15}Pretrained Anchoring (Ours) & 0.8005	& 0.3154 & 0.6661\\
\midrule
% ---------- SD-3.0 ----------
\multirow{2}{*}{{\textbf{SD3}}}
 & Finetuning Anchoring & 0.6182 & 0.2548 & 0.2685 \\		
 & \cellcolor{gray!15}Pretrained Anchoring (Ours) & 0.7579 & 0.3187 & 0.5679\\
\bottomrule
\end{tabular}%
}
\end{table}

\subsection{Additional Comparison of Anchoring Strategy}

Table~\ref{tab:ablation_anchoring} and Fig.~\ref{fig:supp_pretrain_finetune} provide an extended comparison of the two anchoring strategies discussed in Sec.~\ref{sec:anchoring_strategy} to more comprehensively assess the differences between the two strategies. The qualitative results in Fig.~\ref{fig:supp_pretrain_finetune} reveal a clear difference between the two anchoring strategies. Finetuning anchoring preserves subject fidelity well, but it fails to reflect the text information. In contrast, our semantic anchoring personalization captures the text prompt much more faithfully while still maintaining the subject characteristics provided in the references. 
When viewed together with Table~\ref{tab:ablation_anchoring}, the results reveal an inherent trade-off that finetuned anchoring achieves slightly higher subject similarity, but pretrained anchoring consistently improves prompt alignment while maintaining reasonable identity preservation.

However, finetuned anchoring introduces an inherent limitation. Because the anchor branch is updated during personalization, it becomes entangled with the subject and is more susceptible to optimization instability. This effect is particularly evident in the SD3 results, where finetuned anchoring underperforms across all metrics. As further illustrated in Fig.~\ref{fig:supp_pretrain_finetune}, finetuned anchoring often causes mode collapse and fails to represent both the subject and the text prompt. 
For these reasons, we adopt the pretrained anchoring strategy, which provides a more stable reference signal and achieves a balance between subject fidelity and prompt adherence.

%-----------------------------

\begin{table}[h]
\centering
\centering
\caption{Ablation of inference-time guidance strategies on SD1.5 under the model trained with our semantic anchoring objective.}
\label{tab:ablation_inference_method}
\small
\setlength{\tabcolsep}{5pt}
\resizebox{0.85\textwidth}{!}{
\begin{tabular}{l | c c c}
\toprule
\textbf{Method} & \textbf{CLIP-I ${\uparrow}$} & \textbf{CLIP-T ${\uparrow}$} & \textbf{DINO ${\uparrow}$} \\
\midrule\midrule
\cellcolor{gray!15}Ours & 0.7833 & 0.3064 & 0.6051 \\
Pretrained Unconditional Guidance & 0.7869 & 0.2791 & 0.6050 \\		
Finetuned Anchored Guidance & 0.7250 & 0.2497 & 0.4935 \\		
Pretrained Anchored Guidance & 0.7888 & 0.2506 & 0.6049 \\		
\bottomrule
\end{tabular}
}
\end{table}
\subsection{Additional Ablation on Inference-Time Guidance Strategies}

To better understand how inference-time guidance interacts with the proposed training objective, we evaluate four variants of Classifier-Free Guidance (CFG) that differ only in how the unconditional branch is computed. Let $z_t$ denote the noised latent at timestep $t$, and 
$\epsilon_{\theta}(z_t, c)$ be the noise prediction of the personalized model $\theta$ under condition $c$. We denote the target prompt as $c^{\text{sbj}}$ (e.g., "a photo of a [V*] dog") and its anchor prompt as $c^{\text{anc}}$ (e.g., "a photo of a dog"). The pretrained model is represented by $\theta'$.

Using the same guidance scale $s=7.5$, we evaluate four choices of the guidance:\\
(1) \textbf{Unconditional Guidance}: empty prompt using the personalized model
\begin{equation*}
\epsilon_{\text{guide}}
=
\epsilon_{\theta}(z_t, c^{\text{sbj}})
+
s\big(
\epsilon_{\theta}(z_t, c^{\text{sbj}})
-
\epsilon_{\theta}(z_t, c^{\text{uncond}})
\big),
\end{equation*}
(2) \textbf{Pretrained Unconditional Guidance}: empty prompt using the pretrained model
\begin{equation*}
\epsilon_{\text{guide}}
=
\epsilon_{\theta}(z_t, c^{\text{sbj}})
+
s\big(
\epsilon_{\theta}(z_t, c^{\text{sbj}})
-
\epsilon_{\theta'}(z_t, c^{\text{uncond}})
\big),
\end{equation*}
(3) \textbf{Finetuned Anchored Guidance}: prediction of anchor prompt using the personalized model
\begin{equation*}
\epsilon_{\text{guide}}
=
\epsilon_{\theta}(z_t, c^{\text{sbj}})
+
s\big(
\epsilon_{\theta}(z_t, c^{\text{sbj}})
-
\epsilon_{\theta}(z_t, c^{\text{anc}})
\big),
\end{equation*}
(4) \textbf{Pretrained Anchored Guidance}: prediction of anchor prompt using the pretrained model
\begin{equation*}
\epsilon_{\text{guide}}
=
\epsilon_{\theta}(z_t, c^{\text{sbj}})
+
s\big(
\epsilon_{\theta}(z_t, c^{\text{sbj}})
-
\epsilon_{\theta'}(z_t, c^{\text{anc}})
\big).
\end{equation*}

Table~\ref{tab:ablation_inference_method} demonstrates the quantitative results under different inference strategies. Our method achieves the most balanced performance across both image fidelity and text alignment metrics. In contrast, using pretrained information at inference time leads to a noticeable loss in text alignment, and leveraging finetuned class information results in inferior performance across all metrics.

Fig.~\ref{fig:supp_inference_method} further illustrates how each inference strategy interacts with the learned personalized guidance, providing a perspective that is not captured in Table~\ref{tab:ablation_inference_method}. When pretrained information is applied, the generated images contain visible artifacts. This suggests that the fixed pretrained features are not fully compatible with the personalized guidance and therefore introduce inaccurate guidance signals. In comparison, the proposed strategy mitigates these artifacts, suggesting that it provides the most reliable and coherent guidance during inference.

Taken together, inference-time modification does not universally provide benefits with the proposed method. As our training objective already enforces a semantically aligned relation between the subject and its class semantics, the standard unconditional CFG is sufficient and, in practice, produces the most reliable balance of subject fidelity and text alignment. This means that our method does not require any additional guidance strategy at inference time, and replacing the unconditional branch to other guidance can even deteriorate the generated results.

\newpage

\section{User Study}
\label{sec:supp_user_study}
\begin{wraptable}{r}{0.5\textwidth}
    \centering
    \vspace{-15pt} 
    \caption{User study results across four criteria (all values in \%).}
    \label{tab:user_study}
    
    \setlength{\tabcolsep}{3pt} 
    \renewcommand{\arraystretch}{1.3} 
    \resizebox{\linewidth}{!}{%
      \begin{tabular}{@{}lcccc@{}}
        \toprule
        \textbf{Method} & \textbf{\makecell{Identity \\ Preserv.}} & \textbf{\makecell{Prompt \\ Align.}} & \textbf{\makecell{Visual \\ Quality}} & \textbf{Overall} \\
        \midrule\midrule
        \rowcolor{gray!15}\textbf{Ours} & \textbf{64.00} & \textbf{74.50} & \textbf{65.17} & \textbf{78.50} \\
        DreamBooth & 22.67 & 12.50 & 18.00 & 14.33 \\
        DreamBooth-LoRA & 1.33 & 10.50 & 6.83 & 4.50 \\
        IP-Adapter & 12.00 & 2.50 & 10.00 & 2.67 \\
        \bottomrule
      \end{tabular}
    }
    \vspace{-10pt}
\end{wraptable}

To assess perceptual quality beyond quantitative metrics, we conduct a user study with 20 participants along four criteria. For each criterion, evaluators are presented with outputs in a shuffled order and select the best image according to the corresponding metric. The results are summarized in Table~\ref{tab:user_study}. Low-rank adaptation exhibits the lowest identity preservation, suggesting difficulty in incorporating new concepts, consistent with the qualitative examples in Fig.~\ref{fig:SD1.5_final}. Encoder-based methods achieve relatively low prompt alignment, in line with the quantitative results in Table~\ref{tab:main_1}, which also leads to poor overall scores. In contrast, our method attains the best performance across all criteria, indicating a more balanced trade-off between learning new concepts and preserving the pretrained semantics. This suggests that our semantic anchoring enables concept learning    with minimal deviation from the pretrained model.

% \begin{sidewaysfigure*}[p]
%     \centering
%     \includegraphics[width=\linewidth]{figure/supp_sd15_1.jpg}
%     \caption{Additional visual comparison of few-shot personalization and encoder-based methods on the SD1.5 backbone.}
%     \label{fig:supp_sd15_1}
% \end{sidewaysfigure*}

% \input{figure/supp_sd15_1}
\begin{sidewaysfigure}
    \centering

    \setlength{\tabcolsep}{1pt}
    \renewcommand{\arraystretch}{1.0}

    % =========================================================
    % 이미지 크기
    % =========================================================
    \newlength{\suppResultSize}
    \setlength{\suppResultSize}{0.086\textheight}

    % Reference 4장: 간격 없는 2×2 배열
    % 위: ref1, ref2 / 아래: ref3, ref4
    \newcommand{\suppFourRefs}[4]{%
        \vbox{%
            \offinterlineskip
            \hbox{%
                \includegraphics[
                    width=0.5\suppResultSize,
                    height=0.5\suppResultSize
                ]{#1}%
                \includegraphics[
                    width=0.5\suppResultSize,
                    height=0.5\suppResultSize
                ]{#2}%
            }%
            \hbox{%
                \includegraphics[
                    width=0.5\suppResultSize,
                    height=0.5\suppResultSize
                ]{#3}%
                \includegraphics[
                    width=0.5\suppResultSize,
                    height=0.5\suppResultSize
                ]{#4}%
            }%
        }%
    }

    % 결과 이미지
    \newcommand{\suppResultImg}[1]{%
        \includegraphics[
            width=\suppResultSize,
            height=\suppResultSize
        ]{#1}%
    }

    \begin{tabular}{
        @{}
        C{\suppResultSize}
        *{10}{C{\suppResultSize}}
        @{}
    }

    % =========================================================
    % Header
    % =========================================================
    \methodheader{References}
    &
    \methodheader{Ours}
    &
    \methodheader{DreamBooth}
    &
    \methodheader{DreamBooth-LoRA}
    &
    \methodheader{Textual\\Inversion}
    &
    \methodheader{SuDe}
    &
    \methodheader{SVDiff}
    &
    \methodheader{Custom\\Diffusion}
    &
    \methodheader{Steering\\Guidance}
    &
    \methodheader{IP-\\Adapter}
    &
    \methodheader{BLIP-\\Diffusion}
    \\[4pt]

    % =========================================================
    % Row 1: Can
    % =========================================================
    \suppFourRefs
        {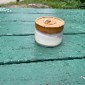}
        {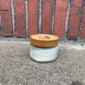}
        {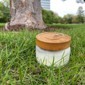}
        {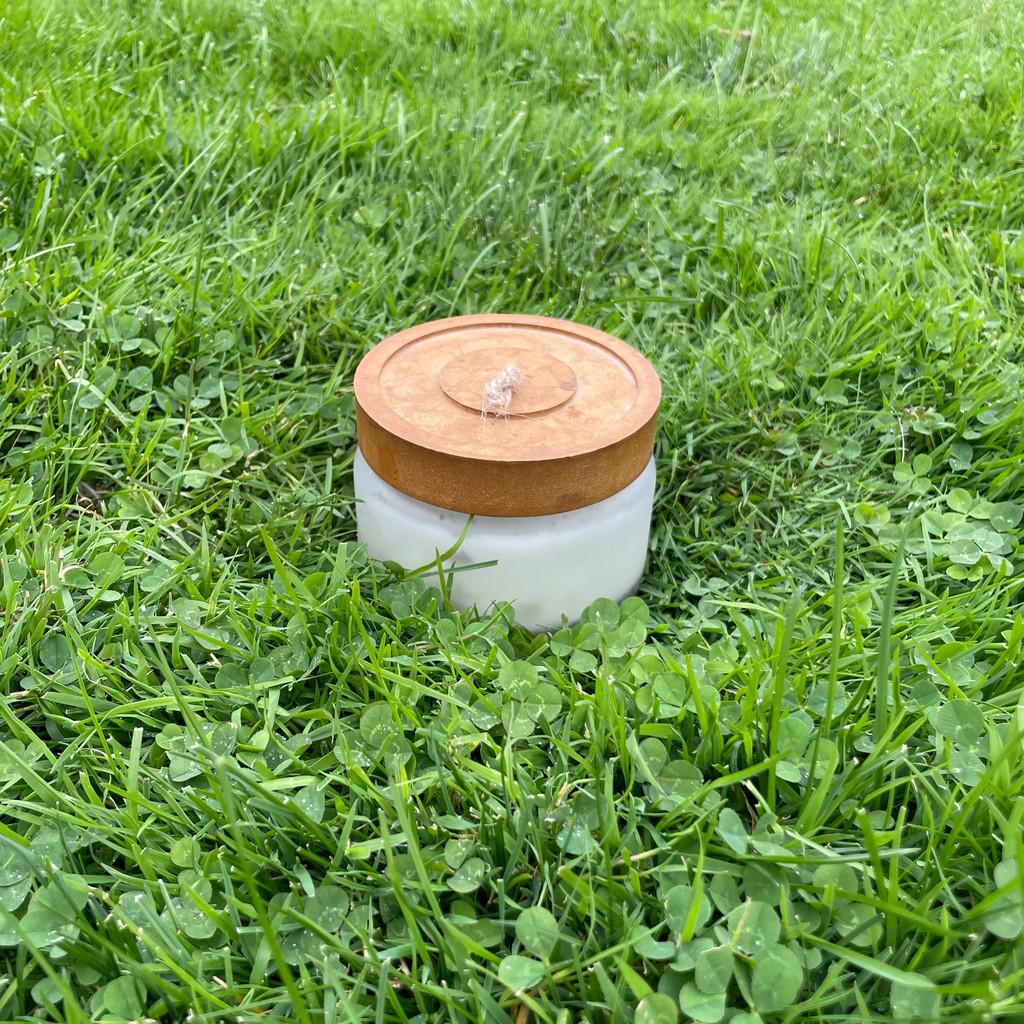}
    &
    \suppResultImg{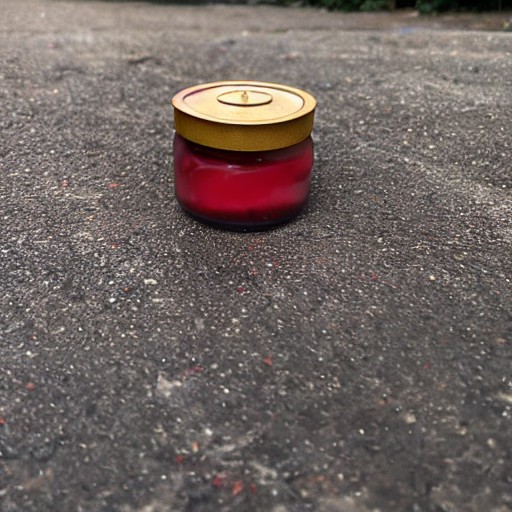}
    &
    \suppResultImg{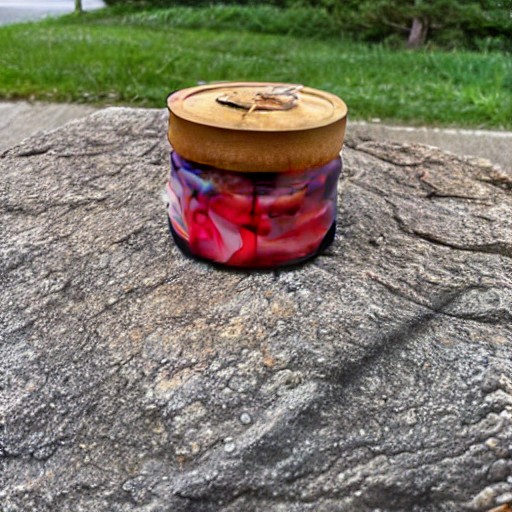}
    &
    \suppResultImg{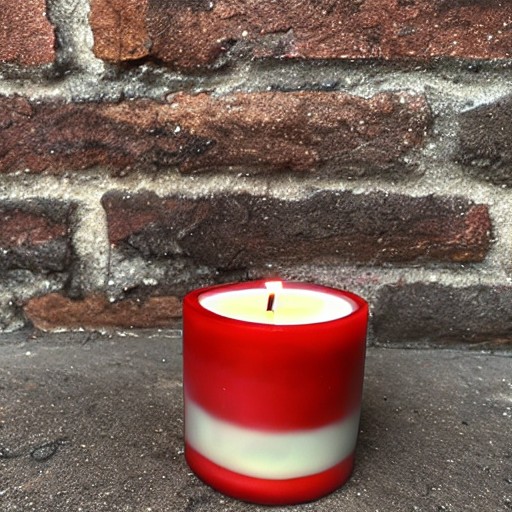}
    &
    \suppResultImg{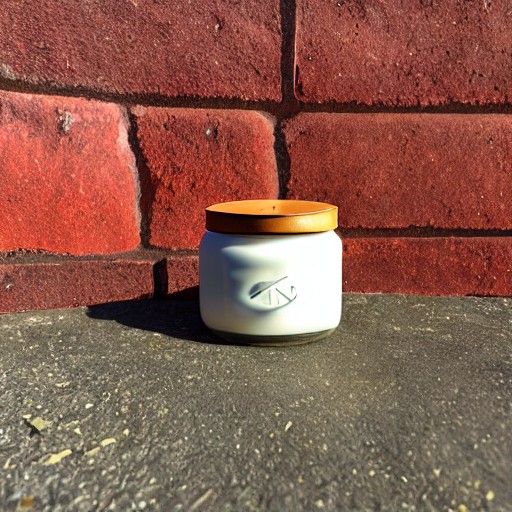}
    &
    \suppResultImg{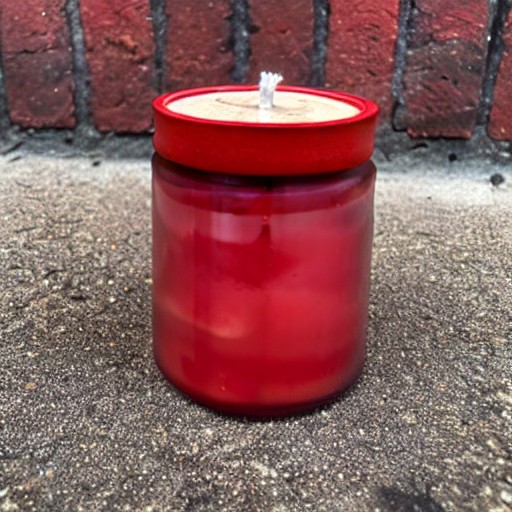}
    &
    \suppResultImg{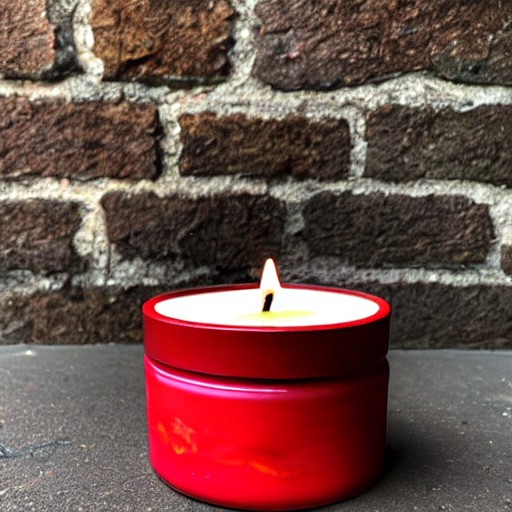}
    &
    \suppResultImg{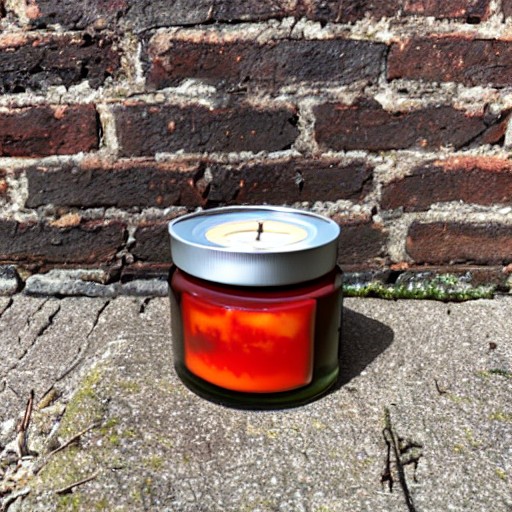}
    &
    \suppResultImg{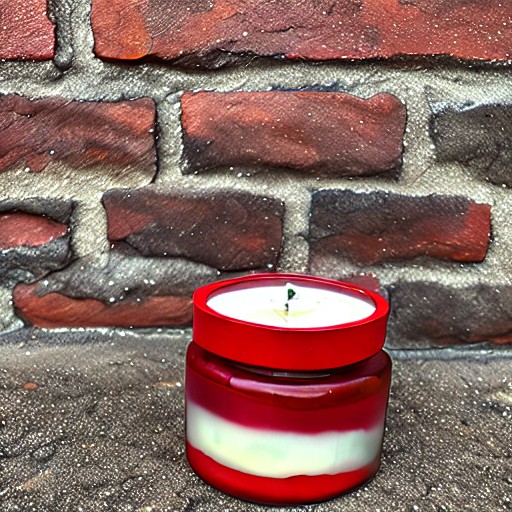}
    &
    \suppResultImg{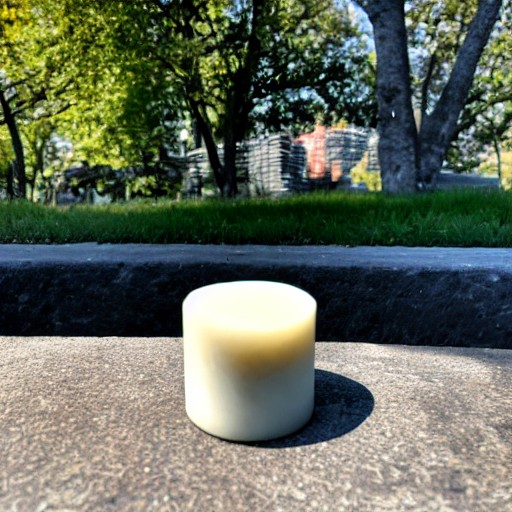}
    &
    \suppResultImg{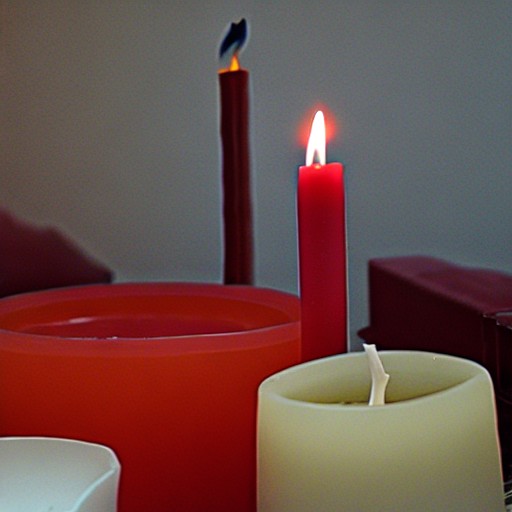}
    \\[-1pt]

    {} &
    \multicolumn{10}{c}{
        {\scriptsize\itshape ``a red [V*]''}
    }
    \\[4pt]

    % =========================================================
    % Row 2: Dog 1
    % =========================================================
    \suppFourRefs
        {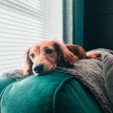}
        {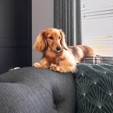}
        {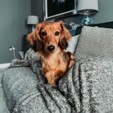}
        {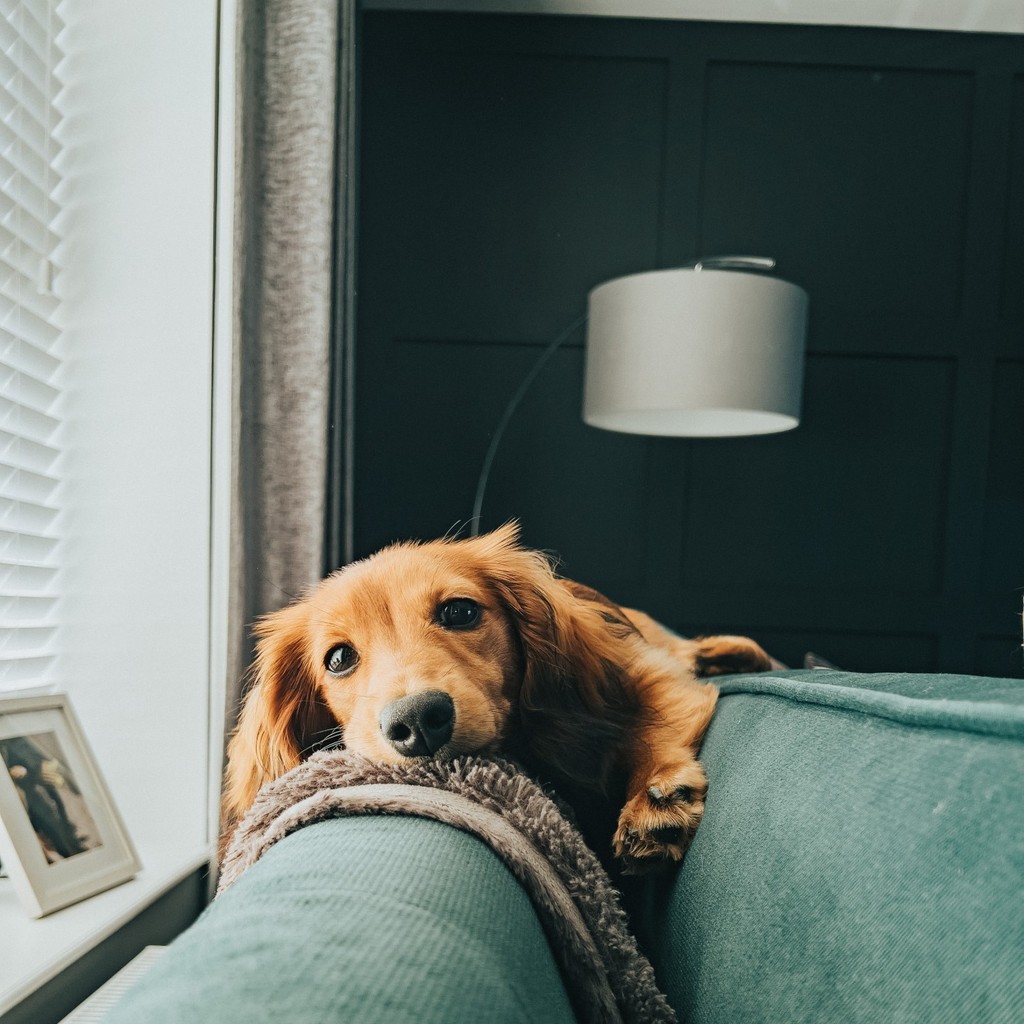}
    &
    \suppResultImg{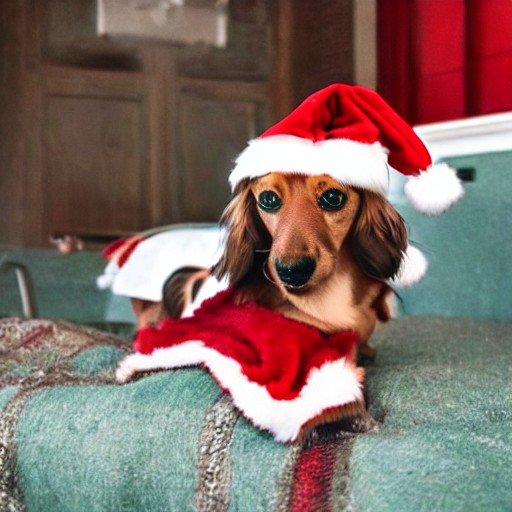}
    &
    \suppResultImg{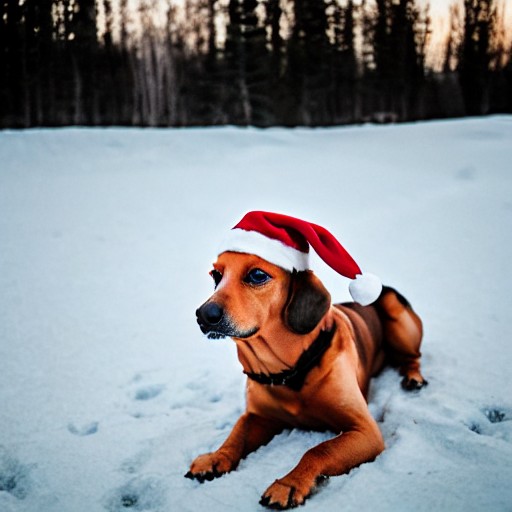}
    &
    \suppResultImg{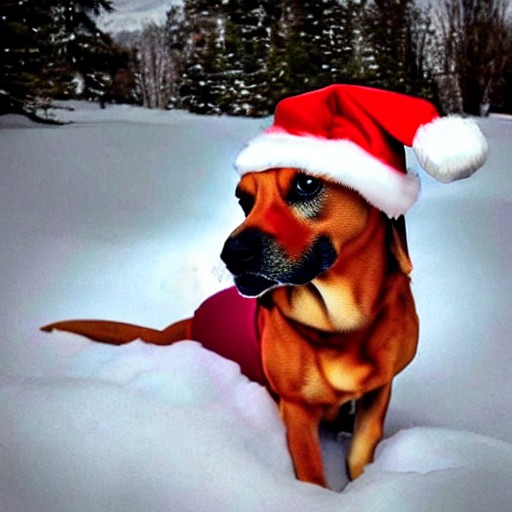}
    &
    \suppResultImg{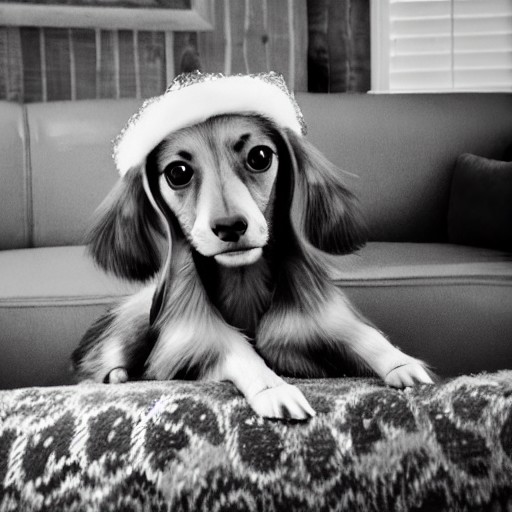}
    &
    \suppResultImg{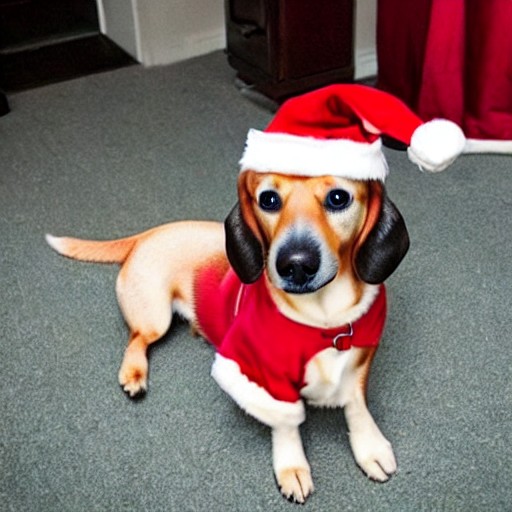}
    &
    \suppResultImg{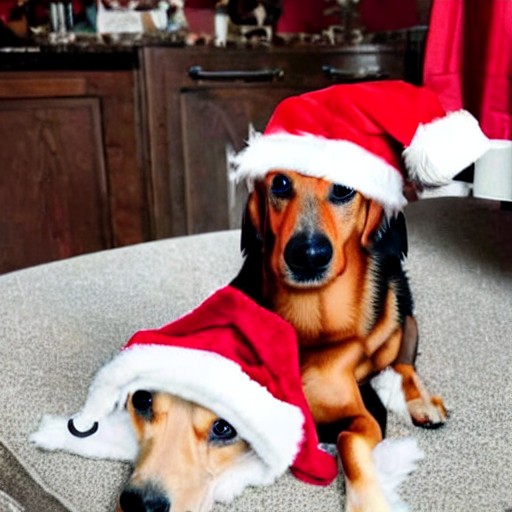}
    &
    \suppResultImg{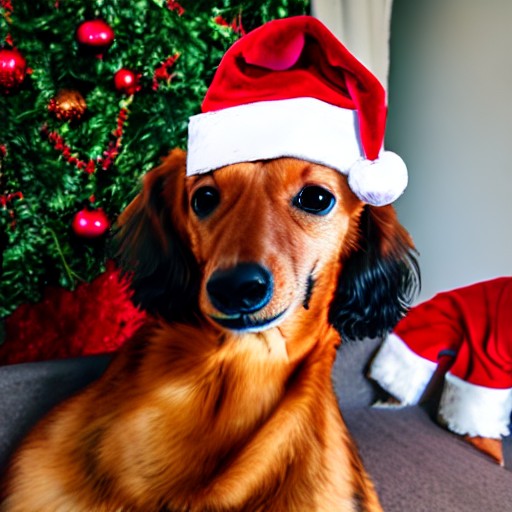}
    &
    \suppResultImg{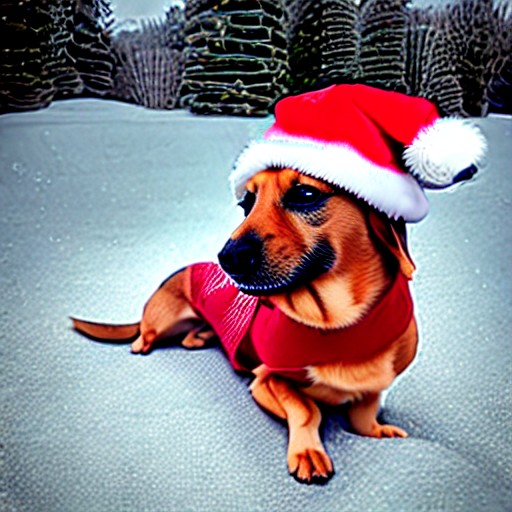}
    &
    \suppResultImg{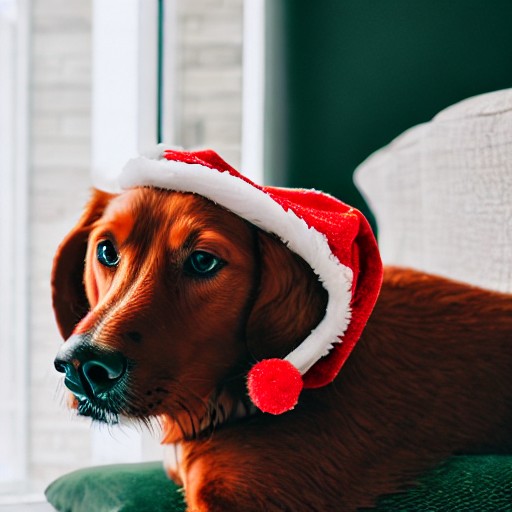}
    &
    \suppResultImg{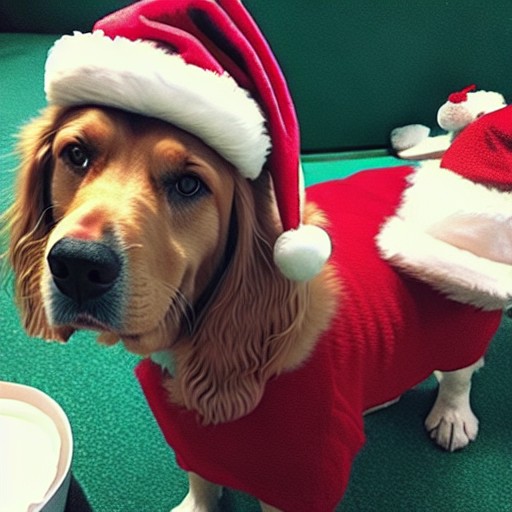}
    \\[-1pt]

    {} &
    \multicolumn{10}{c}{
        {\scriptsize\itshape
        ``a [V*] wearing a santa hat''}
    }
    \\[4pt]

    % =========================================================
    % Row 3: Bag
    % =========================================================
    \suppFourRefs
        {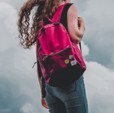}
        {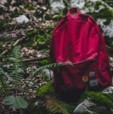}
        {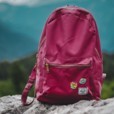}
        {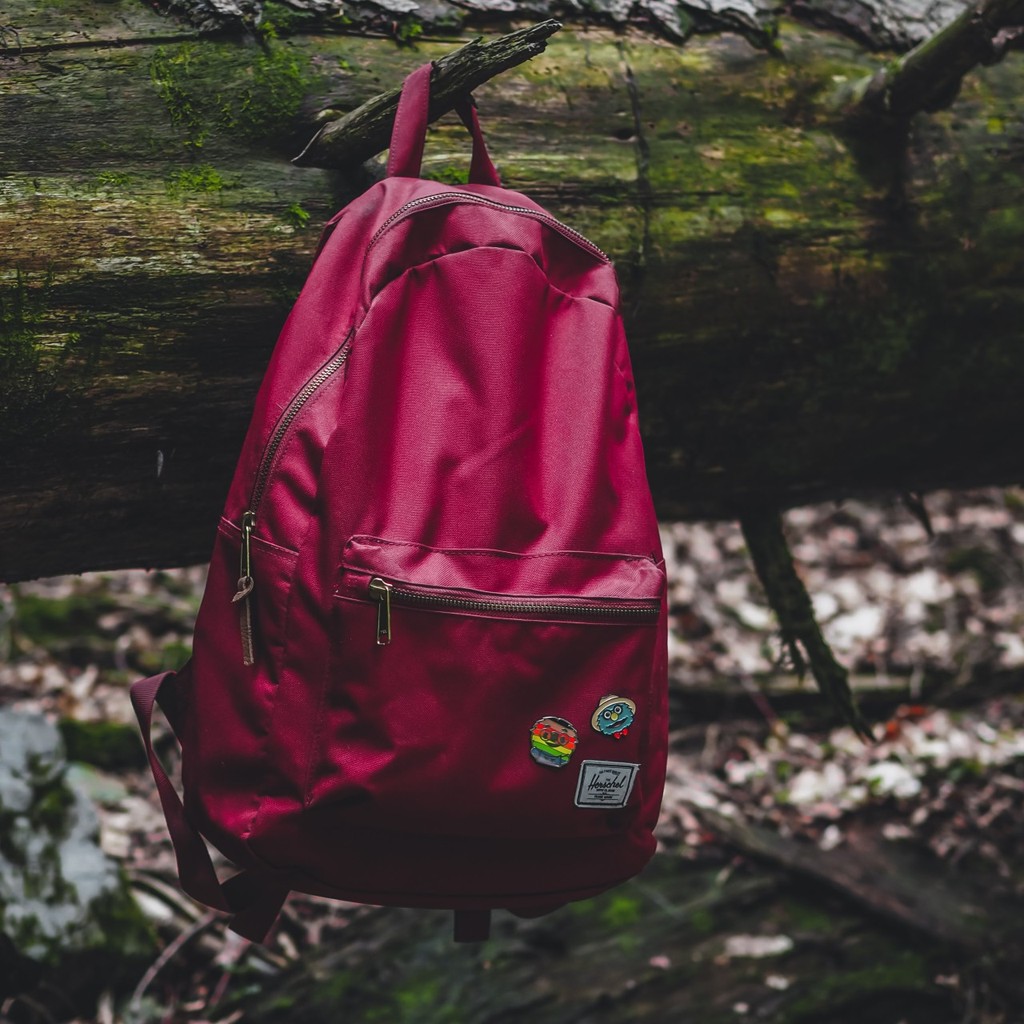}
    &
    \suppResultImg{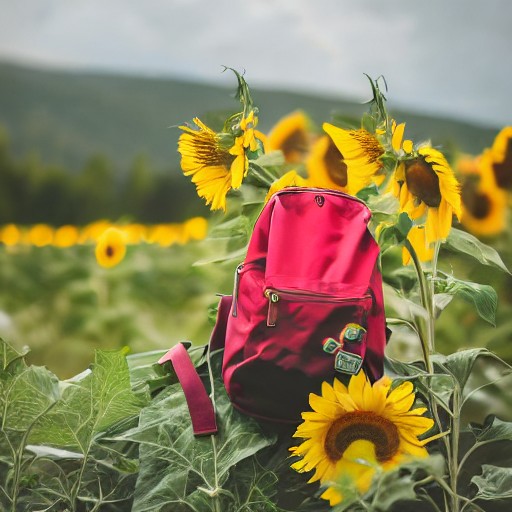}
    &
    \suppResultImg{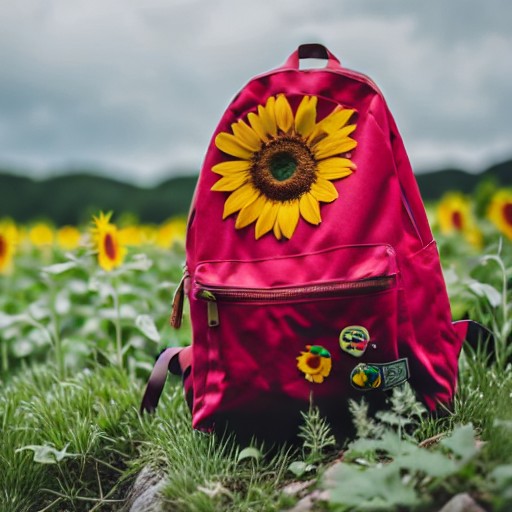}
    &
    \suppResultImg{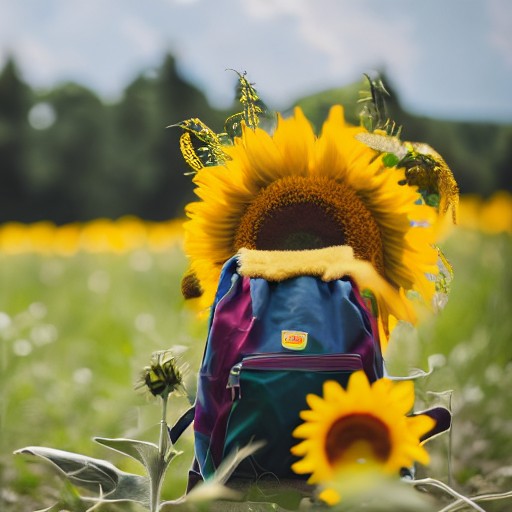}
    &
    \suppResultImg{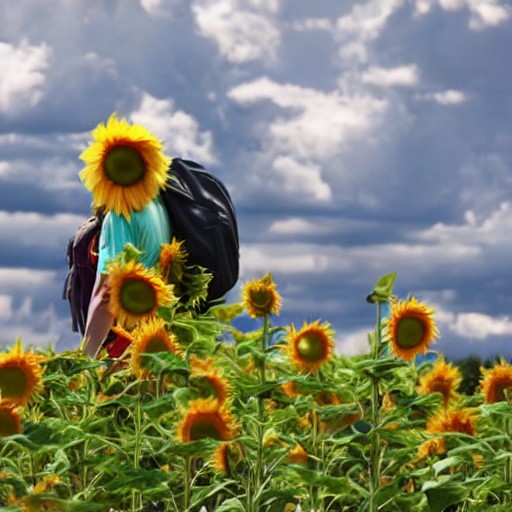}
    &
    \suppResultImg{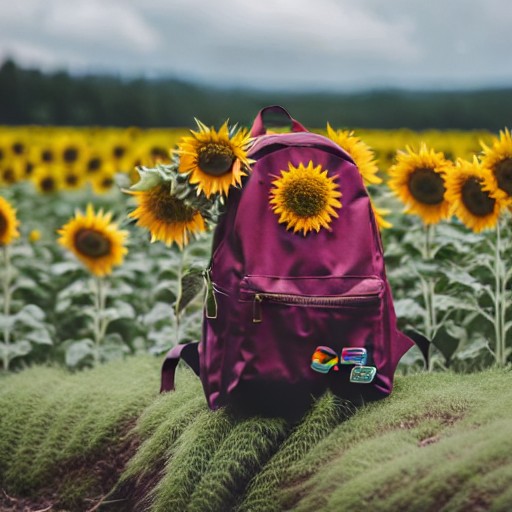}
    &
    \suppResultImg{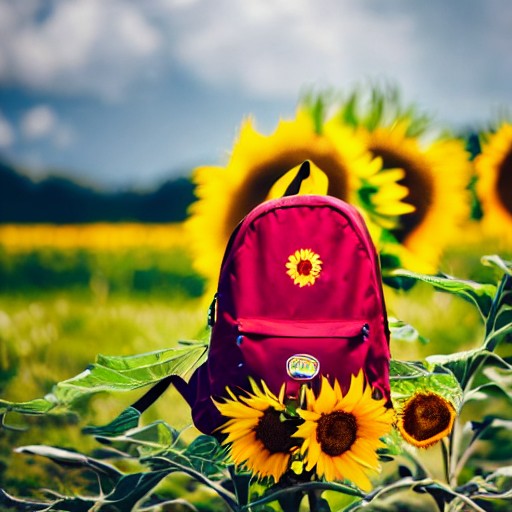}
    &
    \suppResultImg{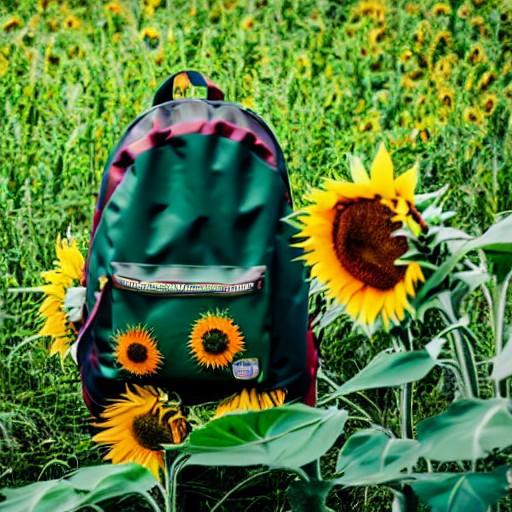}
    &
    \suppResultImg{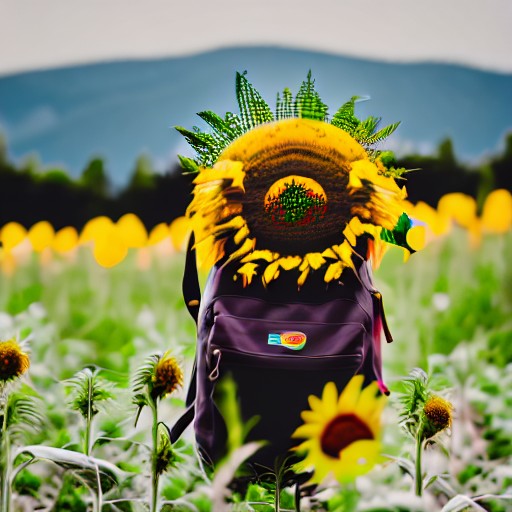}
    &
    \suppResultImg{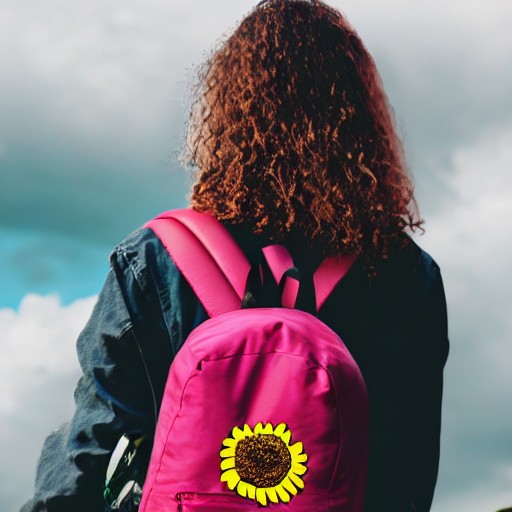}
    &
    \suppResultImg{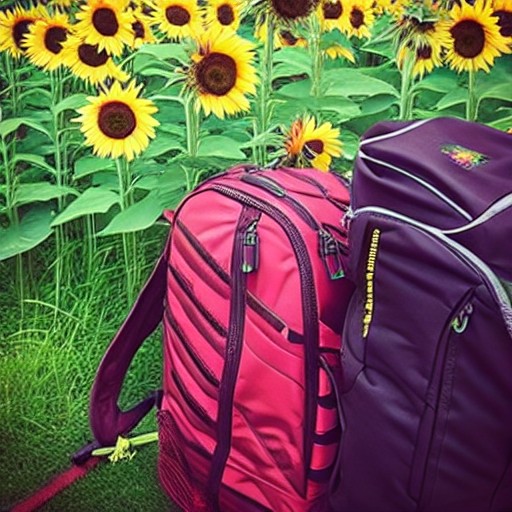}
    \\[-1pt]

    {} &
    \multicolumn{10}{c}{
        {\scriptsize\itshape
        ``a [V*] on top of green grass with sunflowers around it''}
    }
    \\[4pt]

    % =========================================================
    % Row 4: Dog 2
    % =========================================================
    \suppFourRefs
        {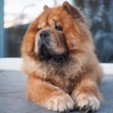}
        {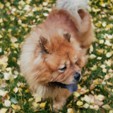}
        {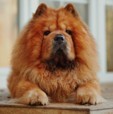}
        {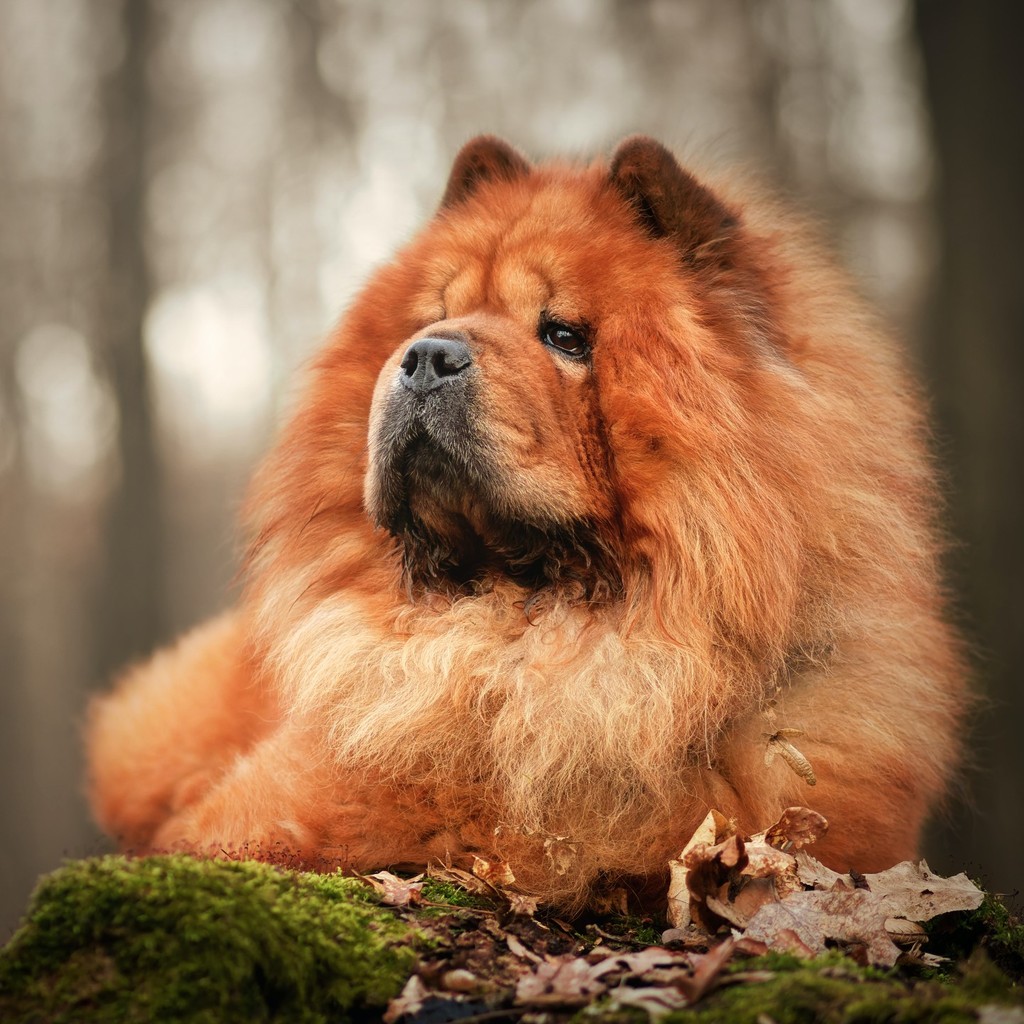}
    &
    \suppResultImg{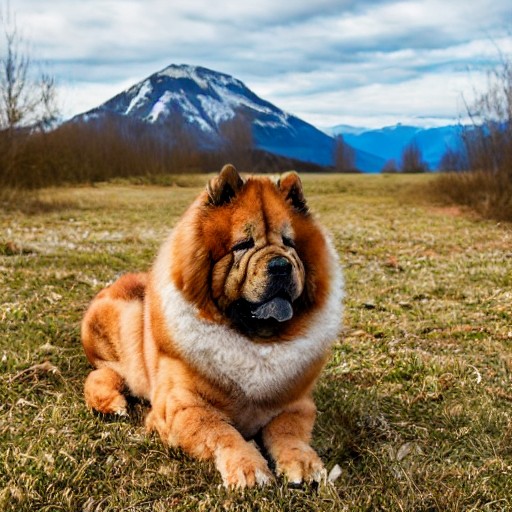}
    &
    \suppResultImg{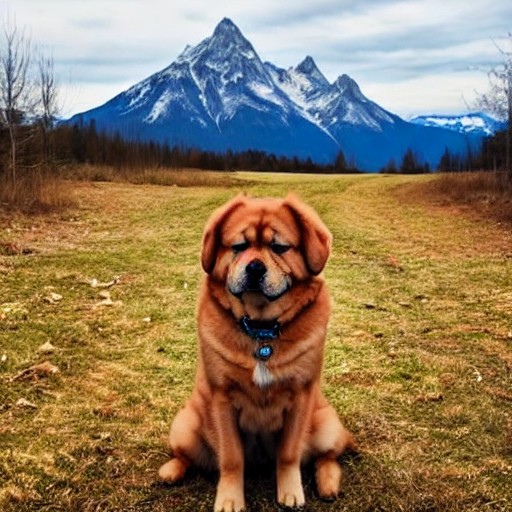}
    &
    \suppResultImg{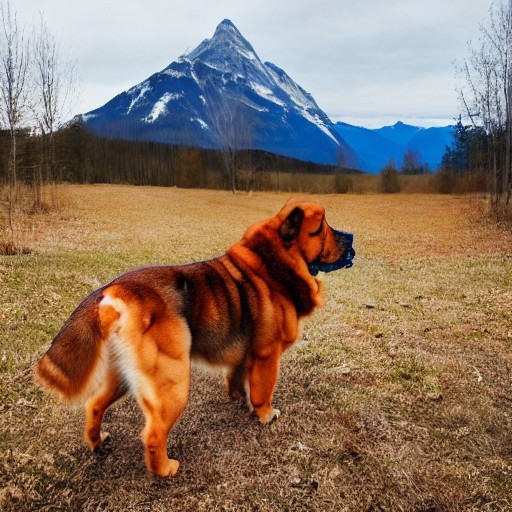}
    &
    \suppResultImg{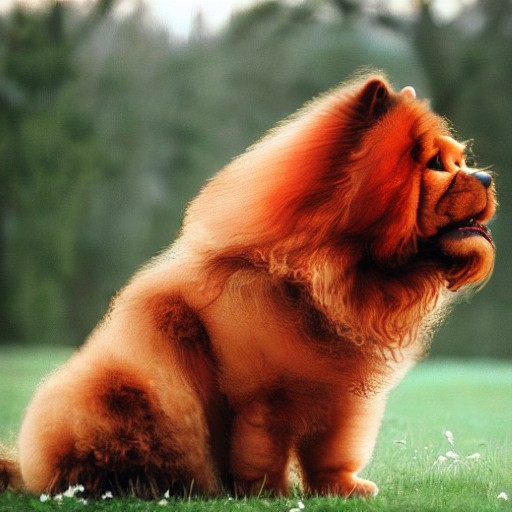}
    &
    \suppResultImg{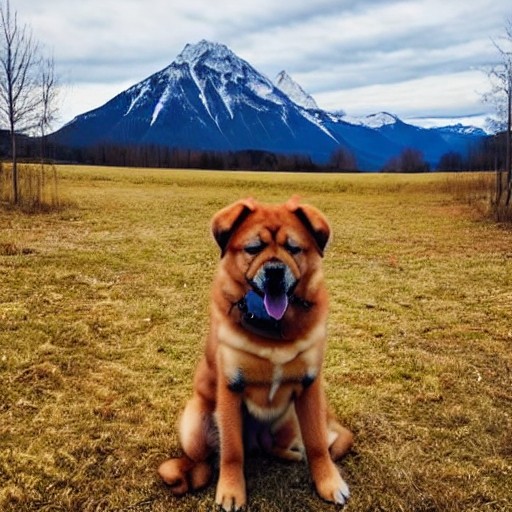}
    &
    \suppResultImg{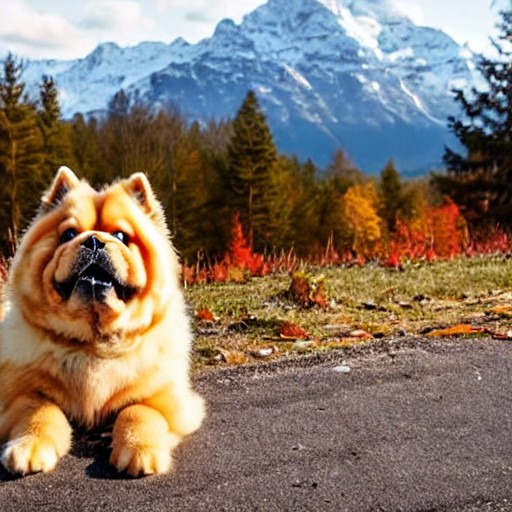}
    &
    \suppResultImg{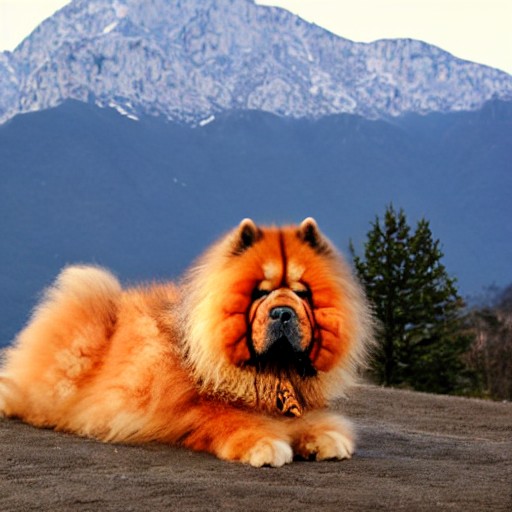}
    &
    \suppResultImg{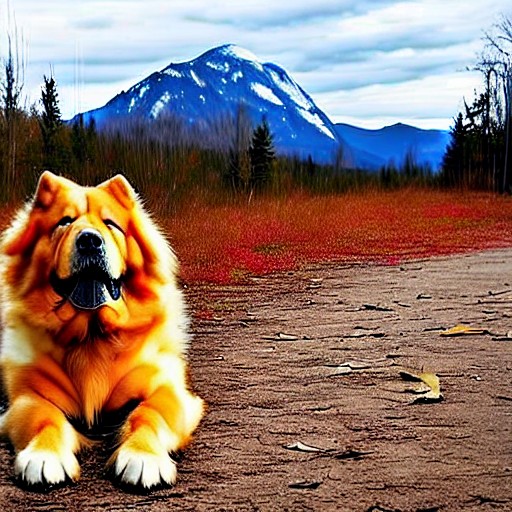}
    &
    \suppResultImg{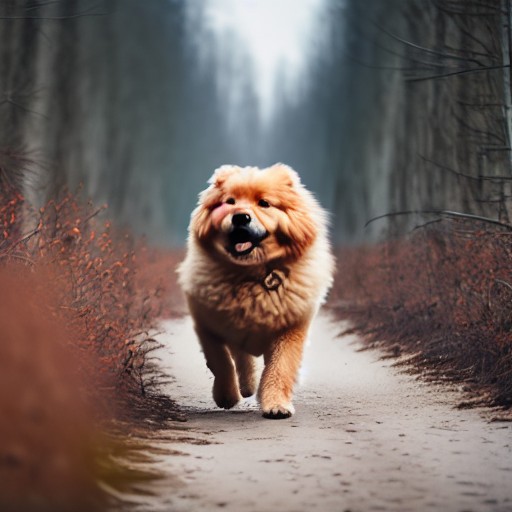}
    &
    \suppResultImg{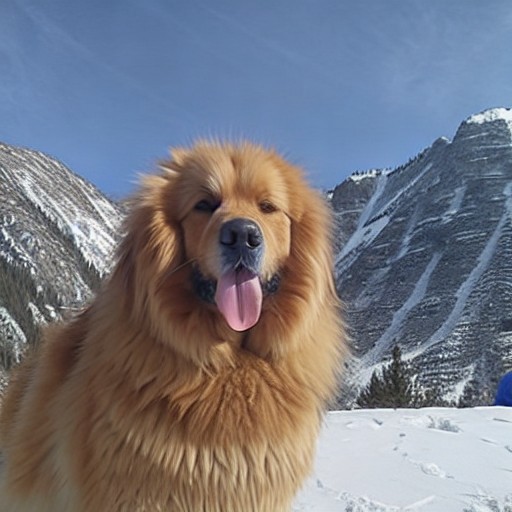}
    \\[-1pt]

    {} &
    \multicolumn{10}{c}{
        {\scriptsize\itshape
        ``a [V*] with a mountain in the background''}
    }
    \\[1pt]

    \end{tabular}

    \caption{
        Additional visual comparison of few-shot personalization and
        encoder-based methods on the SD1.5 backbone.
    }
    \label{fig:supp_sd15_1}

\end{sidewaysfigure}

% \begin{sidewaysfigure*}[p]
%     \centering
%     \includegraphics[width=\linewidth]{figure/supp_sd15_2.png}
%     \caption{Additional visual comparison of few-shot personalization and encoder-based methods on the SD1.5 backbone, including representative failure cases.
%     }
%     \label{fig:supp_sd15_2}
% \end{sidewaysfigure*}

% \input{figure/supp_sd15_2}
\begin{sidewaysfigure}
    \centering

    \setlength{\tabcolsep}{1pt}
    \renewcommand{\arraystretch}{1.0}

    % =========================================================
    % 이미지 크기
    % =========================================================
    \newlength{\suppTwoResultSize}
    \setlength{\suppTwoResultSize}{0.084\textheight}

    % Reference 4장: 간격 없는 2×2 배열
    % 위: ref1, ref2 / 아래: ref3, ref4
    \newcommand{\suppTwoFourRefs}[4]{%
        \vbox{%
            \offinterlineskip
            \hbox{%
                \includegraphics[
                    width=0.5\suppTwoResultSize,
                    height=0.5\suppTwoResultSize
                ]{#1}%
                \includegraphics[
                    width=0.5\suppTwoResultSize,
                    height=0.5\suppTwoResultSize
                ]{#2}%
            }%
            \hbox{%
                \includegraphics[
                    width=0.5\suppTwoResultSize,
                    height=0.5\suppTwoResultSize
                ]{#3}%
                \includegraphics[
                    width=0.5\suppTwoResultSize,
                    height=0.5\suppTwoResultSize
                ]{#4}%
            }%
        }%
    }

    % 결과 이미지
    \newcommand{\suppTwoResultImg}[1]{%
        \includegraphics[
            width=\suppTwoResultSize,
            height=\suppTwoResultSize
        ]{#1}%
    }

    \begin{tabular}{
        @{}
        C{\suppTwoResultSize}
        *{10}{C{\suppTwoResultSize}}
        @{}
    }

    % =========================================================
    % Header
    % =========================================================
    \methodheader{References}
    &
    \methodheader{Ours}
    &
    \methodheader{DreamBooth}
    &
    \methodheader{DreamBooth-LoRA}
    &
    \methodheader{Textual\\Inversion}
    &
    \methodheader{SuDe}
    &
    \methodheader{SVDiff}
    &
    \methodheader{Custom\\Diffusion}
    &
    \methodheader{Steering\\Guidance}
    &
    \methodheader{IP-\\Adapter}
    &
    \methodheader{BLIP-\\Diffusion}
    \\[4pt]

    % =========================================================
    % Row 1: Can
    % =========================================================
    \suppTwoFourRefs
        {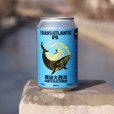}
        {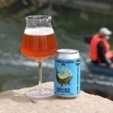}
        {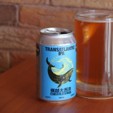}
        {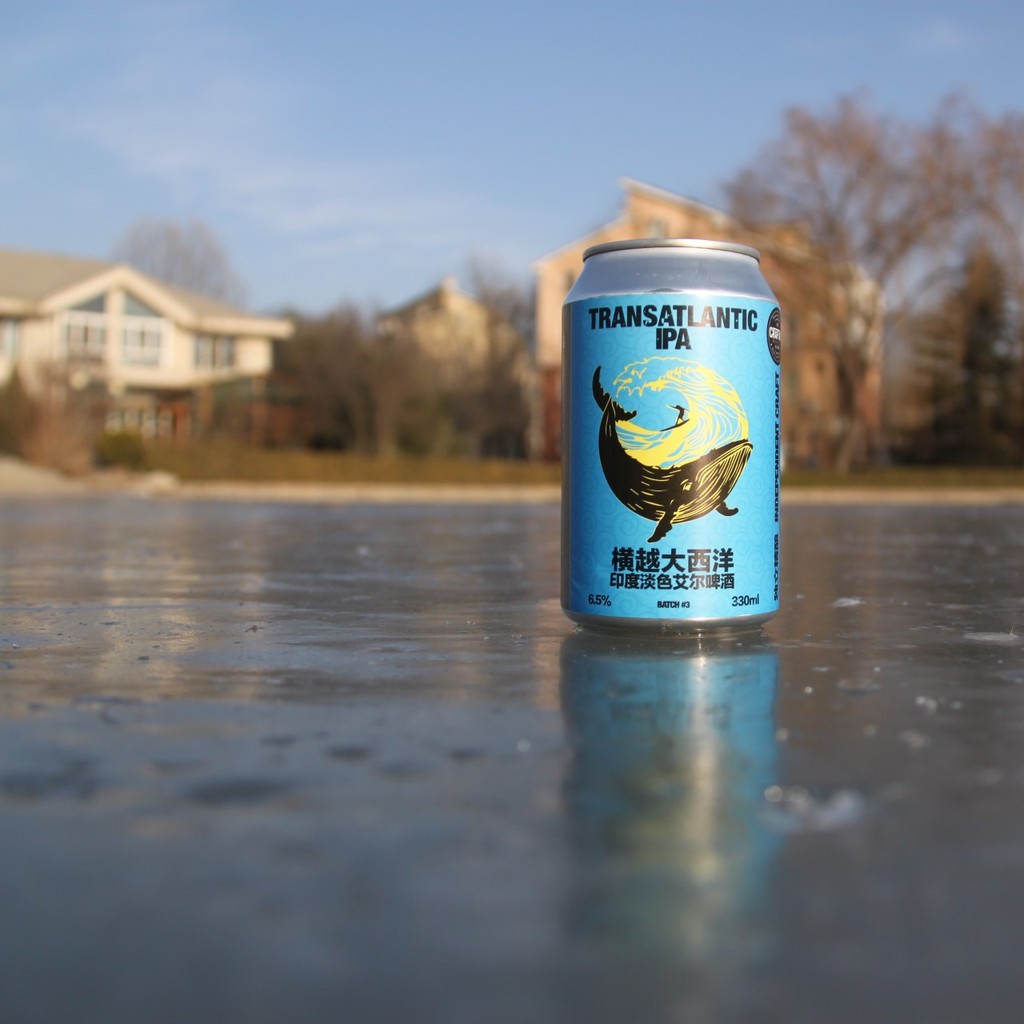}
    &
    \suppTwoResultImg{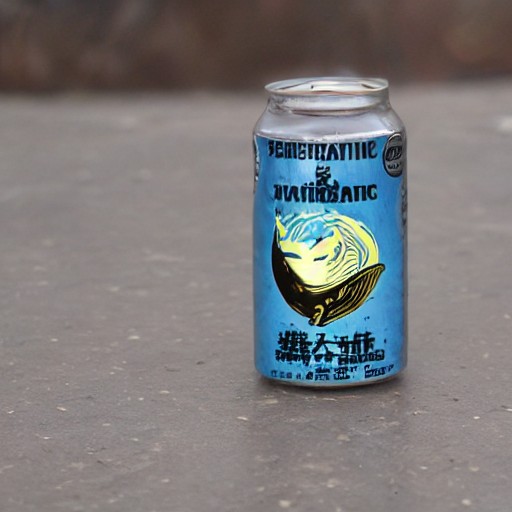}
    &
    \suppTwoResultImg{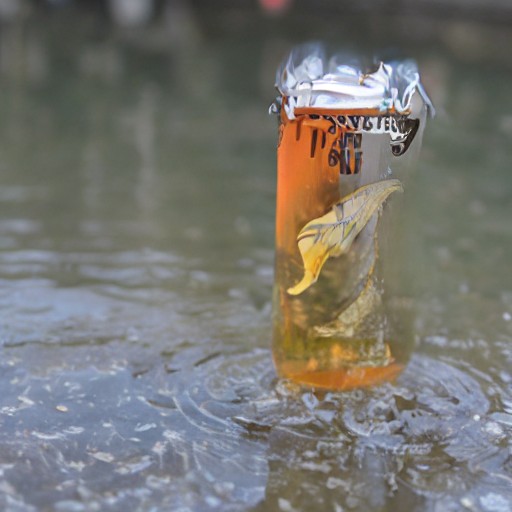}
    &
    \suppTwoResultImg{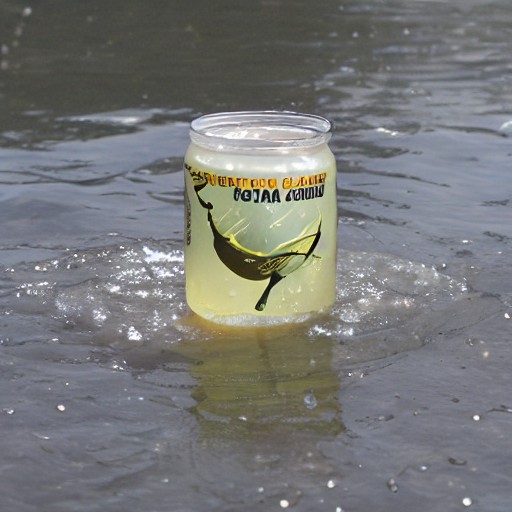}
    &
    \suppTwoResultImg{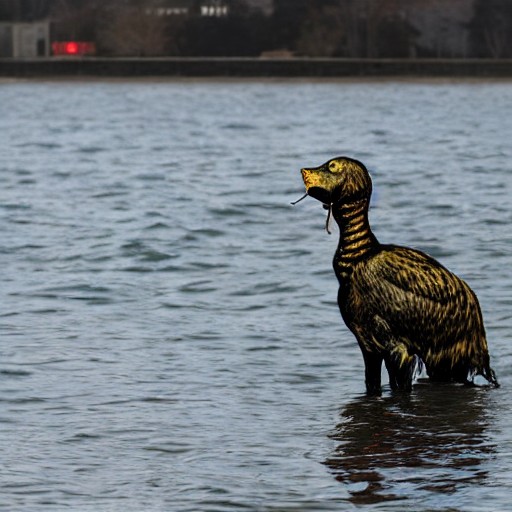}
    &
    \suppTwoResultImg{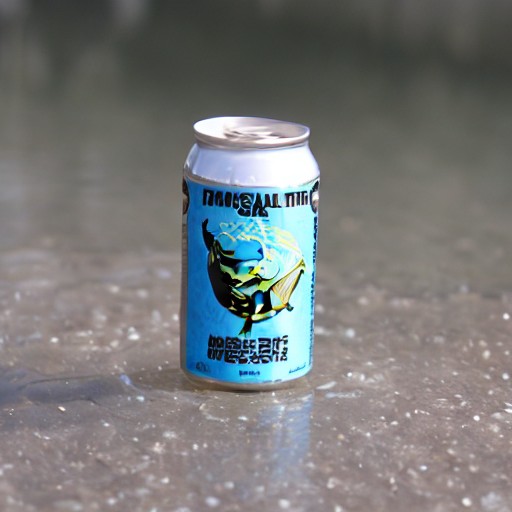}
    &
    \suppTwoResultImg{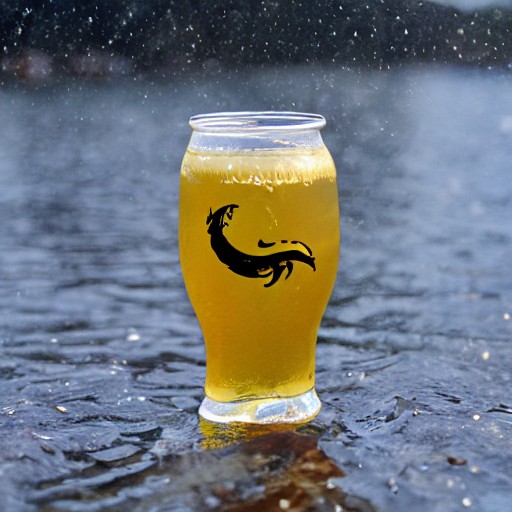}
    &
    \suppTwoResultImg{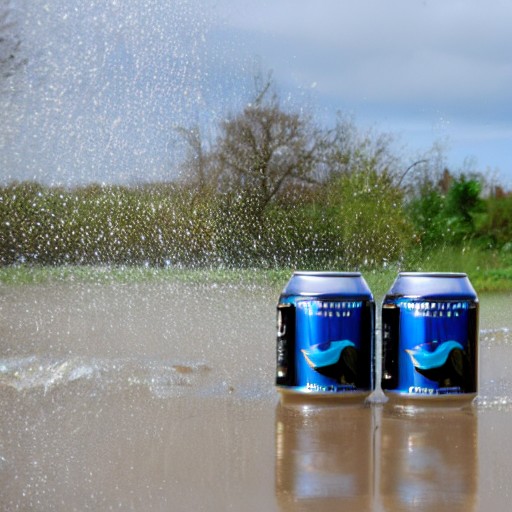}
    &
    \suppTwoResultImg{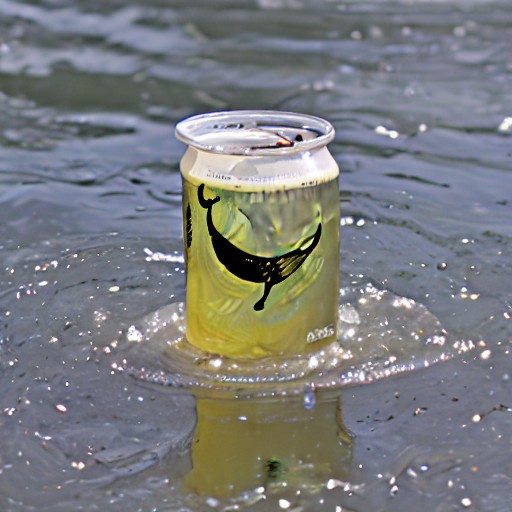}
    &
    \suppTwoResultImg{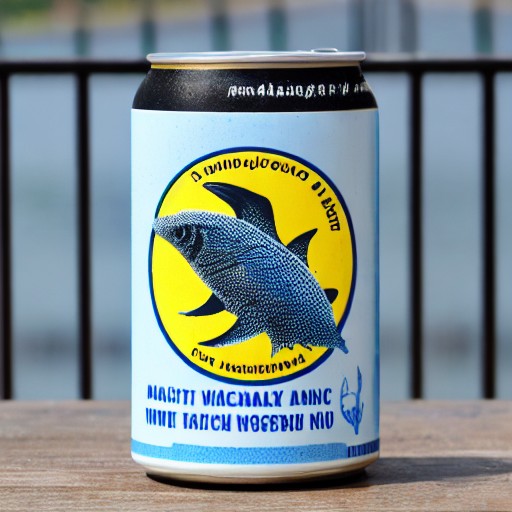}
    &
    \suppTwoResultImg{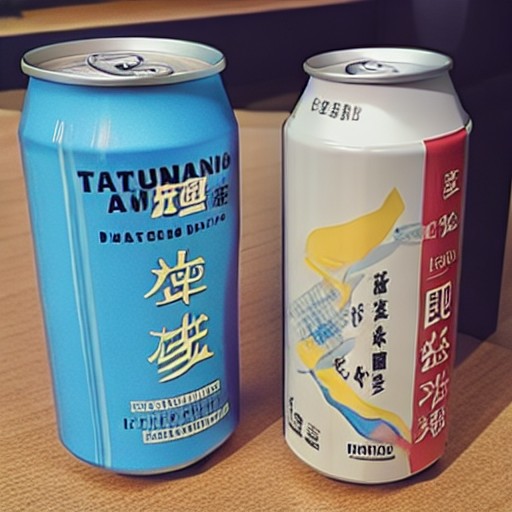}
    \\[-1pt]

    {} &
    \multicolumn{10}{c}{
        {\scriptsize\itshape ``a wet [V*]''}
    }
    \\[5pt]

    % =========================================================
    % Row 2: Toy
    % =========================================================
    \suppTwoFourRefs
        {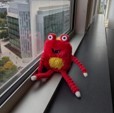}
        {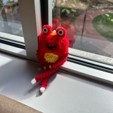}
        {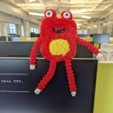}
        {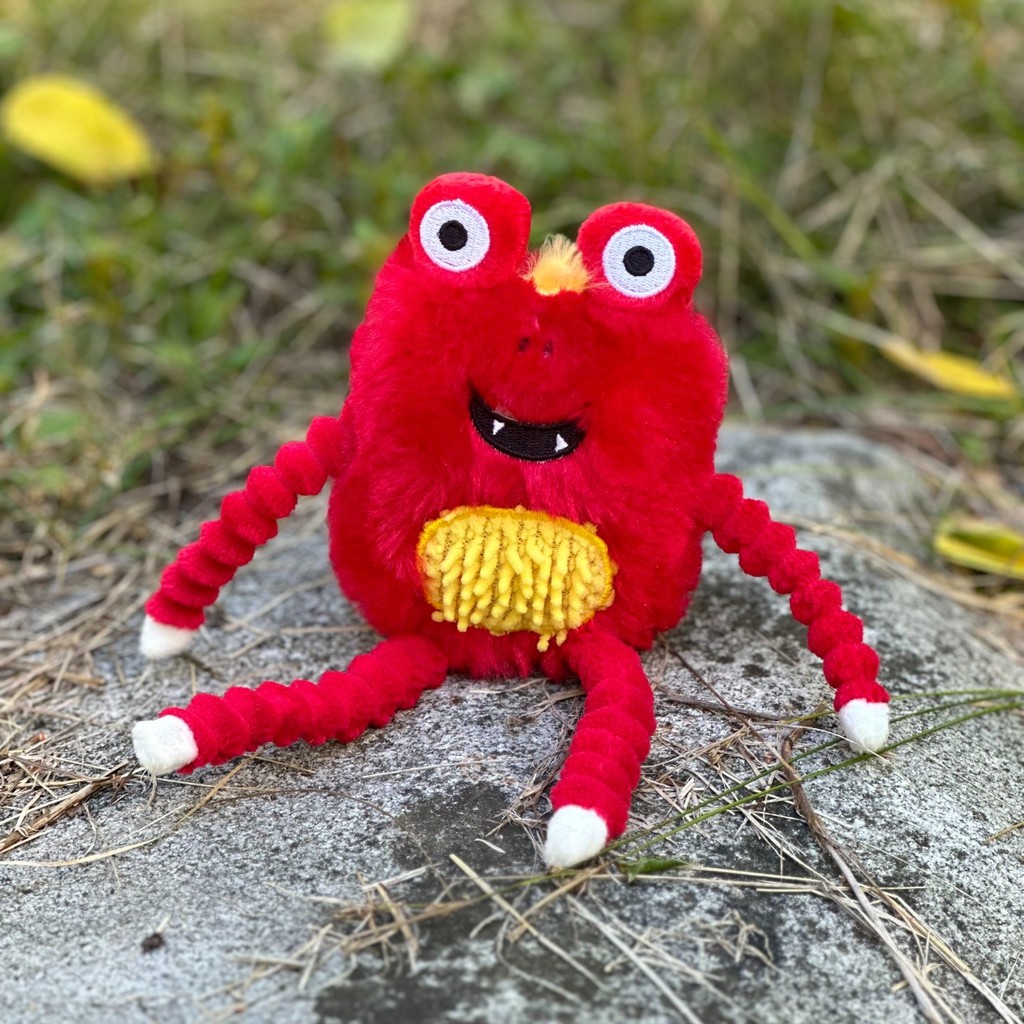}
    &
    \suppTwoResultImg{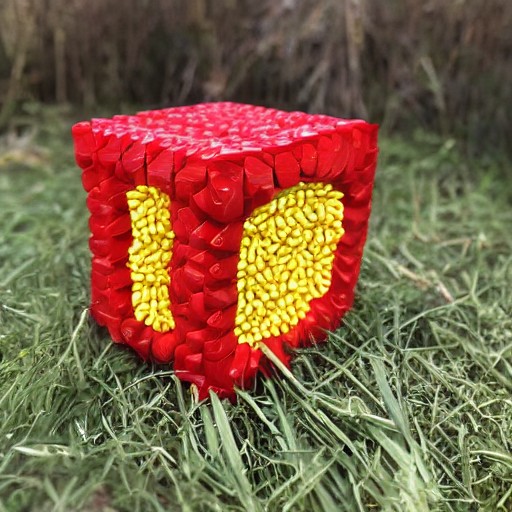}
    &
    \suppTwoResultImg{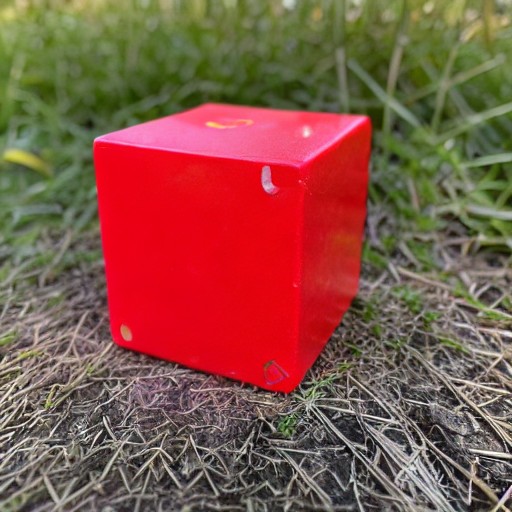}
    &
    \suppTwoResultImg{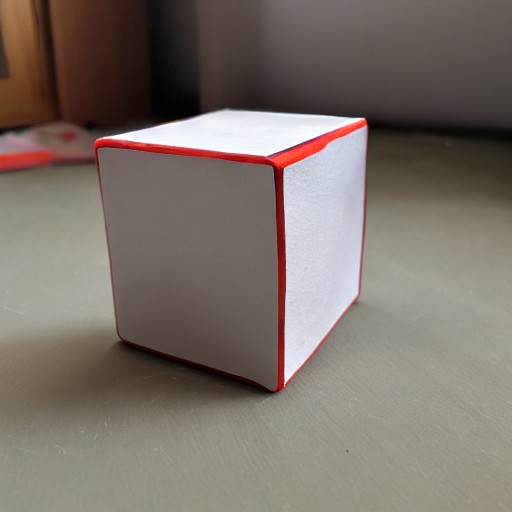}
    &
    \suppTwoResultImg{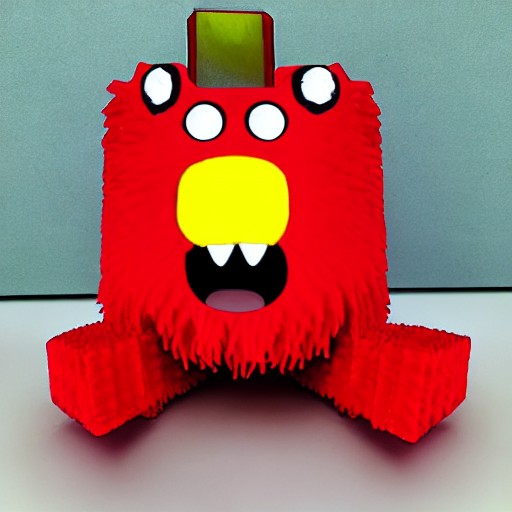}
    &
    \suppTwoResultImg{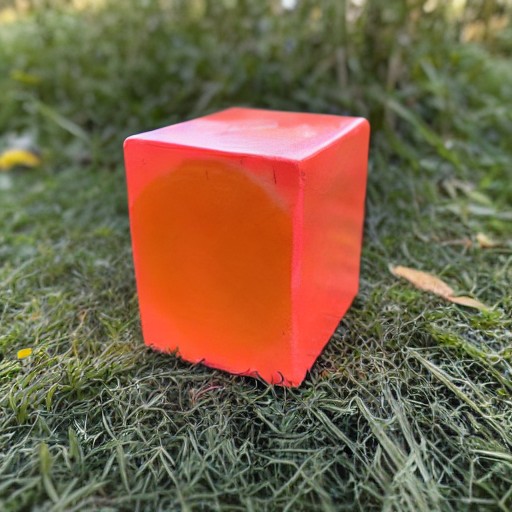}
    &
    \suppTwoResultImg{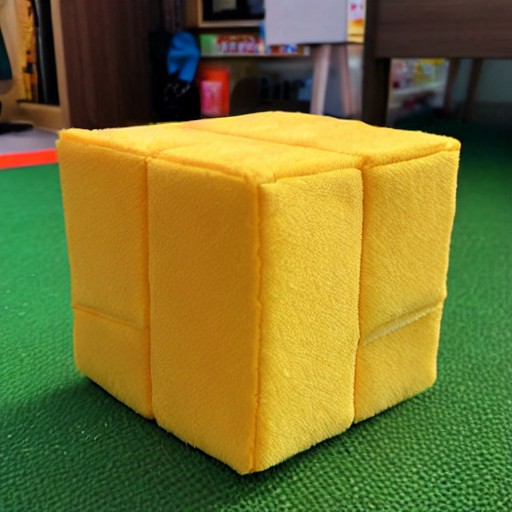}
    &
    \suppTwoResultImg{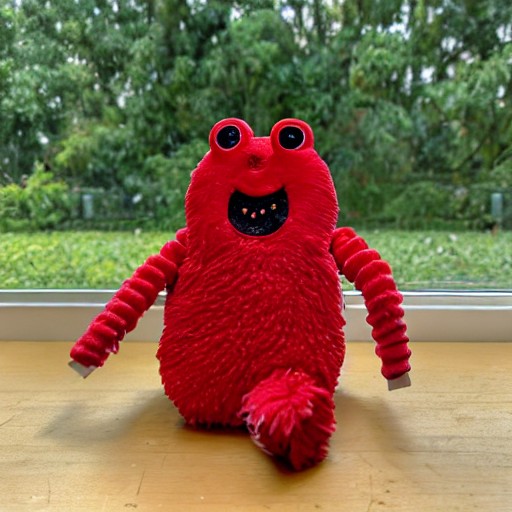}
    &
    \suppTwoResultImg{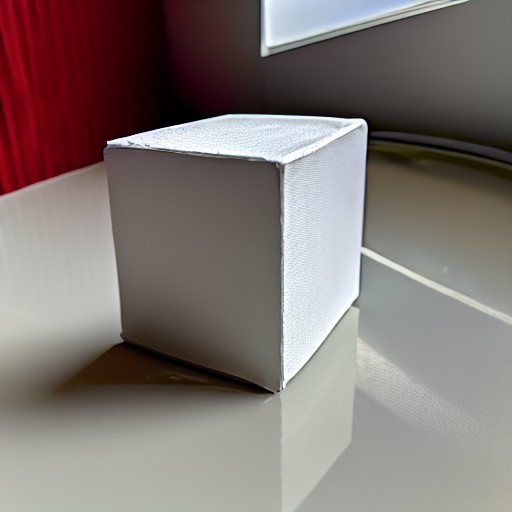}
    &
    \suppTwoResultImg{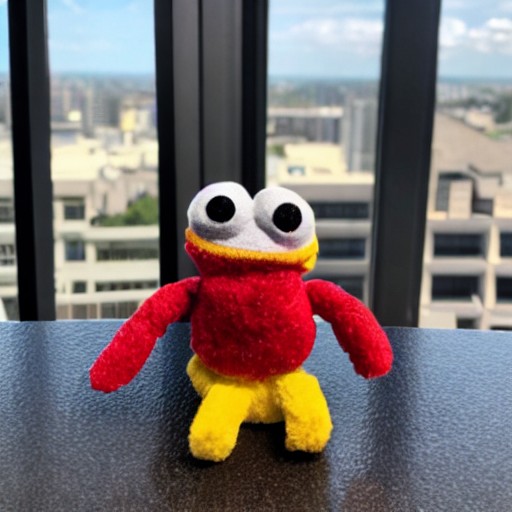}
    &
    \suppTwoResultImg{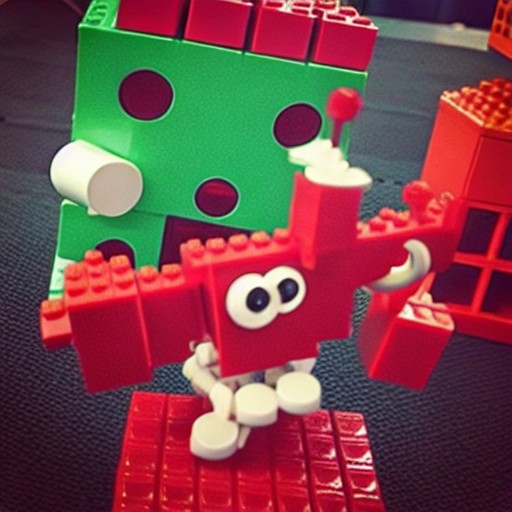}
    \\[-1pt]

    {} &
    \multicolumn{10}{c}{
        {\scriptsize\itshape ``a cube shaped [V*]''}
    }
    \\[5pt]

    % =========================================================
    % Row 3: Monster
    % =========================================================
    \suppTwoFourRefs
        {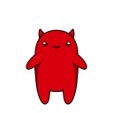}
        {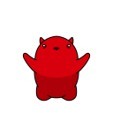}
        {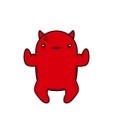}
        {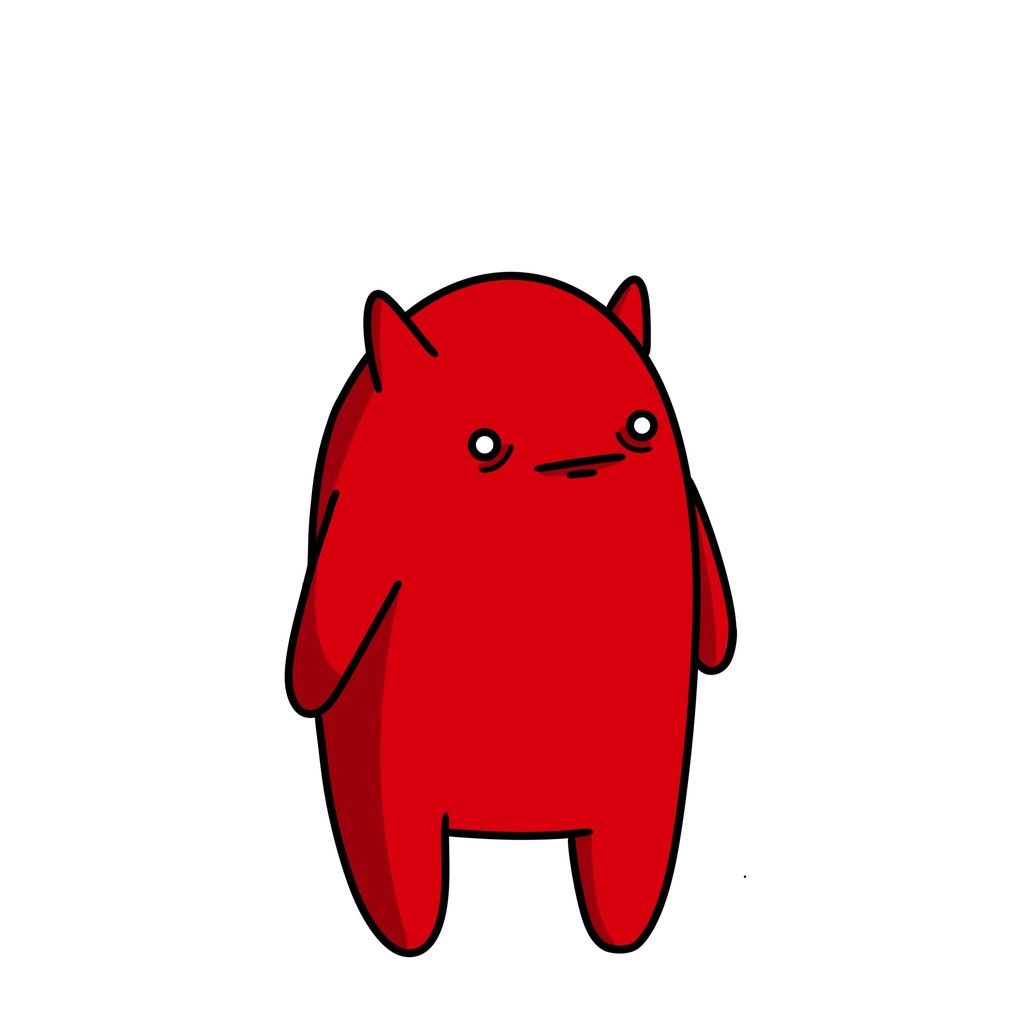}
    &
    \suppTwoResultImg{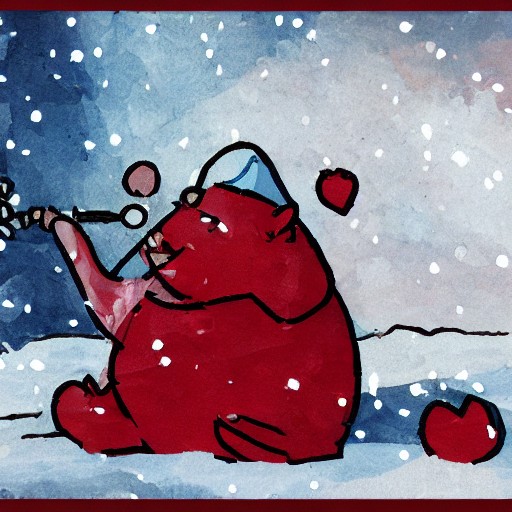}
    &
    \suppTwoResultImg{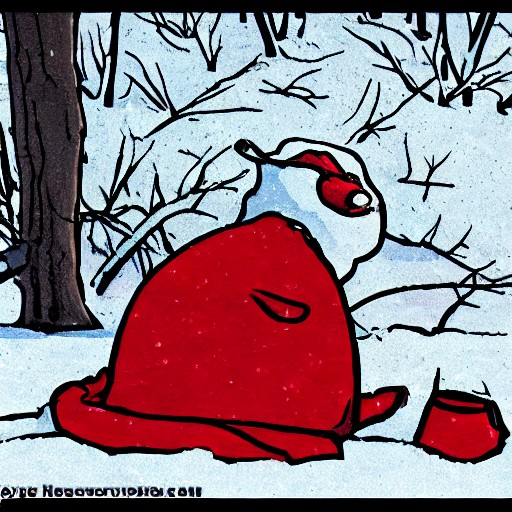}
    &
    \suppTwoResultImg{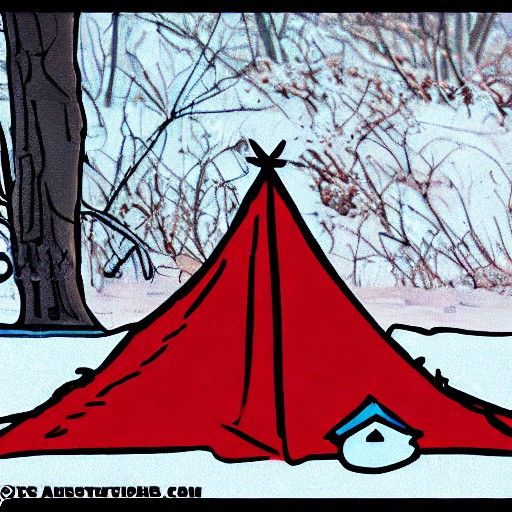}
    &
    \suppTwoResultImg{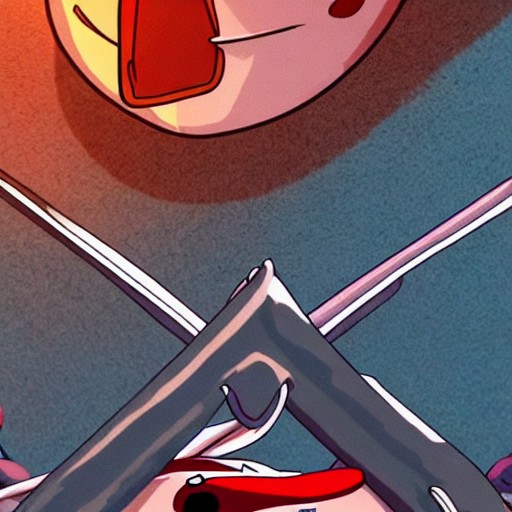}
    &
    \suppTwoResultImg{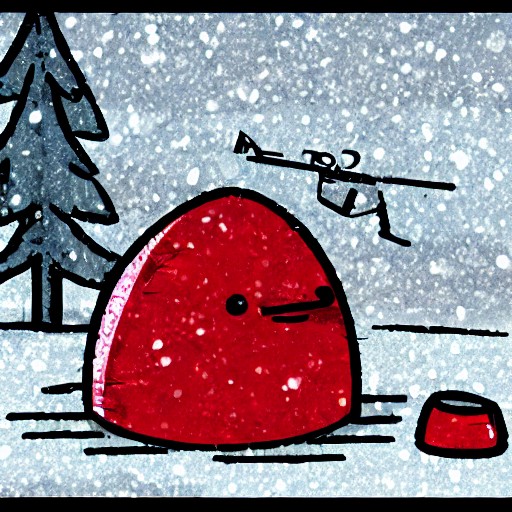}
    &
    \suppTwoResultImg{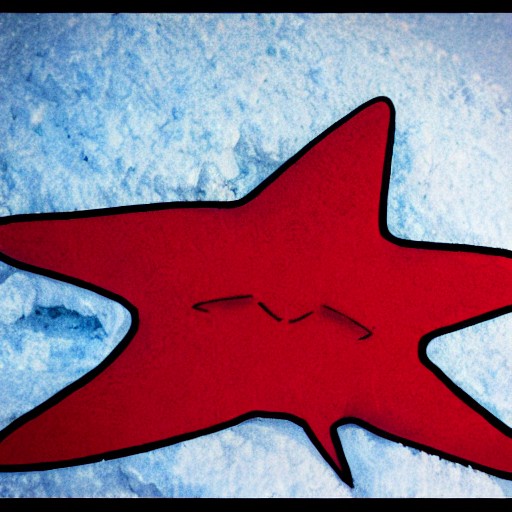}
    &
    \suppTwoResultImg{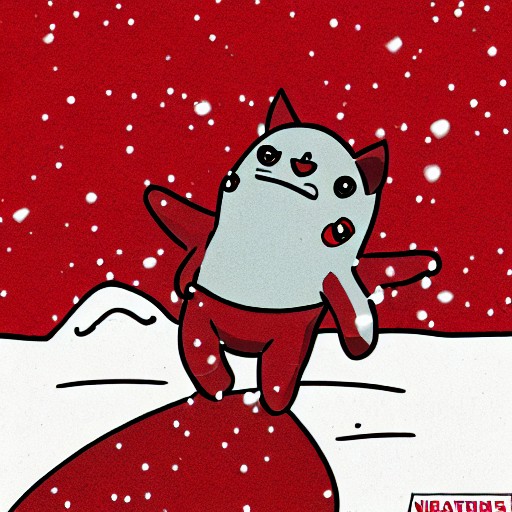}
    &
    \suppTwoResultImg{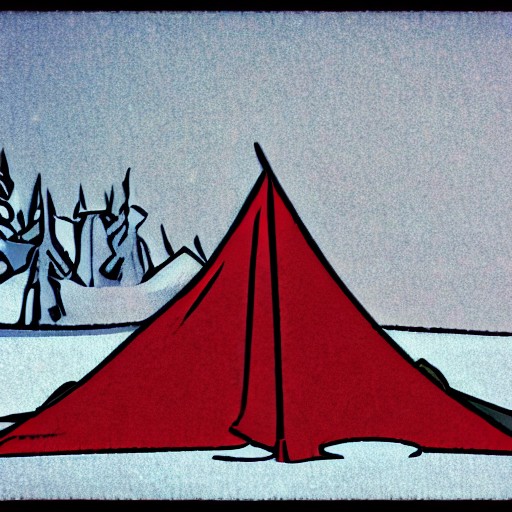}
    &
    \suppTwoResultImg{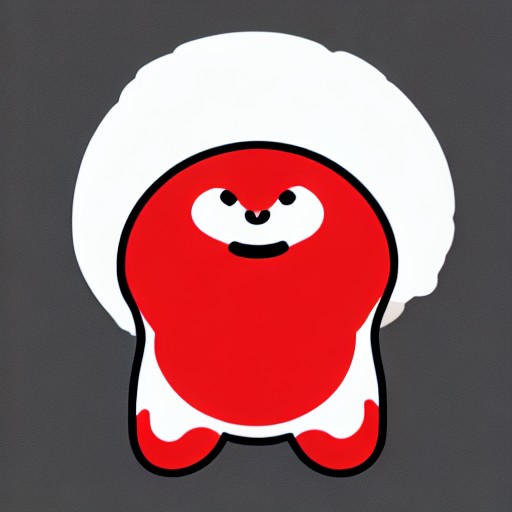}
    &
    \suppTwoResultImg{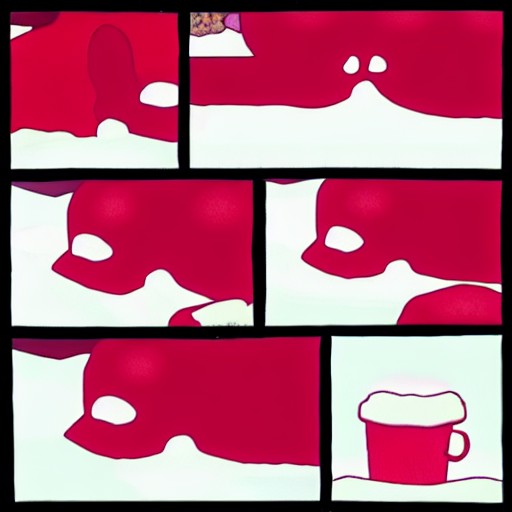}
    \\[-1pt]

    {} &
    \multicolumn{10}{c}{
        {\scriptsize\itshape ``a [V*] in the snow''}
    }
    \\[1pt]

    \end{tabular}

    \caption{
        Additional visual comparison of few-shot personalization and
        encoder-based methods on the SD1.5 backbone, including
        representative failure cases.
    }
    \label{fig:supp_sd15_2}

\end{sidewaysfigure}

%--------------------------------------------------

% \begin{figure*}[t]
%     \centering
%     \includegraphics[width=\linewidth,
%                      height=0.9\textheight,
%                      keepaspectratio]{figure/supp_sd15sdxlsd3.png}
%     \caption{Additional qualitative comparison on SD1.5, SDXL and SD3 backbones among DreamBooth, Beyond-Finetuning, and our method.}
%     \label{fig:supp_sd15sdxlsd3}
% \end{figure*}

% \input{figure/supp_sd15sdxlsd3}
\begin{figure*}[t]
    \centering

    \setlength{\tabcolsep}{2pt}
    \renewcommand{\arraystretch}{1.0}

    % =========================================================
    % References: 간격 없는 2×2 배열
    % 전체 너비 = Reference 열 너비
    % =========================================================
    \newlength{\suppMultiRefTileSize}
    \setlength{\suppMultiRefTileSize}{0.16\textwidth}

    \newcommand{\suppMultiFourRefs}[4]{%
        \vbox{%
            \offinterlineskip
            \hbox{%
                \includegraphics[
                    width=\suppMultiRefTileSize,
                    height=\suppMultiRefTileSize
                ]{#1}%
                \includegraphics[
                    width=\suppMultiRefTileSize,
                    height=\suppMultiRefTileSize
                ]{#2}%
            }%
            \hbox{%
                \includegraphics[
                    width=\suppMultiRefTileSize,
                    height=\suppMultiRefTileSize
                ]{#3}%
                \includegraphics[
                    width=\suppMultiRefTileSize,
                    height=\suppMultiRefTileSize
                ]{#4}%
            }%
        }%
    }

    % =========================================================
    % Keep the entire figure within one page
    % =========================================================
    \begin{adjustbox}{
        max width=\textwidth,
        max totalheight=0.82\textheight,
        center
    }

    \begin{tabular}{
        @{}
        C{0.32\textwidth}
        C{0.070\textwidth}
        *{3}{C{0.205\textwidth}}
        @{}
    }

    % =========================================================
    % Header
    % =========================================================
    \suppheader{References}
    &
    {}
    &
    \suppheader{DreamBooth\\-LoRA}
    &
    \suppheader{Beyond-\\Finetuning}
    &
    \suppheader{Ours}
    \\[1pt]

    % =========================================================
    % BACKPACK
    % =========================================================

    % SD1.5
    {}
    &
    \suppbackbone{SD1.5}
    &
    \suppimg{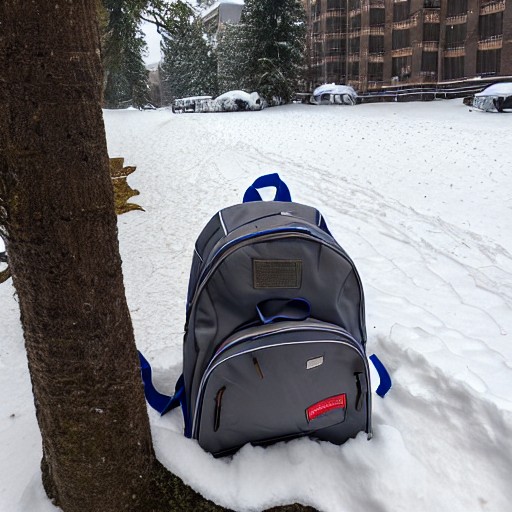}
    &
    \suppimg{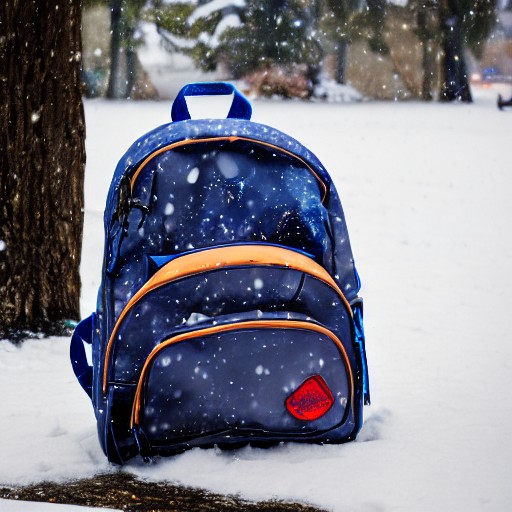}
    &
    \suppimg{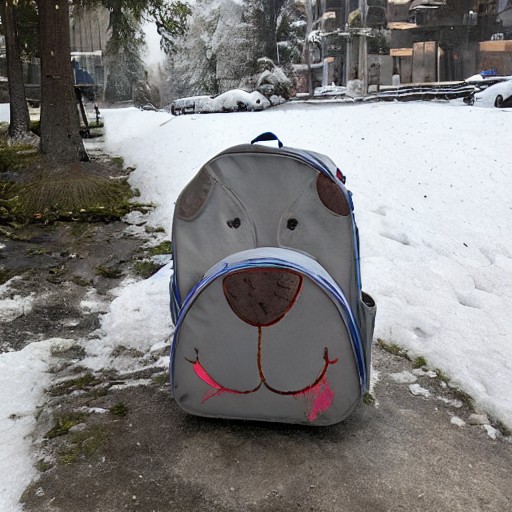}
    \\[0pt]

    % SDXL + References
    \smash{%
        \raisebox{-0.5\height}{%
            \makebox[\linewidth][c]{%
                \suppMultiFourRefs
                    {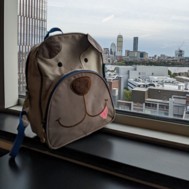}
                    {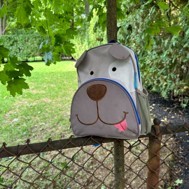}
                    {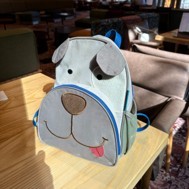}
                    {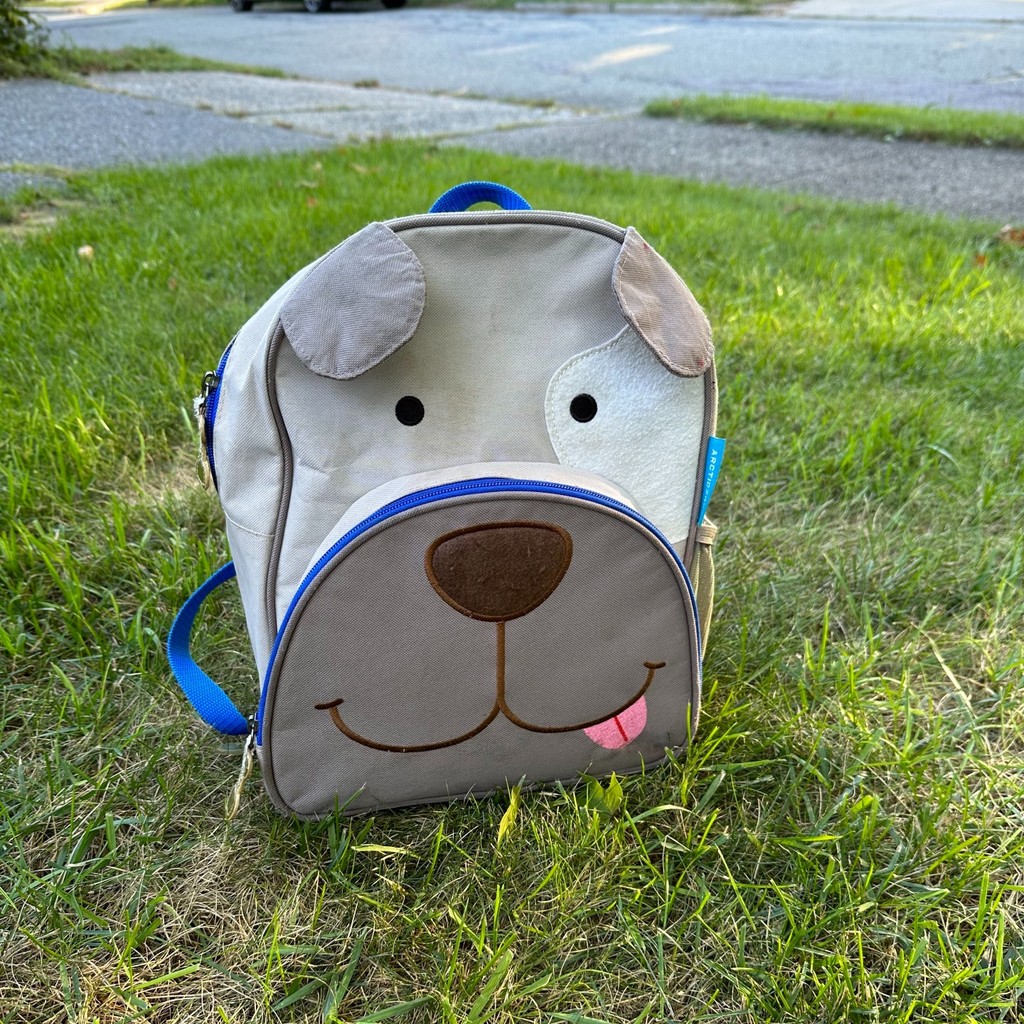}
            }%
        }%
    }
    &
    \suppbackbone{SDXL}
    &
    \suppimg{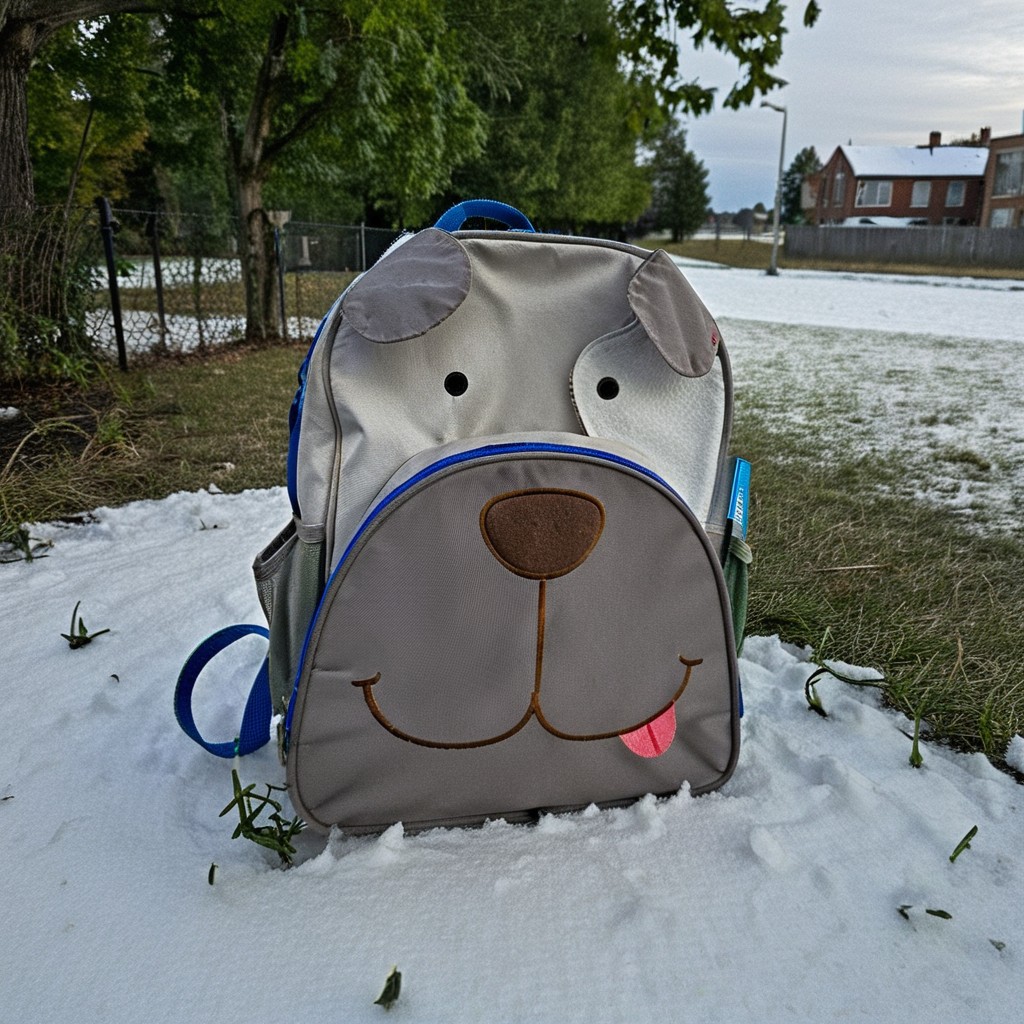}
    &
    \suppimg{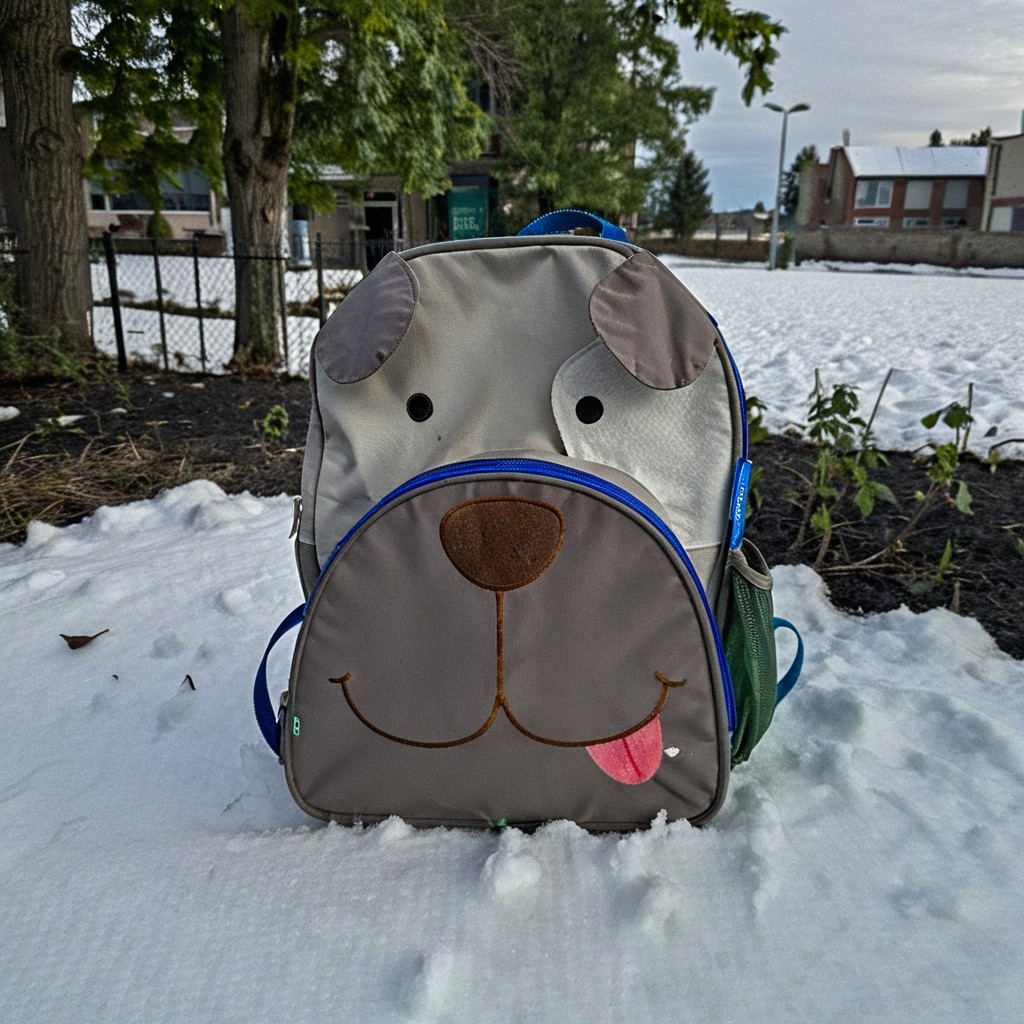}
    &
    \suppimg{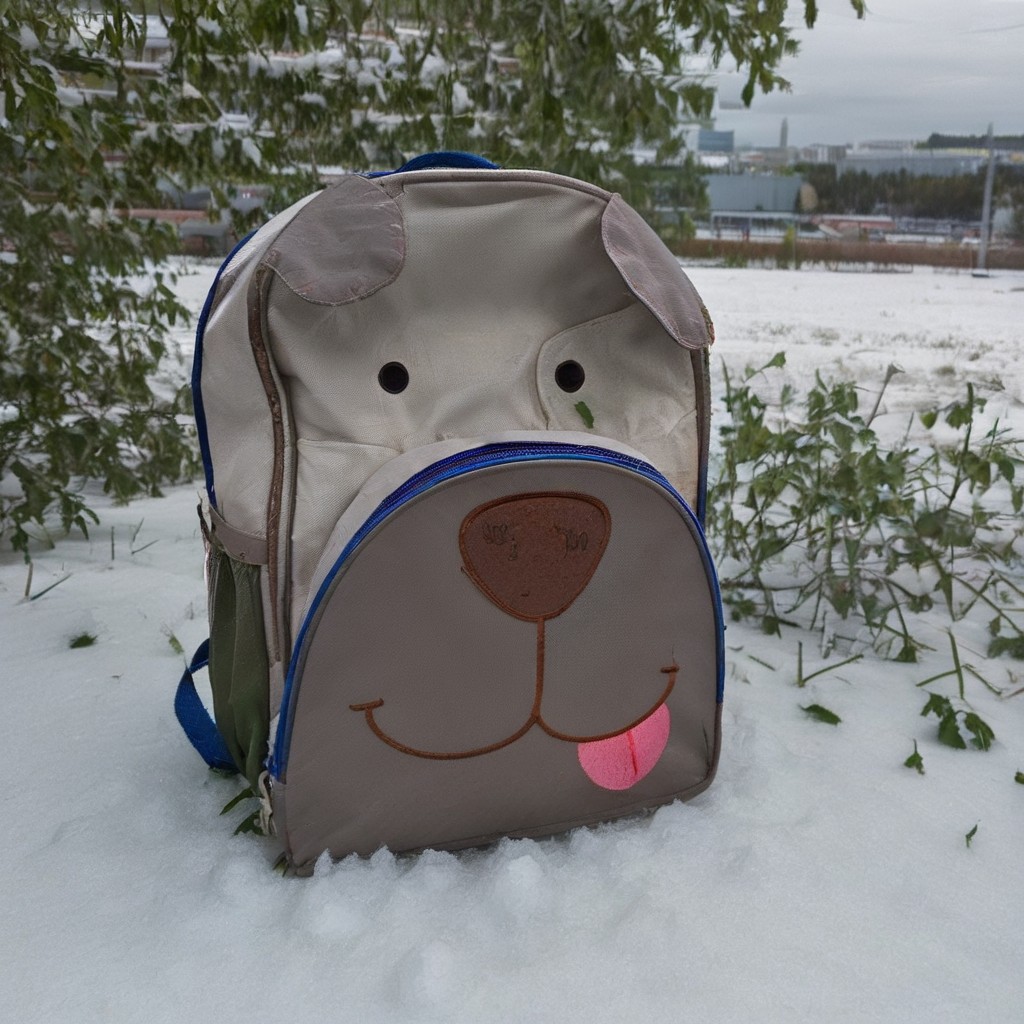}
    \\[0pt]

    % SD3
    {}
    &
    \suppbackbone{SD3}
    &
    \suppimg{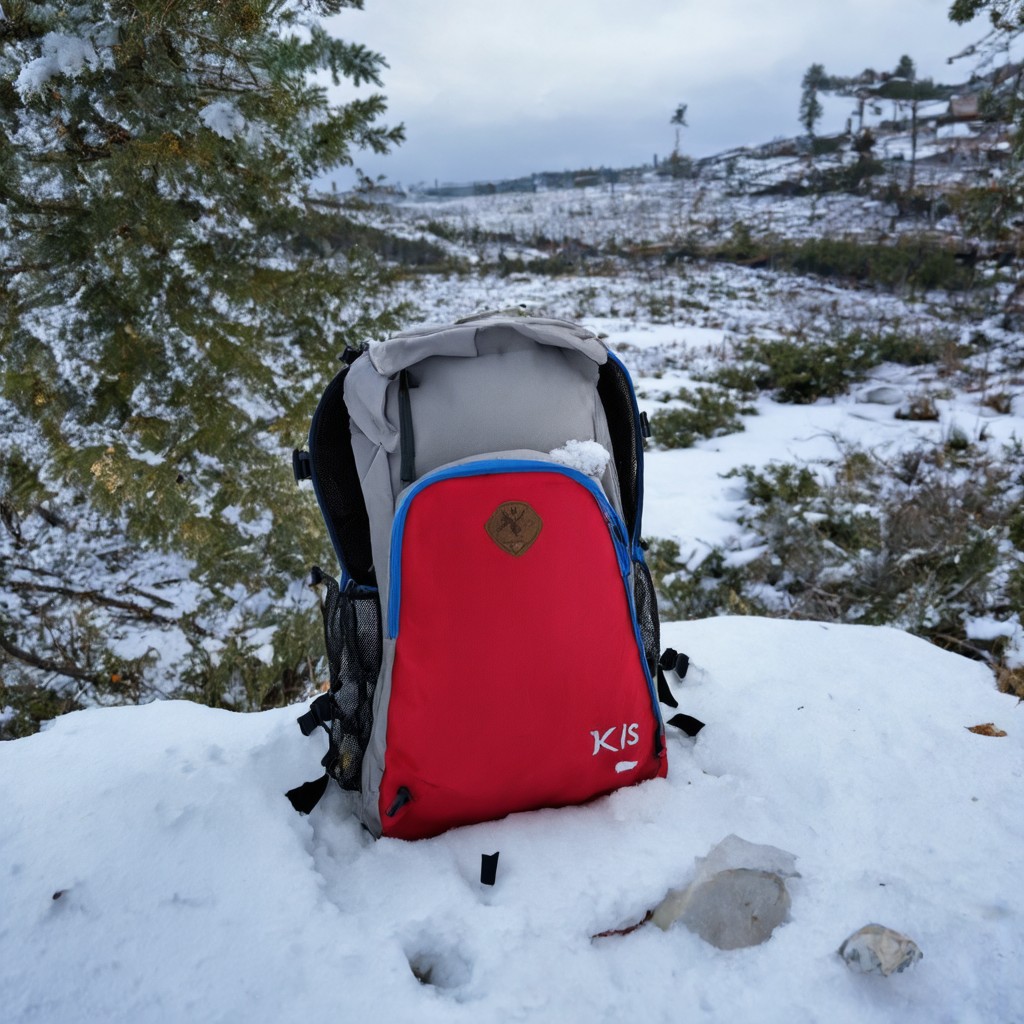}
    &
    \suppimg{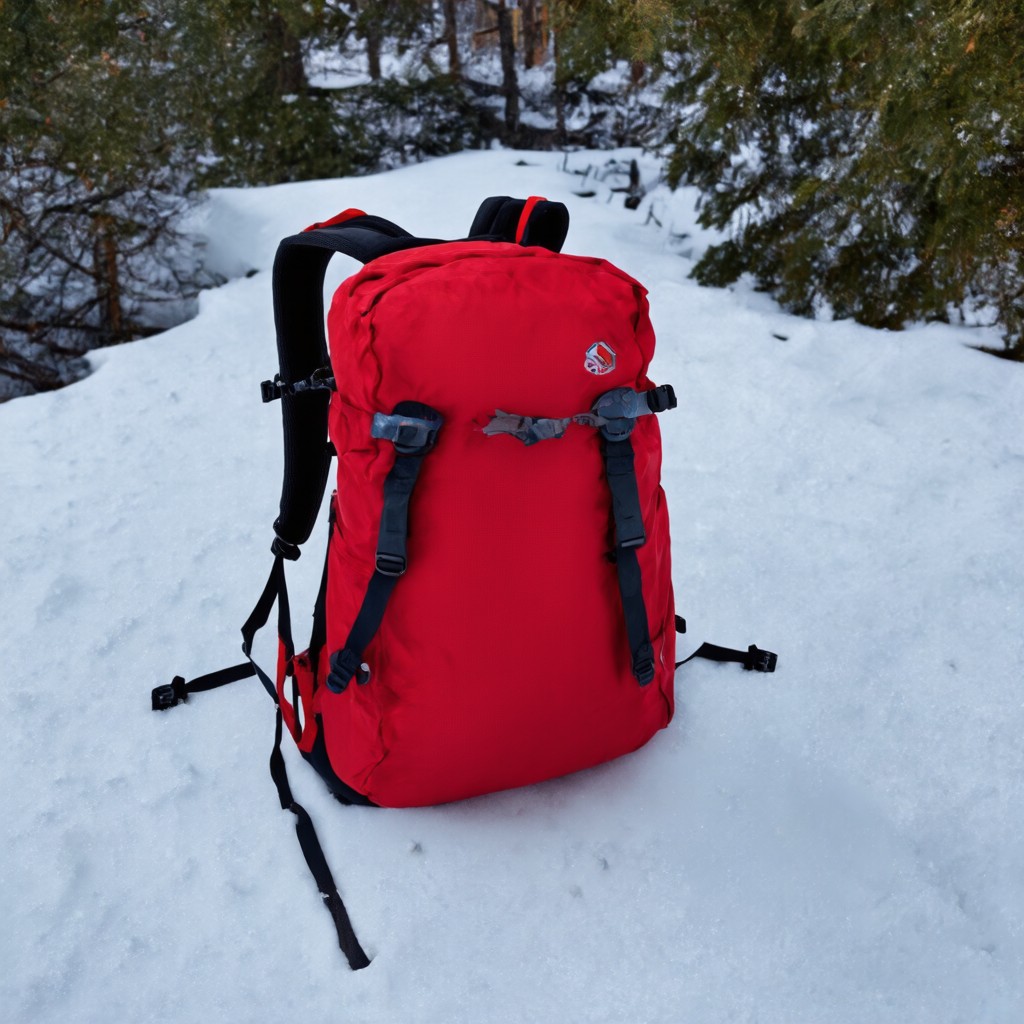}
    &
    \suppimg{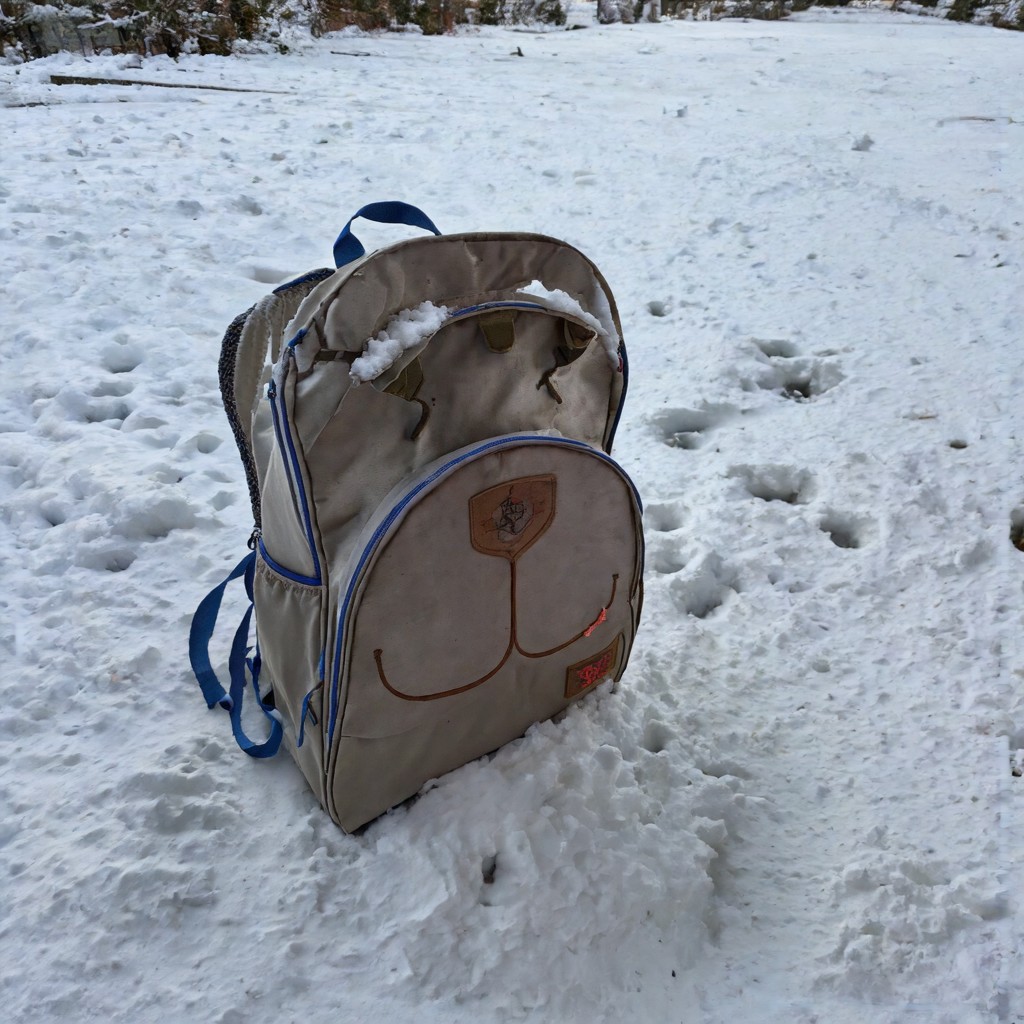}
    \\[-3pt]

    % Prompt
    {}
    &
    {}
    &
    \multicolumn{3}{c}{%
        \suppprompt{``a [V*] in the snow''}
    }
    \\[3pt]

    % =========================================================
    % SLOTH
    % =========================================================

    % SD1.5
    {}
    &
    \suppbackbone{SD1.5}
    &
    \suppimg{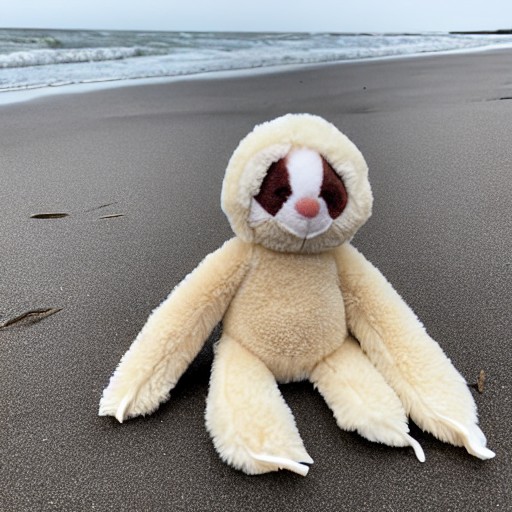}
    &
    \suppimg{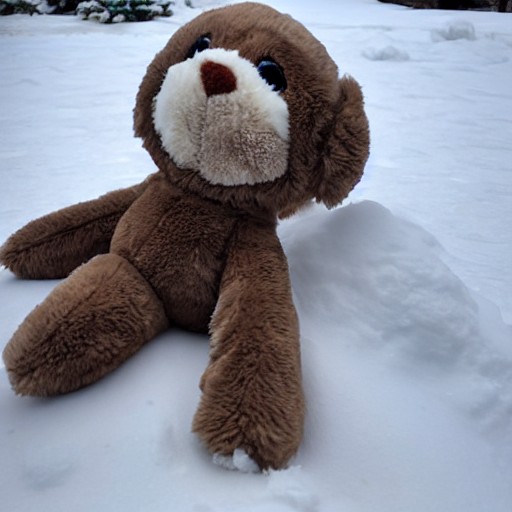}
    &
    \suppimg{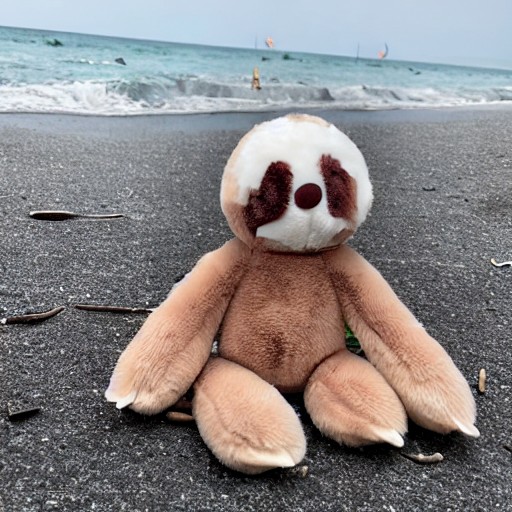}
    \\[0pt]

    % SDXL + References
    \smash{%
        \raisebox{-0.5\height}{%
            \makebox[\linewidth][c]{%
                \suppMultiFourRefs
                    {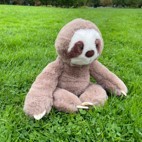}
                    {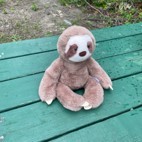}
                    {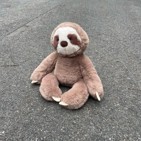}
                    {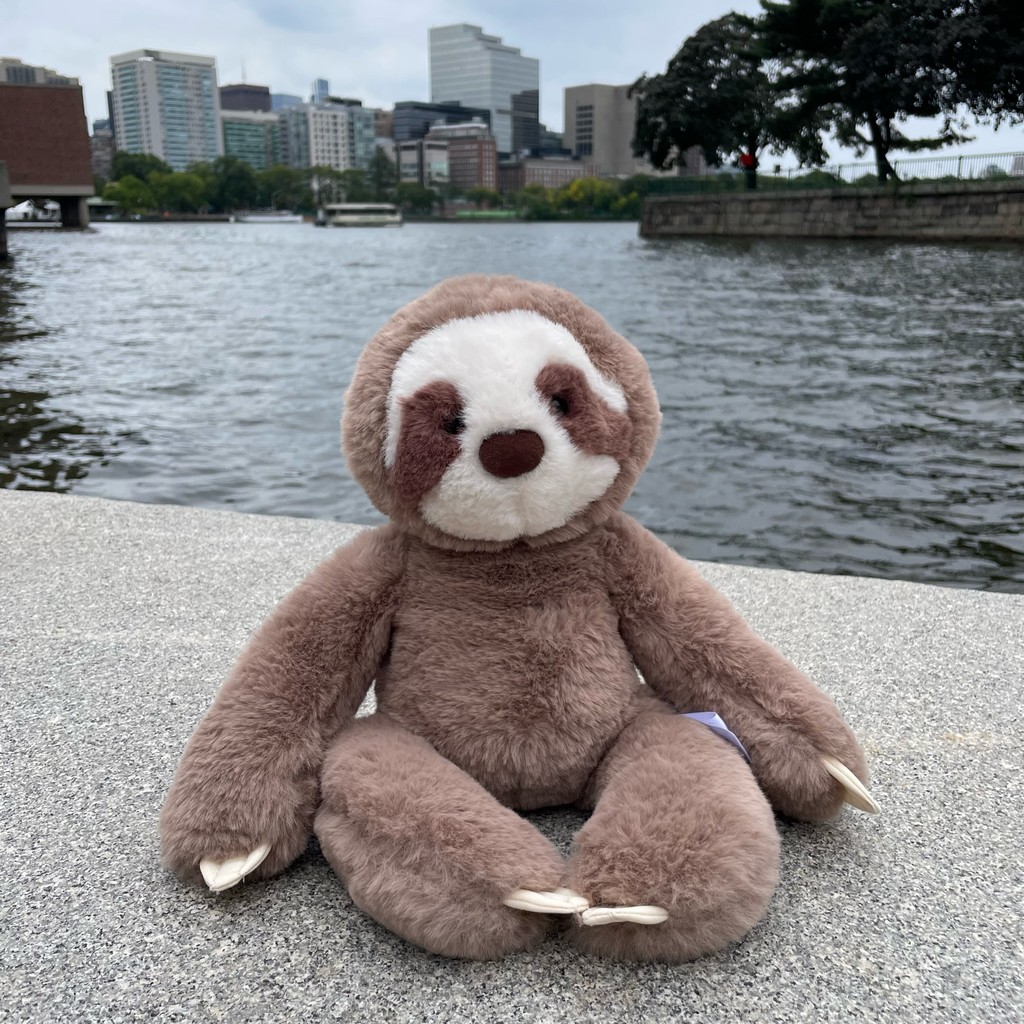}
            }%
        }%
    }
    &
    \suppbackbone{SDXL}
    &
    \suppimg{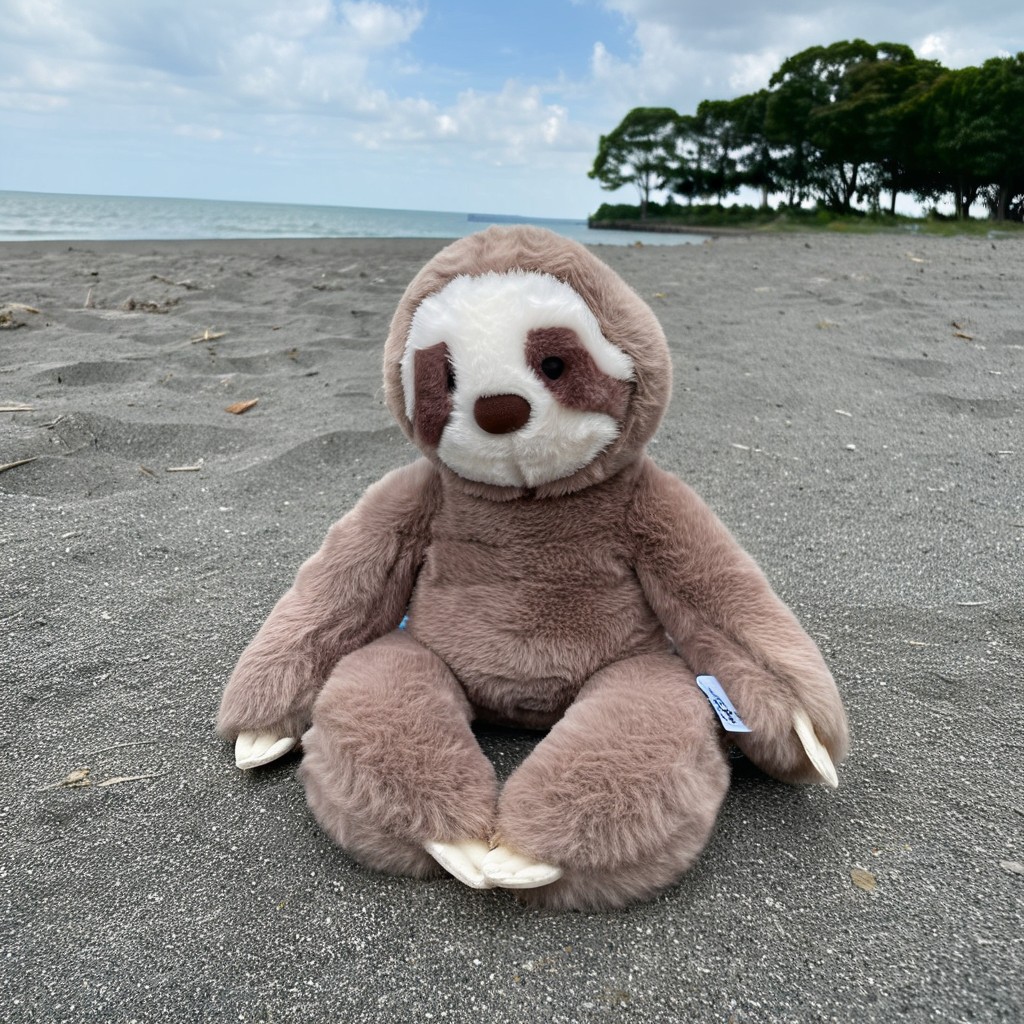}
    &
    \suppimg{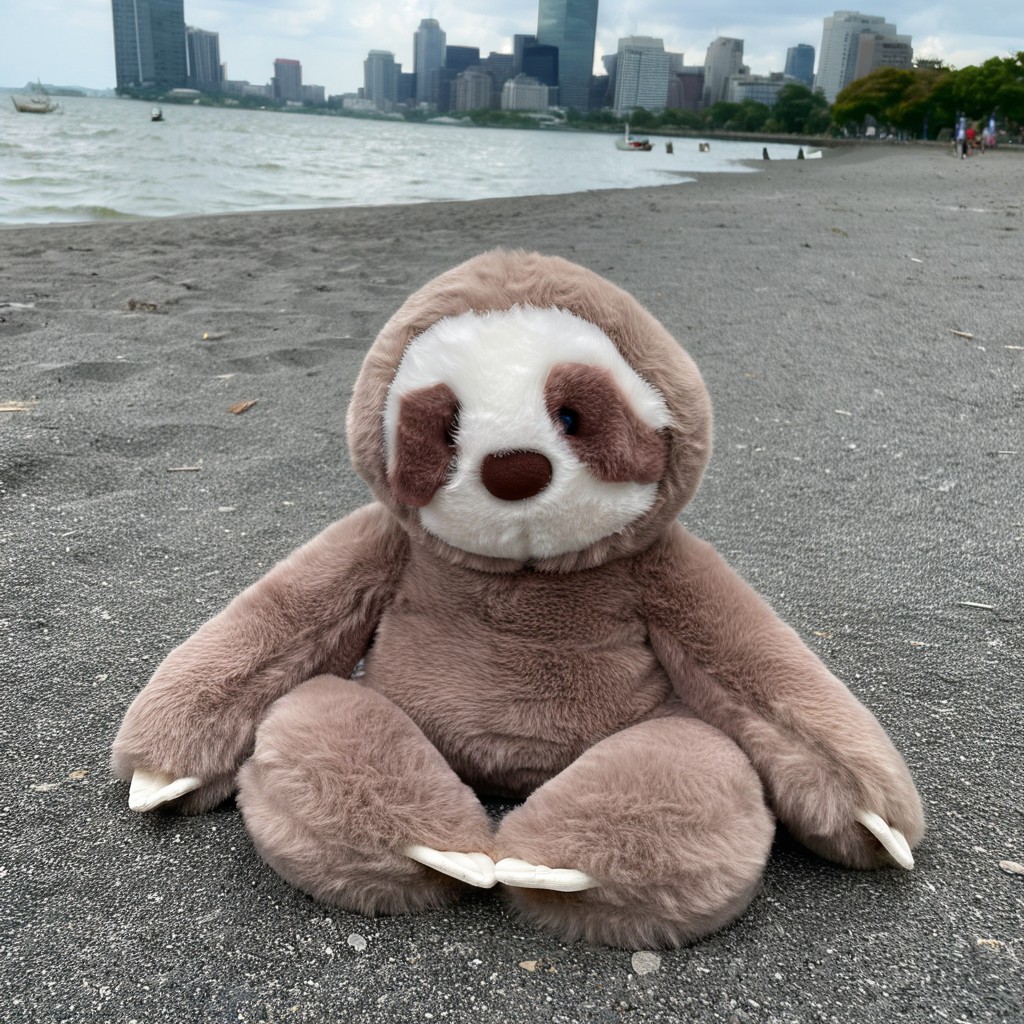}
    &
    \suppimg{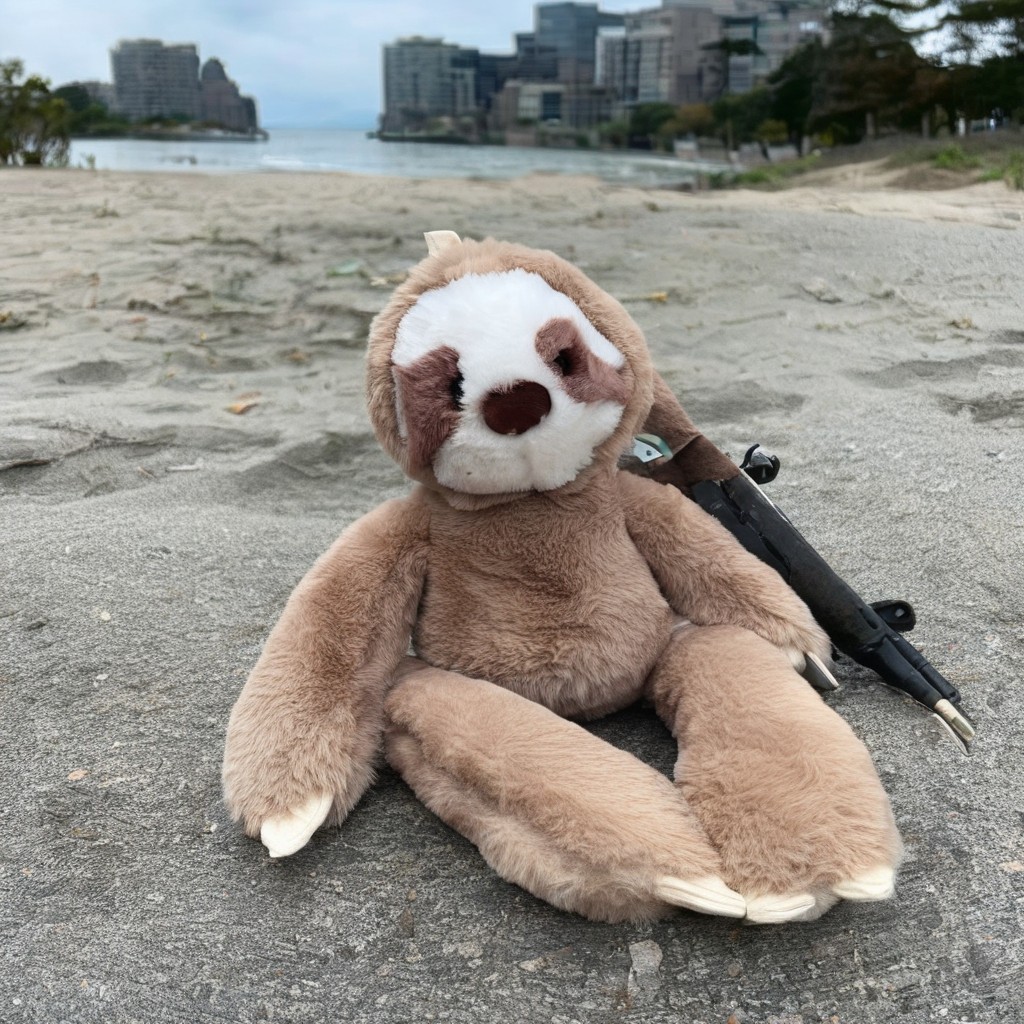}
    \\[0pt]

    % SD3
    {}
    &
    \suppbackbone{SD3}
    &
    \suppimg{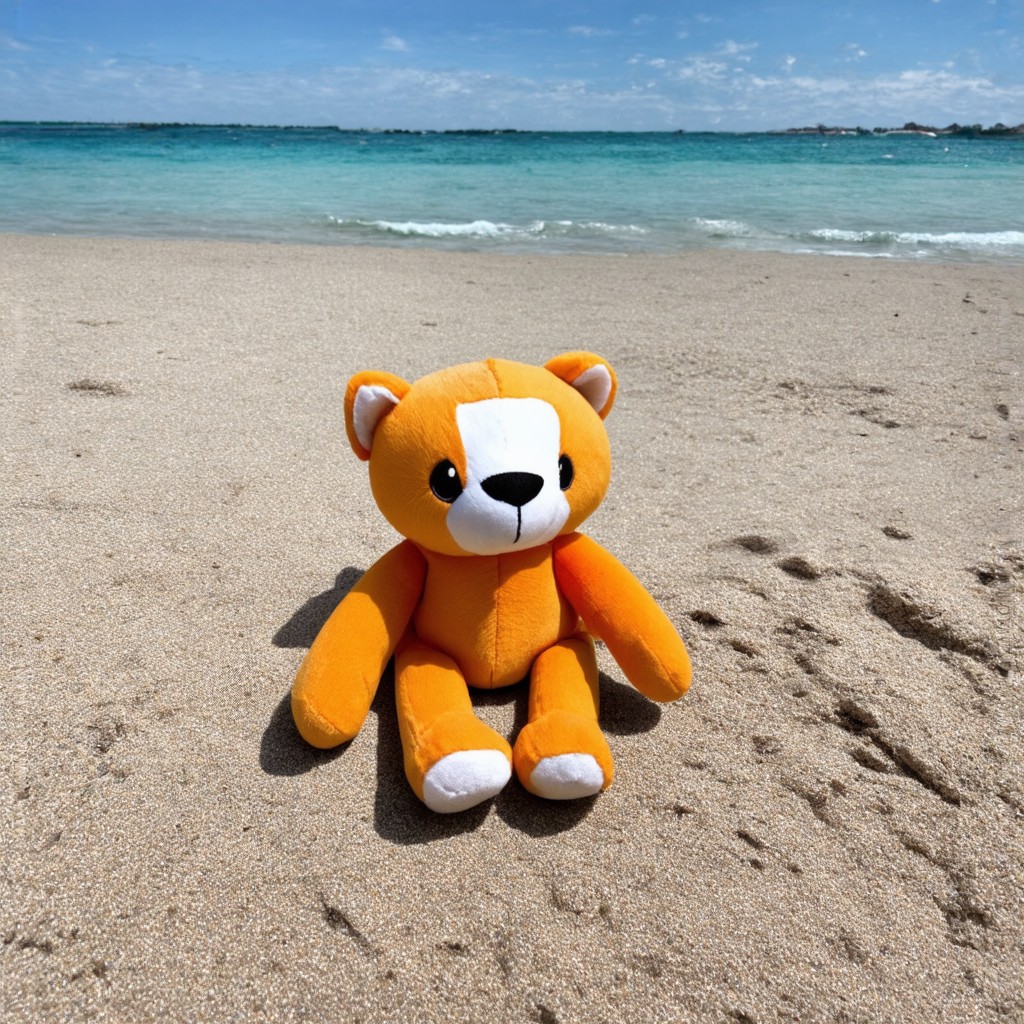}
    &
    \suppimg{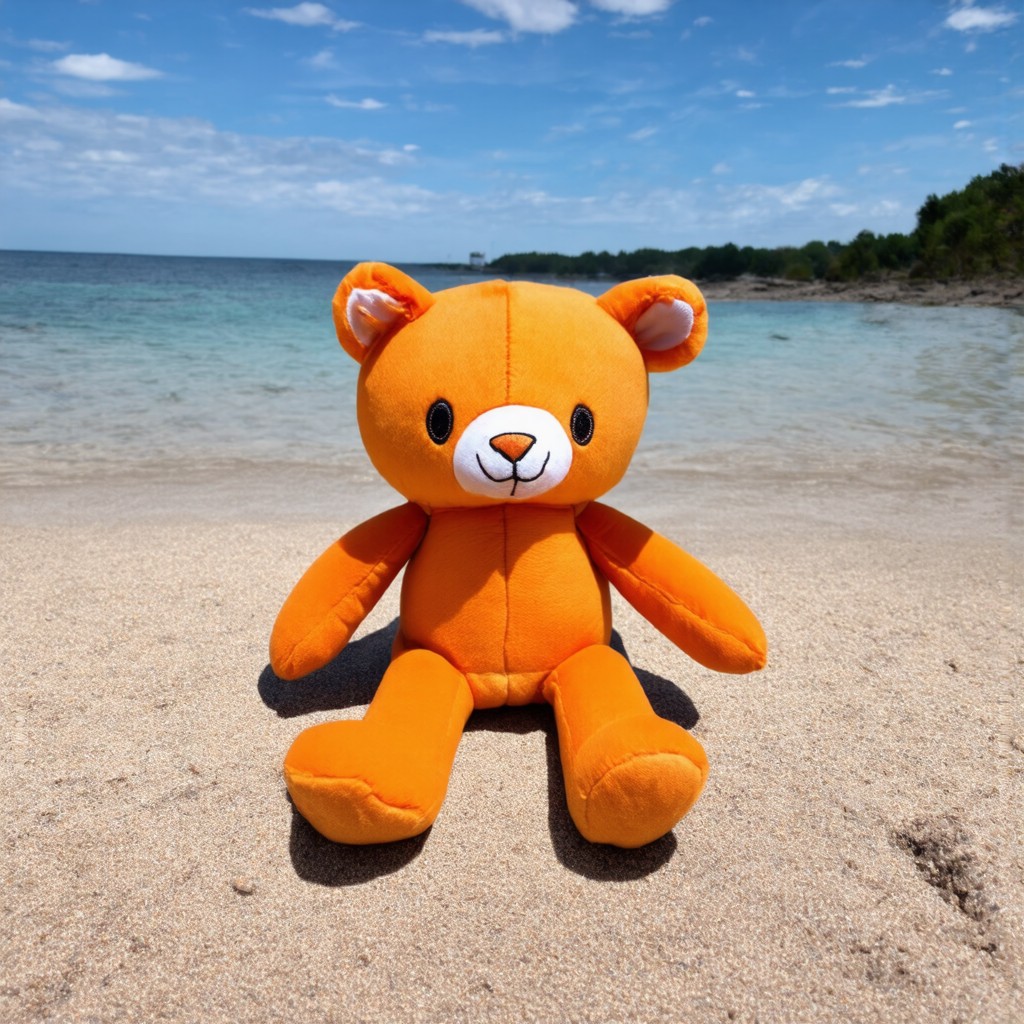}
    &
    \suppimg{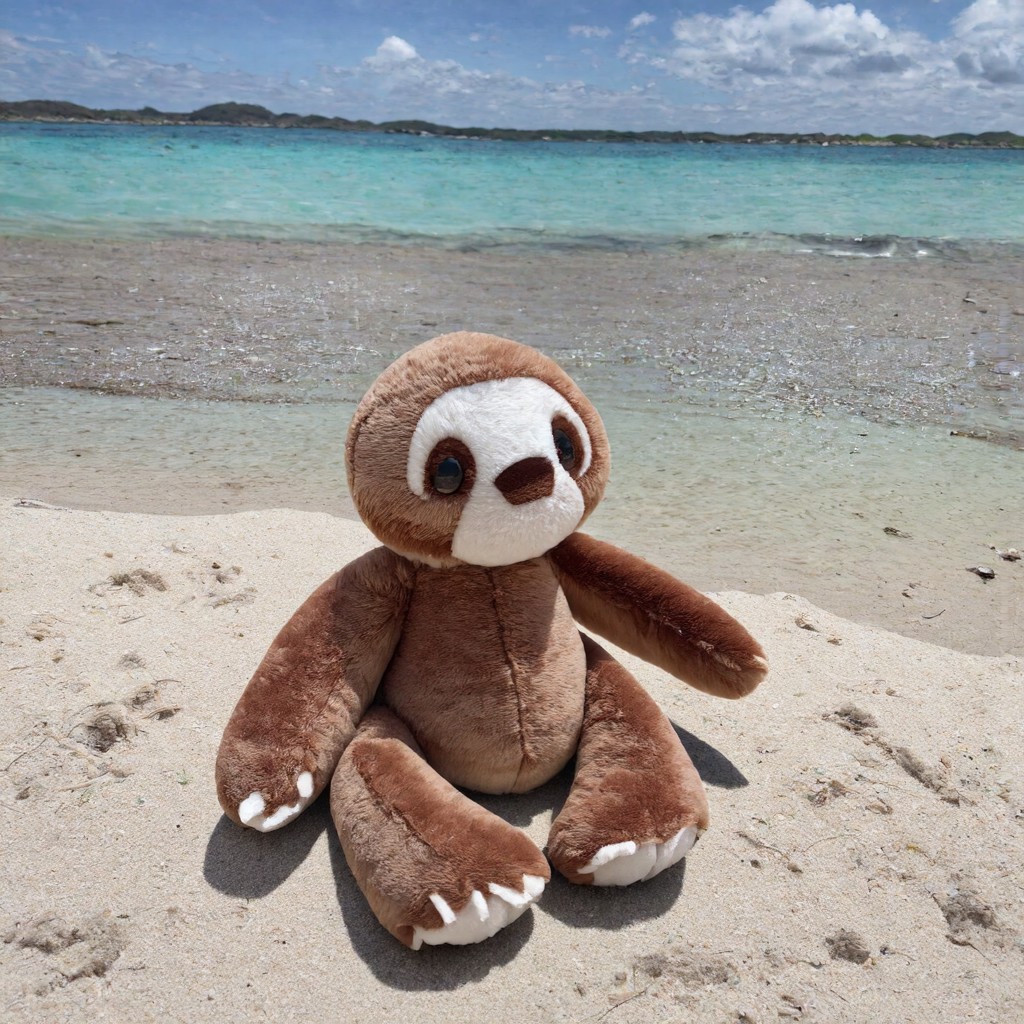}
    \\[-3pt]

    % Prompt
    {}
    &
    {}
    &
    \multicolumn{3}{c}{%
        \suppprompt{``a [V*] on the beach''}
    }
    \\[3pt]

    % =========================================================
    % ROBOT
    % =========================================================

    % SD1.5
    {}
    &
    \suppbackbone{SD1.5}
    &
    \suppimg{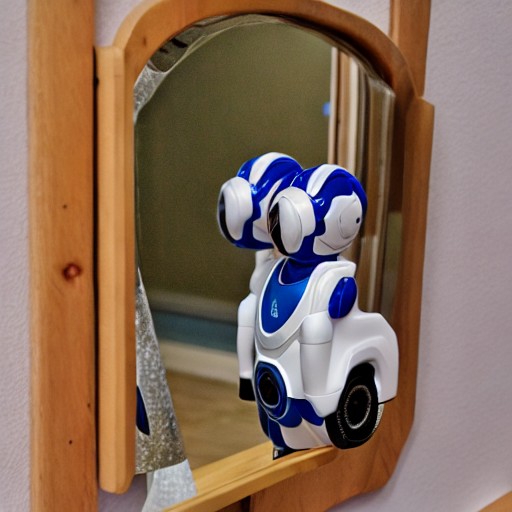}
    &
    \suppimg{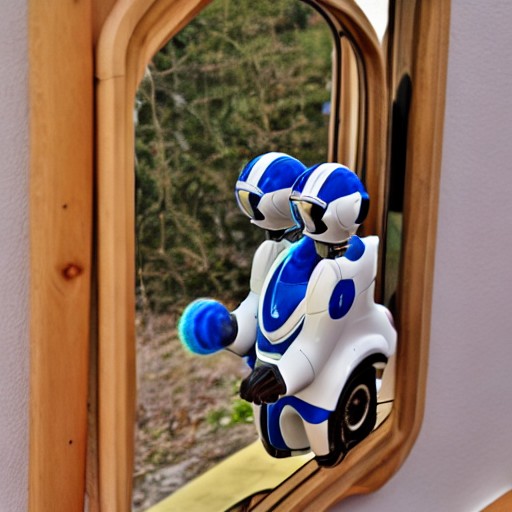}
    &
    \suppimg{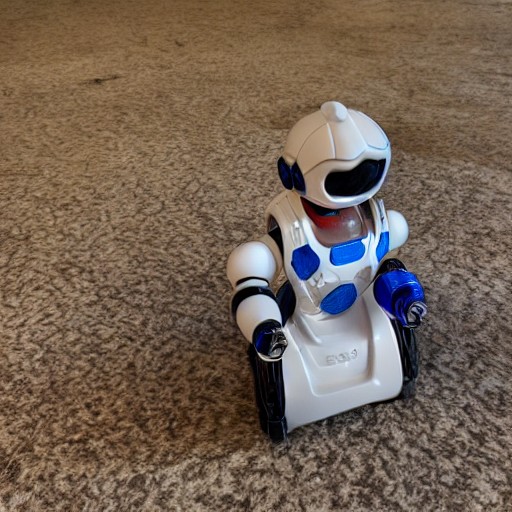}
    \\[0pt]

    % SDXL + References
    \smash{%
        \raisebox{-0.5\height}{%
            \makebox[\linewidth][c]{%
                \suppMultiFourRefs
                    {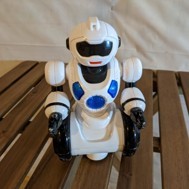}
                    {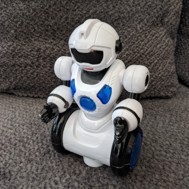}
                    {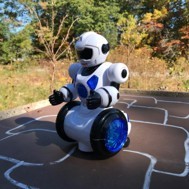}
                    {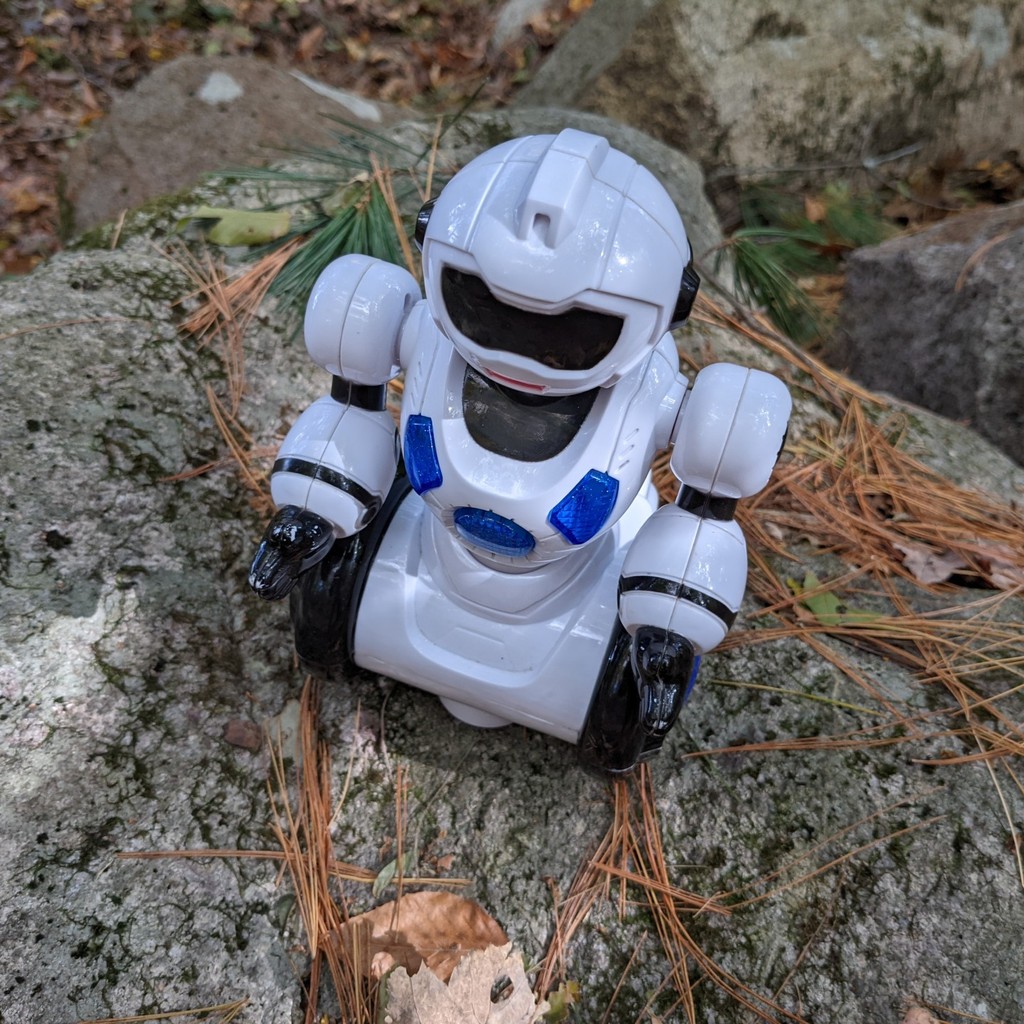}
            }%
        }%
    }
    &
    \suppbackbone{SDXL}
    &
    \suppimg{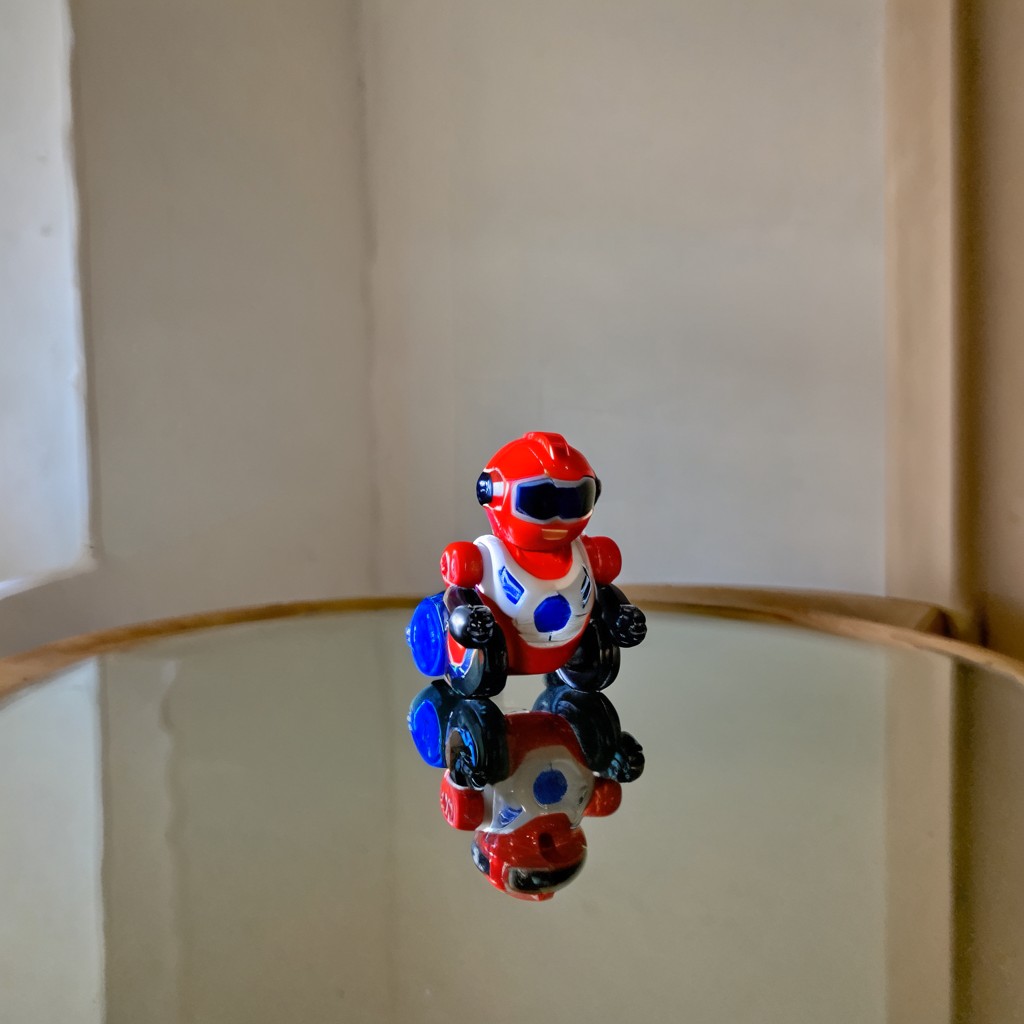}
    &
    \suppimg{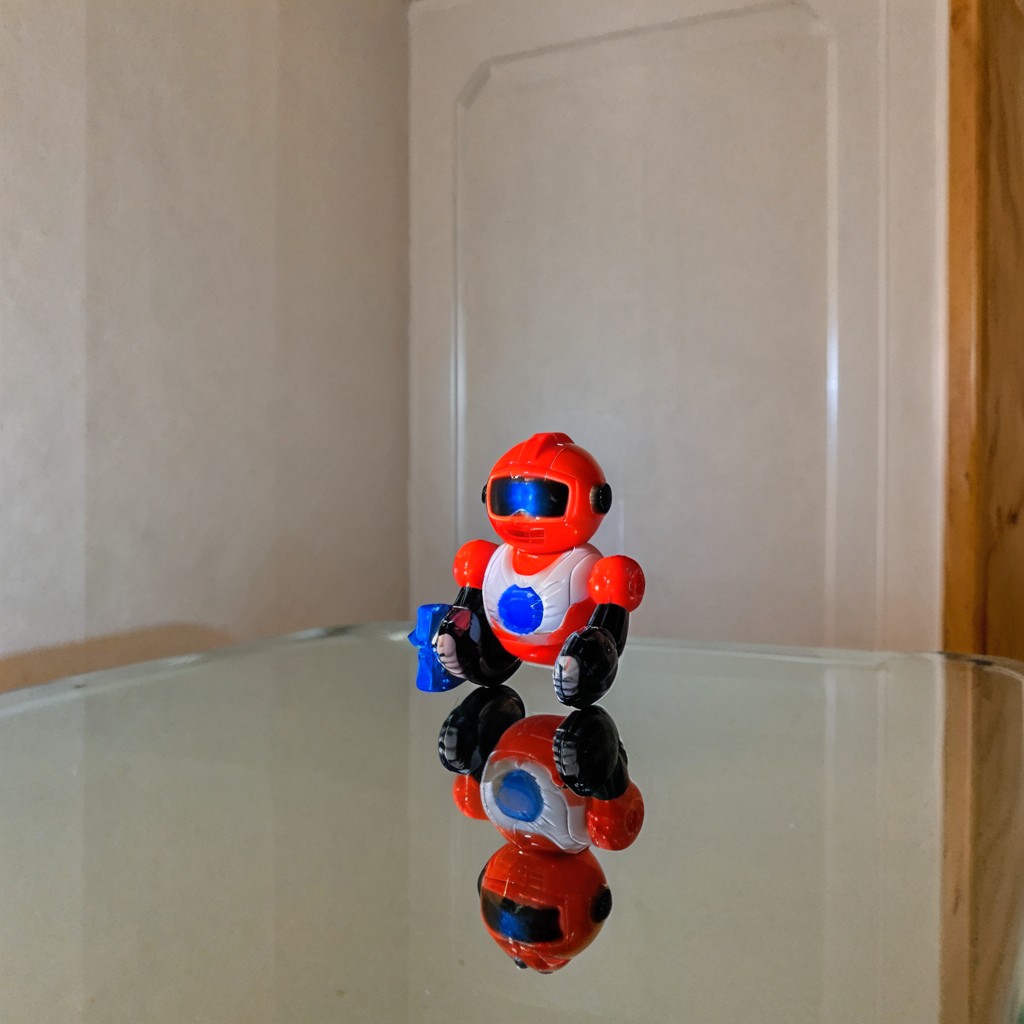}
    &
    \suppimg{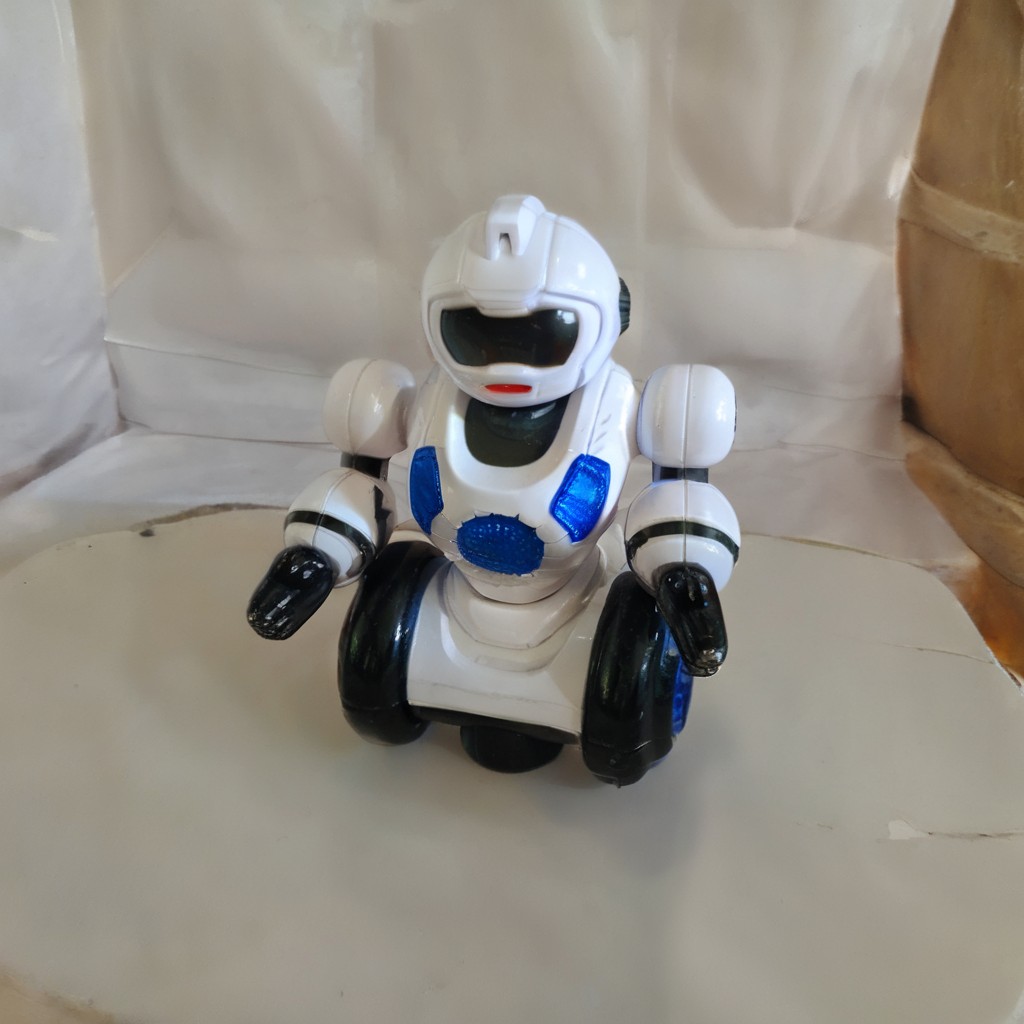}
    \\[0pt]

    % SD3
    {}
    &
    \suppbackbone{SD3}
    &
    \suppimg{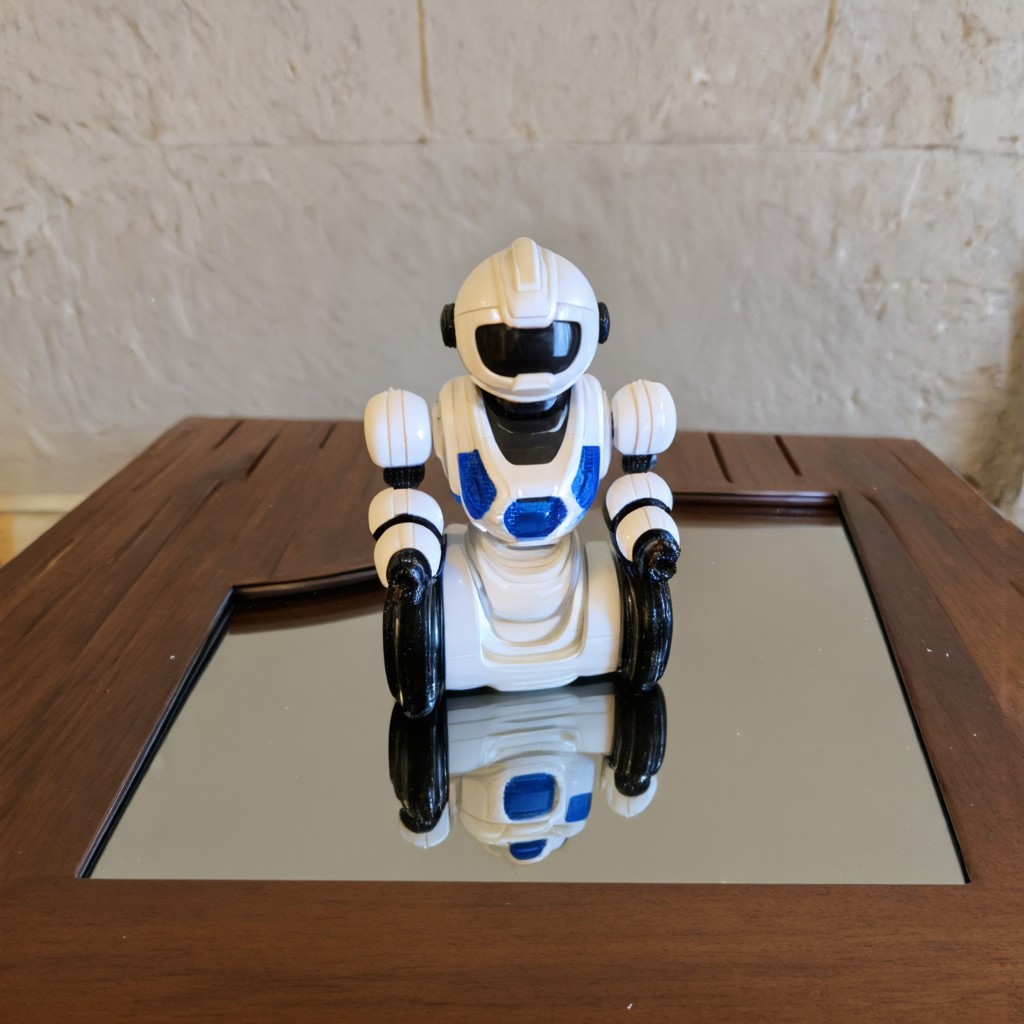}
    &
    \suppimg{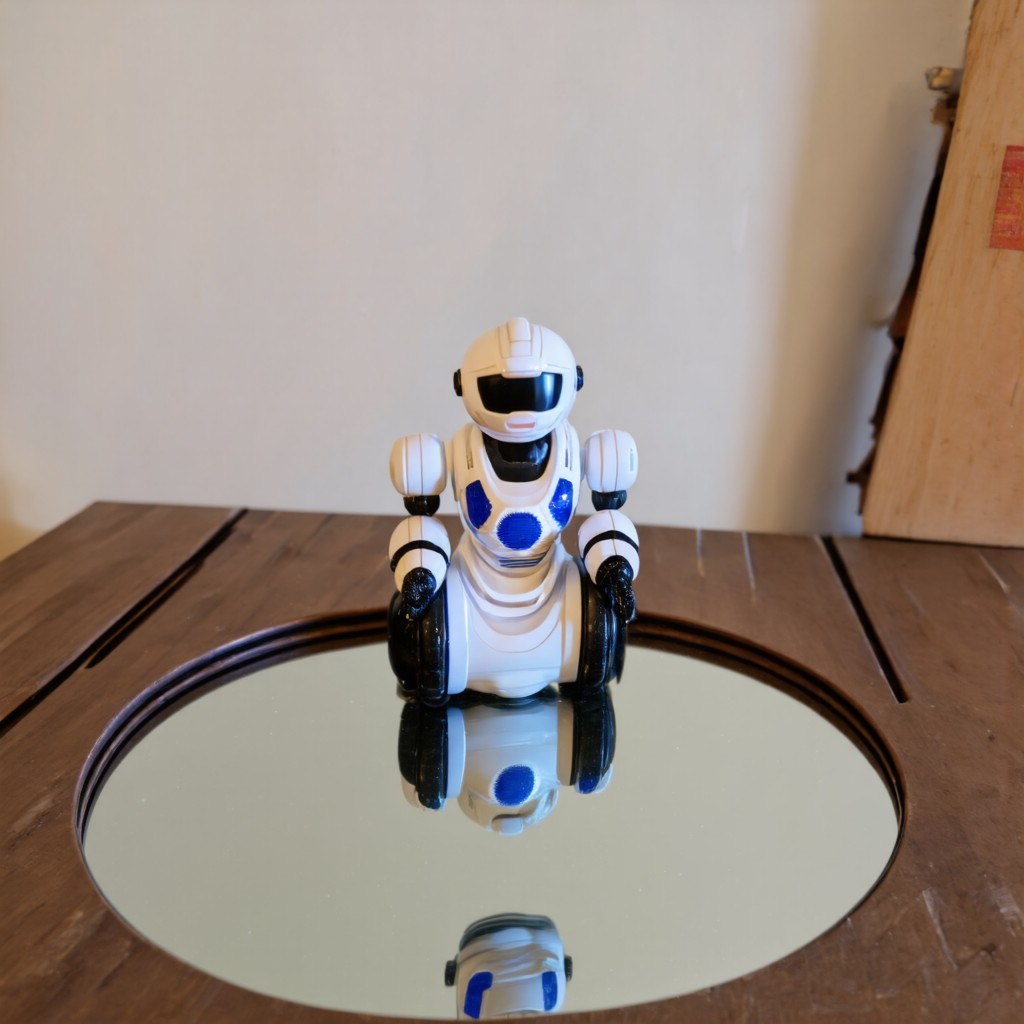}
    &
    \suppimg{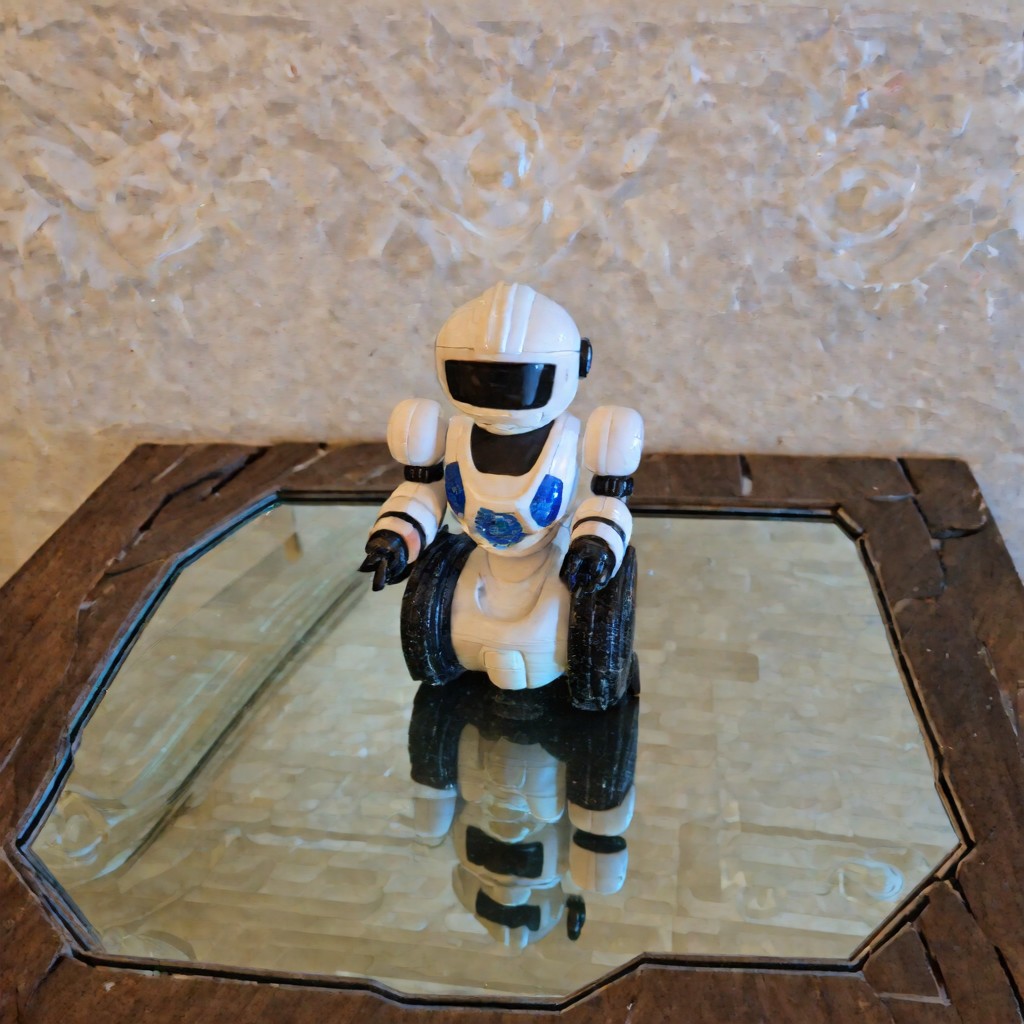}
    \\[-3pt]

    % Prompt
    {}
    &
    {}
    &
    \multicolumn{3}{c}{%
        \suppprompt{``a [V*] on top of a mirror''}
    }

    \end{tabular}

    \end{adjustbox}

    \caption{
        Additional qualitative comparison on SD1.5, SDXL, and SD3
        backbones among DreamBooth-LoRA, Beyond-Finetuning, and our method.
    }
    \label{fig:supp_sd15sdxlsd3}

\end{figure*}

% \begin{figure*}[t]
%     \centering
%     \includegraphics[width=\linewidth]{figure/supp_w_ablation.png}
%     \caption{Comparison of qualitative results under different weighting parameters $w$.
%     }
%     \label{fig:supp_w_ablation}
% \end{figure*}

% \input{figure/supp_w_ablation}
\begin{figure*}[t]
    \centering

    \setlength{\tabcolsep}{1.5pt}
    \renewcommand{\arraystretch}{1.0}

    % =========================================================
    % 이미지 크기
    % =========================================================
    \newlength{\wAblationImageSize}
    \setlength{\wAblationImageSize}{0.132\textwidth}

    % Reference 4장: 간격 없는 2×2 배열
    % 위: ref1, ref2 / 아래: ref3, ref4
    \newcommand{\wFourRefs}[4]{%
        \vbox{%
            \offinterlineskip
            \hbox{%
                \includegraphics[
                    width=0.5\wAblationImageSize,
                    height=0.5\wAblationImageSize
                ]{#1}%
                \includegraphics[
                    width=0.5\wAblationImageSize,
                    height=0.5\wAblationImageSize
                ]{#2}%
            }%
            \hbox{%
                \includegraphics[
                    width=0.5\wAblationImageSize,
                    height=0.5\wAblationImageSize
                ]{#3}%
                \includegraphics[
                    width=0.5\wAblationImageSize,
                    height=0.5\wAblationImageSize
                ]{#4}%
            }%
        }%
    }

    % 결과 이미지
    \newcommand{\wAblationImg}[1]{%
        \includegraphics[
            width=\wAblationImageSize,
            height=\wAblationImageSize
        ]{#1}%
    }

    \begin{tabular}{
        @{}
        C{\wAblationImageSize}  % Reference 묶음과 동일한 너비
        @{\hspace{3pt}}        % References와 프롬프트 사이 간격
        C{0.155\textwidth}
        *{5}{C{\wAblationImageSize}}
        @{}
    }

    % =========================================================
    % Header
    % =========================================================
    \wheader{References}
    &
    {}
    &
    \wheader{$w=0.0$}
    &
    \wheader{$w=0.25$}
    &
    \wheader{$w=0.5$}
    &
    \wheader{$w=0.75$}
    &
    \wheader{$w=1.0$}
    \\[3pt]

    % =========================================================
    % Row 1: Bear
    % =========================================================
    \wFourRefs
        {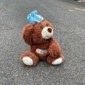}
        {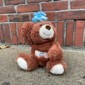}
        {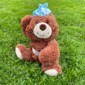}
        {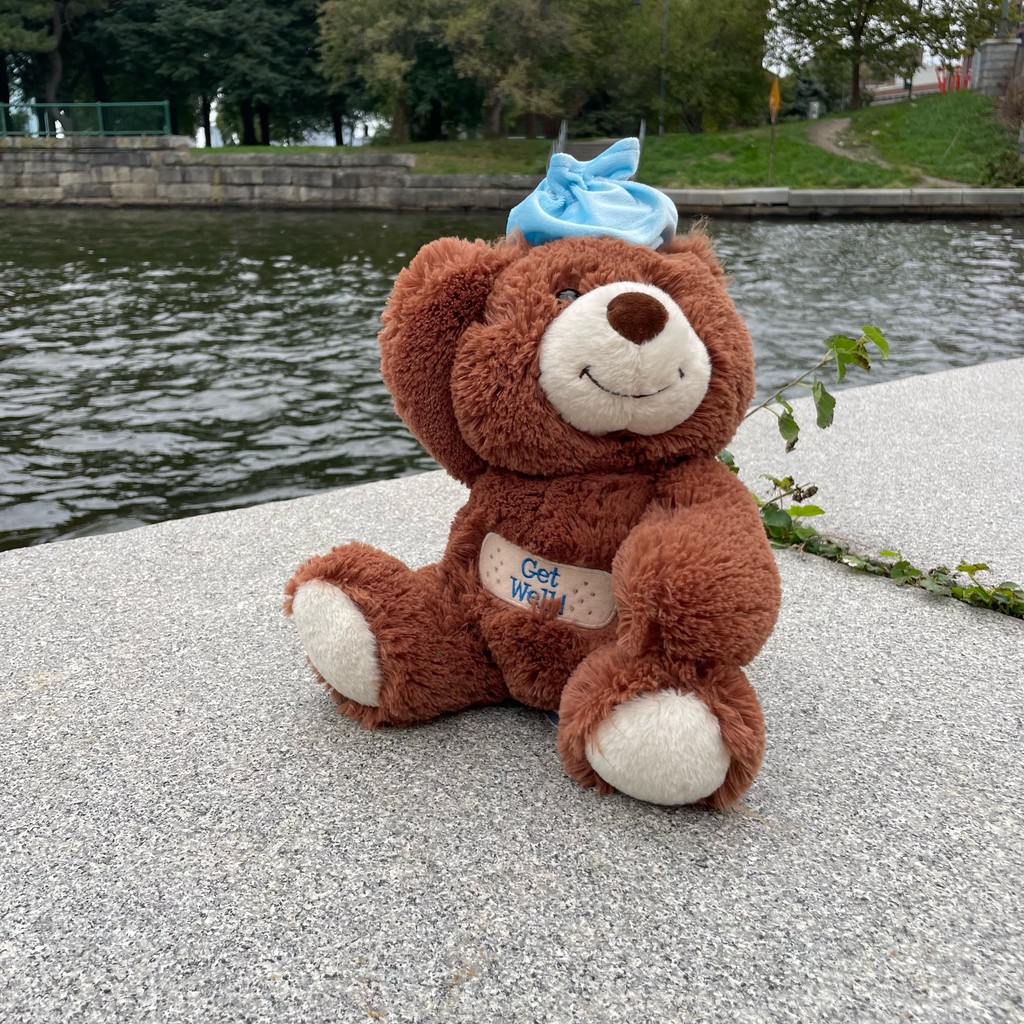}
    &
    \wprompt{``a purple [V*]''}
    &
    \wAblationImg{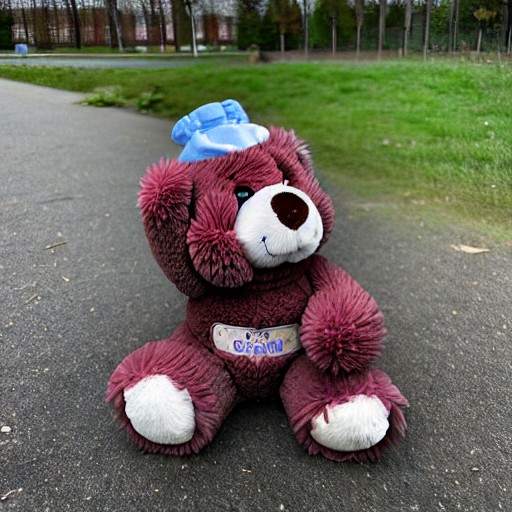}
    &
    \wAblationImg{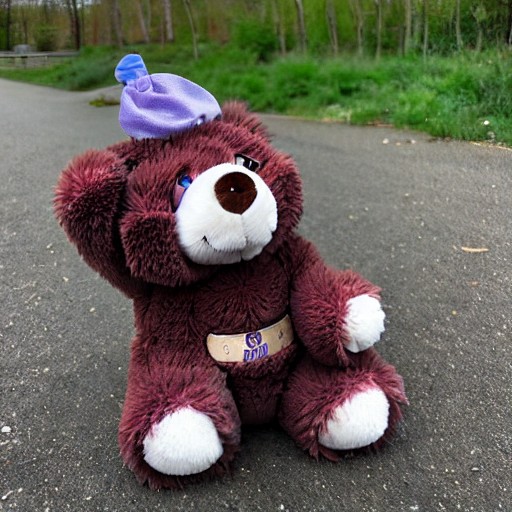}
    &
    \wAblationImg{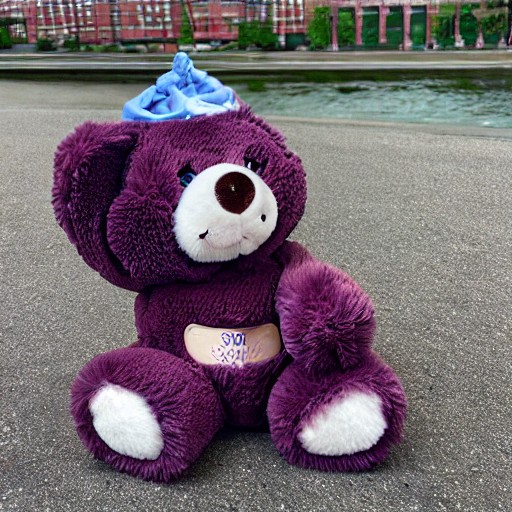}
    &
    \wAblationImg{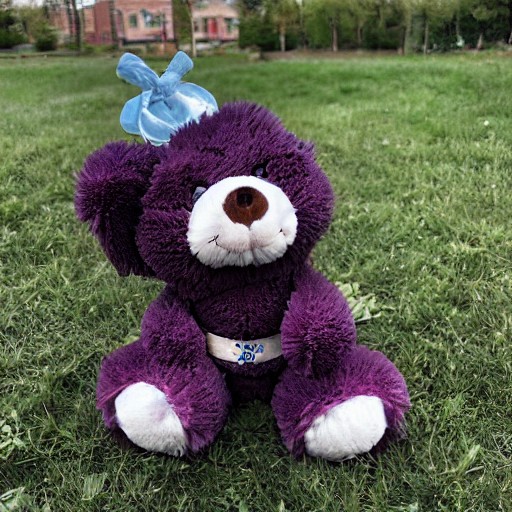}
    &
    \wAblationImg{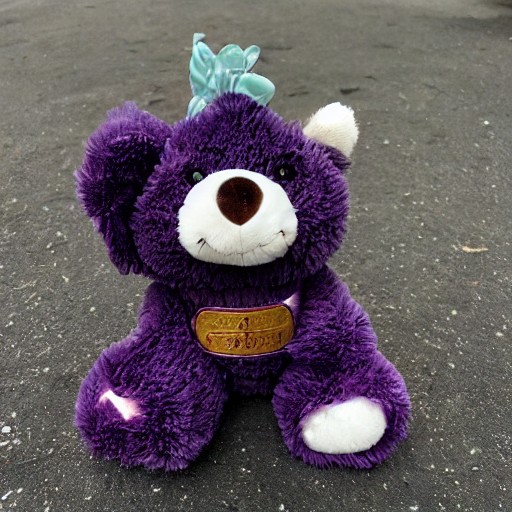}
    \\[3pt]

    % =========================================================
    % Row 2: Cube
    % =========================================================
    \wFourRefs
        {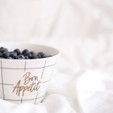}
        {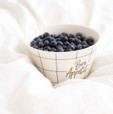}
        {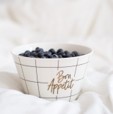}
        {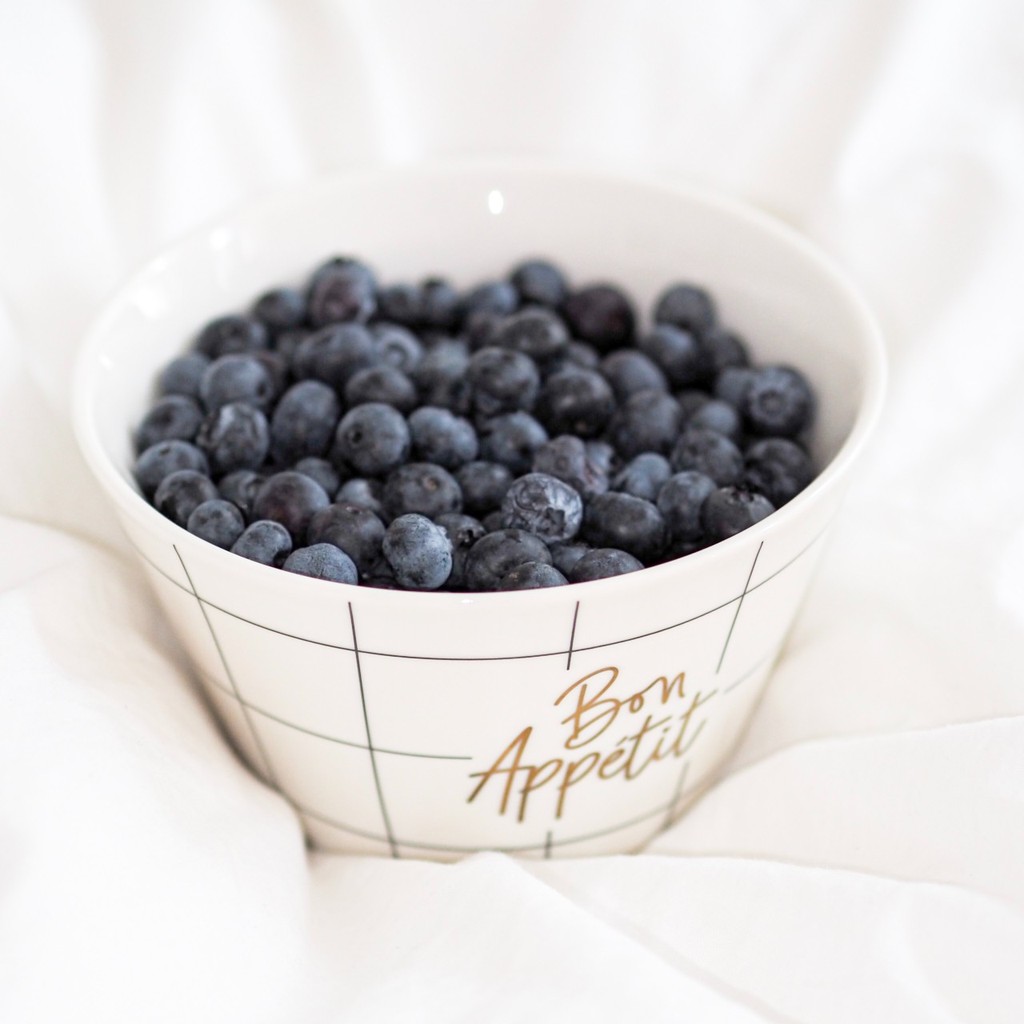}
    &
    \wprompt{
        ``a cube shaped\\
        {}[V*]''
    }
    &
    \wAblationImg{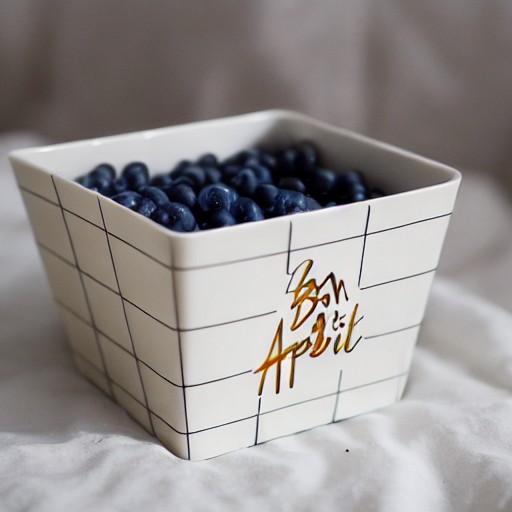}
    &
    \wAblationImg{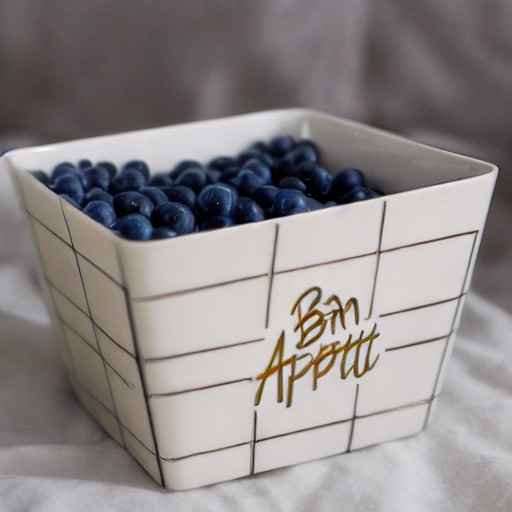}
    &
    \wAblationImg{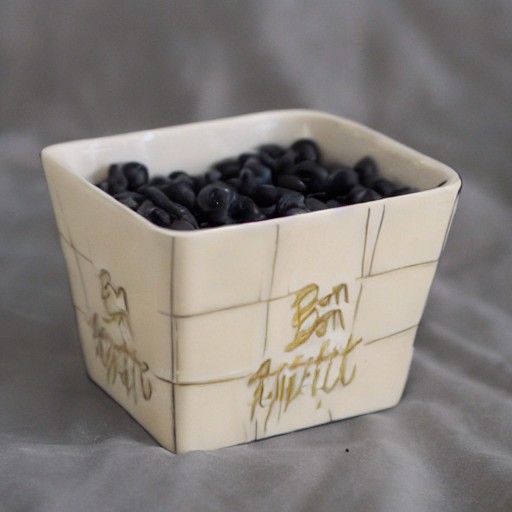}
    &
    \wAblationImg{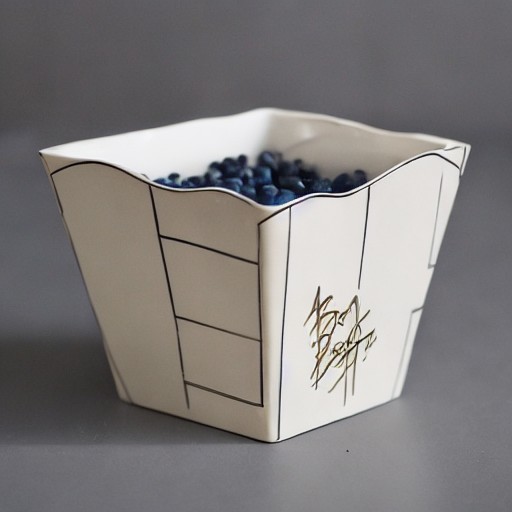}
    &
    \wAblationImg{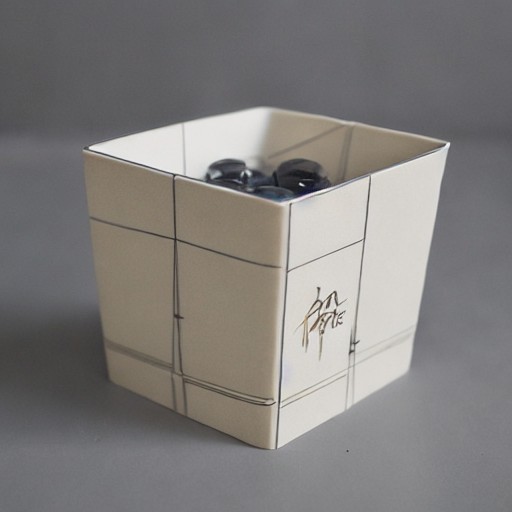}
    \\[3pt]

    % =========================================================
    % Row 3: Dog
    % =========================================================
    \wFourRefs
        {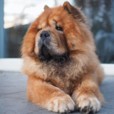}
        {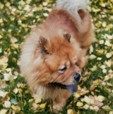}
        {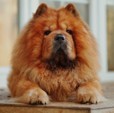}
        {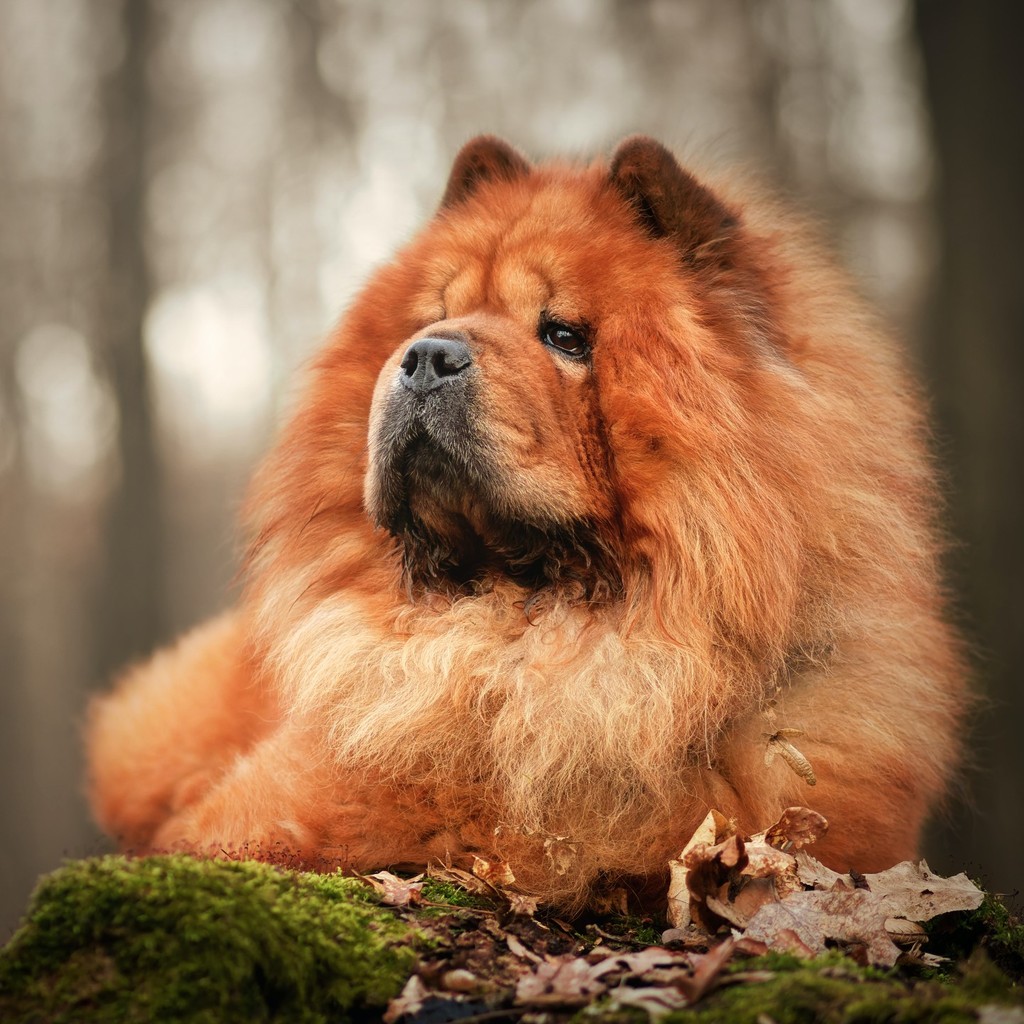}
    &
    \wprompt{
        ``a [V*]\\
        wearing a\\
        black top hat\\
        and a monocle''
    }
    &
    \wAblationImg{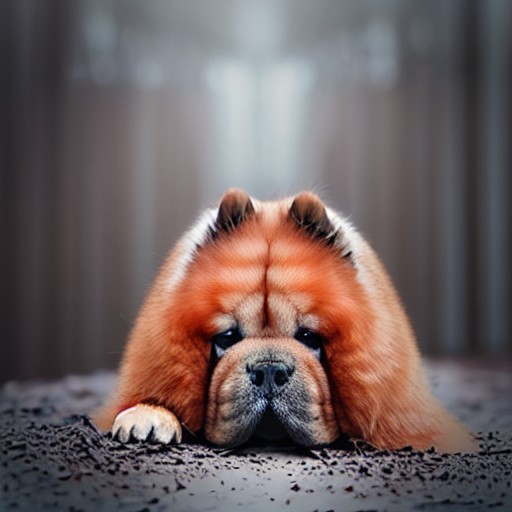}
    &
    \wAblationImg{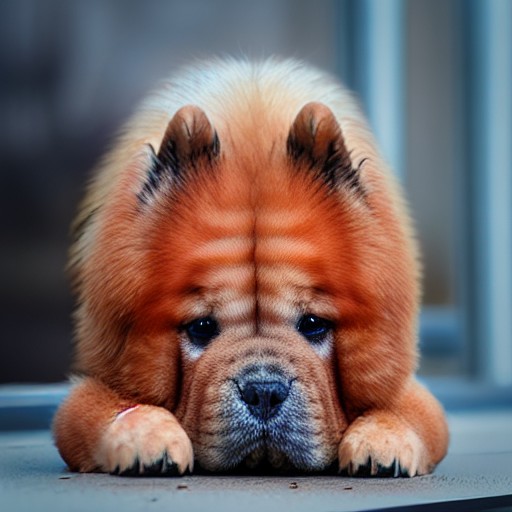}
    &
    \wAblationImg{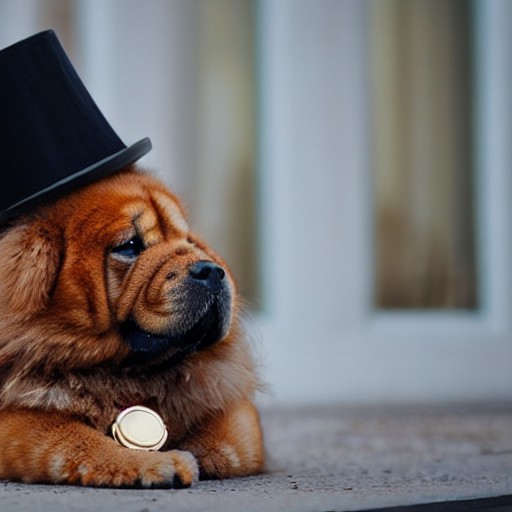}
    &
    \wAblationImg{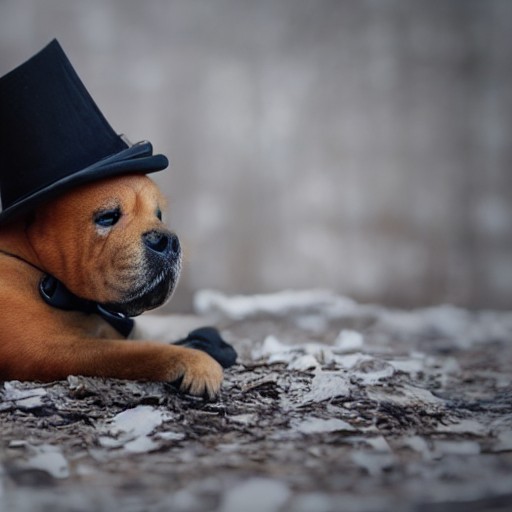}
    &
    \wAblationImg{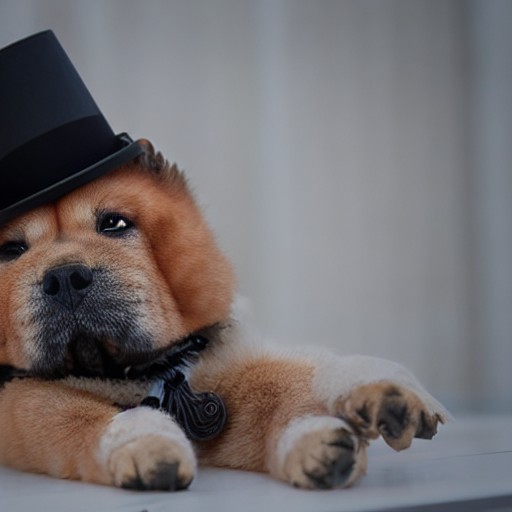}
    \\[3pt]

    % =========================================================
    % Row 4: Cat
    % =========================================================
    \wFourRefs
        {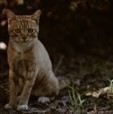}
        {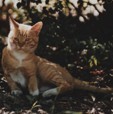}
        {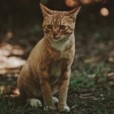}
        {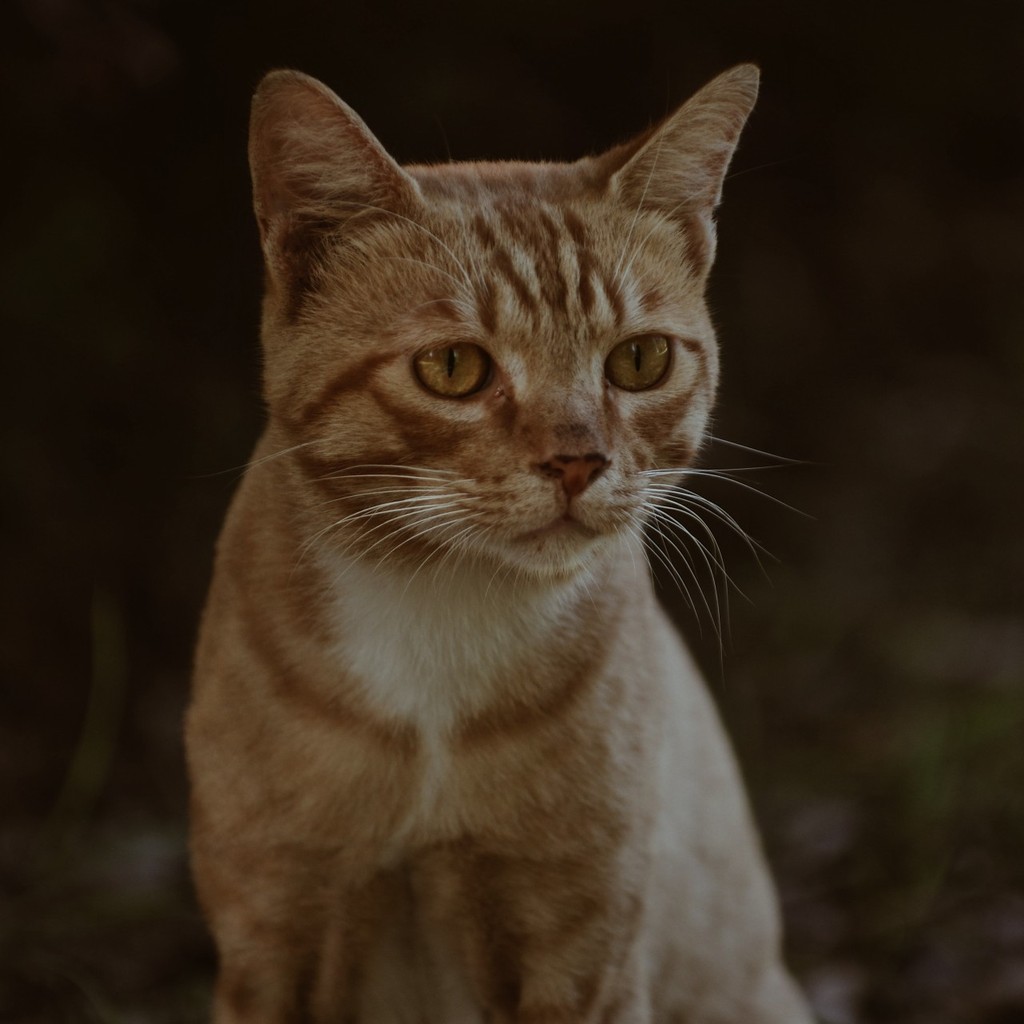}
    &
    \wprompt{
        ``a [V*] in a\\
        chef outfit''
    }
    &
    \wAblationImg{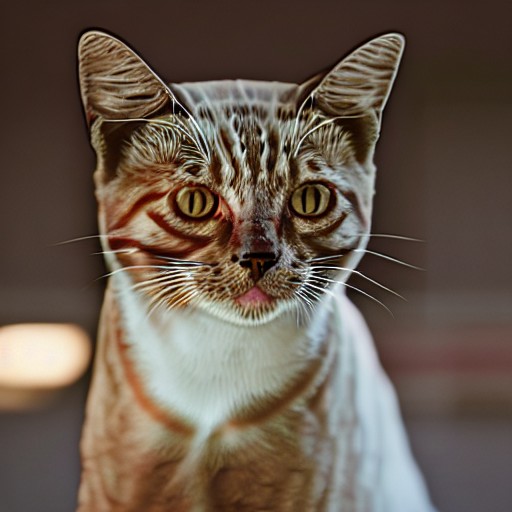}
    &
    \wAblationImg{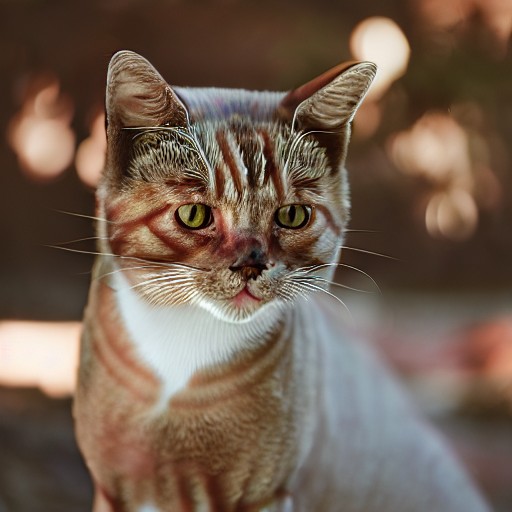}
    &
    \wAblationImg{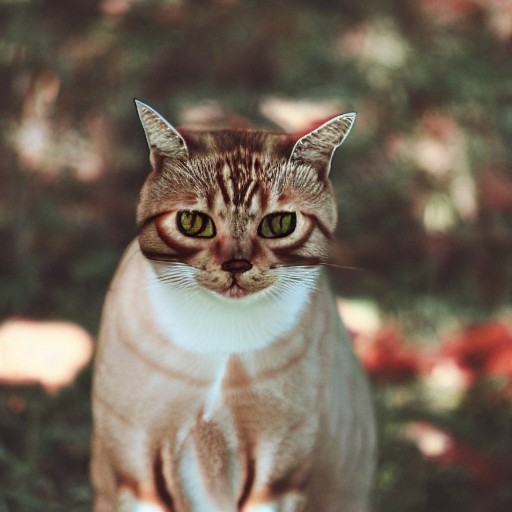}
    &
    \wAblationImg{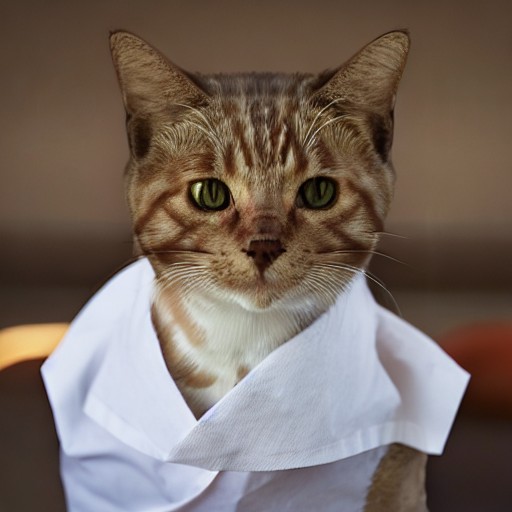}
    &
    \wAblationImg{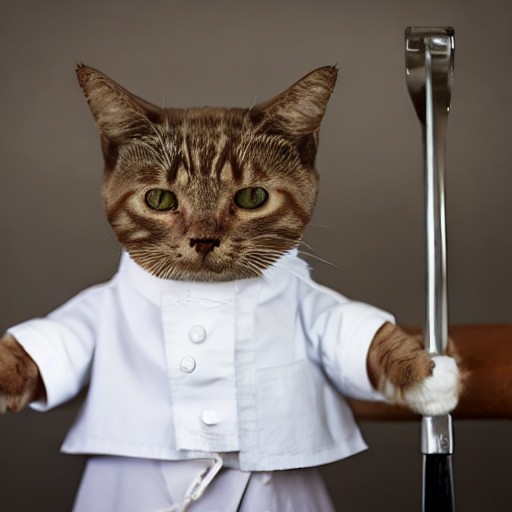}

    \end{tabular}

    \caption{
        Comparison of qualitative results under different weighting
        parameters $w$.
    }
    \label{fig:supp_w_ablation}

\end{figure*}

% \begin{figure*}[t]
%     \centering
%     \includegraphics[width=\linewidth,
%                      height=0.9\textheight,
%                      keepaspectratio]{figure/supp_pretrain_finetune.png}
%     \caption{Comparison of qualitative results using two different anchoring strategies.
%     }
%     \label{fig:supp_pretrain_finetune}
% \end{figure*}

% \input{figure/supp_pretrain_finetune}
\begin{figure*}[t]
    \centering

    \setlength{\tabcolsep}{1.5pt}
    \renewcommand{\arraystretch}{1.0}

    % =========================================================
    % References: 간격 없는 2×2 배열
    % 전체 너비 = 0.225\textwidth
    % =========================================================
    \newlength{\paRefTileSize}
    \setlength{\paRefTileSize}{0.1125\textwidth}

    \newcommand{\paFourRefs}[4]{%
        \vbox{%
            \offinterlineskip
            \hbox{%
                \includegraphics[
                    width=\paRefTileSize,
                    height=\paRefTileSize
                ]{#1}%
                \includegraphics[
                    width=\paRefTileSize,
                    height=\paRefTileSize
                ]{#2}%
            }%
            \hbox{%
                \includegraphics[
                    width=\paRefTileSize,
                    height=\paRefTileSize
                ]{#3}%
                \includegraphics[
                    width=\paRefTileSize,
                    height=\paRefTileSize
                ]{#4}%
            }%
        }%
    }

    \begin{adjustbox}{
        max width=\textwidth,
        max totalheight=0.82\textheight,
        center
    }

    \begin{tabular}{
        @{}
        C{0.225\textwidth}       % References
        C{0.130\textwidth}       % Anchoring label
        *{3}{C{0.190\textwidth}} % SD1.5 / SDXL / SD3
        @{}
    }

    % =========================================================
    % Header
    % =========================================================
    \paheader{References}
    &
    {}
    &
    \paheader{SD1.5}
    &
    \paheader{SDXL}
    &
    \paheader{SD3}
    \\[2pt]

    % =========================================================
    % SLOTH
    % =========================================================

    % Finetuning
    \multirow[c]{2}{0.225\textwidth}{%
        \centering
        \paFourRefs
            {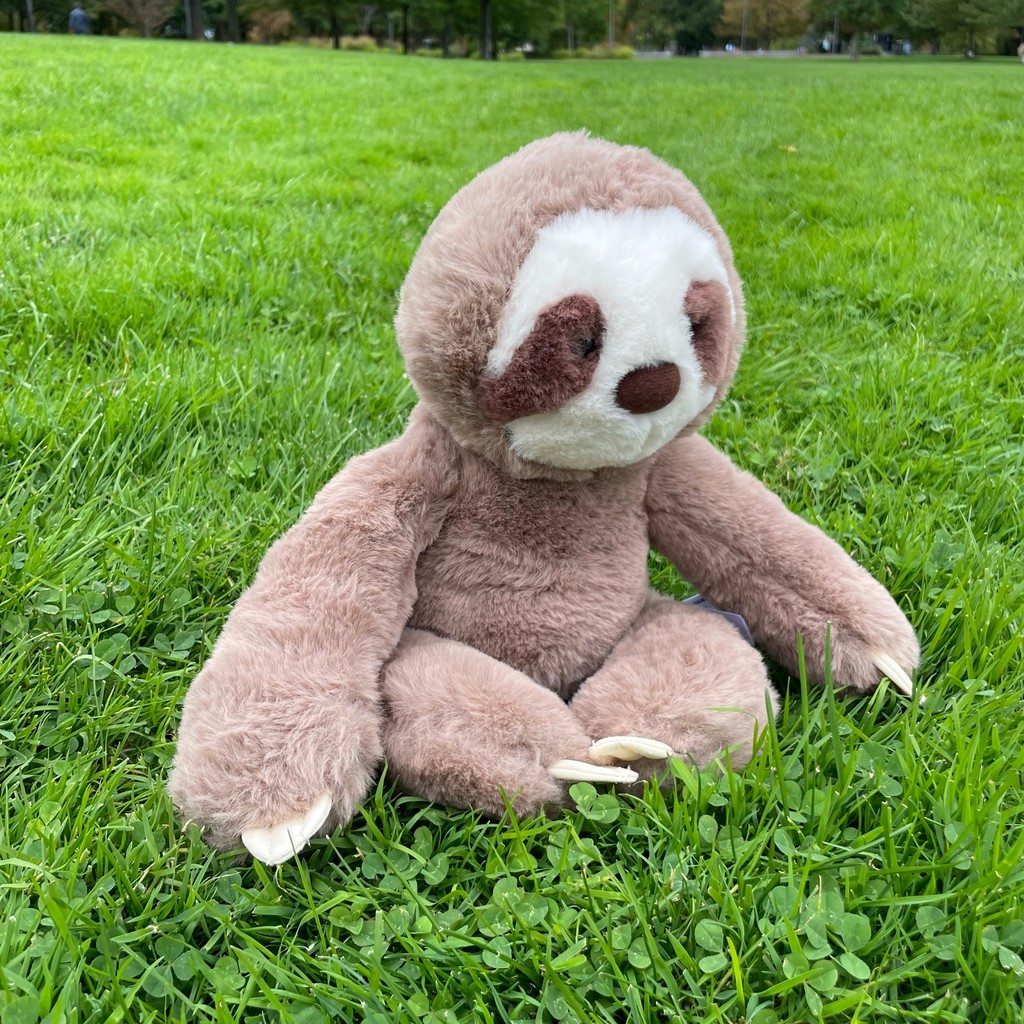}
            {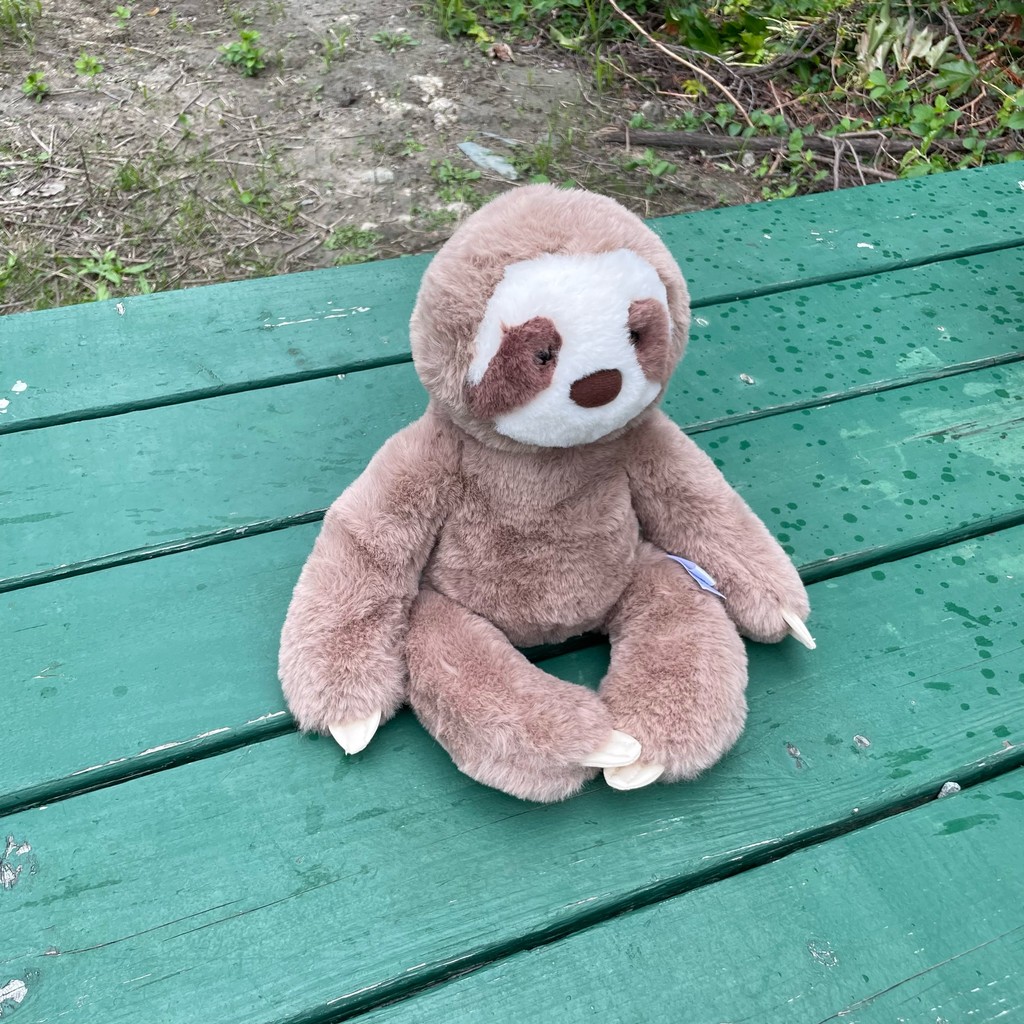}
            {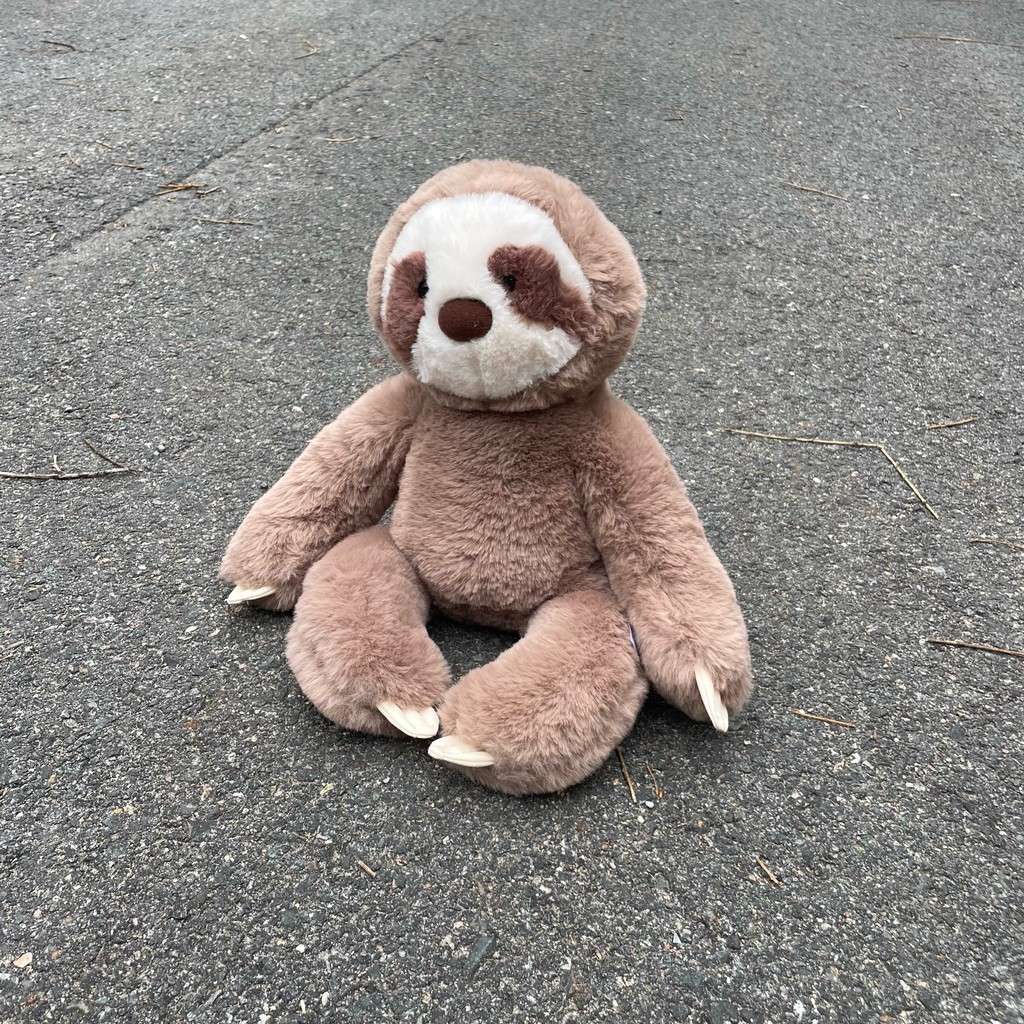}
            {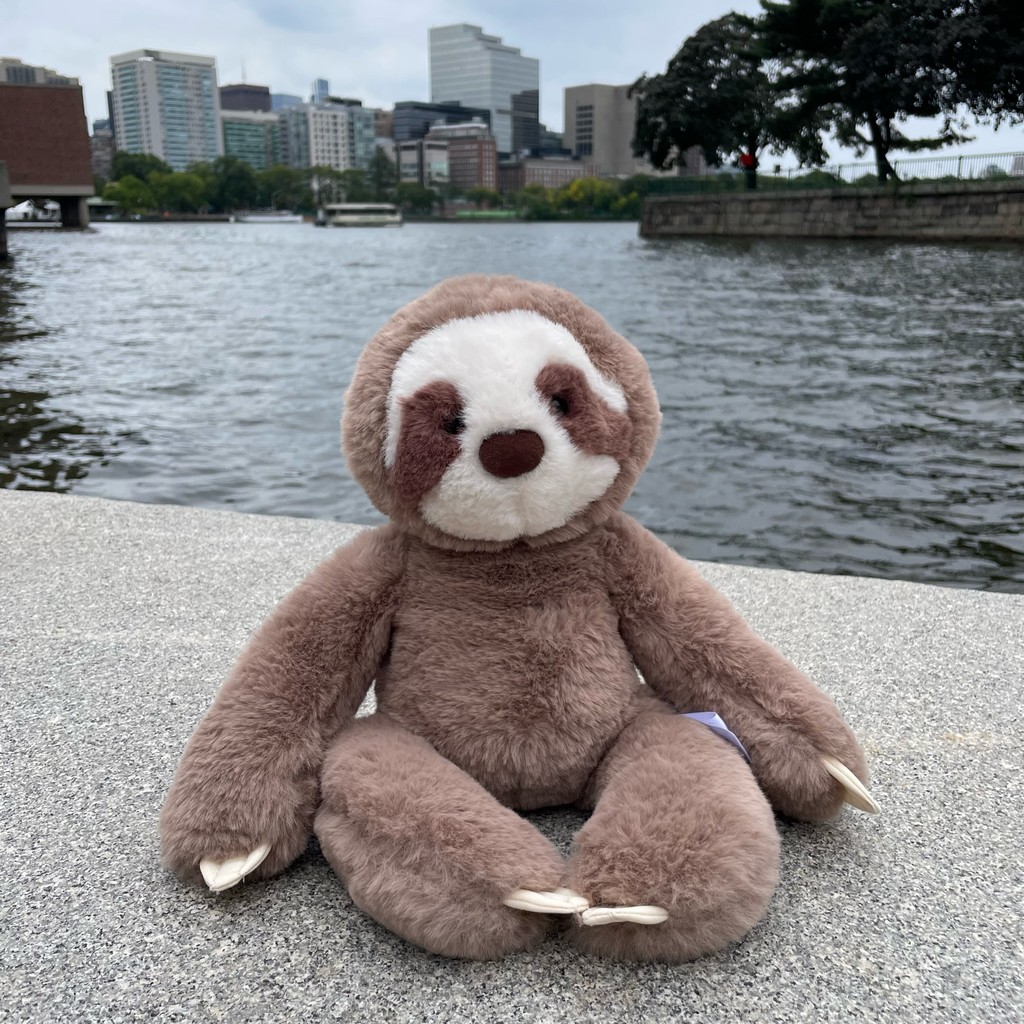}
    }
    &
    \palabel{Finetuning\\Anchoring}
    &
    \paimg{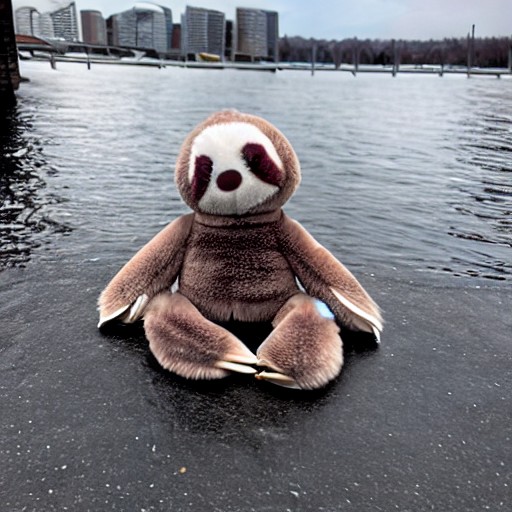}
    &
    \paimg{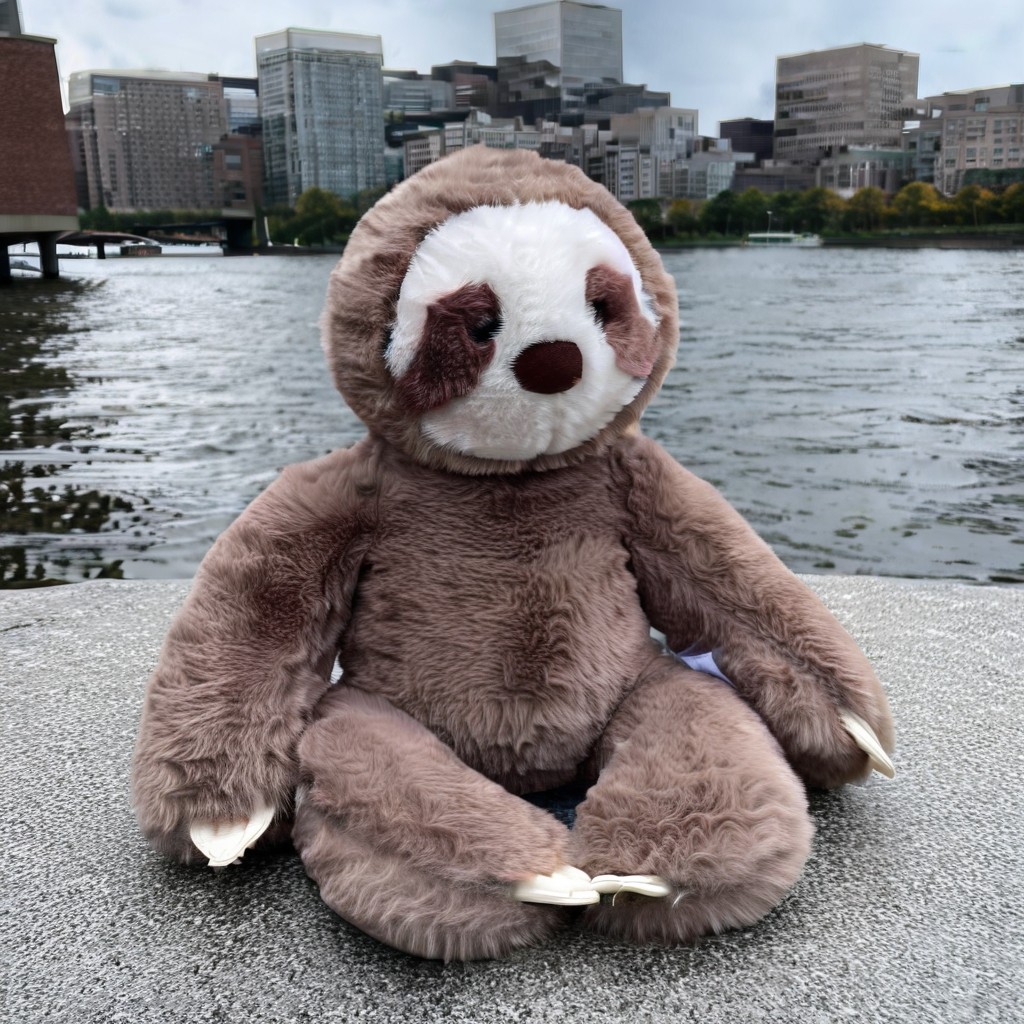}
    &
    \paimg{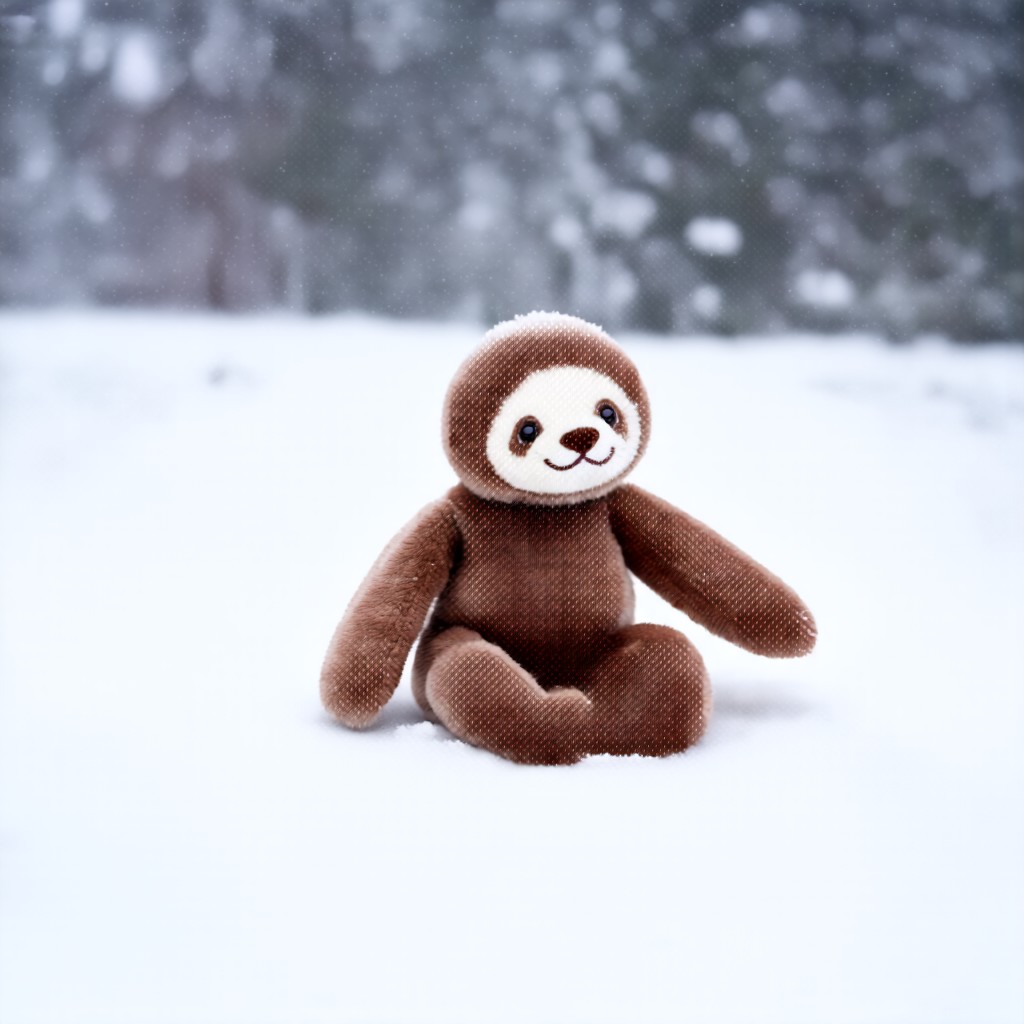}
    \\[1pt]

    % Pretrained
    {}
    &
    \palabel{Pretrained\\Anchoring}
    &
    \paimg{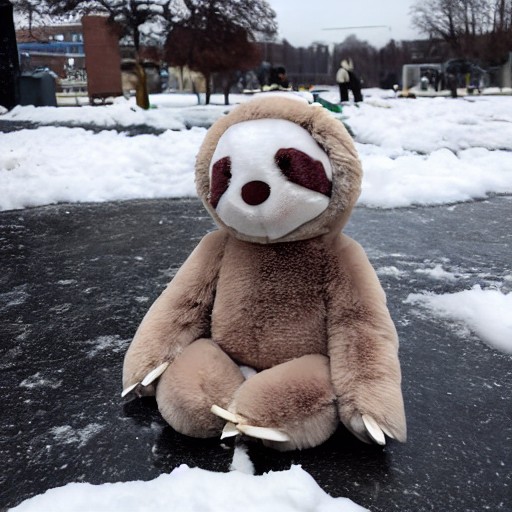}
    &
    \paimg{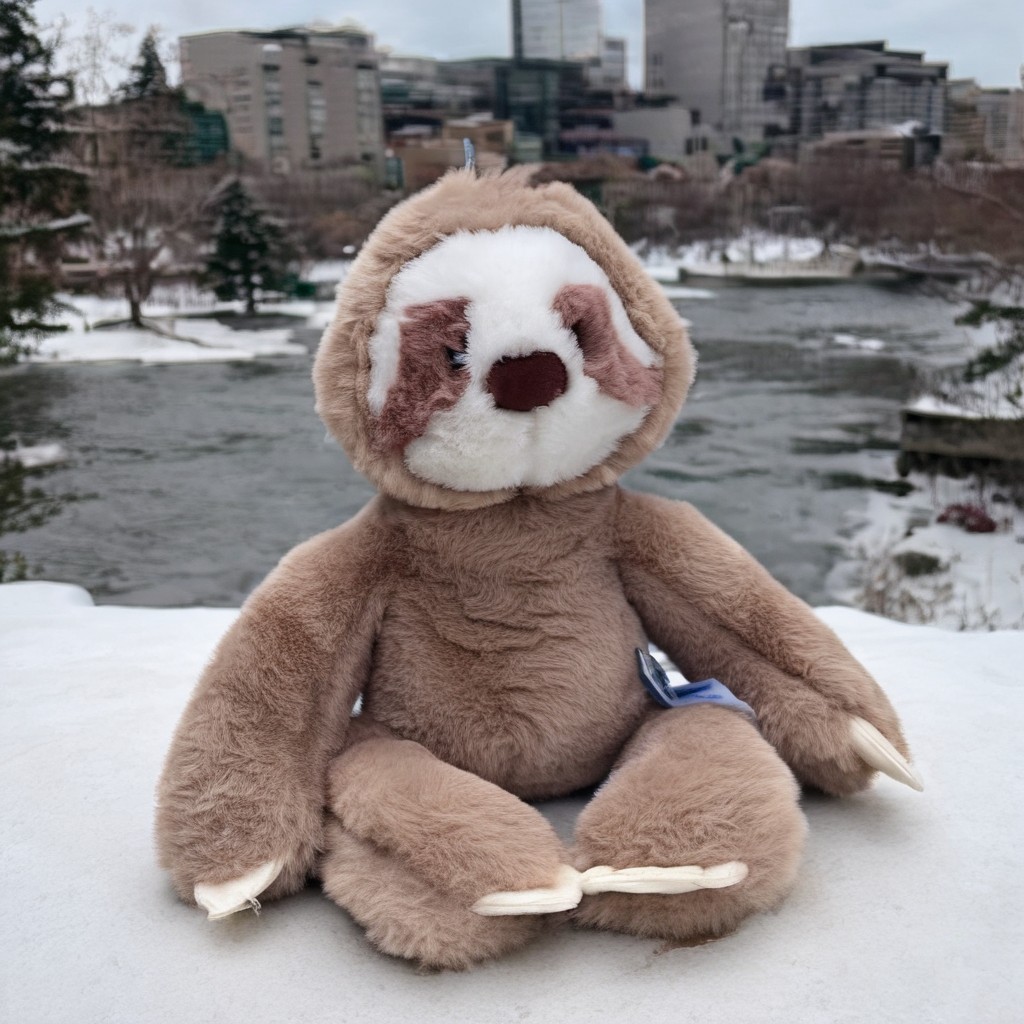}
    &
    \paimg{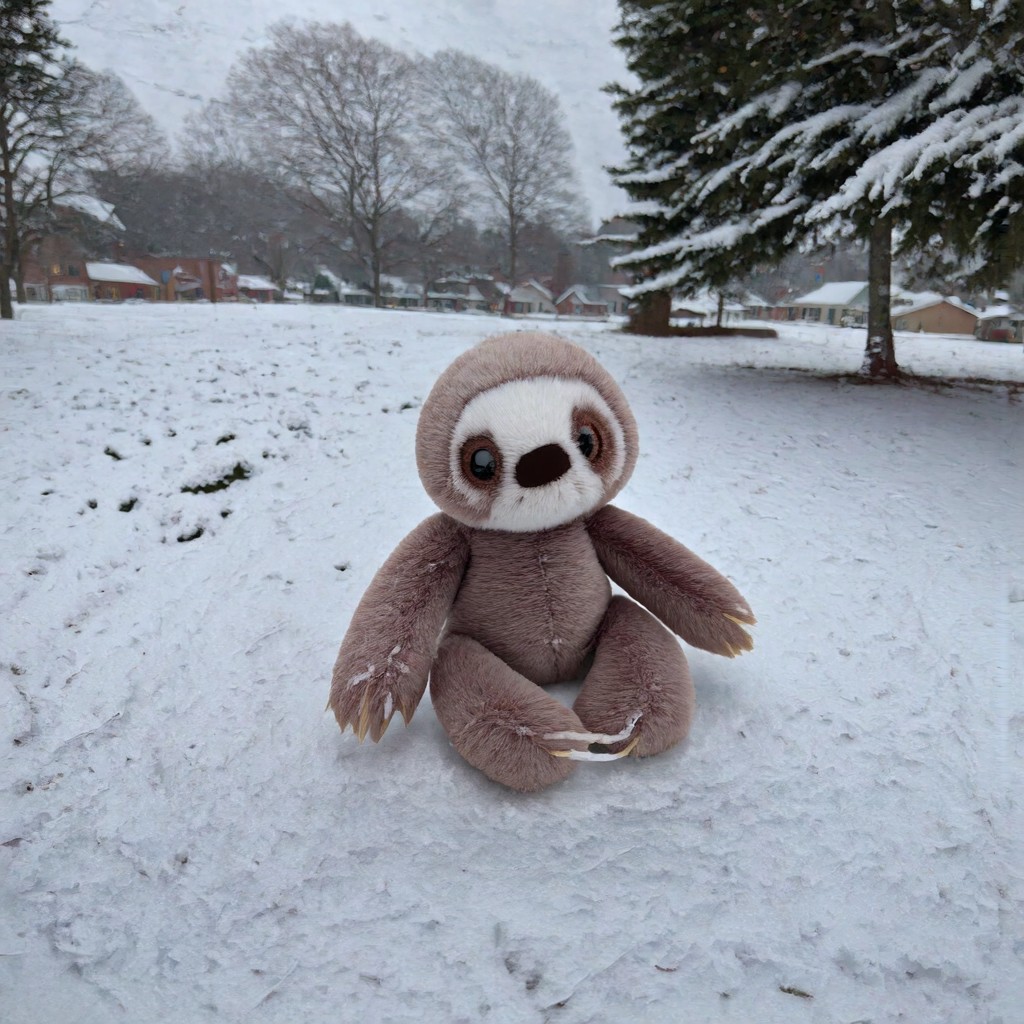}
    \\[-3pt]

    % Prompt
    {}
    &
    {}
    &
    \multicolumn{3}{c}{%
        \paprompt{``a [V*] in the snow''}
    }
    \\[5pt]

    % =========================================================
    % BOOT
    % =========================================================

    % Finetuning
    \multirow[c]{2}{0.225\textwidth}{%
        \centering
        \paFourRefs
            {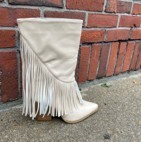}
            {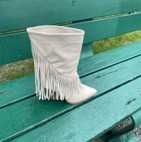}
            {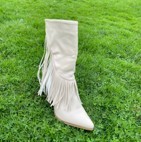}
            {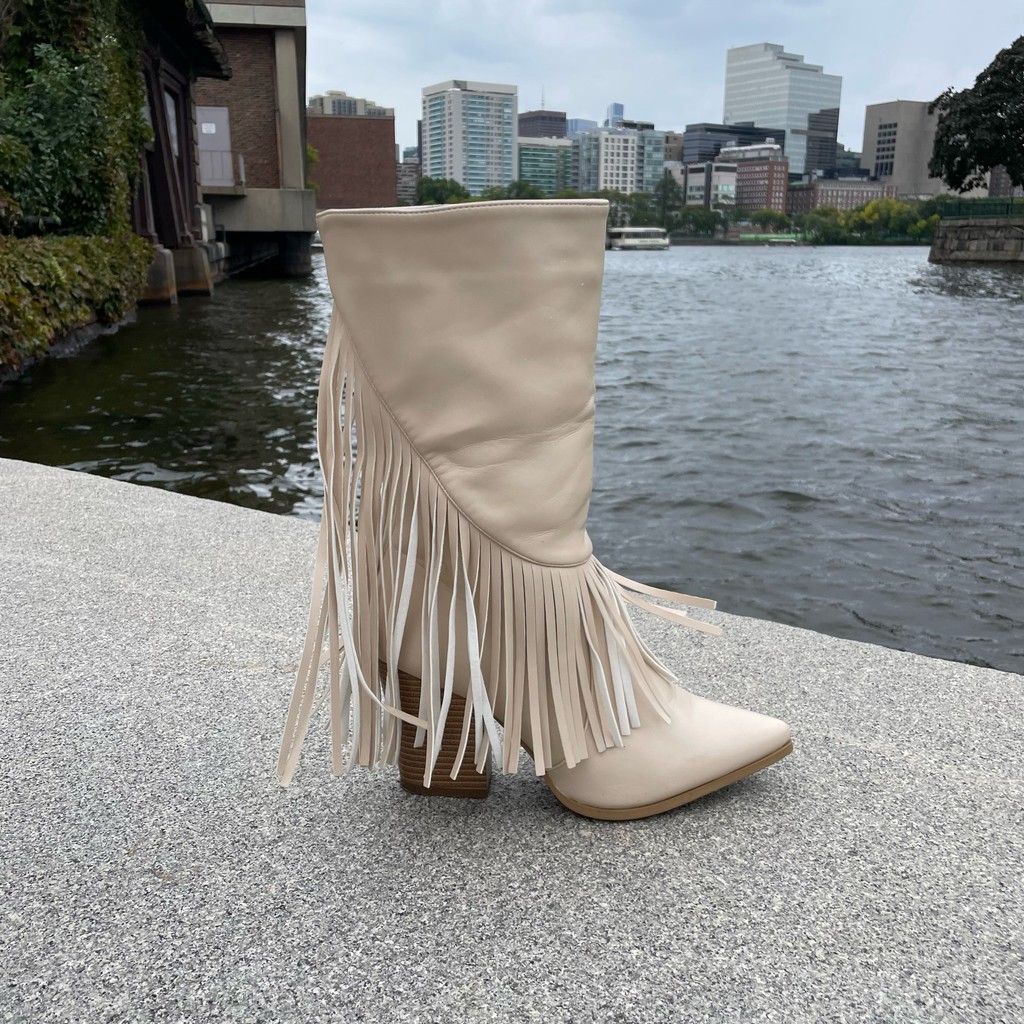}
    }
    &
    \palabel{Finetuning\\Anchoring}
    &
    \paimg{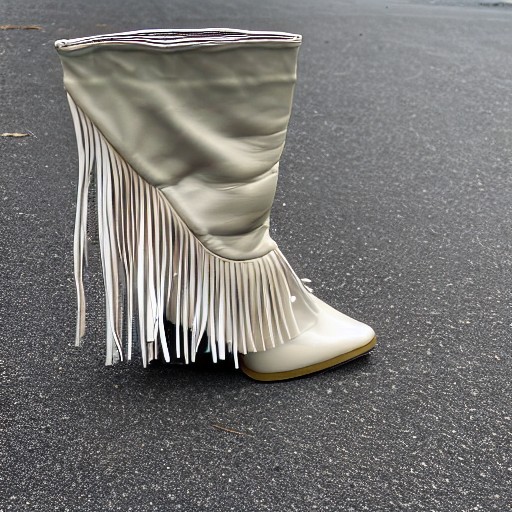}
    &
    \paimg{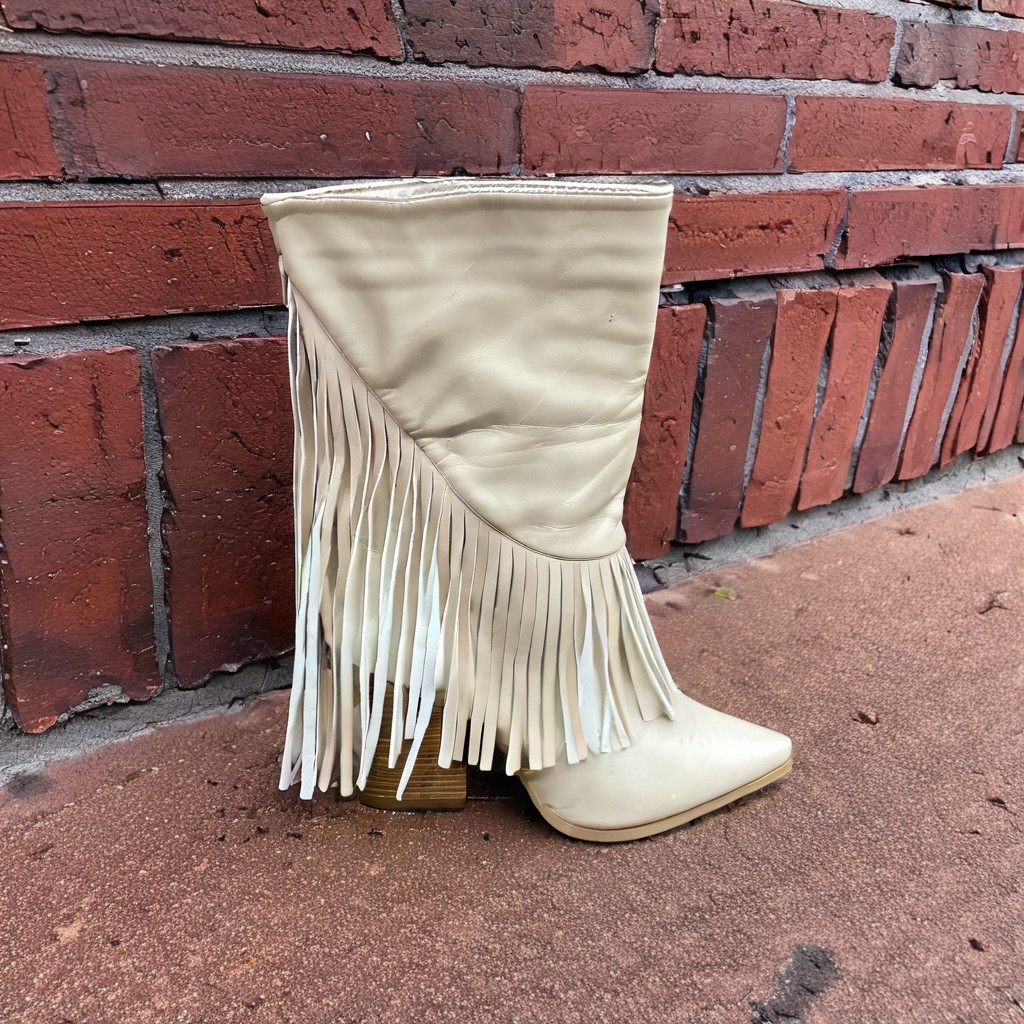}
    &
    \paimg{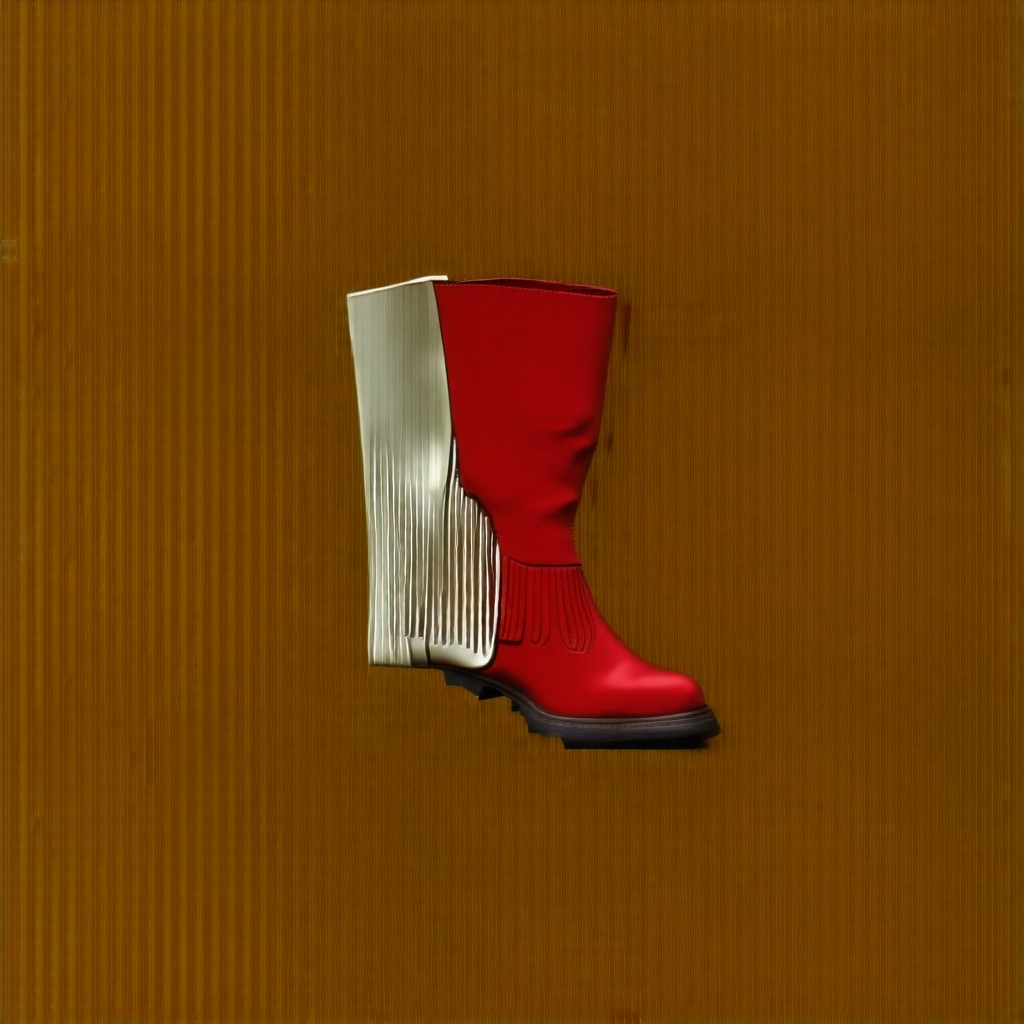}
    \\[1pt]

    % Pretrained
    {}
    &
    \palabel{Pretrained\\Anchoring}
    &
    \paimg{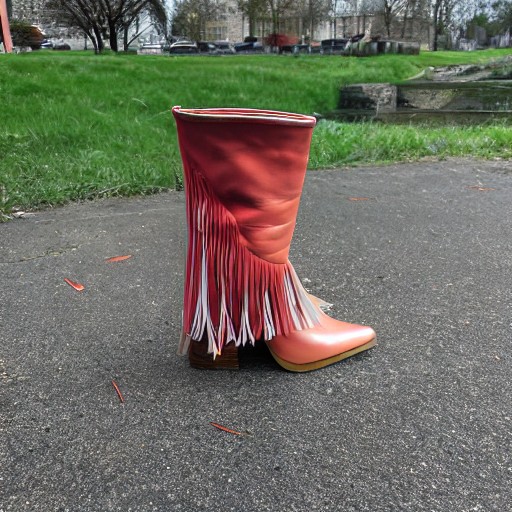}
    &
    \paimg{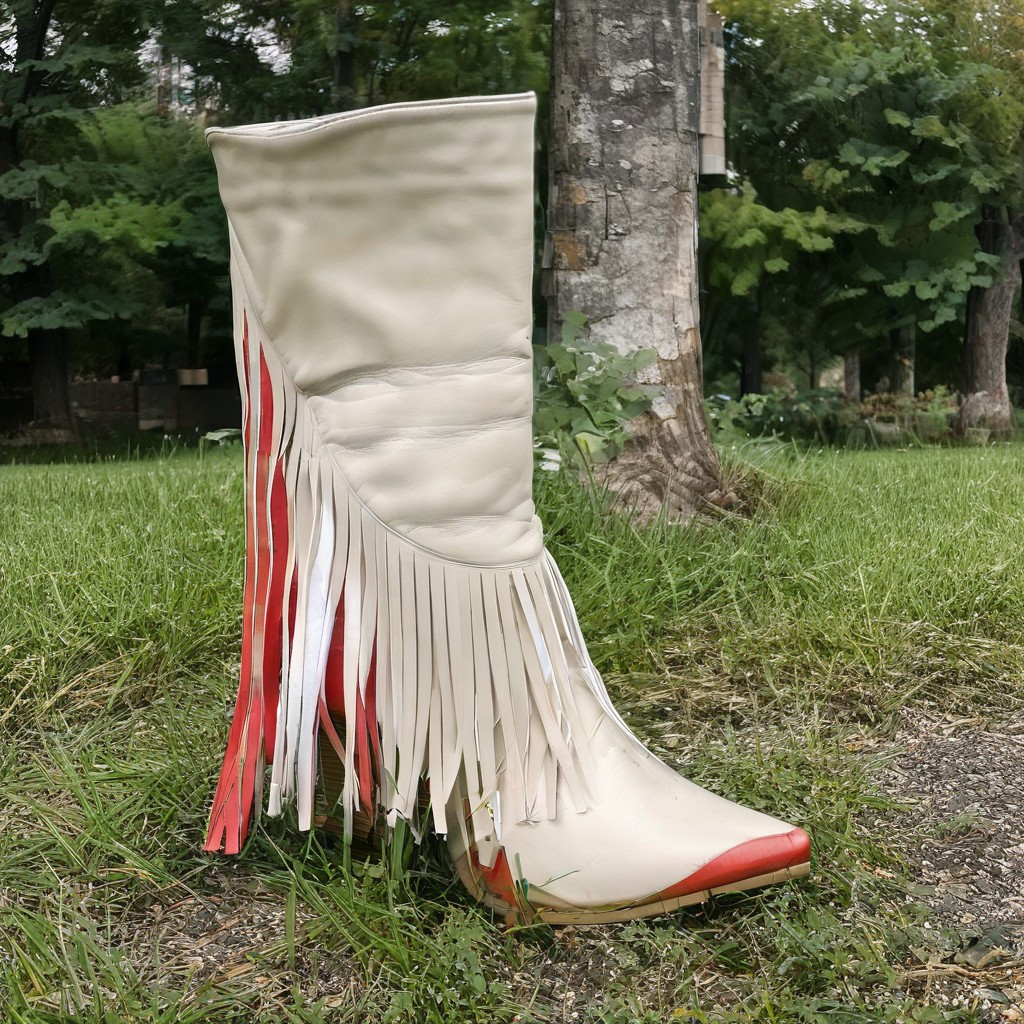}
    &
    \paimg{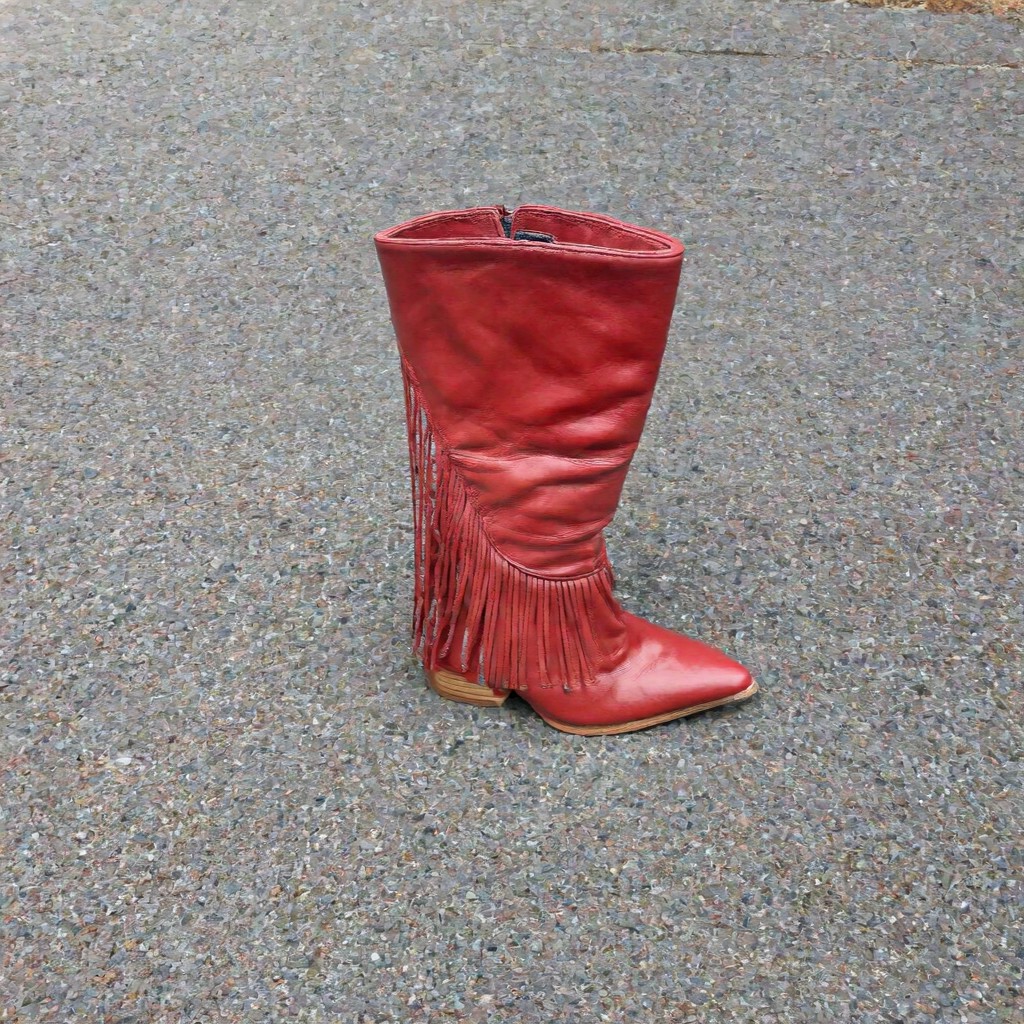}
    \\[-3pt]

    % Prompt
    {}
    &
    {}
    &
    \multicolumn{3}{c}{%
        \paprompt{``a red [V*]''}
    }
    \\[5pt]

    % =========================================================
    % CAT
    % =========================================================

    % Finetuning
    \multirow[c]{2}{0.225\textwidth}{%
        \centering
        \paFourRefs
            {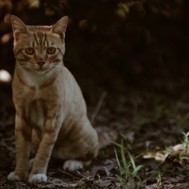}
            {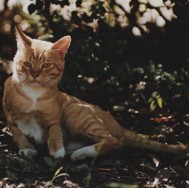}
            {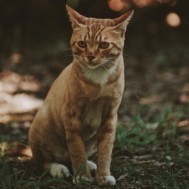}
            {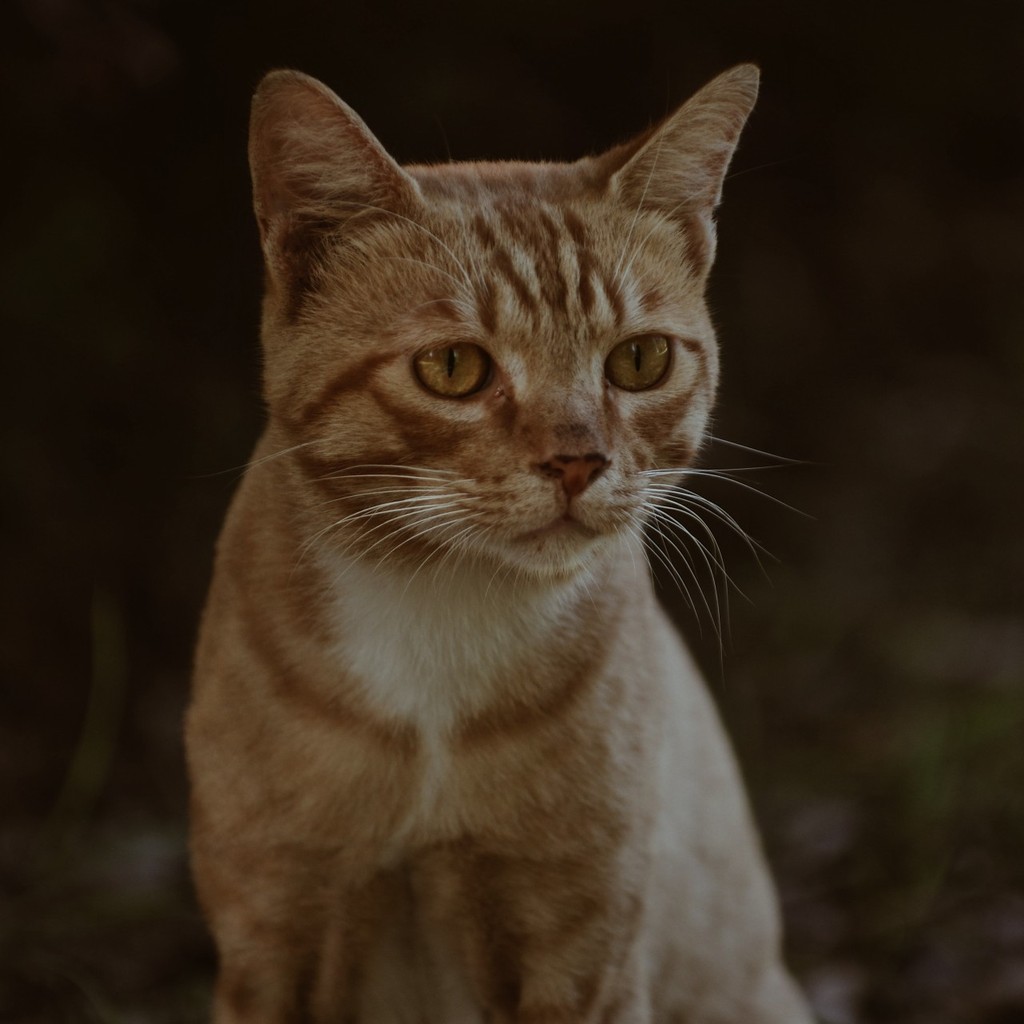}
    }
    &
    \palabel{Finetuning\\Anchoring}
    &
    \paimg{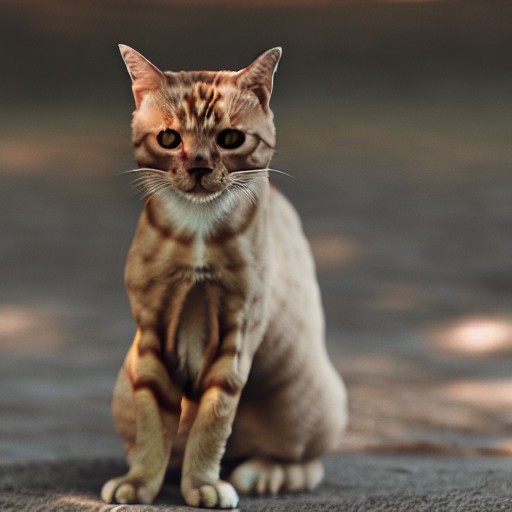}
    &
    \paimg{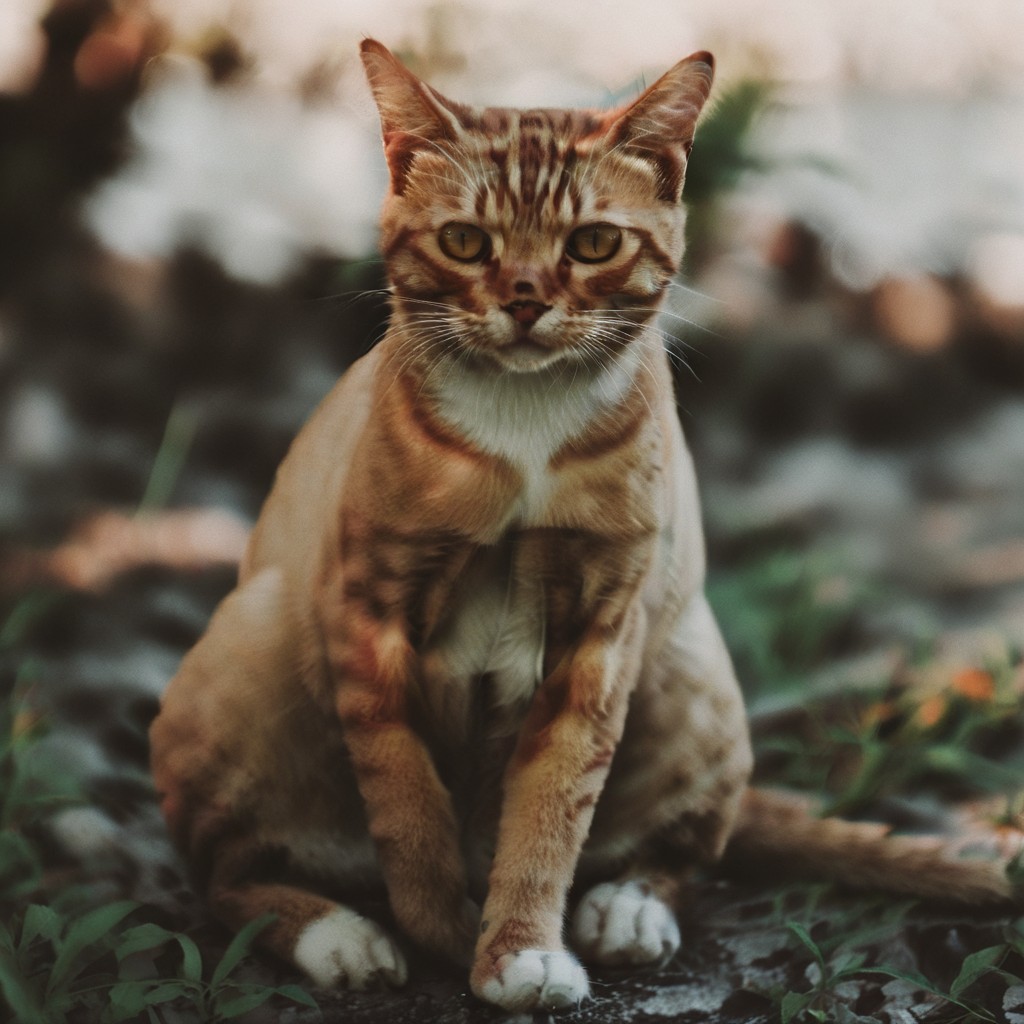}
    &
    \paimg{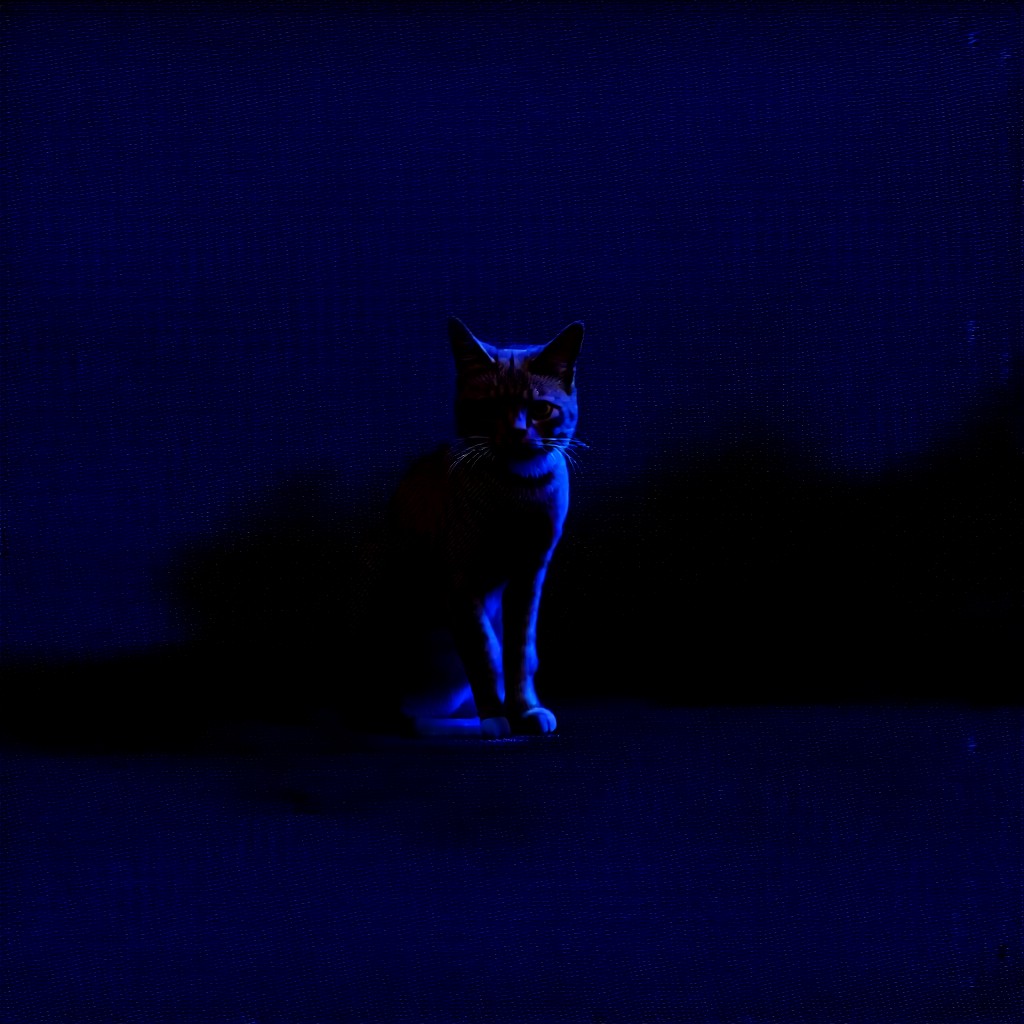}
    \\[1pt]

    % Pretrained
    {}
    &
    \palabel{Pretrained\\Anchoring}
    &
    \paimg{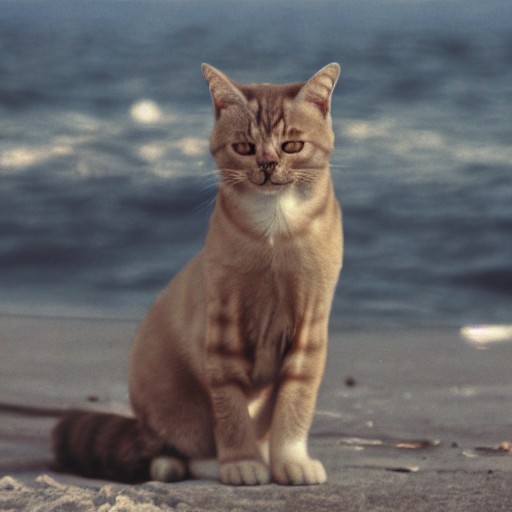}
    &
    \paimg{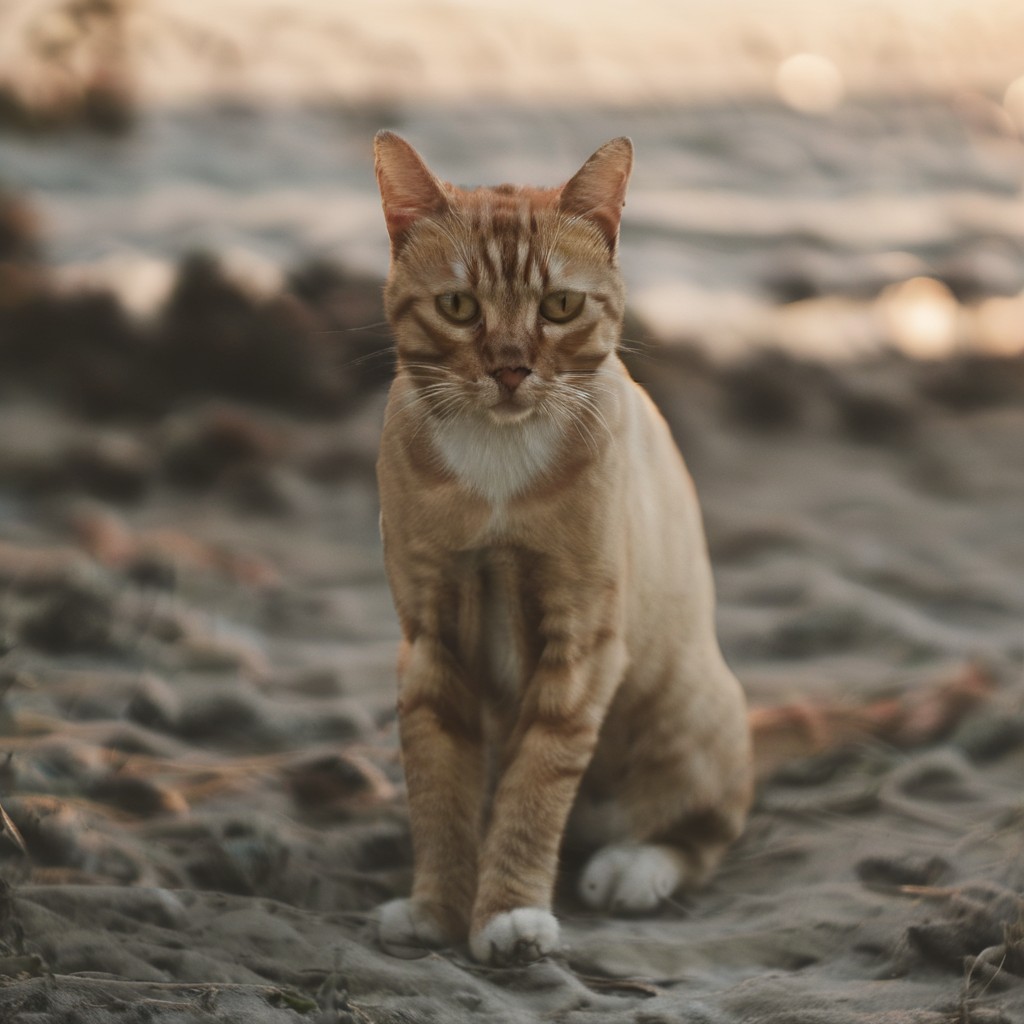}
    &
    \paimg{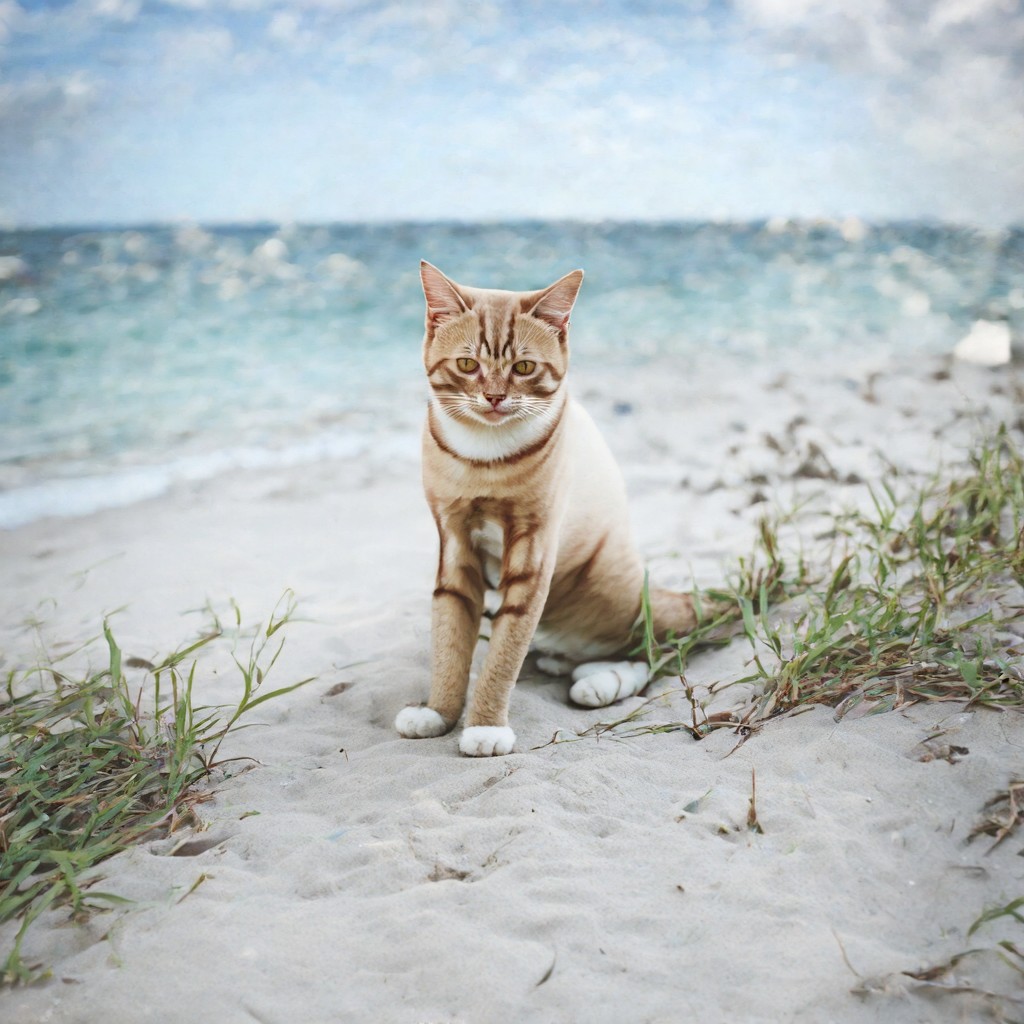}
    \\[-3pt]

    % Prompt
    {}
    &
    {}
    &
    \multicolumn{3}{c}{%
        \paprompt{``a [V*] on the beach''}
    }

    \end{tabular}

    \end{adjustbox}

    \caption{
        Comparison of qualitative results using two different
        anchoring strategies.
    }
    \label{fig:supp_pretrain_finetune}

\end{figure*}

% \begin{figure*}[t]
%     \centering
%     \includegraphics[width=\linewidth,
%                      height=0.9\textheight,
%                      keepaspectratio]{figure/supp_inference_method.png}
%     \caption{Visual comparison of different guidance strategies during inference.
%     }
%     \label{fig:supp_inference_method}
% \end{figure*}

% \input{figure/supp_inference_method}
\begin{figure*}[t]
    \centering

    \setlength{\tabcolsep}{1.5pt}
    \renewcommand{\arraystretch}{1.0}

    % =========================================================
    % References: 간격 없는 2×2 배열
    % 전체 너비 = 0.145\textwidth
    % =========================================================
    \newlength{\infRefTileSize}
    \setlength{\infRefTileSize}{0.0725\textwidth}

    \newcommand{\infFourRefs}[4]{%
        \vbox{%
            \offinterlineskip
            \hbox{%
                \includegraphics[
                    width=\infRefTileSize,
                    height=\infRefTileSize
                ]{#1}%
                \includegraphics[
                    width=\infRefTileSize,
                    height=\infRefTileSize
                ]{#2}%
            }%
            \hbox{%
                \includegraphics[
                    width=\infRefTileSize,
                    height=\infRefTileSize
                ]{#3}%
                \includegraphics[
                    width=\infRefTileSize,
                    height=\infRefTileSize
                ]{#4}%
            }%
        }%
    }

    \begin{tabular}{
        @{}
        C{0.145\textwidth}      % References
        C{0.165\textwidth}      % Prompt
        *{4}{C{0.165\textwidth}} % Methods
        @{}
    }

    % =========================================================
    % Header
    % =========================================================
    \infheader{References}
    &
    {}
    &
    \infheader{Ours}
    &
    \infheader{Pretrained\\Unconditioned}
    &
    \infheader{Finetuned\\Anchored}
    &
    \infheader{Pretrained\\Anchored}
    \\[3pt]

    % =========================================================
    % Row 1: Bear
    % =========================================================
    \infFourRefs
        {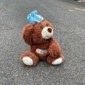}
        {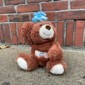}
        {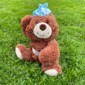}
        {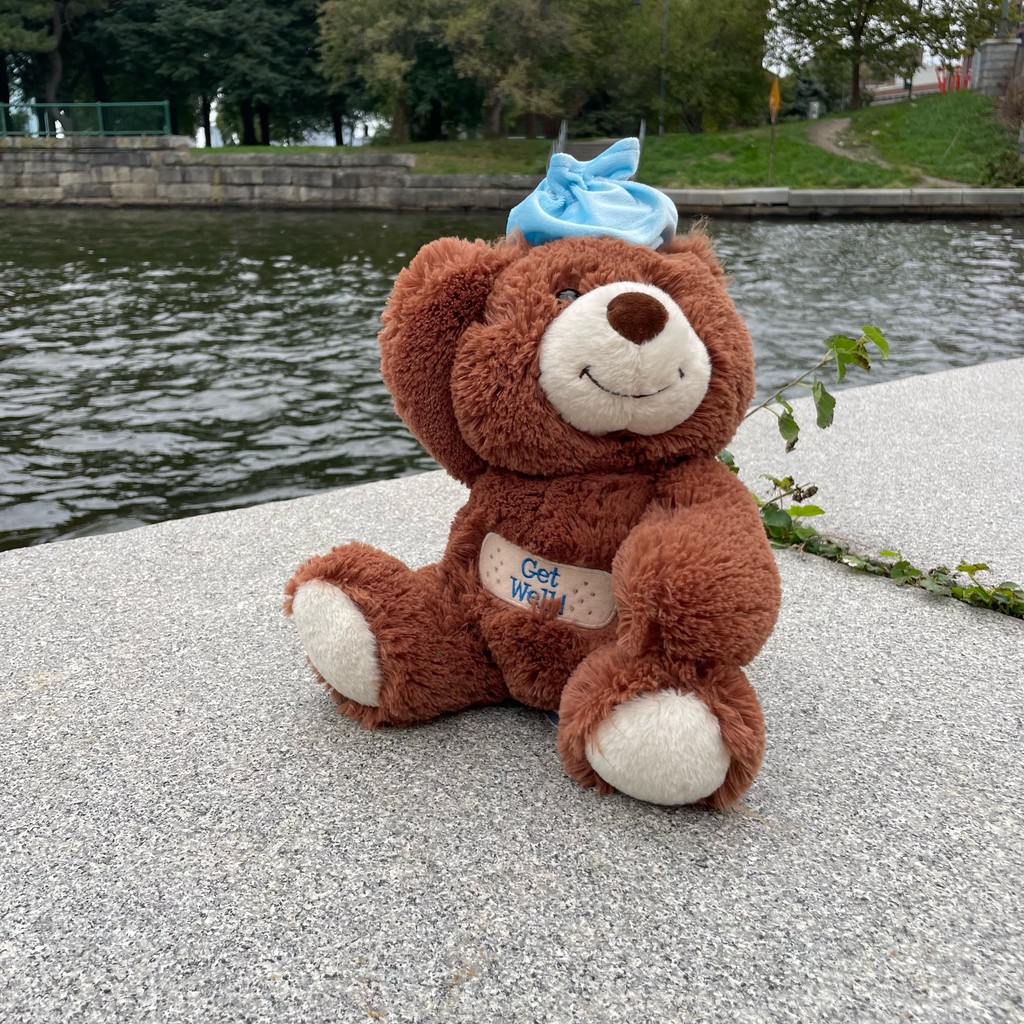}
    &
    \infprompt{
        ``a [V*] with a\\
        mountain in\\
        the background''
    }
    &
    \infimg{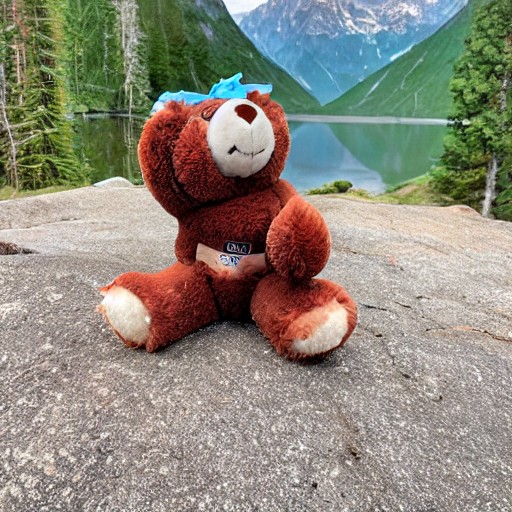}
    &
    \infimg{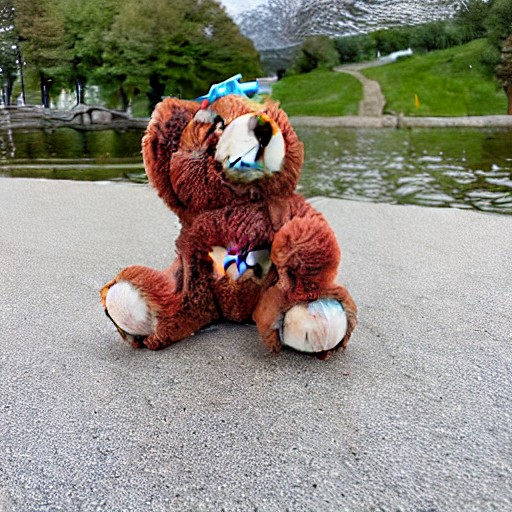}
    &
    \infimg{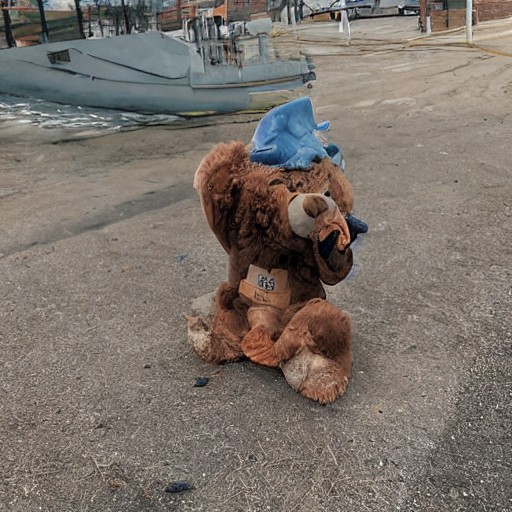}
    &
    \infimg{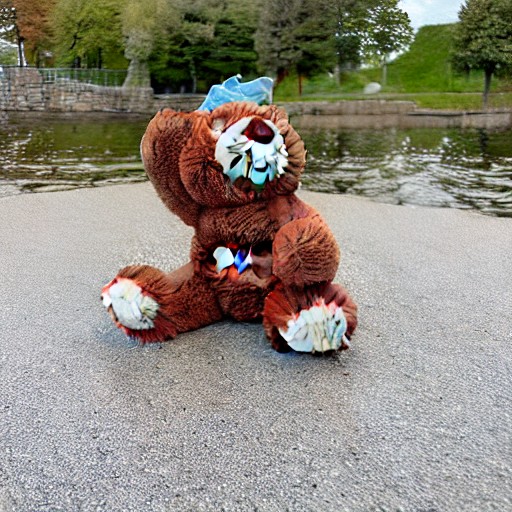}
    \\[3pt]

    % =========================================================
    % Row 2: Dog
    % =========================================================
    \infFourRefs
        {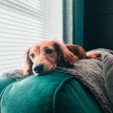}
        {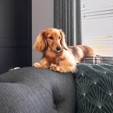}
        {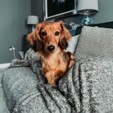}
        {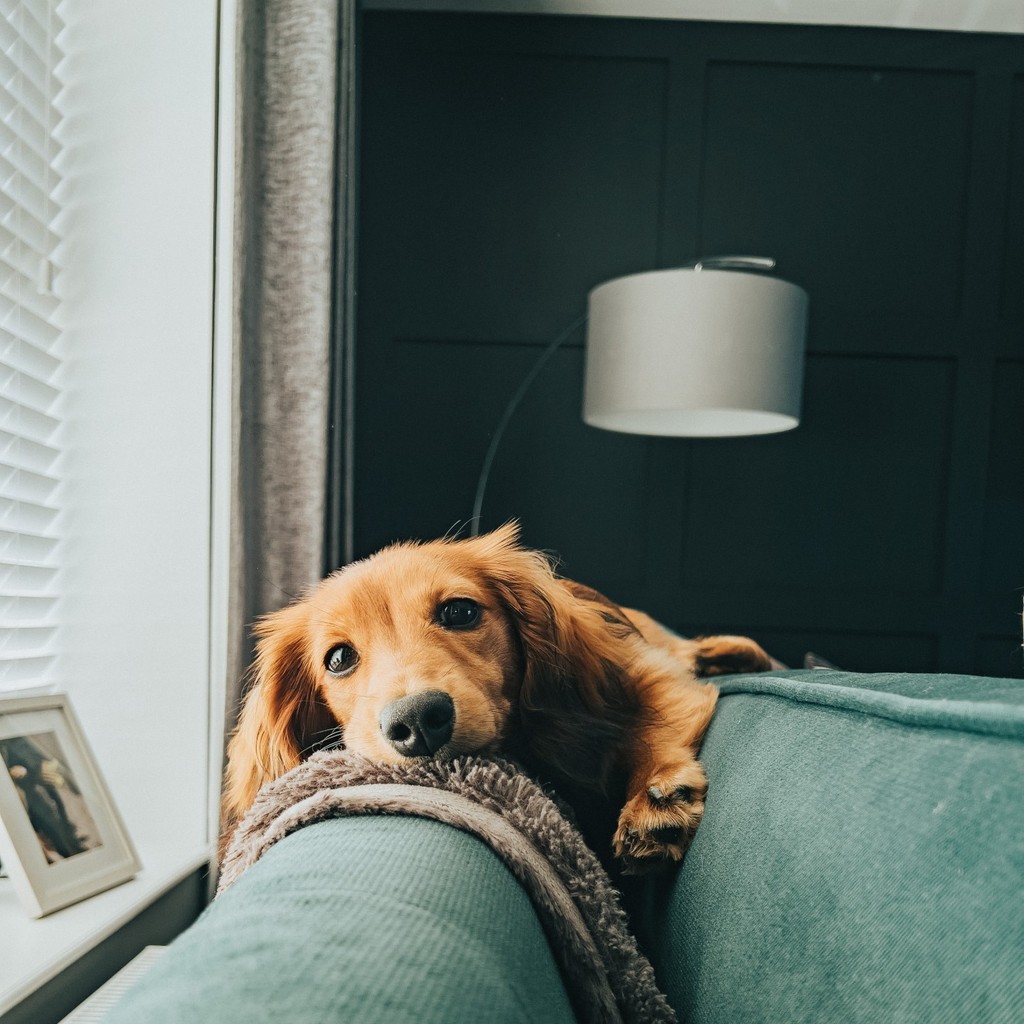}
    &
    \infprompt{
        ``a [V*] on top\\
        of a purple rug\\
        in a forest''
    }
    &
    \infimg{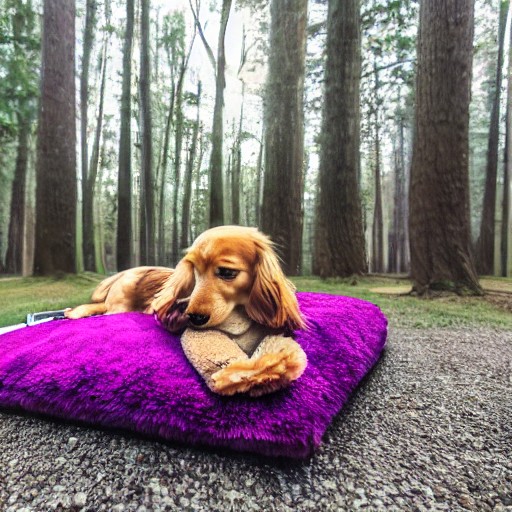}
    &
    \infimg{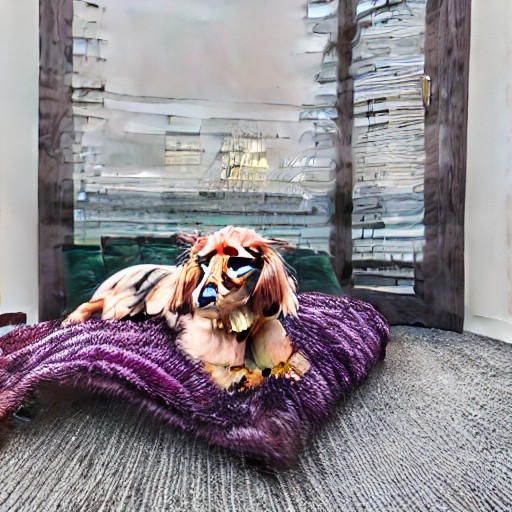}
    &
    \infimg{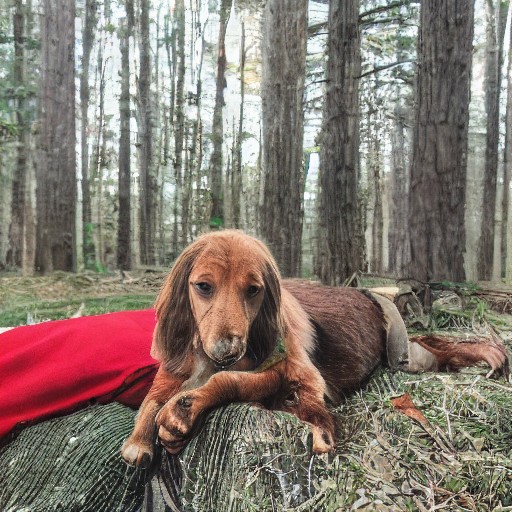}
    &
    \infimg{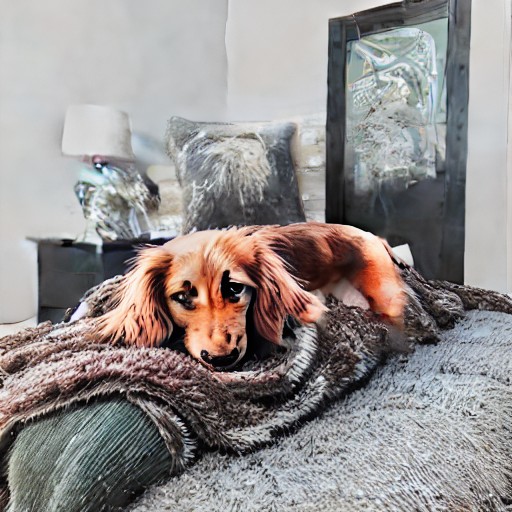}
    \\[3pt]

    % =========================================================
    % Row 3: Cat
    % =========================================================
    \infFourRefs
        {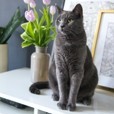}
        {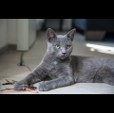}
        {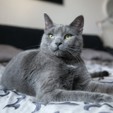}
        {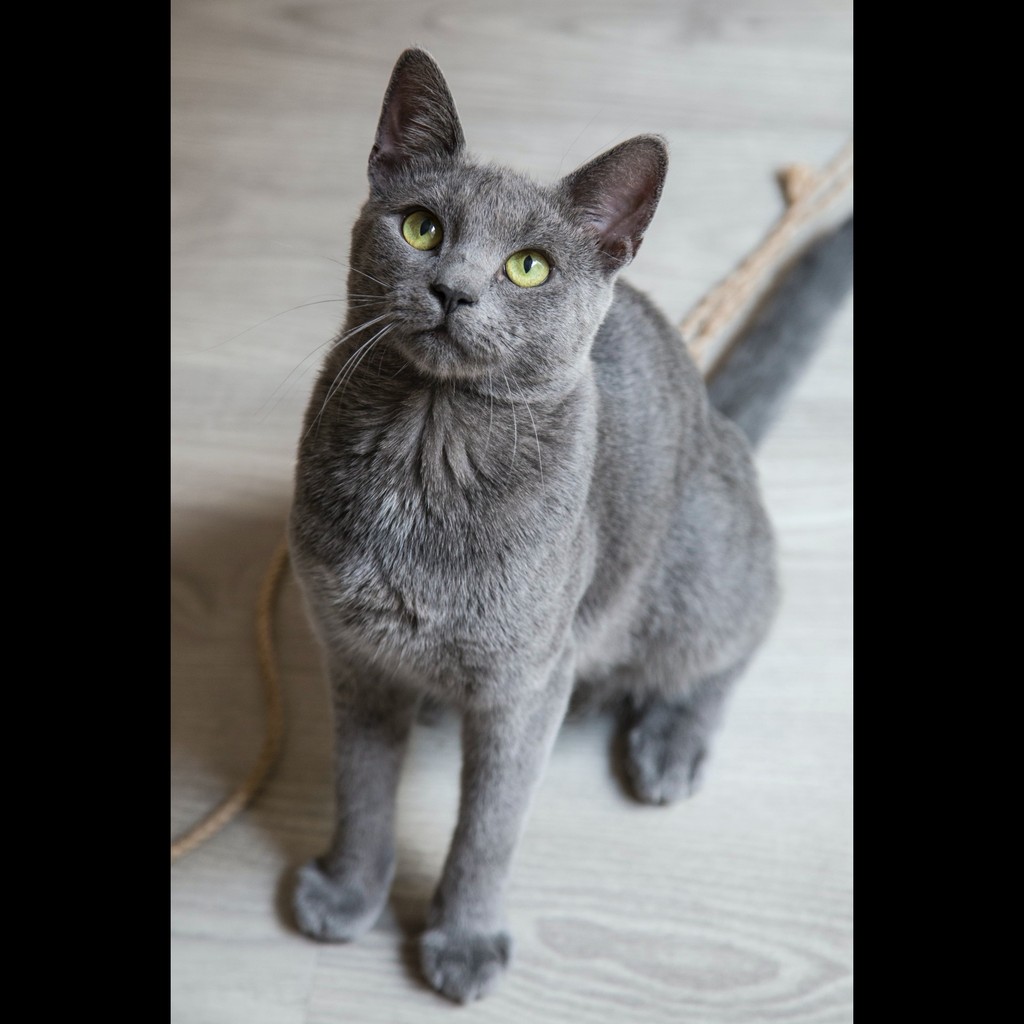}
    &
    \infprompt{
        ``a [V*] with a\\
        mountain in\\
        the background''
    }
    &
    \infimg{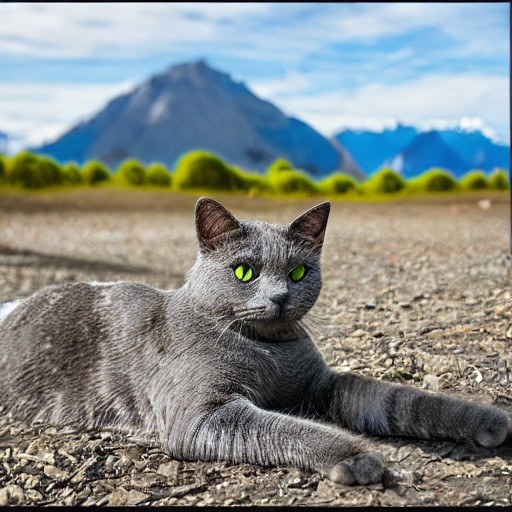}
    &
    \infimg{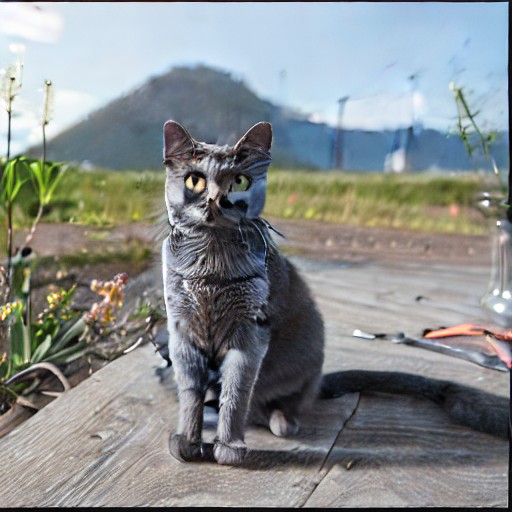}
    &
    \infimg{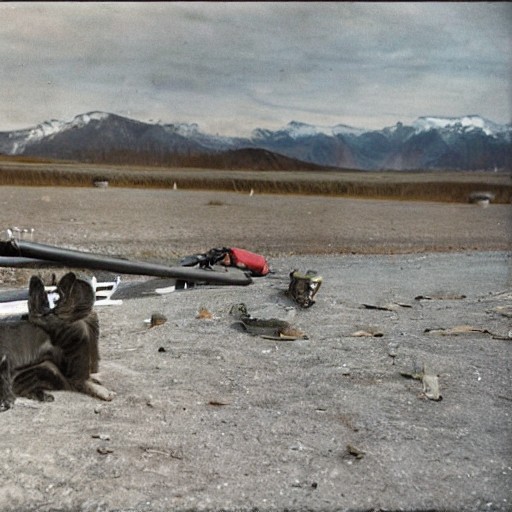}
    &
    \infimg{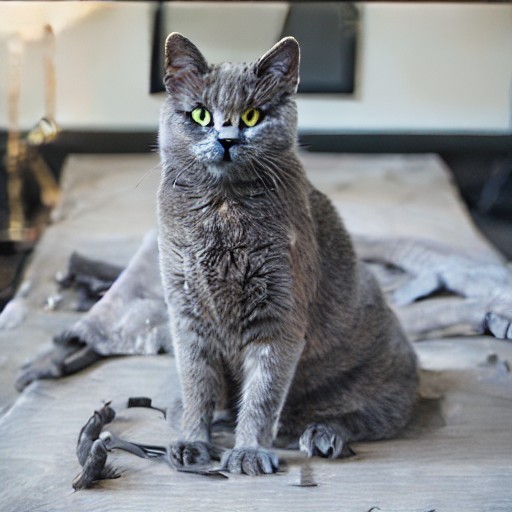}

    \end{tabular}

    \caption{
        Visual comparison of different guidance strategies during inference.
    }
    \label{fig:supp_inference_method}

\end{figure*}

\end{document}